\documentclass[journal, twoside]{IEEEtran}
\usepackage{cite}
\usepackage{graphicx}
\graphicspath{{./Figs/},{./ScoreMap/},{./ResMap/},{./img/}}
\DeclareGraphicsExtensions{.pdf,.jpeg,.png,.tif,.jpg,.pdf}
\usepackage{amsmath}
\usepackage{amssymb}
\usepackage{booktabs} 
\usepackage{pifont} 
\usepackage{bm}
\usepackage{algorithm}
\usepackage{algorithmic}
\usepackage{array}
\usepackage[caption=false,font=footnotesize]{subfig}
\usepackage{dblfloatfix}
\usepackage{colortbl}  
\usepackage{xcolor}
\usepackage{url}
\usepackage[bookmarks=true,colorlinks,linkcolor=red,anchorcolor=cyan,citecolor=magenta]{hyperref}
\usepackage{diagbox}
\usepackage{multirow}
\usepackage{makecell}
\usepackage{utfsym}
\newcommand{\PreserveBackslash}[1]{\let\temp=\\#1\let\\=\temp}
\newcolumntype{C}[1]{>{\PreserveBackslash\centering}p{#1}}
\newcolumntype{R}[1]{>{\PreserveBackslash\raggedleft}p{#1}}
\newcolumntype{L}[1]{>{\PreserveBackslash\raggedright}p{#1}}

\newcommand{\eqrefnew}[1]{Eq.~(\ref{#1})}
\newcommand{\secref}[1]{Section~\ref{#1}}

\usepackage{color}

\usepackage{todonotes}
\presetkeys{todonotes}{inline, bordercolor=white, color=blue!30}{}
\graphicspath{{./Figs/},{./ScoreMap/},{./ResMap/},{./img/}}

\UseRawInputEncoding
\begin{document}
\title{SpecDETR: A transformer-based hyperspectral point object detection network}
\author{Zhaoxu Li, Wei An, Gaowei Guo, Longguang Wang, Yingqian Wang, and Zaiping Lin  
\thanks{
	Zhaoxu Li, Wei An, Gaowei Guo, Yingqian Wang, and Zaiping Lin are with the College of Electronic Science and Technology, National University of Defense Technology, Changsha 410073, China.
	 Longguang Wang is with the Aviation University of Air Force, Changchun 130010, China.(email: lizhaoxu@nudt.edu.cn; anwei@nudt.edu.cn; guogaowei22@nudt.edu.cn; wangyingqian16@nudt.edu.cn; linzaiping@nudt.edu.cn;
	  wanglongguang15 @nudt.edu.cn)}
\thanks{Corresponding author: Wei An }
}

\maketitle

\begin{abstract}
Hyperspectral target detection (HTD) aims to identify specific materials based on spectral information in hyperspectral imagery and can detect  extremely small-sized objects, some of which occupy a smaller than one-pixel area.
However, existing HTD methods are developed based on per-pixel binary classification, neglecting the three-dimensional cube structure of hyperspectral images (HSIs) that integrates both spatial and spectral dimensions.
The synergistic existence of spatial and spectral features in HSIs enable objects to simultaneously exhibit both, yet the per-pixel HTD framework limits the joint expression of these features.
In this paper, we rethink HTD from the perspective of spatial-spectral synergistic representation and propose hyperspectral point object detection as an innovative task framework.
We introduce SpecDETR, the first specialized network for hyperspectral multi-class point object detection, which eliminates dependence on pre-trained backbone networks commonly required by vision-based object detectors.
SpecDETR uses a multi-layer Transformer encoder with self-excited subpixel-scale attention modules to directly extract deep  spatial-spectral joint features from hyperspectral cubes.
During feature extraction, we introduce a self-excited mechanism to enhance object features through self-excited amplification, thereby accelerating network convergence.
Additionally, SpecDETR regards point object detection as a one-to-many set prediction problem, thereby achieving a concise and efficient DETR decoder that surpasses the state-of-the-art (SOTA) DETR decoder.
We develop a simulated hyper$\textit{S}$pectral $\textit{P}$oint $\textit{O}$bject $\textit{D}$etection benchmark termed SPOD, and for the first time, evaluate and compare the performance of visual object detection networks and  HTD methods on hyperspectral point object detection.
Extensive experiments demonstrate that our proposed SpecDETR outperforms SOTA visual object detection networks and HTD methods. 
Our code and dataset are available at \href{https://github.com/ZhaoxuLi123/ SpecDETR}{https://github.com/ZhaoxuLi123/SpecDETR}.  
\end{abstract}                                                              


\IEEEpeerreviewmaketitle

\section{Introduction}
Hyperspectral imagery (HSI) has gained significant attention in Earth observation due to its rich spectral information in recent decades. HSI usually provides hundreds of narrow spectral bands, allowing for the land cover differentiation at the pixel level.
With advancements in hyperspectral classification \cite{PAOLETTI2018120,THOREAU2024323}, unmixing \cite{ZHU2014101}, anomaly detection \cite{SU2020195,ZHAO2016126,li2023you}, change detection \cite{HU2024465}, biomass estimation \cite{GUO2023120,XU2023169}  and target detection \cite{JIAO2018235,DONG2015138}, HSI has assumed a crucial role in various remote sensing applications such as agricultural survey, mineral exploration, and greenhouse gas detection.
Hyperspectral target detection (HTD), which utilizes prior spectral information to locate specific materials,can identify extremely small objects that are undetectable in conventional optical images, with some objects areas being even smaller than a single pixel area.

\begin{figure}[t]
	\centering
	\includegraphics[width=\linewidth]{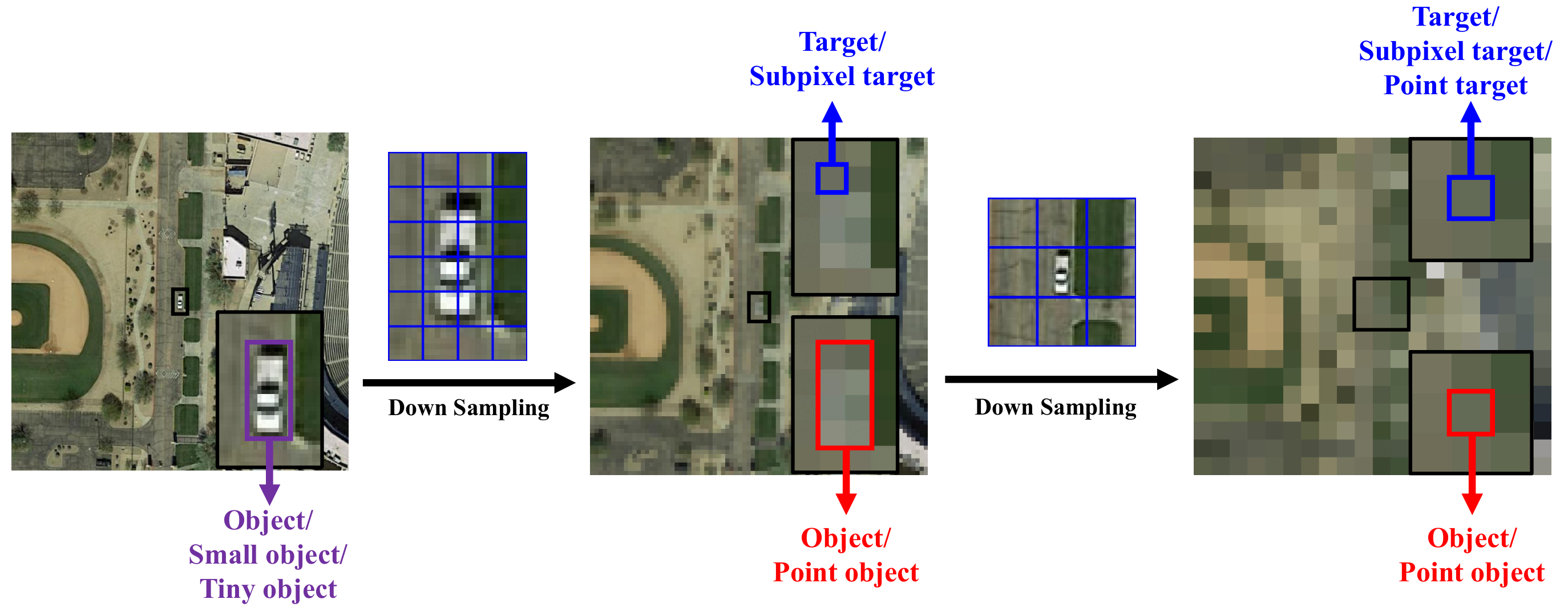}
	\caption{{Illustration of the concept of point object.}
		\label{pointcar}}
\end{figure}

HTD is generally regarded as a per-pixel binary classification problem based on prior spectral information, where each pixel in the HSI is categorized as either a target (containing the target spectrum) or background (lacking the target spectrum) based on known target spectra. 
In recent years, significant advancements have been made in the field of HTD.
Feng et al. \cite{feng2024transformer} introduced a cross-domain few-shot learning method tailored for HTD tasks, addressing the challenge of scarce target prior information.
Chang \cite{chang2024constrained} delved into the mathematical principles of Constrained Energy Minimization (CEM) widely used in sub-pixel HTD and proposed several novel CEM generalizations.
These HSI methods treat the 3D hyperspectral cube as independent spectral vectors.
Notably, some HTD approaches incorporate spatial characteristics. 
For instance, Feng et al. \cite{feng2023coarse} proposed a coarse-to-fine HTD algorithm based on low-rank tensor decomposition, leveraging the spatial-temporal properties of HSI to extract pure background information.
But these HTD methods still operate within a per-pixel classification framework, utilizing only the prior spatial information of targets, and fail to fully represent the  spatial-spectral joint features of objects.
The unique attribute of hyperspectral images, which integrate spatial and spectral features, allows objects to exhibit both simultaneously.
We believe that such a per-pixel HTD framework can restrict the representation of instance-level  spatial-spectral joint features of objects, leading to limitations in detection capability.
On one hand, the current HTD framework primarily utilize prior target spectral information, often neglecting the extraction and utilization of instance-level object spatial characteristics. On the other hand, existing HTD methods typically output detection scores for each pixel, failing to directly provide instance-level prediction results. The success of visual object detection networks such as Faster R-CNN \cite{girshick2015fast}, YOLO \cite{redmon2016you}, and Detection Transformer (DETR) \cite{carion2020end} offers a new perspective. Compared to the per-pixel binary classification framework in HTD, visual object detection networks can directly yield instance-level category and location predictions. Furthermore,  HTD relies exclusively on prior spectra and test images for detection, while visual object detection networks can effectively learn robust, high-level, instance-level features from large-scale training images annotated with instance-level object information. During training, the learning frameworks of visual object detection networks can focus on difficult-to-classify samples, such as easily confused objects categories and background regions similar to objects. Therefore, in this paper, we introduce visual object detection into hyperspectral target detection and expand hyperspectral target detection to hyperspectral point object detection.
Considering the differences between HTD and visual object detection, we will redefine the concepts of target and object, providing a clear definition of point object. Furthermore, we will elucidate the relationship between spatial information and spectral information, thereby clarifying the motivation behind our proposed framework for point object detection.

We have observed differences in similar concepts between the fields of visual object detection and HTD. In Fig.~\ref{pointcar}, we illustrate the distinctions among the object, the target in HTD, and the point object of focus in our paper using a vehicle object as an example. Object detection aims to locate and classify all pixels belonging to a single instance as a whole. In the object detection filed, small objects or tiny objects are typically defined by the pixel number and can still be localized and classified based on their spatial information. For instance, the COCO dataset \cite{lin2014microsoft} and the SODA dataset \cite{cheng2023towards} consider objects with a bounding box (bbox) area equal to or smaller than 1024 pixels as small objects, which still contain certain texture and shape detail information. In the field of HTD, most literature defines a target as a pixel containing spectral features of interest, and a subpixel target as a pixel whose target spectral proportion, also termed as target abundance, is less than 1. Some literature \cite{pointtarget1,pointtarget2,pointtarget3} specifically uses the terms point target or subpixel target to refer to instances that occupy only a single pixel. However, these works still employ the per-pixel classification framework and pixel-level evaluation metrics, essentially adhering to the pixel concept. We focus on shifting from pixel-based to instance-based concept, treating target pixels belonging to the same instance as a collective unit. We refer to these instances, which lack spatial detail features and are composed of very few pixels, as point objects. The definition of point objects is not based on pixel number, owing to the presence of many mixed pixels within their composition. Any objects whose sizes are close to or smaller than the spatial resolution or ground sample distance (GSD) of the image, and whose shape information alone is insufficient to support their classification, can be considered as point objects.

\begin{figure}[t]
	\centering
	\subfloat[\label{sfig:aircraft}]{\includegraphics[width=0.33\linewidth]{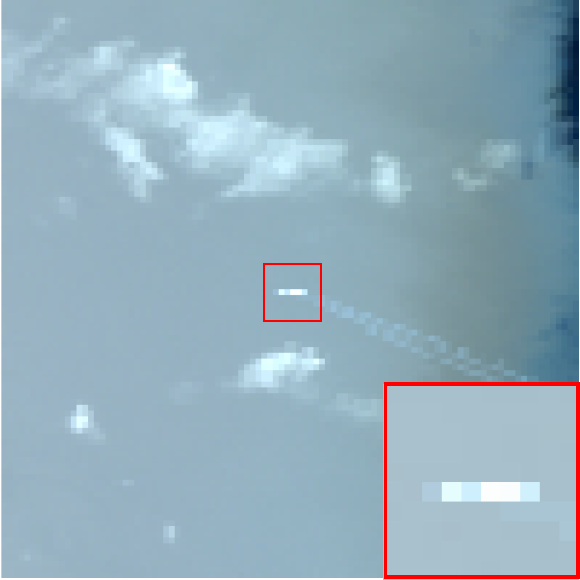}}
	\hfil
	\subfloat[\label{sfig:vehicle}]{\includegraphics[width=0.33\linewidth]{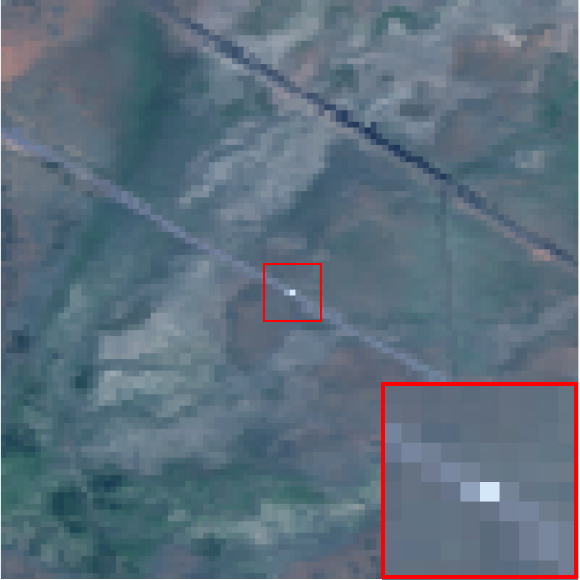}}
	\hfil
	\subfloat[\label{sfig:ship}]{\includegraphics[width=0.33\linewidth]{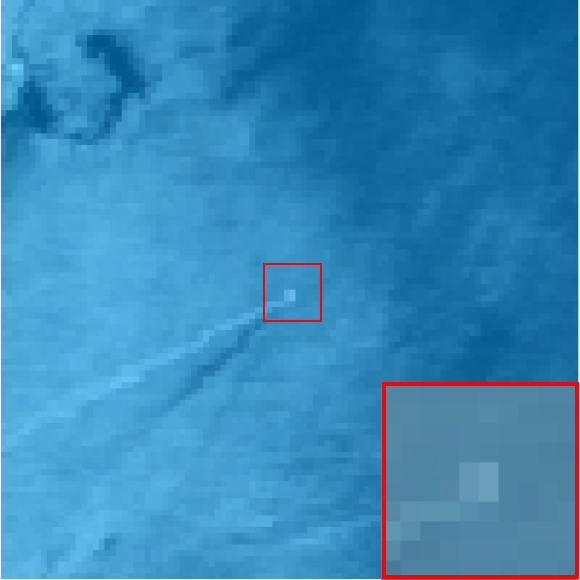}}
	\hfil
	\\
	\subfloat[\label{sfig:spec}]{\includegraphics[width=1\linewidth]{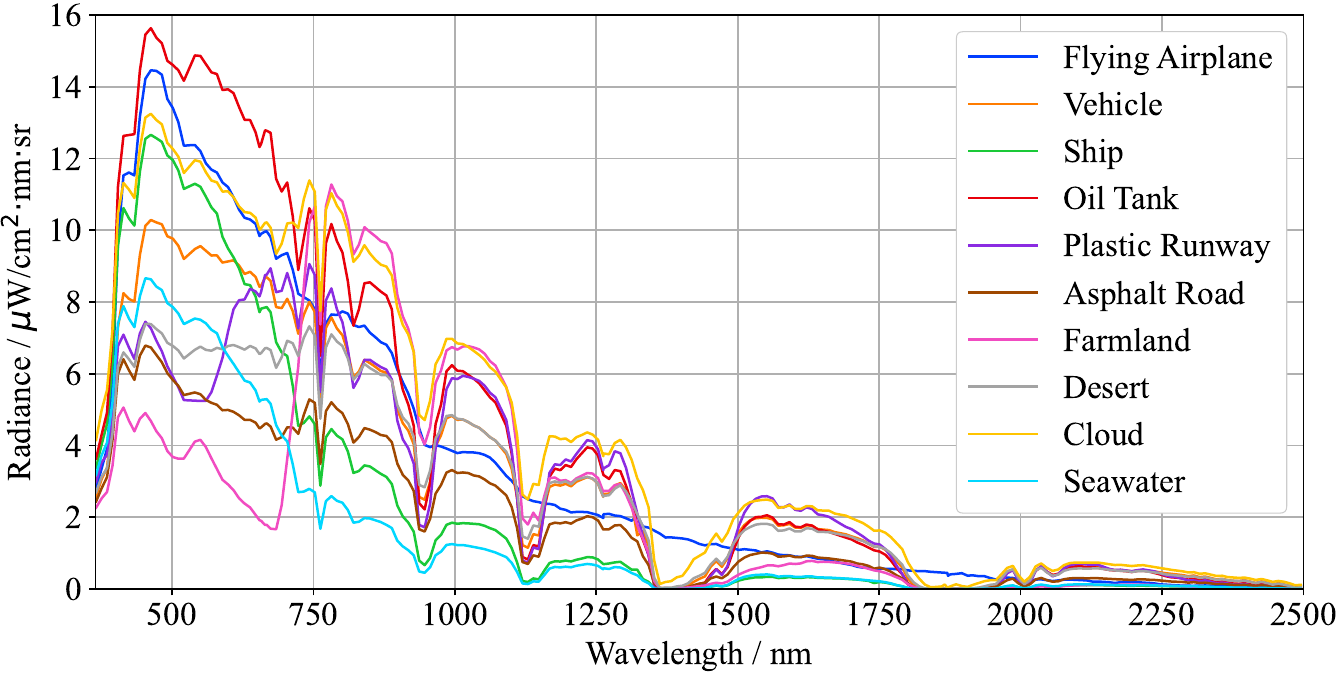}}
	\hfil
	\caption{Real-world point objects in HSI collected by AVIRIS. (a) Flying airplane. (b) Vehicle. (c) Ship. (d) Spectral radiance curves of the shown point objects and other ground objects.}
	\label{fig:pointobject}
\end{figure}

In the field of computer vision (CV), researchers typically employ feature extraction networks to extract spatial features  as shape and texture from RGB images for visual downstream tasks such as classification, object detection, and instance segmentation.
In contrast, hyperspectral imaging systems generate data cubes containing hundreds of spectral bands, prompting HTD methods to prioritize spectral information utilization
However, hyperspectral images are inherently spatial-spectral integrated, forming a three-dimensional cube that encompasses both spatial and spectral dimensions. 
Fig.~\ref{fig:pointobject} illustrates the pseudo-color images and spectral curves of three real-world point objects: an airplane in flight, a car on the road, and a ship on the sea surface.
The limited spatial resolution results in deficient shape and texture details for these objects, as shown in Fig.~\ref{fig:pointobject}(a-c). 
While spatial information permits coarse localization of these point objects, it proves insufficient for reliable classification.
Fig.~\ref{fig:pointobject}(d) presents the spectral curves of the three point objects and other common ground objects, revealing distinct differences among  their spectral signatures due to physical mechanisms. For instance, the shown airplane, owing to its high altitude and proximity to the sensor, exhibits less spectral radiation attenuation in the two atmospheric absorption bands at 1.4 $\mu $m and 1.8 $\mu $m, resulting in significantly higher radiation values at these bands compared to those of other curves. 
This observation motivates our investigation: Can we effectively extract both spatial and spectral features of point objects?
We believe that current object detection frameworks inherently can represent spatial-spectral joint features.
The spatial features manifested in single channel imagery or RGB imagery fundamentally arise from spectral radiance variations across pixel locations. As shown in Fig.~\ref{fig:pointobject}(a-c), the observed spatial characteristics essentially represent a limited  spatial-spectral joint representation constrained by limited spectral dimensionality.
From this perspective, current visual detection networks already demonstrate foundational capacity for  spatial-spectral joint feature learning from hyperspectral cube data.
Nevertheless, employing the object detection framework to achieve end-to-end localization and classification of subpixel-level point objects remains, to our knowledge, an unexplored challenge—the focal point of this paper.

Inspired by the current DETR-like detectors, we propose the first specialized network for hyperspectral point object detection,  SpecDETR. We treat the spectral features of each pixel in the HSI as a token serving as the fundamental operation unit in Transformer \cite{vaswani2017attention}. In existing visual object detection networks, the pretrained feature extraction network, referred to as the backbone, is used to extract deep features from input RGB images, which are then fed into the detection head for object localization and classification. Unlike these visual object detection networks, we bypass the backbone and directly utilize a Transformer encoder to extract deep features from spectral tokens.
Building upon the deformable attention module \cite{zhu2020deformable}, we propose a self-excited subpixel-scale attention module (S2A module) for the extraction of  spatial-spectral joint features of point objects, which efficiently achieves subpixel-scale deformable sampling while enabling self-excited amplification of object features.
Furthermore, based on the DINO \cite{zhang2022dino} decoder, we design a concise decoder framework tailored for point object detection, integrating one-to-many label assignment and the end-to-end advantages of DETR. With the same attention operations and decoder layer number, our decoder achieves superior point object detection accuracy with fewer parameters than the DINO decoder.
To advance the development of hyperspectral point object detection, we developed a hyper\textbf{S}pectral multi-class \textbf{P}oint \textbf{O}bject \textbf{D}etection dataset, termed \textbf{SPOD}. 
Using the SPOD dataset, we thoroughly evaluated the mainstream visual object detection networks and  HTD methods. 
Our SpecDETR outperforms state-of-the-art (SOTA) visual object detection networks and HTD methods on the SPOD dataset.
The contributions of this paper can be summarized as follows: 

\begin{enumerate}
	\item We extend hyperspectral target detection to hyperspectral point object detection and propose the first specialized hyperspectral point object detection network, SpecDETR.
	\item We regard SpecDETR as a one-to-many set prediction problem and develop a concise and efficient DETR decoder framework that surpasses the current SOTA DETR decoder framework on point object detection.
	\item We develop a hyperspectral point object detection benchmark and evaluate visual object detection networks and HTD methods on the hyperspectral point object detection task for the first time.
\end{enumerate}

The rest of this paper is organized as follows. \secref{sec:RelateWork} reviews the fields related to our work. \secref{sec:method} describes the details of SpecDETR. The datasets and experiments are presented in \secref{sec:datasets} and \secref{sec:experiments}, respectively, followed by the conclusion in \secref{sec:Cnl}.

\section{Related Work}
\label{sec:RelateWork}
\subsection{Hyperspectral Target Detection}
HTD aims to locate targets based on the spectral characteristics in HSIs \cite{Ovrv13-TD}. Classic HTD methods can be categorized into matching methods \cite{SMF-TD}, statistical analysis methods \cite{CEM-TD, ACE-TD, HSS-TD}, and subspace-based methods \cite{ASD99-TD, OSP-TD, TCIMF-TD}. While these methods demonstrate effective target-background separation in homogeneous scenarios, they have limitations in handling complex scenarios.
In response to these limitations, machine learning-based techniques such as sparse representation \cite{SJSR-TD, SR-TD, KSR-C, SRBBH-TD, CSRBBH-TD, KSRBBH-TD,zhu2023learning} and kernel methods \cite{KCEM, KOSP, kwon2006comparative, KMSD-TD, KTCIMF-TD, SVMCK-C} have gained popularity due to their ability to handle complex data and achieve impressive performance. 
Recently, many deep learning methods have been specifically designed to address the characteristics of HTD. Traditional HTD typically provide only a few, or even a single, prior spectral curve along with a test image, which results in a small sample number and may lead to network overfitting. Some methods \cite{jiao2023triplet,sun2023ablal,luo2024agms} employ the data generation technique to enhance sample diversity. Others, like MLSN \cite{wang2022MLSN} and TLNSS \cite{shi2023TLNSS}, introduce transfer learning to mitigate the problem of insufficient training samples.
In addition, some method incorporate prior spectra into their network structure by adopting siamese networks for contrastive learning, including TSCNTD  \cite{zhu2020two}, STTD \cite{rao2022siamese}, AGMS \cite{luo2024agms}, and CS-TTD \cite{yang2024cs}. While other methods design networks based on physical models, such as JSPEN \cite{dong2024JSPEN} and IRN \cite{shen2023hyperspectral}.
Given that HTD involves the processing of one-dimensional spectral curves, Transformer, which originated in the NLP domain, have gained popularity in HTD research. Notable examples include STTD,CS-TTD, TSTTD \cite{jiao2023triplet}, HTDFormer \cite{li2023htdformer}, MCLT \cite{wang2024MCLT},  and SGT \cite{chen2024SGT}.
However, current HTD methods are based on a per-pixel binary classification framework which restricts the representation of instance-level object features.
In this paper, we explore the feasibility of adopting an instance-level object detection framework as an alternative to the current spectral-level HTD framework.

\subsection{Object Detection}

In recent years, the field of visual object detection has made significant progress driven by deep learning, encompassing two-stage detectors, single-stage detectors, and DETR-like detectors. 
Two-stage detectors, such as Faster R-CNN \cite{girshick2015fast}, provide more precise object detection through a two-step process but are computationally intensive, whereas single-stage detectors like YOLO \cite{redmon2016you} offer faster, albeit slightly less accurate, detection.
Carion et al. \cite{carion2020end} proposed an end-to-end object detector based on Transformer \cite{vaswani2017attention}, named DETR, which reformulates object detection as a one-to-one set prediction problem. Deformable DETR \cite{zhu2020deformable} significantly improves the training efficiency of DETR by introducing deformable attention mechanisms and multi-scale feature fusion. DAB-DETR \cite{liu2022dab} decomposes the object query into content and positional queries and introduces explicit positional representation. DN-DETR \cite{li2022dn} and Group DETR \cite{chen2023group} introduce denoising training and grouped queries, respectively, to address the instability of bipartite matching and accelerate model convergence. Building on these works, DINO \cite{zhang2022dino}, through the introduction of contrastive denoising training, mixed query selection, and look forward twice techniques, surpasses CNN detectors on the visual object detection benchmark COCO dataset \cite{lin2014microsoft}, achieving the best results. 
CO-DETR \cite{zong2023detrs} further improves detection accuracy by introducing more dense supervision signals to enhance the discriminability of encoder features.
However, these DETR-like detectors are typically computationally expensive. RT-DETR \cite{rtdetr} achieves real-time detection performance comparable to that of the YOLO series. YOLOv10 \cite{yolov10} combines the label assignment strategies of DETR  with CNN architecture, achieving an optimal balance between detection accuracy and inference speed.
These visual object detection networks are designed for objects with distinct shape and texture details. In this work, we aim to further optimize and customize the DETR framework for the point object detection task.

\subsection{Hyperspectral Object Detection}
Recently, object detection has been introduced to hyperspectral image processing, showing promising results in fields such as camouflage object detection \cite{yan2021object}, vehicle detection \cite{lee2021channel, rangnekar2022semi, 10225553}, and methane gas detection \cite{kumar2023methanemapper}.
For instance, S2ADet \cite{10225553} employs a dual-backbone network architecture, where one backbone processes color images while the other operates on three-channel images derived from hyperspectral data through Principal Component Analysis, extracting spatial and spectral semantic information, respectively. MethaneMapper \cite{kumar2023methanemapper} uses two ResNet networks to extract features from visible and shortwave infrared bands, respectively.
These works rely on visual backbone networks pre-trained on large RGB datasets.
Similarly, recent popular hyperspectral object tracking networks also follow this approach \cite{SEE-Net23,SiamHYPER22,HA-Net24,RawTrack}.
SEE-Net uses pre-trained backbone networks to extract features from different pseudo-color image combinations \cite{SEE-Net23}.
SiamHYPER adds a hyperspectral awareness module to the RGB-based Siamese tracker, extracting features through dual-inputs of RGB and hyperspectral data \cite{SiamHYPER22}.
HA-Net also employs dual-inputs of RGB and hyperspectral data to extract highly discriminative semantic features \cite{HA-Net24}.

On one hand, the aforementioned researches primarily focus on large-sized objects and are not well-suited to address the challenge of subpixel object detection, which is a critical concern in the traditional HTD task. 
On the other hand, these works still add spectral information extraction designs on top of pre-trained backbone networks based on RGB inputs. 
In this paper, we propose the hyperspectral point object detection task, aiming to tackle the problem of detecting extremely small objects at the subpixel scale within the object detection framework. Furthermore, we explore whether the current object detection framework can directly extract spatial-spectral joint features from hyperspectral cube data.

\section{Methodology}
\label{sec:method}
\subsection{Task Definition}
Traditional HTD methods use a prior target spectral library to perform binary classification on pixels.
Our work extends visual object detection to the hyperspectral point object detection task, aiming to achieve object-level detection of point objects.
Our network learns object-level features on a set of training HSIs with bounding box (bbox) annotations of point objects.
If annotations include category labels, the trained network can predict the categories of point objects. In contrast, the HTD methods always treat all target prior spectra as the same category and neglect classification ability.

For the hyperspectral point object detection task, we adopt the GT label format of the common object detection such as the COCO dataset \cite{lin2014microsoft}. 
For each hyperspectral image in both the training and testing sets, the GT label comprises the true bbox information and category label for each object. 
Each object is accompanied by a bbox, which precisely indicates the object's location in the image.
This rectangular box is typically represented by four parameters: the coordinates of the top-left corner (x, y), and the width (w) and height (h) of the box, or alternatively, the center coordinates (x, y), width (w), and height (h). These two forms of representation are usually equivalent.
The GT label for an object also includes a category label.

\begin{figure*}[t]
	\centering
	\includegraphics[width=0.9\linewidth]{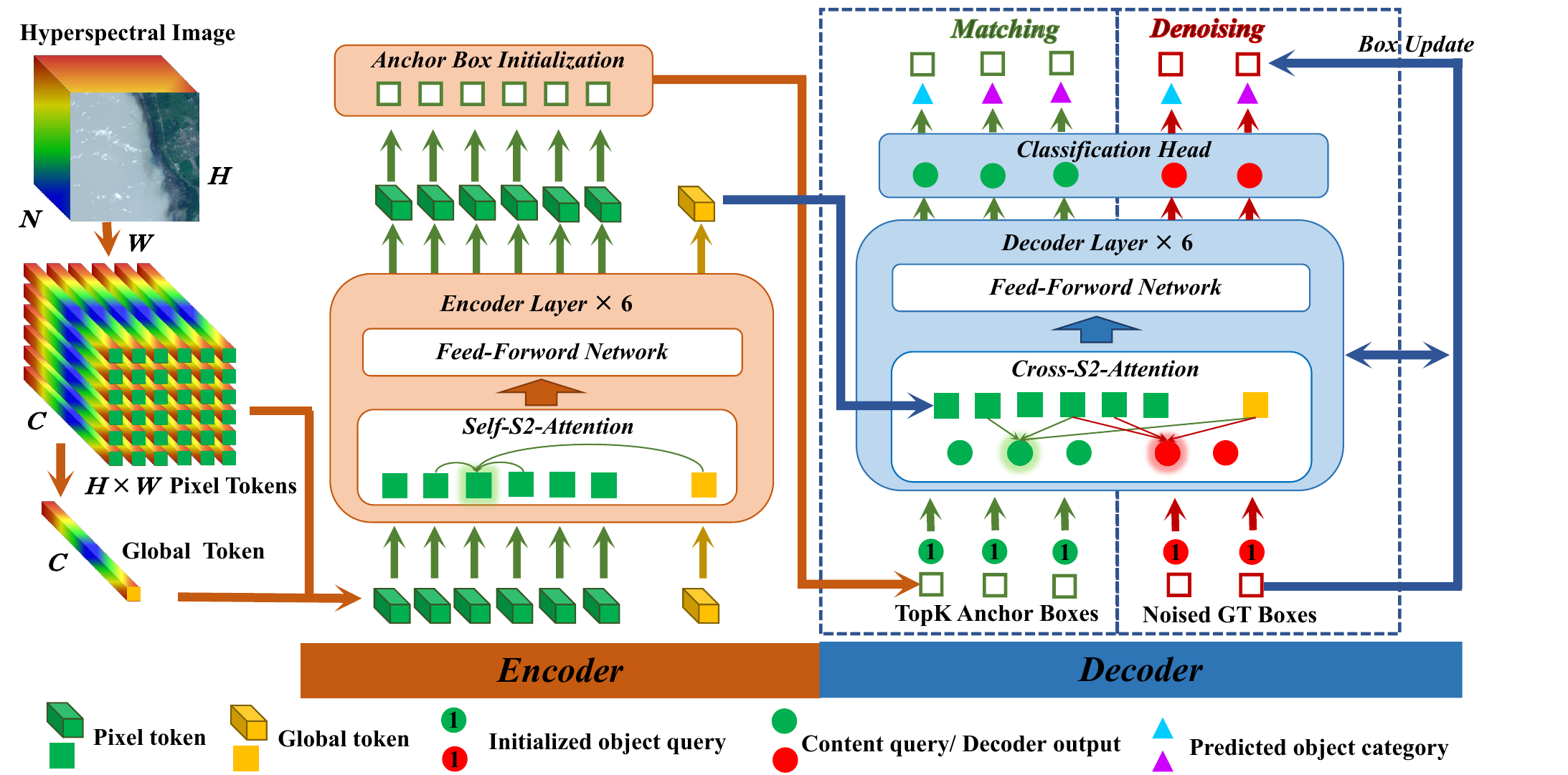}
	\caption{{An overview of our SpecDETR. For clear visualization, we omit positional embedding and only display the classification prediction head of the final decoder layer.}
		\label{overview}}
\end{figure*}


\subsection{Overview}
As shown in Fig.~\ref{overview}, our SpecDETR is built upon DINO \cite{zhang2022dino} and removes the backbone network of DINO. It only consists of a multi-layer Transformer encoder, a multi-layer Transformer decoder, and multiple prediction heads. 
Given an HSI, we use a linear layer to transform the channel number and treat each pixel as a token. Then, we feed them into the Transformer encoder to extract deep features. 
Based on the output of the multi-layer encoder, a candidate anchor box is generated at each pixel location, from which a fixed number of anchor boxes are selected. For each selected anchor box, an initial object query vector is generated. The initial object queries and their corresponding anchor boxes are then fed into the multi-layer decoder and updated layer-by-layer.
The output of SpecDETR consist of refined bboxes and classification results generated by updated object queries.
Following DINO, SpecDETR introduces an additional denoising branch alongside the original decoder pipeline which is termed the matching branch.
Unlike current DETR-like detectors, SpecDETR regards point object detection as a one-to-many set prediction problem, and accordingly, it has been specifically refined.
We propose Center-Shifting Contrastive DeNoising (CCDN) training to replace Contrastive DeNoising (CDN) training in DINO. 
In the matching branch, we use a hybrid one-to-many label assigner instead of a Hungarian matching-based one-to-one label assignment in current DETR-like detectors. Based on the one-to-many matching, we simplify the decoder by removing cross-attention between object queries and initializing object queries as non-learnable, uniform vectors.
Additionally, SpecDETR uses non-maximum suppression (NMS) to remove overlapping prediction boxes. 
The loss function of SpecDETR is consistent with that of DINO.
We will elaborate on the details of our improvements in the following sections.

\subsection{Data Tokenization}
Given an HSI data cube $\boldsymbol{X} \in \mathbb{R} ^{H\times W\times N}$, where $H$, $W$, and $N$ represent the width, height, and band number, we preprocess $\boldsymbol{X}$ before feeding it into the encoder by:
\begin{equation}
	\label{eq_1}
	\boldsymbol{P}_0=\text{LN}\left(  \text{linear}\left( \frac{\boldsymbol{X}}{V} \right) \right) ,
\end{equation}
where $\text{linear} \left( \cdot \right)$ represents a linear layer, $\text{LN} \left( \cdot \right)$ represents a layer normalization layer \cite{ba2016layer}, and $V$ is a constant. The output data cube $\boldsymbol{P}_0\in \mathbb{R} ^{H\times W\times C}$ has an initialized feature vector at each pixel position, serving as a pixel token for the following Transformer, where $C$ represents the dimension of tokens.
Subsequently, a initialized global token $\boldsymbol{g}_0$ is initialized by:
\begin{equation}
	\label{eq_2}
	\boldsymbol{g}_0=\frac{1}{H\times W}\sum_{i=1}^{H\times W}{\boldsymbol{p}_{0,i}},
\end{equation}
where $\boldsymbol{p}_{0,i}$ represents the $i-$th initialized pixel token.

Normalization is a crucial preprocessing step in image processing. For general vision tasks, 8-bit quantized RGB images are typically Z-score normalized for each band before being input into  neural networks. For HSI processing tasks, HSIs are usually subjected to Max-Min normalization.
However, both normalization methods are unsuitable for our hyperspectral point object detection framework. HSIs possess higher quantization bits and more bands than conventional RGB images, with each band exhibiting distinct data distributions. Some bands have minor standard deviations, resulting in excessively large values after Z-score normalization, which can adversely affect network convergence.
Furthermore, traditional HTD work primarily validates on datasets containing only a single HSI, whereas our proposed  framework is designed for datasets with a vast number of HSIs. On one hand, the numerical range varies between HSIs, and performing Max-Min normalization on each HSI individually can result in inconsistencies in the spectral numerical  characteristics of HSIs. On the other hand, actual hyperspectral radiance images sometimes contain pixels with extremely high values; after Max-Min normalization, most image regions are concentrated within a very small numerical range, causing quantization loss of spectral information.
Therefore, in \eqrefnew{eq_1}, we normalize all HSIs within the same dataset by dividing by the same constant $ V $. We adopt different settings for $ V $ depending on whether the data is radiance or reflectance. For the SanDiego and Gulfport datasets used in our experiments, which are reflectance data subjected to radiometric correction, the theoretical numerical range is [0,1], and $V$ is set to 1. The numerical range of radiance data is influenced by various factors, including band range, acquisition scenario, unit system, and producer processing methods. For radiance data, we employ a lenient criterion for selecting $V$: ensuring that most values of the normalized HSIs within the same dataset are smaller than 1.
The other two datasets used in our experiments, SPOD and AVON, are based on radiance data collected by different hyperspectral sensor in different data campaigns.
For the SPOD dataset, we set $ V $ to 3000; for the AVON dataset, we set $ V $ to 5000.  This setting can also be automated by setting $ V $ to the value corresponding to the 60\%, 80\%, or other specified percentiles of all values in the dataset, sorted in ascending order.

\subsection{Transformer Encoder}

Deformable DETR and its successors, such as DN-DETR, DINO, and Co-DETR, employ a deformable Transformer encoder on the multi-scale feature maps outputted by the backbone network to further refine features. The deformable Transformer encoder replaces the Transformer attention module in the original DETR with a deformable multi-scale attention module, which focuses solely on a small subset of key sampling points around reference points, serving as a pre-filter to highlight critical elements among all feature map pixels.
For the point objects of interest in this paper, given their extremely small size, they only interact with nearby local background pixels, while distant background pixels have negligible impact. The deformable attention module, as a local sparse sampling operator, is particularly well-suited for feature extraction of point objects. Therefore, we propose the self-excited subpixel-scale attention module (S2A module) based on the deformable attention module for the extraction of  spatial-spectral joint features of point objects.
Furthermore, following the DETR-like detectors, we cascade multiple Transformer encoder layers equipped with the S2A module to achieve a progressive refinement of object and background features from shallow to deep levels. The input to the multi-layer Transformer encoder consists of $\boldsymbol{P}_0$ from \eqrefnew{eq_1} and $\boldsymbol{g}_0$ from \eqrefnew{eq_2}, with the set of all initialized tokens denoted as $\boldsymbol{F}_0$. For the $j$-th Transformer encoder layer, it receives the output $\boldsymbol{F}_{j-1}$ from the previous encoder layer and processes it as follows:
\begin{equation}
	\label{eq_3}
	\tilde{\boldsymbol{F}}_j=\text{LN}\left( \text{Self-S2A}\left( \boldsymbol{F}_{j-1} \right) + \boldsymbol{F}_{j-1} \right),
\end{equation}
\begin{equation}
	\label{eq_4}
	\boldsymbol{F}_j=\text{LN}\left( \text{FFN}\left( \tilde{\boldsymbol{F}}_j \right) +\tilde{\boldsymbol{F}}_j \right),
\end{equation}
where $\text{S2A}\left( \cdot \right)$ represents the S2A module, and $\text{FFN} \left( \cdot \right)$ denotes the feed-forward network, comprising linear layers and ReLU activation layers, also known as the multi-layer perceptron (MLP). During the classic Transformer encoder, the query and value elements of the attention module are identical, hence the attention module in the encoder is termed a self-attention module. Adhering to this convention, we refer to the S2A module in \eqrefnew{eq_3} as the self-S2A module.

The self-S2A module integrates two feature extraction operators based on deformable attention. The first is a subpixel-scale deformable sampling operator performed on the feature map $\boldsymbol{P}_{j-1}$ at the original image resolution. In the traditional DETR framework, deep feature extraction is jointly accomplished by the pre-trained backbone network and the encoder. However, current backbone networks like ResNet are designed for general image understanding tasks, where object shape information is rich and sizes vary, with some objects even occupying the entire image. Consequently, these backbone networks extract feature maps at different downsampling ratios from the input image, while the encoder in current DETR-like detectors is used for further feature refinement.
For the point objects focused on in this paper, the emphasis of current backbone networks on larger spatial features and the downsampling operations are not appropriate. Therefore, we directly perform deformable attention computation on each pixel token at the original image feature resolution, replacing the traditional DETR framework's pipeline of collaborative feature extraction between the backbone network and the encoder. The updated feature map retains the original resolution, thereby avoiding the loss of features for extremely small-sized objects due to downsampling operations. Deformable sampling is based on bilinear interpolation, which aligns well with the linear mixing model assumption of mixed pixels, enabling subpixel-scale deformable sampling on $\boldsymbol{P}_{j-1}$. In addition, we introduce a global token which combined with the subpixel-scale deformable sampling operator, and implements the self-excited operator. Although the self-excited operator is also based on deformable attention, it functions to self-amplify object features rather than feature aggregation.  We term this mechanism as the self-excited mechanism. The subpixel-scale deformable sampling operator and the self-excited operator represent two extremes: one computes deformable attention on the feature map at the original resolution, while the other does so on a single-pixel feature map. Consequently, the multi-scale deformable attention module from deformable DETR can be directly utilized for their implementation.

\begin{figure*}[t]
	\centering
	\includegraphics[width=1\linewidth]{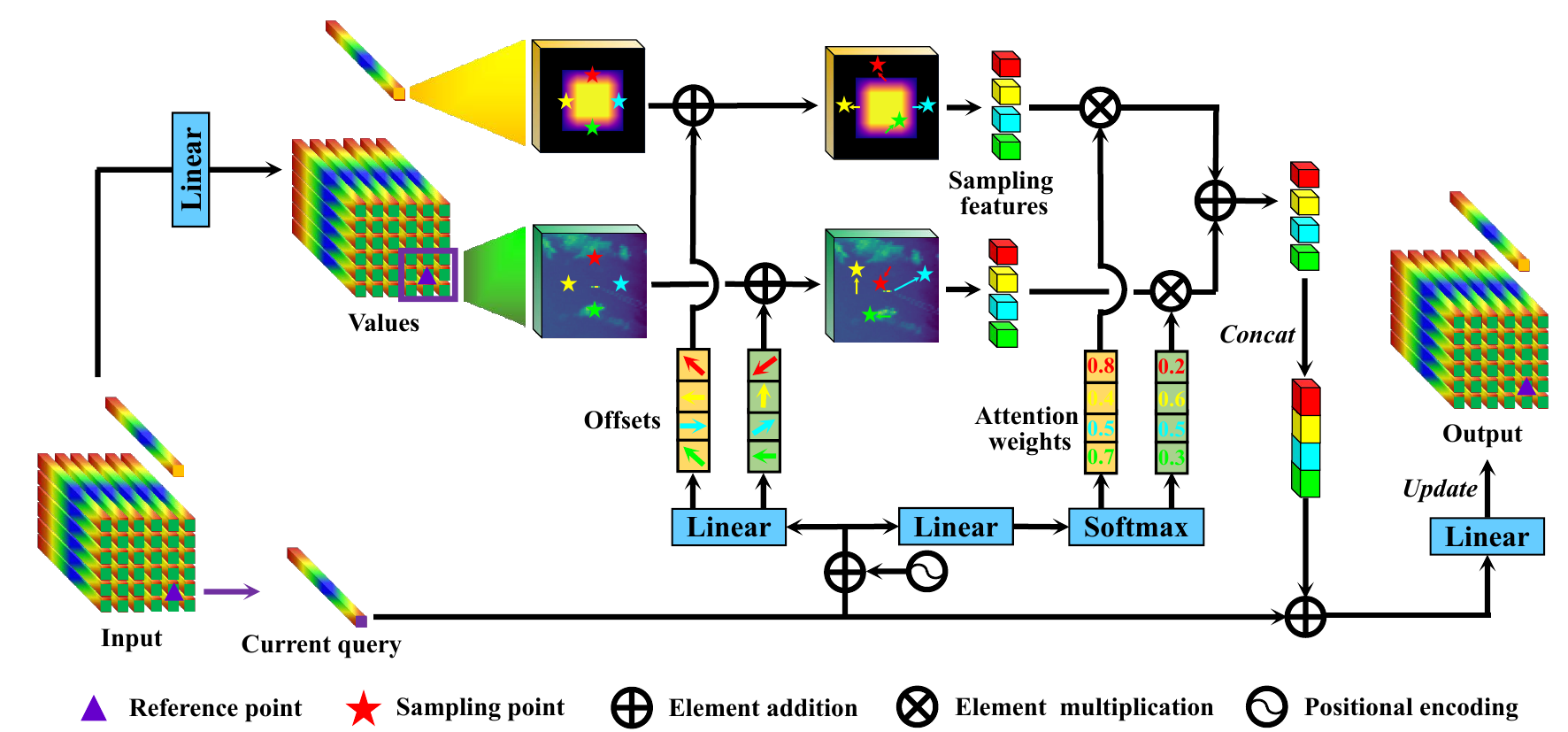}
	\caption{{A schematic diagram of the self-S2A module. For clear visualization, the number of sampling points for each attention head is set to 1. Different colors of the sampling points represent different attention heads.}
		\label{attention}}
\end{figure*}

The schematic diagram of the self-S2A module is shown in Fig.~\ref{attention}. The number of deformable attention heads in the self-S2A module is denoted as $M$, with each attention head having $K$ sampling points, and $M$ divides the token dimension $C$ evenly.
Taking the $i$-th input pixel token $\boldsymbol{p}_{j-1,i}$ as the current query element, its 2D pixel position is set as the current reference point $\boldsymbol{r}$. $\boldsymbol{p}_{j-1,i}$, augmented with its position embedding, is passed through two independent linear layers to obtain $M \times K \times 2$ sampling point position offsets $\Delta \boldsymbol{r}$ relative to the reference point and attention weights $\boldsymbol{A}$, respectively:
\begin{equation}
	\label{eq_5}
	\varDelta \boldsymbol{r}=\text{linear}\left( \boldsymbol{p}_{j-1,i}+\text{pos}\left( \boldsymbol{p}_{j-1,i} \right) \right) +\varDelta \boldsymbol{r}_{\text{int}},
\end{equation}
\begin{equation}
	\label{eq_6}
	\boldsymbol{A}=\text{linear}\left( \boldsymbol{p}_{j-1,i}+\text{pos}\left( \boldsymbol{p}_{j-1,i} \right)\right), 
\end{equation} 
where $\text{pos}\left( \boldsymbol{p}_{j-1,i} \right)$ is the position encoding of the same dimension as $\boldsymbol{r}$, $\Delta \boldsymbol{r}_{\text{int}}$ represents the initialized position offsets, $\Delta \boldsymbol{r} \in \mathbb{R}^{2 \times M \times K \times 2}$, and $\boldsymbol{A} \in \mathbb{R}^{M \times K \times 2}$. $\Delta \boldsymbol{r}$ is split into the pixel token component $\Delta \boldsymbol{r}_P \in \mathbb{R}^{2 \times M \times K}$ and the global token component $\Delta \boldsymbol{r}_G \in \mathbb{R}^{M \times K \times 2}$. Similarly, $\boldsymbol{A}$ is also split into $\boldsymbol{A}_P \in \mathbb{R}^{M \times K}$ and $\boldsymbol{A}_G \in \mathbb{R}^{M \times K}$.
The $\boldsymbol{P}_{j-1}$ and $\boldsymbol{g}_{j-1}$ are fed into the same linear layer to generate the feature maps to be sampled (which can be understood as the value elements in the classical Transformer):
\begin{equation}
	\label{eq_7}
	\hat{\boldsymbol{P}}_j, \hat{\boldsymbol{g}}_j = \text{linear}\left( \boldsymbol{P}_{j-1}, \boldsymbol{g}_{j-1} \right).
\end{equation}
$\hat{\boldsymbol{P}}_j$ and $\hat{\boldsymbol{g}}_j$ are both evenly split along the channel dimension into $M$ sub-feature maps to be sampled, resulting in a channel dimension of $C/M$.
For the $m$-th attention head, the sampling value of the current query element $\boldsymbol{p}_{j-1,i}$ is:
\begin{equation}
	\label{eq_8}
	\begin{aligned}
	\boldsymbol{v}^m = \sum_{k=1}^K \left( A_{P}^{m,k} \cdot \text{bili}\left( \hat{\boldsymbol{P}}_{j}^{m}, \boldsymbol{r} + \Delta \boldsymbol{r}_{P}^{m,k} \right) \right)
	\\ +\sum_{k=1}^K \left( A_{G}^{m,k} \cdot \text{bili}\left( \hat{\boldsymbol{g}}_{j}^{m}, \boldsymbol{r} + \Delta \boldsymbol{r}_{G}^{m,k} \right) \right),
  \end{aligned}
\end{equation}

where $k$ is the index of the sampling point, $m$ is the index of the attention head, $i$ is the index of the encoder layer, $\boldsymbol{v}^m \in \mathbb{R}^{C/M}$, and $\text{bili} (\cdot)$ represents the sampling operation based on bilinear interpolation.
The scalar attention weights $A_{P}^{m,k}$ and $A_{G}^{m,k}$ are normalized by $\sum_{k=1}^K (A_{P}^{m,k} + A_{G}^{m,k})$.
After obtaining the sampled feature values from the $M$ attention heads, the current query element $\boldsymbol{p}_{j-1,i}$ is updated as follows:
\begin{equation}
	\label{eq_9}
	\tilde{\boldsymbol{p}}_{j,i} = \text{linear}\left( \boldsymbol{p}_{j-1,i} + \text{concat}\left( \boldsymbol{v}^1, \boldsymbol{v}^2, \cdots, \boldsymbol{v}^M \right) \right),
\end{equation}
where $\text{concat} (\cdot)$ denotes the concatenation operation along the channel dimension for the vectors.
The same operations from \eqrefnew{eq_5} to \eqrefnew{eq_9} are applied to the remaining pixel tokens and the global token $\boldsymbol{g}_{j-1}$, resulting in the updated $\tilde{\boldsymbol{P}}_j$ and $\tilde{\boldsymbol{g}}_j$. The union of these two sets constitutes the $\tilde{\boldsymbol{F}}_j$ in \eqrefnew{eq_3}.

\begin{figure}[t]
	\centering
	\includegraphics[width=0.6\linewidth]{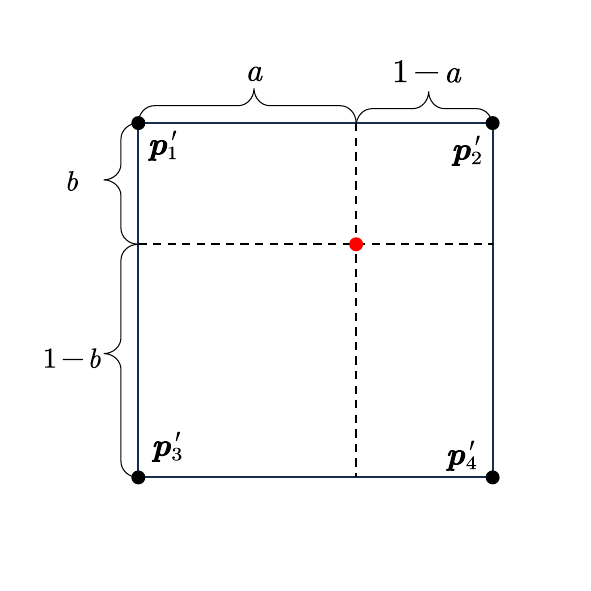}
	\caption{{A schematic diagram of bilinear interpolation.}
		\label{BilinearInterpolation}}
\end{figure}

Compared to ordinary convolution operation or the ordinary Transformer attention module, the deformable sampling operation based on bilinear interpolation extends the feature sampling space from the pixel scale to the subpixel scale.
In \eqrefnew{eq_8}, the position range of the sampling points, $\boldsymbol{r}+\varDelta \boldsymbol{r}_{P}^{m,k}$, includes components from the linear layer output, and therefore is within the real number domain. 
This approach allows the sampling point offsets to become learnable and to be smoothly updated during backpropagation, enabling the adaptive adjustment of the sampling point distribution during training. Additionally, sampling based on bilinear interpolation ensures effective extraction of spectral features at subpixel-scale locations.
As shown in Fig.~\ref{BilinearInterpolation}, $\boldsymbol{p}{1}^{\prime}$, $\boldsymbol{p}{2}^{\prime}$, $\boldsymbol{p}{3}^{\prime}$, and $\boldsymbol{p}{4}^{\prime}$ are the vectors at the integer coordinate points in $\hat{\boldsymbol{P}}{j}^{m}$ closest to $\boldsymbol{r}+\varDelta \boldsymbol{r}{P}^{m,k}$. The horizontal and vertical distances from $\boldsymbol{r}+\varDelta \boldsymbol{r}{P}^{m,k}$ to $\boldsymbol{p}{1}^{\prime}$ are $a$ and $b$, respectively. The sampled value by bilinear interpolation is given by:
\begin{equation}
	\label{eq_10}
	\begin{aligned}
		\text{bili} \left( \hat{\boldsymbol{P}}{i}^{m},\boldsymbol{r}+\varDelta \boldsymbol{r}{P}^{m,k} \right) =a\boldsymbol{p}{1}^{\prime}+\left( b-ab \right) \boldsymbol{p}{2}^{\prime}
		\\+\left( a-ab \right) \boldsymbol{p}{3}^{\prime}+\left( 1-a-b+ab \right) \boldsymbol{p}{4}^{\prime}.
	\end{aligned}
\end{equation}
By denoting the coefficients of $\boldsymbol{p}{1}^{\prime}$, $\boldsymbol{p}{2}^{\prime}$, $\boldsymbol{p}{3}^{\prime}$, and $\boldsymbol{p}{4}^{\prime}$ as $\varepsilon_1$, $\varepsilon_2$, $\varepsilon_3$, and $\varepsilon_4$, respectively, the above equation can be rewritten in the form of the linear mixing model (LMM) \cite{LMM}:
\begin{equation}
	\label{eq_11}
	\text{bili} \left( \hat{\boldsymbol{P}}_{j}^{m},\boldsymbol{r}+\varDelta \boldsymbol{r}_{P}^{m,k} \right) =\sum_{z=1}^4{\varepsilon _z\boldsymbol{p}_{z}^{\prime},\,\,s.t.\begin{cases}
			\sum_{z=1}^4{\varepsilon _z}=1\\
			\varepsilon _z>0\\
	\end{cases}}.
\end{equation}

In remote sensing images, a pixel may contain multiple types of ground objects. The LMM handles this mixed phenomenon by assuming that the reflectance or radiance spectrum of a pixel is a linear combination of the spectra of different ground objects. The abundance of each spectral bases being non-negative and the sum of abundances equal to 1. \eqrefnew{eq_11} implies that the sub-pixel scale feature aggregation operator samples in a spectral feature space that is continuous in positional dimensions. Particularly for the first encoder layer, $\hat{\boldsymbol{P}}_{1}$ is generated by applying two fully connected operations along the feature dimension to the original spectral image, maintaining spatial features consistent with the original image. The sampling points can be viewed as mixed pixels generated by considering the nearest four spectra as spectral bases, with the feature proportion of each spectrum related to the spatial distance.  The feature sampled at any given subpixel-scale position are capable of reflecting the spectral characteristics of neighboring pixels, thereby ensuring effective spectral feature extraction.

\begin{figure}[t]
	\centering
	\includegraphics[width=1\linewidth]{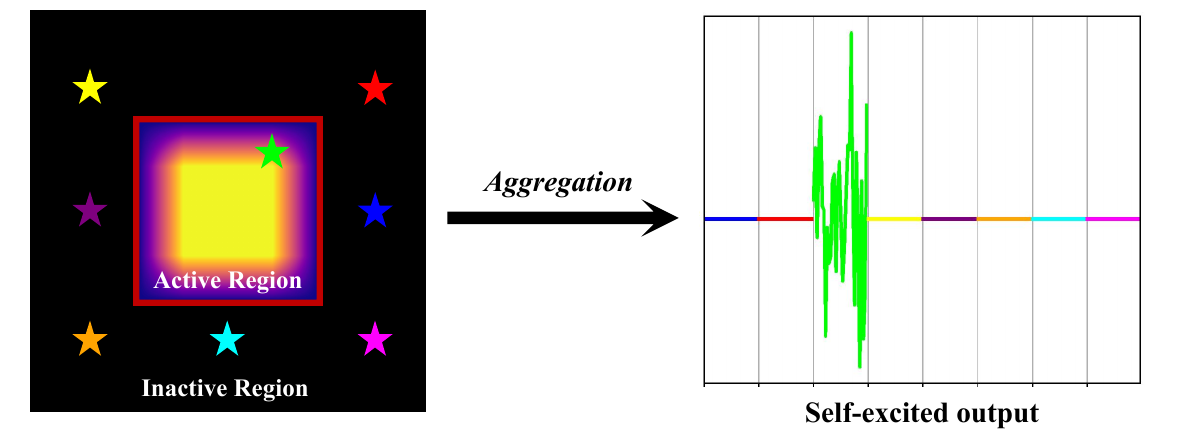}
	\caption{{A schematic diagram of  the self-excited mechanism. For clear visualization, the number of sampling points for each attention head is set to 1. Different colors of the sampling points represent different attention heads.}
		\label{Selfexcited}}
\end{figure}

Fig.~\ref{Selfexcited} illustrates the self-excited mechanism. 
The self-excited operator and the subpixel-scale deformable sampling operator are consistent in structure, but they implement different functions. This difference arises from the variation in the value elements to be sampled. The subpixel-scale deformable sampling operator treats the feature map as the value elements to be sampled, achieving  spatial-spectral joint feature aggregation in the local region where the query element is located. In contrast, the self-excited operator uses the global token as the value element to be sampled, leveraging this single value element to amplify its own semantic information.
In the deformable attention operation based on bilinear interpolation, the global token can also be transformed into a feature space that is continuous in positional dimensions. The global token fills the origin and its eight adjacent integer coordinate points, while other integer coordinates are filled with zero vectors. After bilinear interpolation, a non-zero region of 4$\times$4 in size, centered at the origin, is generated, with all other regions being zero vectors. We refer to the non-zero square region as the active region, and the remaining area of the feature space as the inactive region. In deformable DETR and its successors, the number of attention heads is set to 8 by default.
In the initialization for deformable attention, the initial positions of the $k$-th sampling point for the 8 attention heads are sequentially set to the four vertices and the midpoints of the four sides of a square with a size of $2k\times2k$ centered at the reference point. The reference point is scaled to a square region of $2\times2$ in size centered at the origin. This ensures that the initial positions of the first sampling points for the 8 attention heads are located within the active region, while other sampling points are located in the inactive region.
The final sampling positions are the initial positions plus the spatial offsets generated by the pixel token serving as the query element through linear projection. When all sampling points of an attention head are located in the inactive region, the output of these this attention head is a zero vector. However, if any sampling point of an attention head is in the active region, that attention head is activated, and its output is non-zero. The output vector of the self-excited operator is the concatenation of the outputs from the 8 attention heads along the channel dimension.
As illustrated in the right figure of Fig.~\ref{Selfexcited}, different combinations of activated attention heads generate special sampling outputs. As previously analyzed, the activation of different attention heads is determined by the pixel tokens of the query elements. Consequently, the semantic differences among various types of pixel markers can lead to different combinations of activated attention heads. The sampling output form of the self-excited operator accentuates these semantic differences.
The implementation of the self-excitation mechanism is straightforward and can directly utilize existing multi-scale deformable attention modules.


\subsection{Transformer Decoder}

\begin{figure}[t]
	\centering
	\subfloat[\label{decode1}]{\includegraphics[height=0.7\linewidth]{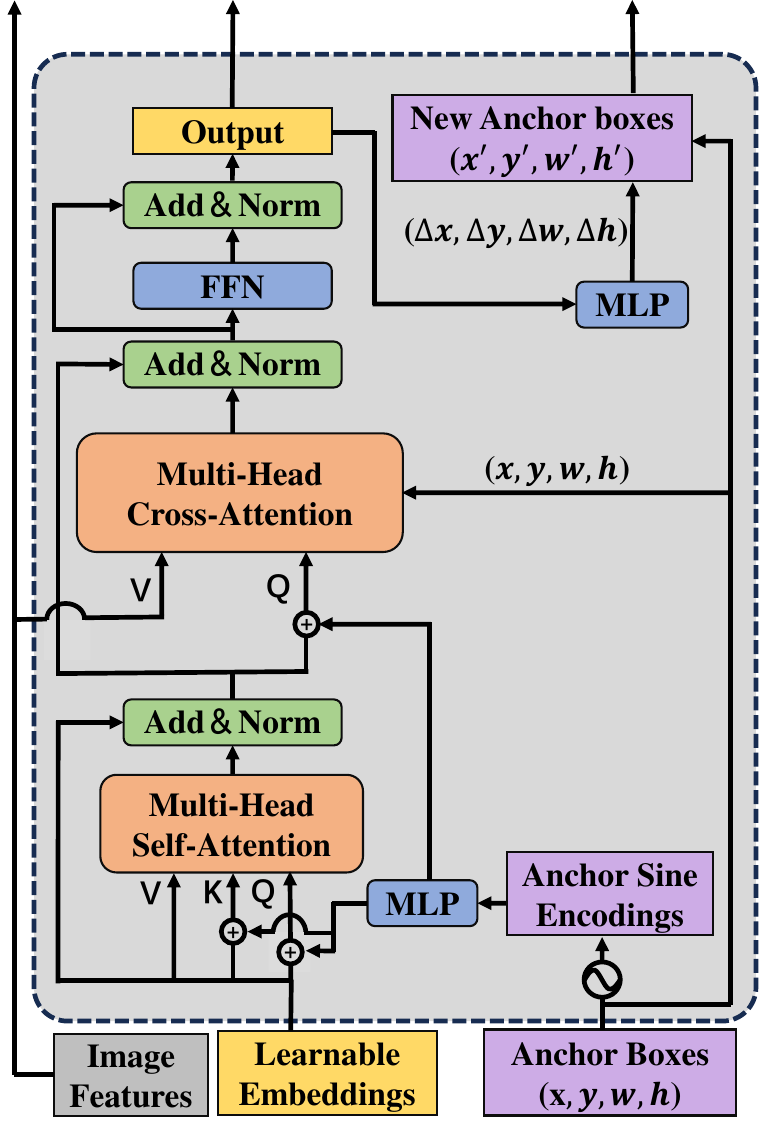}}
	\hfil
	\subfloat[\label{decode2}]{\includegraphics[height=0.7\linewidth]{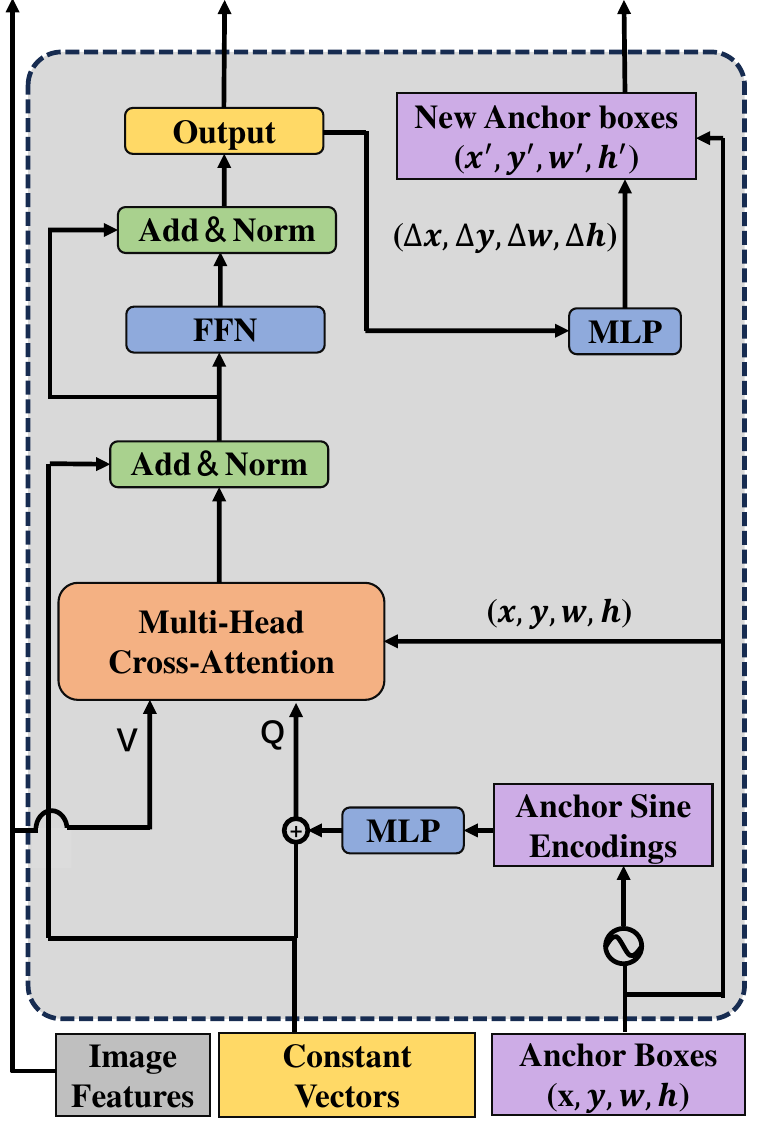}}
	\hfil
	\caption{Comparison of decoder layer structures. (a) DINO. (b) SpecDETR. Only the first layer is illustrated. The structure of the remaining layers is identical to that of the first layer, while the object queries are the output of the preceding layers.}
	\label{fig:decode}
\end{figure}

In the design of the transformer decoder, we optimize based on DINO's decoder, retaining the iterative bbox refinement and two-stage framework from deformable DETR, as well as the contrastive denoising training and a box prediction method named ``look forward twice" from DINO.
DETR-like detectors model object detection as a one-to-one set prediction problem, primarily to eliminate the need for traditional detectors' NMS post-processing operation. General object detection datasets, such as COCO \cite{lin2014microsoft}, contain numerous highly overlapping objects, while detectors prior to DETR can produce multiple redundant predictions for the same object. These two phenomena create a parameter setting dilemma for the IoU threshold of NMS, making the removal of NMS in DETR-like detectors particularly meaningful.
However, our experiments reveal that DETR-like detectors still generate redundant predictions for point objects. Additionally, overlapping point objects, due to their lack of significant shape and texture details, cannot be distinguished like overlapping large objects in the COCO \cite{lin2014microsoft} dataset. This implies that overlapping point objects should be treated as a single object for detection. Therefore, we believe that removing NMS is not suitable for point objects, and we model point object detection as a one-to-many set prediction problem. Based on this, we optimize the decoder structure used by DINO, as shown in Fig.~\ref{fig:decode}, and improve the contrastive  denoising training and label assignment method. This section mainly introduces the forward propagation pipeline of the decoder.

The decoder is composed of multiple cascaded decoder layers, each of which uses object queries as query elements and the feature maps output by the encoder as value elements. These layers iteratively update the object queries and refine the  object anchor boxes. Let the number of encoder layers be $ E $ and the number of decoder layers be $ D $.
Initially, each pixel token output by the last encoder layer generates a candidate anchor box. For the $i$-th pixel token $ \boldsymbol{p}_{E,i} $, the initial center position of the corresponding anchor box is set to the pixel location $(p_{xi}, p_{yi}) $, with both the initial width and height set to $s$. A normalized bbox vector $\boldsymbol{b}_{\text{int},i} $ is constructed as $ \left[ \frac{p_{xi}}{W}, \frac{p_{yi}}{H}, \frac{s}{W}, \frac{s}{H} \right] $.
The pixel token $ \boldsymbol{p}_{E,i} $ is fed into a bbox regression head to refine the initial bbox vector:
\begin{equation}
	\label{eq_12}
	\boldsymbol{b}_{0,i} = \text{Sig}\left( \text{breg}_0\left( \boldsymbol{p}_{E,i} \right) + \text{InSig}\left( \boldsymbol{b}_{\text{int},i} \right) \right)
\end{equation}
where the bbox regression head $ \text{breg}_0(\cdot) $ is a 3-layer MLP network, $\text{Sig}(\cdot)$ represents the Sigmoid activation function that maps the entire real number domain to the (0,1) interval, and $ \text{InSig}(\cdot) $ represents the inverse of the Sigmoid activation function, mapping the (0,1) interval to the entire real number domain.
Subsequently, $ \boldsymbol{p}_{E,i} $ is fed into a classification head:
\begin{equation}
	\label{eq_13}
	\boldsymbol{c}_{0,i} = \text{Sig}\left(\text{cls}_0\left(\boldsymbol{p}_{E,i}\right)\right)
\end{equation}
where the classification head $ \text{cls}_0(\cdot) $ is a linear layer, and $ \boldsymbol{c}_{0,i} \in \mathbb{R}^{N_{\text{cat}}} $ represents the category predictions, also known as object confidence scores, with $ N_{\text{cat}} $ being the number of object categories. After the Sigmoid activation function, the values of $ \boldsymbol{c}_{0,i} $ range between (0,1), where values closer to 1 indicate a higher probability that the prediction belongs to the corresponding category.
From all $ H \times W $ candidate anchor boxes, the top $ Q_{\text{match}} $ boxes with the highest confidences are selected as the initial bboxes corresponding to the object queries of the decoder. During training, the denoising training branch additionally provides $ Q_{\text{DN}} $ noised GT boxes, the details of which will be introduced in the following text. Let the number of object queries for the decoder be $ Q $; during training, $ Q = Q_{\text{match}} + Q_{\text{DN}} $, and during inference, $ Q = Q_{\text{match}} $.
DINO uses the index embedding to generate initial object queries, endowing them with learnability. Under one-to-one set prediction modeling, learnable and distinguishable object queries can enhance detection performance. However, in the context of point object detection with one-to-many set prediction modeling, we believe that distinguishable object queries is unnecessary. On one hand, we remove the self-attention computation between object queries, so they do not interfere with each other during forward pass. On the other hand, DINO aims for a single prediction per object, while our SpecDETR does not impose this constraint. Therefore, we initialize all object queries as all one vectors of dimension $C$.

Let $d $ index the decoder layers and $i $ index the object queries. In the $d $-th decoder layer, we take the object queries fed from the previous layer as the query elements and the output $\boldsymbol{F}_{E} $ of the last encoder layer as the value elements to compute the cross-attention between the object queries and $\boldsymbol{F}_{E} $, thereby updating the object queries as follows:
\begin{equation}
	\label{eq_14}
	\tilde{\boldsymbol{q}}_{d,i}=\text{LN}\left( \text{Cross-S2A}\left( \boldsymbol{q}_{d-1,i}, \boldsymbol{F}_E \right) +\boldsymbol{q}_{d-1,i} \right) ,
\end{equation}
\begin{equation}
	\label{eq_15}
	\boldsymbol{q}_{d,i}=\text{LN}\left( \text{MLP}\left( \tilde{\boldsymbol{q}}_{d,i} \right) +\tilde{\boldsymbol{q}}_{d,i} \right) ,
\end{equation}
where $\boldsymbol{q}_{d,i} \in \mathbb{R}^C $, $d \in \left\{ 1,2,\cdots ,D \right\} $, and $i \in \left\{ 1,2,\cdots ,Q \right\} $.
Compared to \eqrefnew{eq_3}, the query elements and value elements  of \eqrefnew{eq_14} are not the same, hence the S2A module in \eqrefnew{eq_14} is termed a cross-S2A module. The cross-S2A module is used to capture the object category and boundary information near the anchor box $\boldsymbol{b}_{d-1,i} $ corresponding to $\tilde{\boldsymbol{q}}_{d-1,i} $. Therefore, the reference point of the cross-S2A module is set to the center of the anchor box, and the initial positions of the sampling points are scaled within the anchor box, with the outermost sampling points placed on the box boundary. Other details are consistent with the self-S2A module.

The vector $\boldsymbol{q}_{d,i}$ is fed into a bbox regression head to update the bbox vector $\boldsymbol{b}_{d-1,i}$ as follows:
\begin{equation}
	\label{eq_16}
	\hat{\boldsymbol{b}}_{d,i}=\text{Sig}\left( \text{breg}_d\left( \boldsymbol{q}_{d,i} \right) +\text{InSig}\left( \boldsymbol{b}_{d-1,i} \right) \right),
\end{equation}
\begin{equation}
	\label{eq_17}
	\boldsymbol{b}_{d,i}=\text{detach}\left( \hat{\boldsymbol{b}}_{d,i} \right),
\end{equation}
\begin{equation}
	\label{eq_18}
	\boldsymbol{b}_{d,i}^{\text{pred}}=\text{Sig}\left( \text{breg}_d\left( \boldsymbol{q}_{d,i} \right) +\text{InSig}\left( \hat{\boldsymbol{b}}_{d-1,i} \right) \right) ,
\end{equation}
where $\text{breg}_d(\cdot)$ is a bbox regression head with the same structure as $\text{breg}_0(\cdot)$, and $\text{detach}(\cdot)$ represents the gradient detachment operation which can interrupt the backpropagation. Both $\boldsymbol{b}_{d,i}^{\text{pred}}$ and $\boldsymbol{b}_{d,i}$ are the bbox prediction for the $i$-th object query at the $d$-th decoder layer, with $\boldsymbol{b}_{d,i}$ used for bbox updates in the $d+1$-th decoder layer and $\boldsymbol{b}_{d,i}^{\text{pred}}$ used to compute the loss function for the $d$-th decoder layer. \eqrefnew{eq_16}-\eqrefnew{eq_18} collectively implement DINO's look forward twice method, which ensures that the parameter updates of $\text{breg}_d(\cdot)$ are influenced by both the bbox prediction error at the $d$-th decoder layer and the bbox prediction error at the $d+1$-th decoder layer.
$\boldsymbol{q}_{d,i}$ is then fed into a classification head to obtain the class prediction results corresponding to $\boldsymbol{b}_{d,i}^{\text{pred}}$:
\begin{equation}
	\label{eq_19}
	\boldsymbol{c}_{d,i}=\text{Sig}\left(\text{cls}_d\left(\boldsymbol{q}_{d,i}\right)\right)
\end{equation}
where $\text{cls}_d(\cdot)$ is a classification head with the same structure as $\text{cls}_0(\cdot)$. The confidence scores for each class are combined with $\boldsymbol{b}_{d,i}^{\text{pred}}$, resulting in $Q \times N_C$ predictions. The refined bboxes and class confidence scores output from the last decoder layer serve as the prediction output of SpecDETR, while the prediction results from other layers are used only for loss function computation during the training phase. The output $ Q \times N_C$ predictions can be reduced for redundancy through post-processing techniques such as NMS and confidence score filtering.

In the classical Transformer network, the decoder contains a self-attention module among query elements as well as a cross-attention module between query elements and encoder output elements. As shown in Fig.~\ref{fig:decode}(a), the decoder in current DETR-like detectors still follows this pipeline to facilitate information interaction between object queries, thereby enabling low-redundancy predictions without the need for NMS post-processing. However, as discussed at the beginning of this section, point object detection is more suitably modeled as a one-to-many rather than a one-to-one set prediction problem, making the information interaction between object queries unnecessary. Therefore, SpecDETR removes the self-attention module between object queries, simplifying the processing pipeline and reducing computational complexity.

\subsection{Hybrid Label Assigner}

Group  DETR, DN-DETR, and DINO have demonstrated that the one-to-one label assigner employed by the original DETR leads to a lack of supervisory signals during training, thereby resulting in slower convergence.
As discussed above, removing NMS is not suitable for point objects, and point object detection should be modeled as a one-to-many set prediction problem. Therefore, we replace the Hungarian matching in current DETR-like detectors with a one-to-many label assigner in the matching branch.
Traditional object detection networks often use some IoU-based label matching \cite{girshick2015fast} as the one-to-many label assigners. However, due to the extremely small size of point objects, even the deviation of a few pixels  can significantly reduce the IoU between the predicted bbox and the GT box. Therefore, we propose a hybrid one-to-many label assigner that combines the Hungarian matching in DETR \cite{carion2020end}  and Max IoU matching \cite{girshick2015fast}. Initially, we use the Hungarian matching to forcibly assign a predicted bbox to each GT sample, a step we refer to as forced matching. Subsequently, we allocate an additional variable number of high-quality predicted bboxes to each GT sample based on IoU between the predicted boxes and GT boxes, a step termed dynamic matching.

\noindent \textbf{Forced Matching.} 
During training, when the bbox regression head is inaccurate or encounters challenging samples, the predicted bbox can deviate from the GT box. Given that point objects are only a few or even one pixel in size, pixel-level position deviations can result in extremely low IoU or even an IoU of zero. The Hungarian matching in DETR  achieves bipartite matching between GT boxes and predicted bboxes based on a combined loss of the bbox GIoU loss \cite{GIOULoss}, bbox L1 loss, and classification loss. Therefore, we first employ Hungarian matching to forcibly assign a positive sample to each GT box, ensuring the network can learn the hard samples.

\noindent \textbf{Dynamic Matching.} 
Following forced matching, we utilize the Max IoU assigner \cite{girshick2015fast} to further allocate positive samples among the remaining anchor boxes. In addition to the IoU threshold, we introduce a number threshold $T$. If a GT box is assigned to more than $T$ anchor boxes, we retain only the top $T$ anchor boxes with the highest IoU as positive samples.
We seek to ensure that predicted bboxes which closely match the GT boxes have higher confidence scores from the classification head and can be retained after NMS. 
Dynamic matching leverages both the IoU threshold and the number threshold to provide high-quality positive samples for the classification head. The introduction of a matching number threshold allows the actual matching IoU threshold to dynamically increase as the network converges, thereby continuously improving the quality of positive samples.

\begin{figure}[t]
	\centering
	\includegraphics[width=1\linewidth]{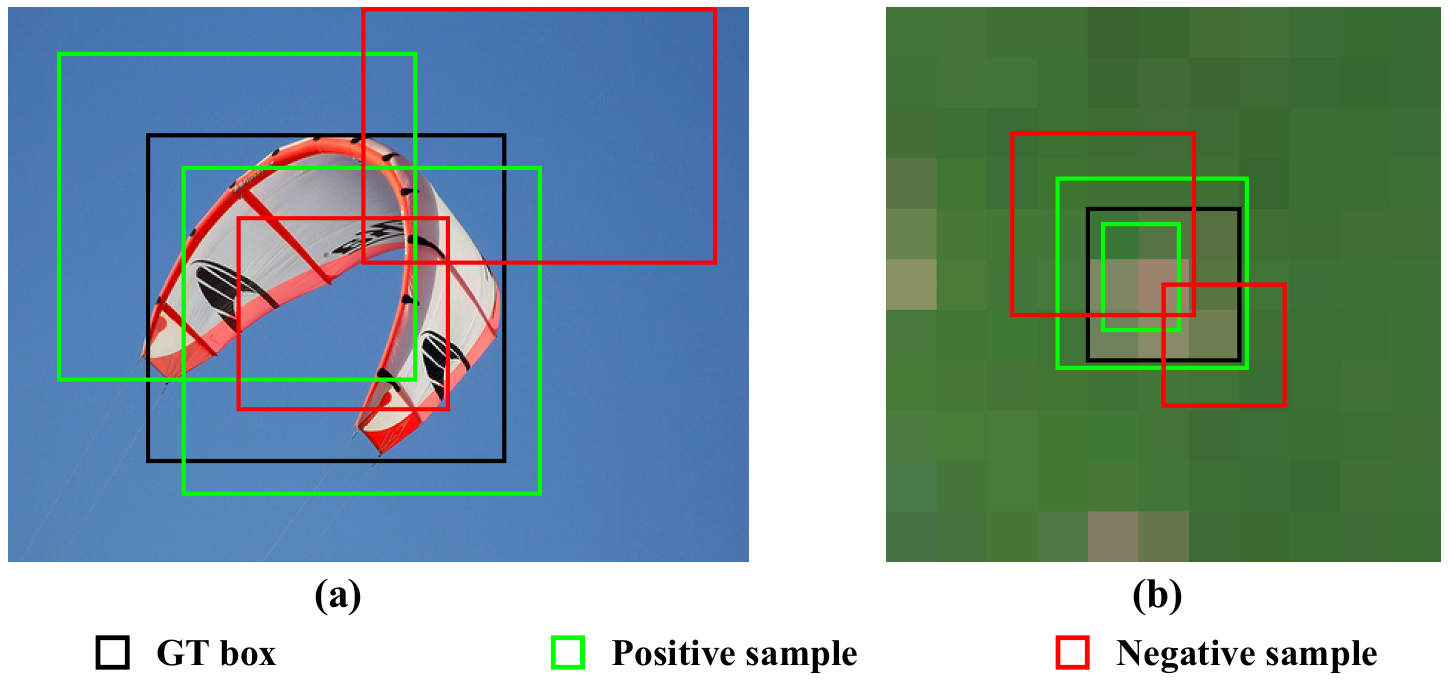}
	\caption{{Illustration of positive and negative positional queries of CDN and CCDN training. (a) CDN. (b) CCDN.}
		\label{ccdn}}
\end{figure}

\subsection{Center-Shifting Contrastive DeNoising Training}

DN-DETR introduces a DeNoising (DN) training branch in the decoder to stabilize training and accelerate convergence.
DINO improved DN training to Contrastive DeNoising (CDN) training, incorporating negative queries to reject useless anchors. CDN generates the anchor boxes of positive and negative queries by adding two levels of noise (position offsets) to the corners of the GT boxes. However, CDN is designed for one-to-one matching, where the anchor box range of negative queries includes the positive queries, as shown in Fig.~\ref{ccdn}(a). However, negative samples with high overlap with GT boxes can disrupt the classification heads of our SpecDETR. Therefore, we adopt the query generation method of DN \cite{li2022dn}, consisting of center shifting and box scaling steps controlled by hyperparameters $\tau_1$ and $\tau_2$, with $\tau_1 > 0$ and $\tau_2 > 0$. 
Our denoising approach adds two levels of shifting to the centers of the GT boxes to generate the anchor boxes of positive and negative  queries, termed as Center-Shifting Contrastive DeNoising (CCDN).

\noindent \textbf{Center Shifting.} Given a GT box $\left(x,y,w,h\right)$, random noises $\left( n_x, n_y \right) $ are added to the center$\left(x,y\right)$. For positive queries, the noise satisfies  $-0.5\tau _1w<n_x<0.5\tau _1w$ and $-0.5\tau _1h<n_y<0.5\tau _1h$. For negative samples,  $ n_x$ is randomly sampled within the range  $\left[ -\tau _1w,-0.5\tau _1w \right] \cup \left[ 0.5\tau _1w,\tau _1w \right] $, and $ n_y$ is randomly sampled within the range  $\left[ -\tau _1h,-0.5\tau _1h \right] \cup \left[ 0.5\tau _1h,\tau _1h \right]$.

\noindent \textbf{Box Scaling.}  Regardless of positive or negative queries, the width and height are randomly sampled within the ranges $[\min(0$, $ w-\tau_2w )$, $ w+\tau_2w ]$  and 
$ [\min(0$, $ h-\tau_2h )$, $h+\tau_2h ]$, respectively.

\section{Datasets and Evaluation Metrics}
\label{sec:datasets}

\begin{figure}[t]
	\centering
	\includegraphics[width=1\linewidth]{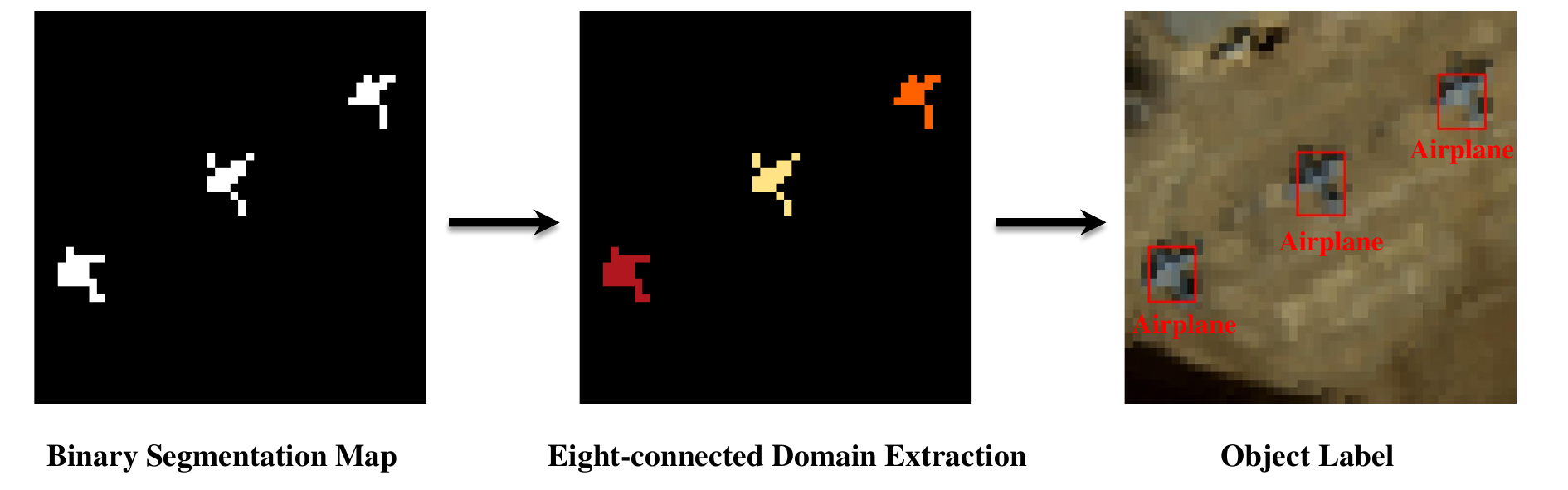}
	\caption{The schematic diagram for generating object labels on binary segmentation maps.}
	\label{fig:object_label}
\end{figure}	

\begin{figure}[t]
	\centering
	\subfloat[]{\includegraphics[height=0.6\linewidth]{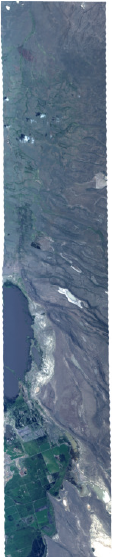}}
	\hfil
	\subfloat[]{\includegraphics[height=0.6\linewidth]{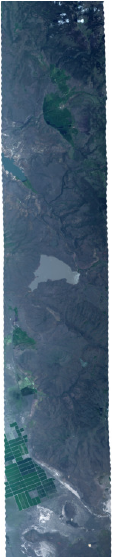}}
	\hfil
	\subfloat[]{\includegraphics[height=0.6\linewidth]{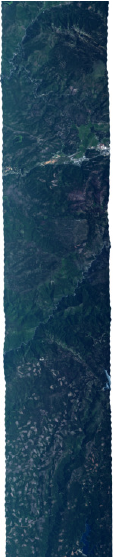}}
	\hfil
	\subfloat[]{\includegraphics[height=0.6\linewidth]{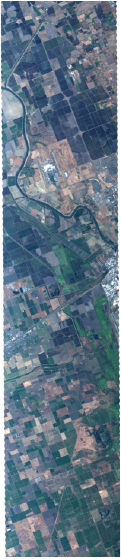}}
	\hfil
	\subfloat[]{\includegraphics[height=0.6\linewidth]{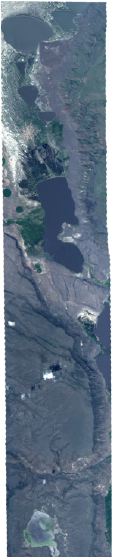}}
	\hfil
	\subfloat[]{\includegraphics[height=0.6\linewidth]{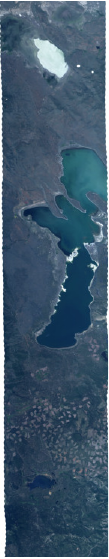}}
	\hfil
	\subfloat[]{\includegraphics[height=0.6\linewidth]{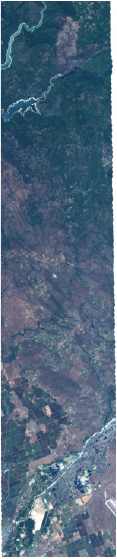}}
	\hfil
	\subfloat[]{\includegraphics[height=0.6\linewidth]{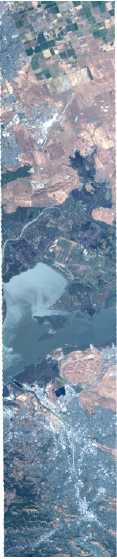}}
	\caption{Partial regions of AVIRIS flight lines used to construct the SPOD dataset. (a)-(d) The flight line f180601t01p00r06 for training data. (e)-(h) The flight line f180601t01p00r10 for test data.}
	\label{fig:flightline}
\end{figure}	

\begin{figure}[t]
	\centering
	\includegraphics[width=1\linewidth]{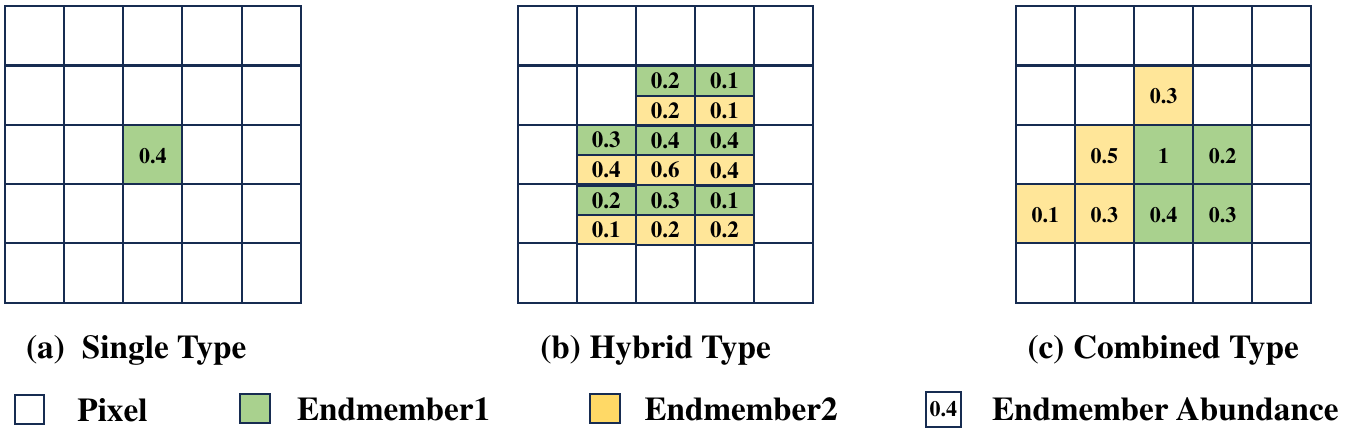}
	\caption{Diagrams of the three types of object templates on the SPOD dataset.}
	\label{fig:object_type}
\end{figure}	

\begin{figure*}[t]
	\centering
	\subfloat[]{\includegraphics[width=0.33\linewidth]{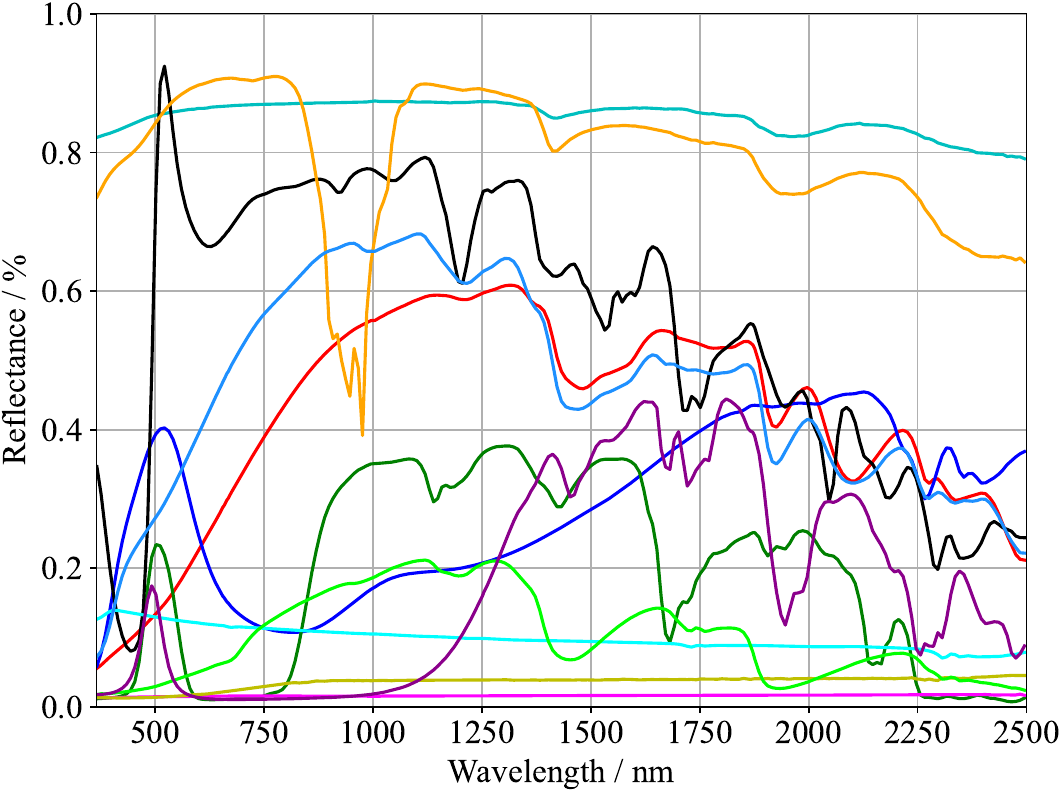}}
	\hfil
	\subfloat[]{\includegraphics[width=0.33\linewidth]{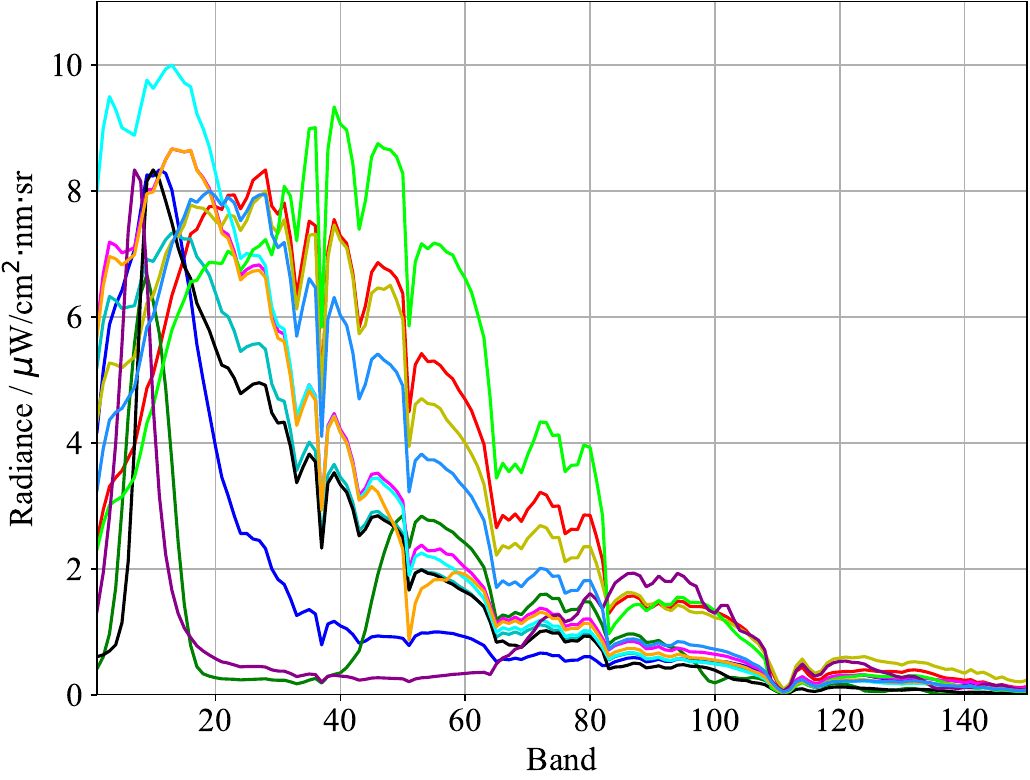}}
	\hfil
	\subfloat[]{\includegraphics[width=0.33\linewidth]{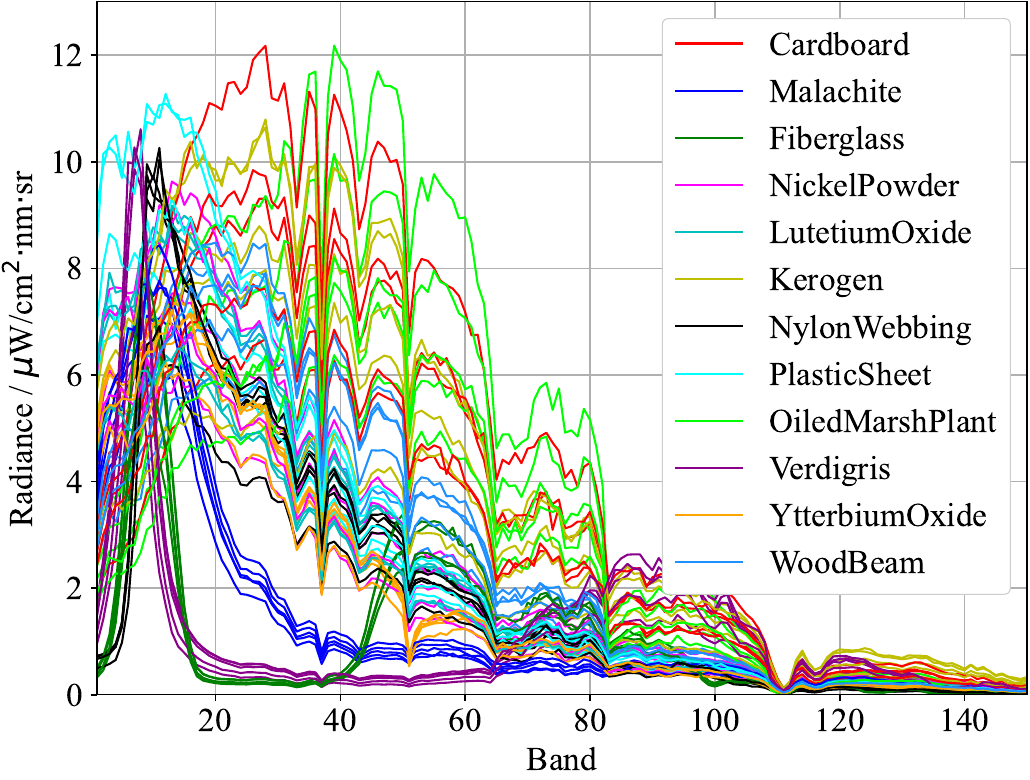}}
	\hfil
	\caption{Spectral curves of different endmembers. (a) Reflectance curves.
		(b) Radiance baselines.
		(c) Radiance curves after adding noise.}
	\label{fig:spotspec}
\end{figure*}	

\begin{table}[t]
	\centering
	\caption{The object simulation setup on the SPOD dataset.}	
	\renewcommand\arraystretch{1} 
	\scriptsize  
	\setlength{\tabcolsep}{1mm}{   
		\begin{tabular}{ccccccc}
			\hline
			\multirow{2}{*}{Name} & \multirow{2}{*}{Type} & Endmember & Object  & Maximum & Training  & Test   \\
			&       &components & pixels  &  abundance & samples & samples \\
			\hline
			\rowcolor[rgb]{ .886,  .937,  .855} C1    & Single & Cardboard  & 1     & 0.05-0.2 & 122   & 686 \\
			\hline
			\rowcolor[rgb]{ .886,  .937,  .855} C2    & Single & Malachite  & 1     & 0.05-0.2 & 118   & 674 \\
			\hline
			\rowcolor[rgb]{ .867,  .922,  .969} C3    & Single & Fiberglass  & 1-2   & 0.2-1 & 136   & 633 \\
			\hline
			\rowcolor[rgb]{ .867,  .922,  .969} C4    & Single & NickelPowder & 3-5   & 0.2-1 & 120   & 664 \\
			\hline
			\rowcolor[rgb]{ .867,  .922,  .969} C5    & Single & LutetiumOxide & 3-5   & 0.2-1 & 108   & 688 \\
			\hline
			\rowcolor[rgb]{ .988,  .894,  .839}  & & Kerogen,  &  & &  &  \\
			
			\rowcolor[rgb]{ .988,  .894,  .839} \multirow{-2}{*}{C6}      &   \multirow{-2}{*}{Hybrid}     & NylonWebbing & \multirow{-2}{*}{6-10}      &    \multirow{-2}{*}{0.2-1}    &  \multirow{-2}{*}{122}     &\multirow{-2}{*}{669}  \\
			\hline
			\rowcolor[rgb]{ .988,  .894,  .839} &  & PlasticSheet, & & &&  \\
			\rowcolor[rgb]{ .988,  .894,  .839} \multirow{-2}{*}{C7}       &  \multirow{-2}{*}{Combined}     & OiledMarshPlant  & \multirow{-2}{*}{6-10}       &    \multirow{-2}{*}{0.2-1}    &    \multirow{-2}{*}{106}     &\multirow{-2}{*}{686}  \\
			\hline
			\rowcolor[rgb]{ .988,  .894,  .839}  &  & Verdigris, &  &  &&  \\
			\rowcolor[rgb]{ .988,  .894,  .839}      &       & YtterbiumOxide, &       &       &       &  \\
			\rowcolor[rgb]{ .988,  .894,  .839}  \multirow{-3}{*}{C8}     & \multirow{-3}{*}{ Combined }    & WoodBeam &\multirow{-3}{*}{11-16}       & \multirow{-3}{*}{0.2-1}      &   \multirow{-3}{*}{123}      & \multirow{-3}{*}{629} \\
			\hline
		\end{tabular}%
	}
	\label{tab:object_type}%
\end{table}%

\begin{figure}[t]
	\centering
	\subfloat[C1]{\includegraphics[width=0.24\linewidth]{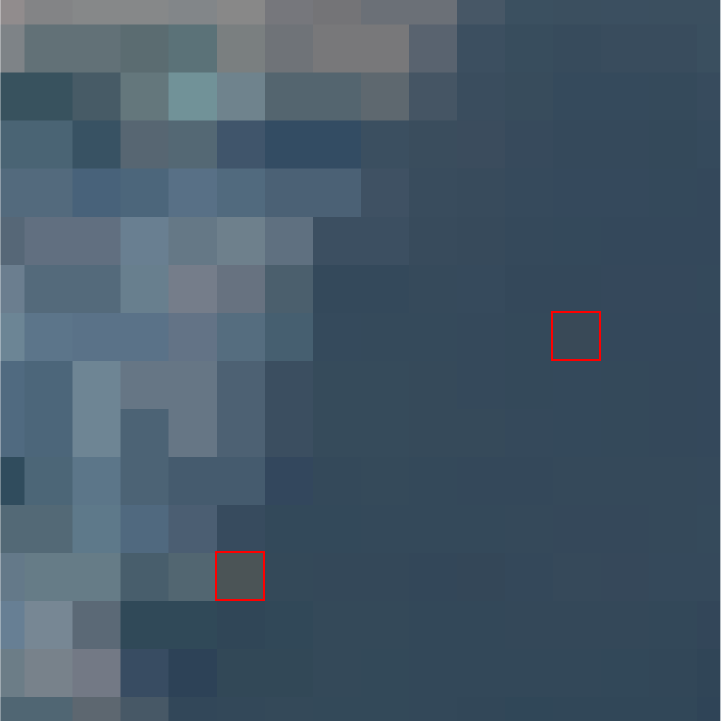}}
	\hfil
	\subfloat[C2]{\includegraphics[width=0.24\linewidth]{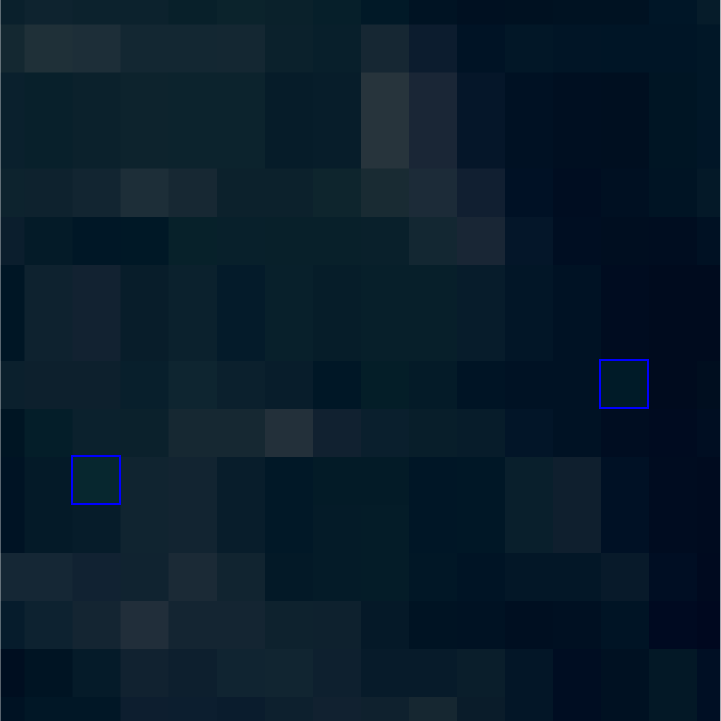}}
	\hfil
	\subfloat[C3]{\includegraphics[width=0.24\linewidth]{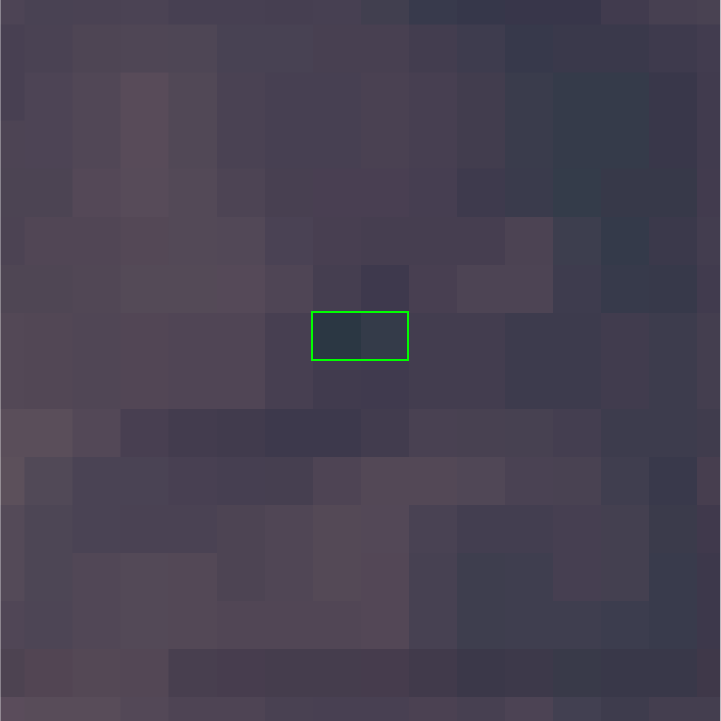}}
	\hfil
	\subfloat[C4]{\includegraphics[width=0.24\linewidth]{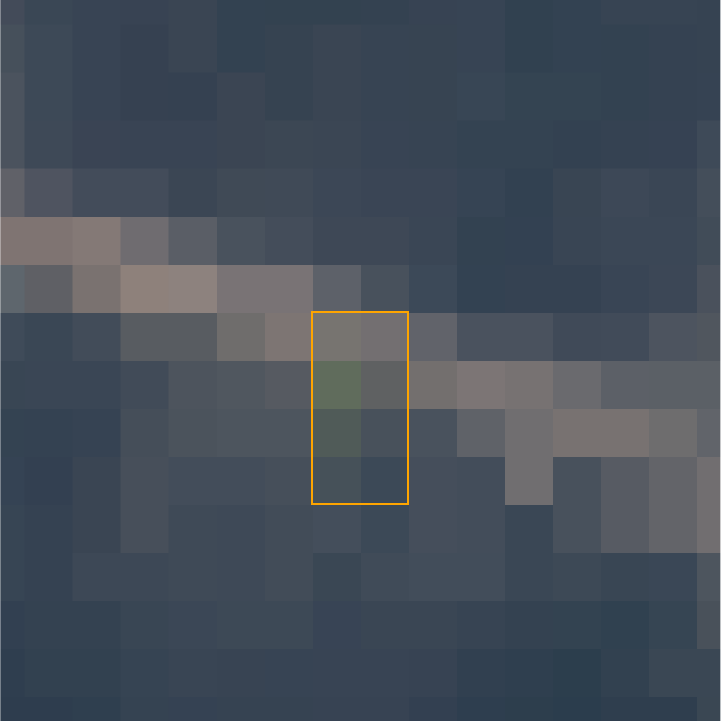}}
	\hfil
	\\
	\subfloat[C5]{\includegraphics[width=0.24\linewidth]{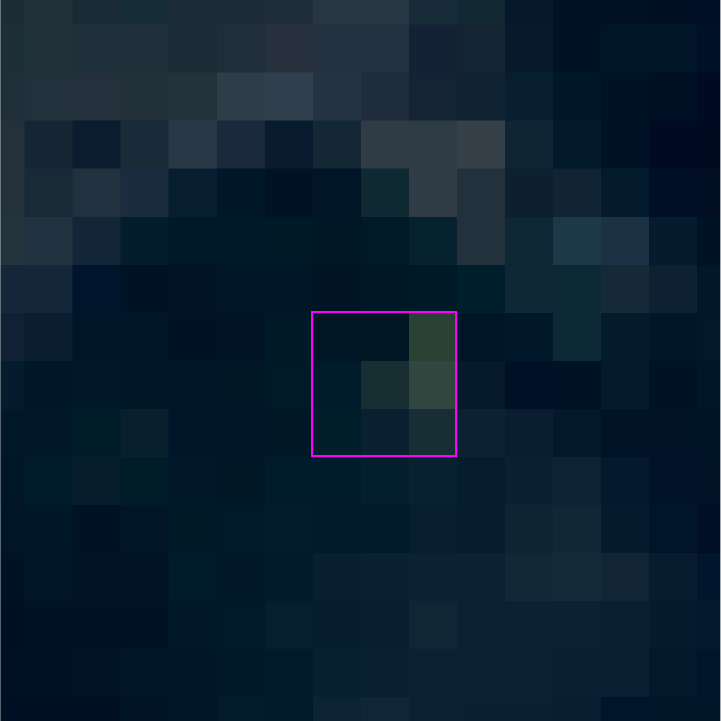}}
	\hfil
	\subfloat[C6]{\includegraphics[width=0.24\linewidth]{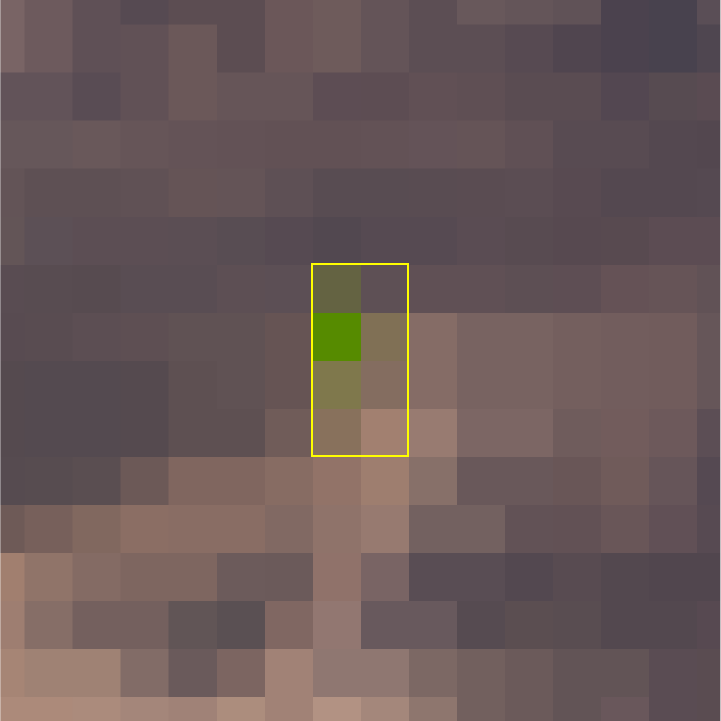}}
	\hfil
	\subfloat[C7]{\includegraphics[width=0.24\linewidth]{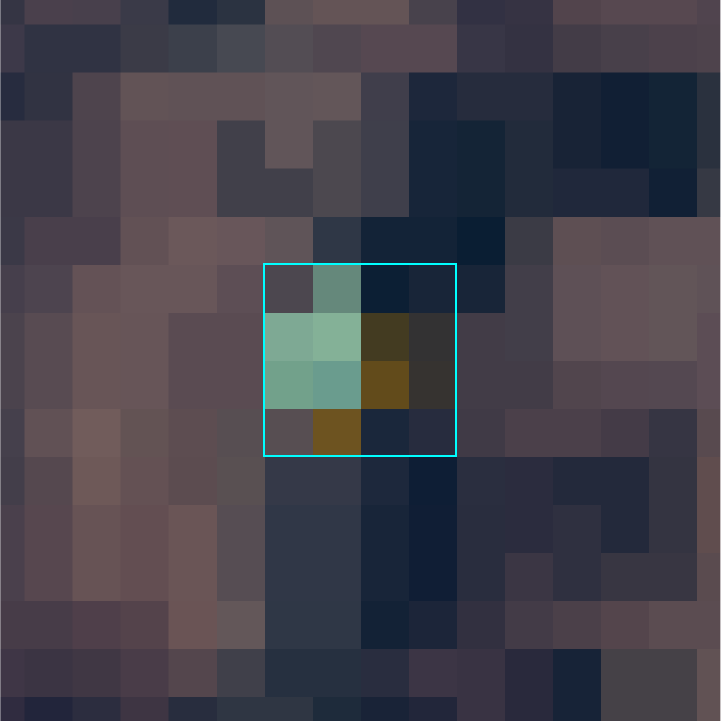}}
	\hfil
	\subfloat[C8]{\includegraphics[width=0.24\linewidth]{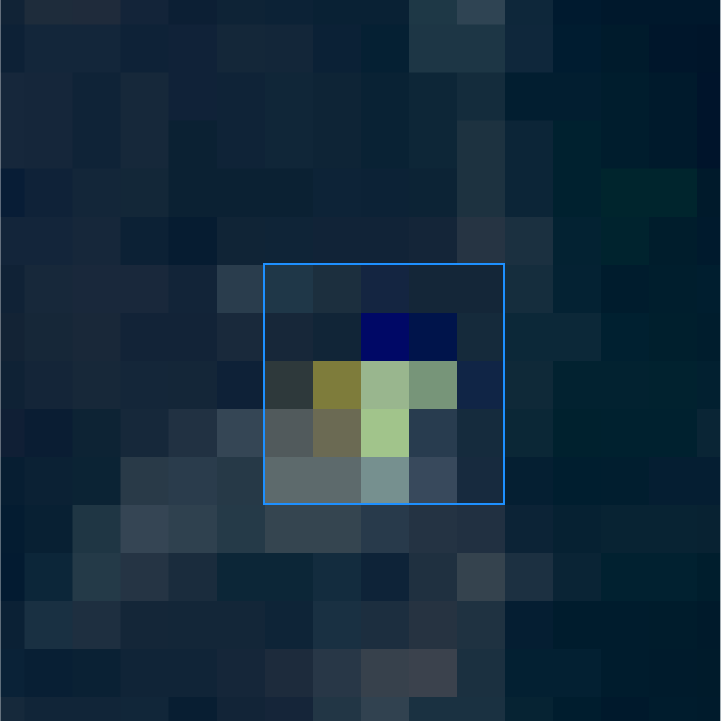}}
	\hfil
	\caption{Examples of objects on the SPOD dataset on pseudo-color images.}
	\label{fig:spotobject}
\end{figure}	

This section will introduce the hyperspectral multi-class point object detection dataset that we develop, as well as three public hyperspectral target detection datasets the Avon dataset, SanDiego dataset and Gulfport dataset. In addition, we will also discuss the construction of  GT labels and evaluation criteria for the point object detection task.

\subsection{Ground-truth Label Construction}
\label{subsec:datasets_gt}

Object detection datasets are typically annotated manually, where professional annotators view the images and draw the bounding boxes of the objects, while also marking the object categories. 
However, for point objects, many constituent pixels are mixed pixels, making manual annotation challenging to ensure the accuracy of mixed pixel labeling. Therefore, some real HTD datasets, such as the Avon and Gulfport datasets, provide rough GT positions obtained from GPS, which have significant errors.
To accurately evaluate the performance of different methods on hyperspectral multi-class point object detection, we have constructed two datasets with precise labels: SPOD and Avon. These datasets provide 8 classes and 2 classes of point objects, respectively. The SPOD dataset is simulation-based, and precise GT labels can be directly obtained from the object simulation configuration.
In contrast, the Avon dataset is a real-world dataset comprising 24 multi-pixel objects. Each object contains both pure object pixels and  object pixels mixed with background. For the Avon dataset, we first perform target detection, followed by manual selection of object pixels for annotation.
For each class of object in the Avon dataset, we selected a pure object spectrum and a mixed object spectrum as a prior spectrum to perform hyperspectral target detection, respectively. We then manually selected pure object pixels and mixed object pixels from the two HTD score maps, respectively. The pixels selected from both score maps were deemed valid object pixels, and they were utilized to create the binary pixel mask label map for the Avon dataset.
Traditional HTD methods typically use this binary pixel mask as the GT labels.
However, for the point object detection task, this binary segmentation map must be further transformed into bboxes and category labels, as illustrated in Fig.~\ref{fig:object_label}.
On the binary segmentation map of the specified category, we utilized eight-connected domain analysis to cluster the object pixels, considering each eight-connected domain as an independent object entity. The maximum enclosing rectangle of the eight-connected domain is used as the bbox, and each bbox is assigned the corresponding object category.


The prior information required for the hyperspectral point object detection is fundamentally different from that of traditional HTD. Traditional HTD require only a few target spectra (often just one), whereas the hyperspectral point object detection necessitates numerous training images with object labels. Classic HTD datasets like the SanDiego dataset do not come with a prepared training image set.
To enable our SpecDETR to be evaluated on the Avon, SanDiego, and Gulfport datasets, we employed a data simulation method from the SPOD dataset. Using target prior spectra and background images without objects, we generated training image sets for these three public HTD datasets. Since the objects in these three public HTD datasets are all single-material, we used only one prior spectrum per object category to generate the simulated training samples.
Since the training sets for these three datasets are simulated, the GT labels are precise, ensuring effective training of the object detection network.

\subsection{The SPOD Dataset}

Due to the spectral mixing phenomenon, acquiring a large number of high-quality manually annotated point objects is extremely challenging. Therefore, we develop a simulated high-spectral multi-class point object detection benchmark based on observed real-world patterns to evaluate the accuracy performance of various methods in point object detection.
As shown in Fig.~\ref{fig:flightline}, we selected two flight lines\footnote{https://aviris.jpl.nasa.gov/} captured by the AVIRIS sensor over California and Nevada, covering 21,000 $km^2$ of the Earth's surface with significant geographic variation. One flight line provides 150 128$\times$128 training background images, while the other provides 500 128$\times$128 testing background images. After removing the atmospheric absorption bands, we retained 150 bands with the band IDs [8-57, 66-79, 86-103, 123-145, 146-149, 173-214].

To ensure the realism of the simulation, we develop a simulation method based on the spatial characteristics of point objects and their spectral variation properties. This method is divided into three parts: object template generation, endmember spectrum generation, and simulated image synthesis.  
In the object template generation phase, we incorporate the point spread morphological characteristic of point objects and the co-existence of multiple endmember spectra. Additionally, since spectra of the same material can be affected by various factors such as illumination, we analyze the local spectral variation property and wide-area spectral variation property of water areas in AVIRIS images. These properties are used to generate simulated endmember spectra.

\noindent \textbf{Object Template Generation}:
For each simulated object, we randomly select the pixel number $N_P$ from the predefined range which is shown in Table~\ref{tab:object_type}. Then, a clustered random combination of $N_P$  pixels is generated to serve as a blank object template. Next, endmembers and abundances are assigned to each object pixel.  
Object pixels not adjacent to background pixels are considered pure pixels, with their object spectral abundance set to 1. Object pixels adjacent to background pixels are treated as mixed pixels, with their object spectral abundance randomly generated from the range (0.01, 1). The abundance of mixed pixels is assigned based on their distance to the object center, with closer pixels receiving higher abundance values. Additionally, if all object pixels are mixed pixels, the abundance of the center pixel is randomly selected from the maximum abundance range, as specified in Table~\ref{tab:object_type}.  
To account for the fact that real object spectra may contain multiple endmembers, we define three types of object templates: single-type, hybrid-type, and combined-type, as illustrated in Fig.~\ref{fig:object_type}.  
The single-type object template contains only one object endmember, so the object spectral abundance is equivalent to the endmember abundance.  
Hybrid-type and combined-type object templates contains two or more object endmembers. 
For the hybrid-type object template, each object pixel includes all types of object endmembers, and the object spectral abundance is randomly split into the abundances of each endmember.  
For the combined-type object template, each object pixel is assigned only one object endmember, and the object spectral abundance is equivalent to the endmember abundance. The combined-type object template represents a combination of multiple single-spectral objects.  

\noindent \textbf{Endmember Spectrum Generation}: As shown in Fig.~\ref{fig:object_type}(a), we select 12 types of artificial object spectral reflectance curves $\boldsymbol{r}_t$ from the USGS spectral library \cite{kokaly2017usgs} as initial object endmembers. Based on the radiance curve $\boldsymbol{r}_t$ of seawater and an average radiance curve $\boldsymbol{s}_w$ of a seawater region in the AVIRIS data, $\boldsymbol{r}_t$ is converted into an object endmember baseline $\boldsymbol{s}_t$ by:
\begin{equation}
	\boldsymbol{s}_t=\frac{M_t}{\max \left( \frac{\boldsymbol{r}_t}{\boldsymbol{r}_w}\circ \boldsymbol{s}_w \right)}\boldsymbol{r}_t \circ \frac{1}{\boldsymbol{r}_w}\circ \boldsymbol{s}_w,
\end{equation}
where $\circ$ denotes the pixel-wise  product, ${M_t}$ is the correction to ensure that the numerical range of the object endmember baseline is close to the HSIs to be synthesized, and ${M_t}$ has slight differences for different endmember. Fig.~\ref{fig:spotspec}(b) shows the spectral curves of endmember baselines.
Then, we generate endmember spectra by adding noise into endmember baselines based on the statistical variation properties of water area spectra in AVIRIS data, as shown in  Fig.~\ref{fig:spotspec}(c). 
Spectra of the same object only exhibit the local fluctuation property, while different objects exhibit the wide-area variation property.
Further details can be found in the Supplementary Material.

\noindent \textbf{Simulated Image Synthesis}:
For a simulated HSI to be synthesized, we first randomly select the number of object to be added from the range (1, 20). Each object is assigned a category randomly chosen from the eight object types listed in Table~\ref{tab:object_type}. For each object, we generate the corresponding object template and endmember spectra. The object template is then placed at a random location on the HSI, and the endmember spectra are injected into the HSI according to the object template. The injection process follows the linear mixing model \cite{LMM}, where the synthesized object spectrum is the abundance-weighted sum of the endmember spectra and the background pixel spectra.

As shown in Table~\ref{tab:object_type}, we define eight types of point objects, including five single-type objects (labeled C1, C2, C3, C4, and C5), one hybrid-type object (labeled C6), and two combined-type objects (labeled C7 and C8). Among these, C1 and C2 are single-pixel objects with object spectral abundances smaller than 0.2. The categories and quantities of objects in each simulated image are randomly determined, with the maximum number of objects set to 20.
Examples of each object type on pseudo-color images are shown in Fig.~\ref{fig:spotobject}, where C1 and C2 are barely perceptible to the human eye. To increase the challenge, we average every 5 adjacent bands, reducing the 150 bands to 30 bands to reduce spectral information.

\subsection{The Avon Dataset}
\begin{figure}[t]
	\centering
	\subfloat[]{\includegraphics[height=0.33\linewidth]{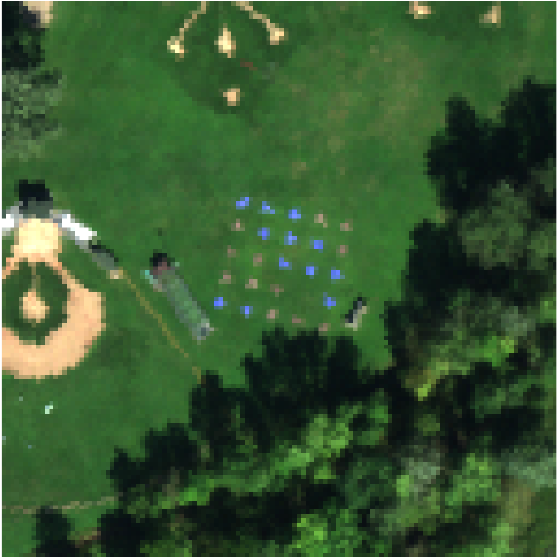}}
	\hfil
	\subfloat[]{\includegraphics[height=0.33\linewidth]{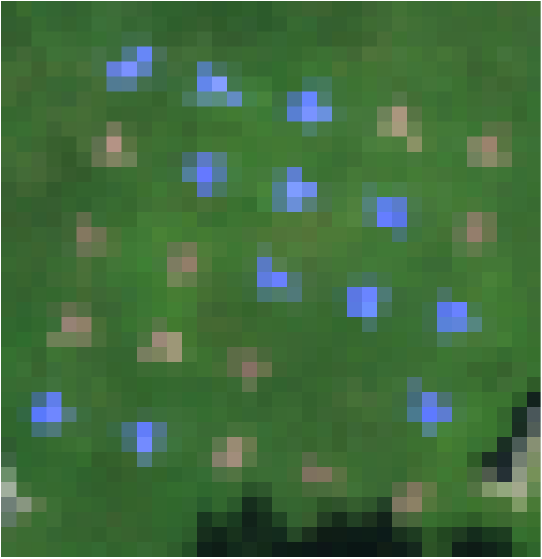}}
	\hfil
	\subfloat[]{\includegraphics[height=0.33\linewidth]{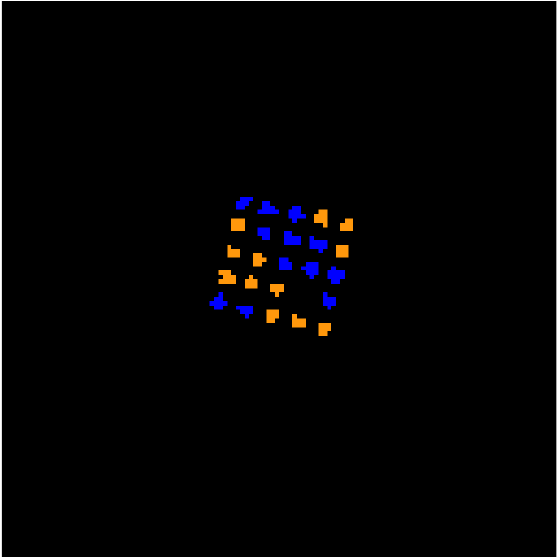}}
	\hfil
	\\
	\subfloat[]{\includegraphics[width=1\linewidth]{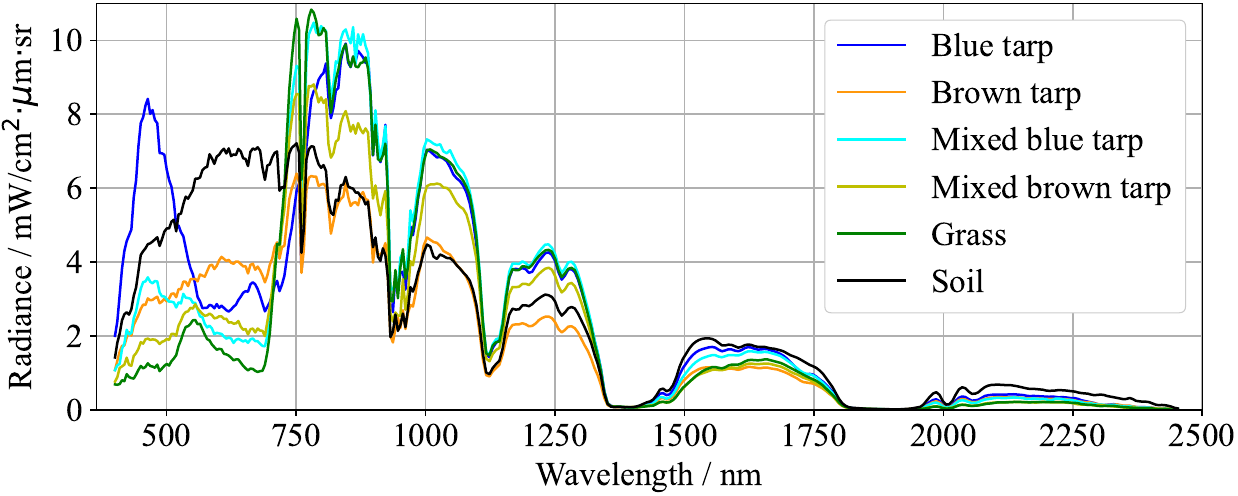}}
	\hfil
	\caption{The Avon dataset. (a) Pseudo-color image. (b) Zoomed-in image of the surrounding area of objects. (c) GT. (d) Radiance spectral curves of objects and background.}
	\label{fig:avon}
\end{figure}

\begin{figure}[t]
	\centering
	\subfloat{\includegraphics[width=0.25\linewidth]{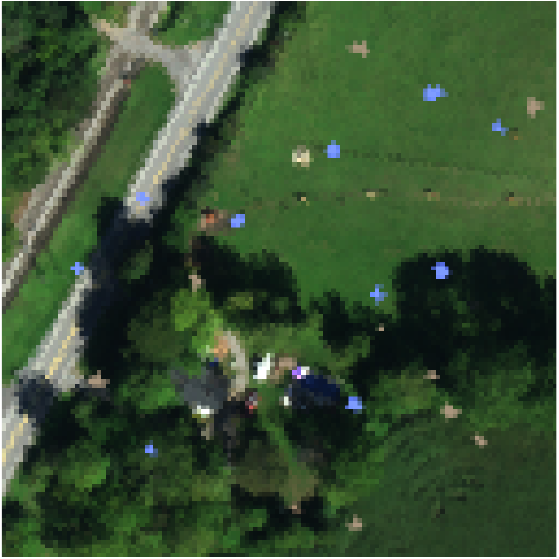}}
	\hfil
	\subfloat{\includegraphics[width=0.25\linewidth]{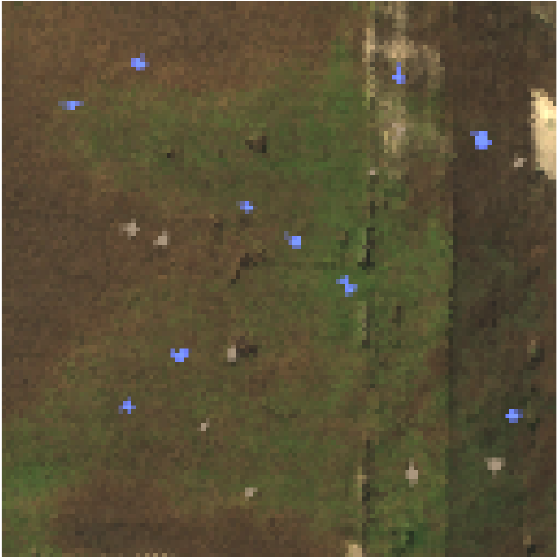}}
	\hfil
	\subfloat{\includegraphics[width=0.25\linewidth]{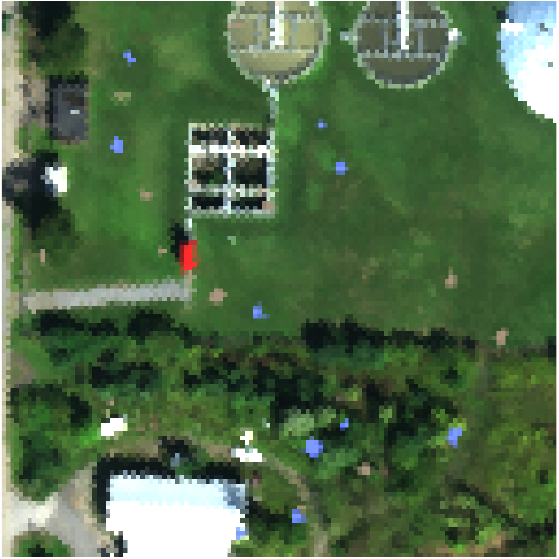}}
	\hfil
	\subfloat{\includegraphics[width=0.25\linewidth]{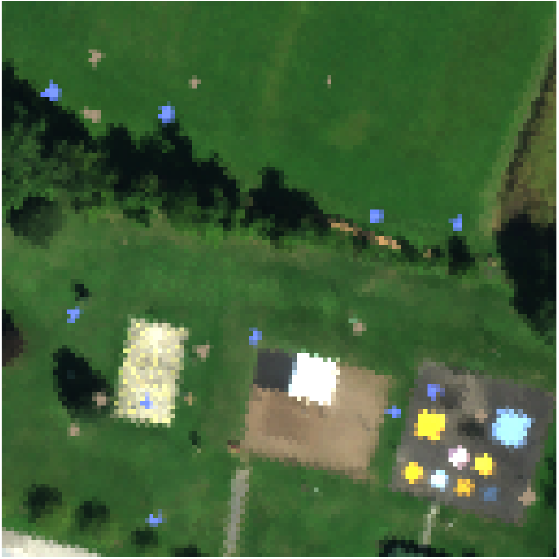}}
	\hfil
	\caption{Simulated training images of the Avon dataset.}
	\label{fig:avontrain}
\end{figure}	

The Avon dataset\footnote{https://dirsapps.cis.rit.edu/share2012/} was collected by a ProSpecTIR-VS sensor during the SHARE 2012 data campaign \cite{giannandrea2013share}, providing hyperspectral images from multiple flights with different flight times.
It has a spatial resolution of 1m and a spectral resolution of 5nm, with 360 bands between 400-2450nm. 
As shown in Fig.~\ref{fig:avon}, we crop a test image of size 128$\times$128 from the flight  0920-1701, which contains 12 blue tarps and 12 brown tarps. The spectra of the edge pixels of the tarps are a compromise between the pure tarp spectra and the grass background spectrum. Between 800-1200nm, the spectrum of brown tarps is similar to that of the soil background.
We use the method described in \secref{subsec:datasets_gt} to form the per-pixel annotations shown in Fig.~\ref{fig:avon}(c) and convert them into the label format for point object detection.
Additionally, we use the pure blue tarp spectrum and brown tarp spectrum from Fig.~\ref{fig:avon}(d) as endmember spectra and cut background images from another fight 0920-1631, generating 150 training images. Each training image contains 10 simulated blue tarps and 10 simulated brown tarps, as shown in Fig.~\ref{fig:avontrain}.
To evaluate the impact of illumination on the performance of the proposed method, we extract the same area as the test image from four other flights and manually annotate the object labels.

\subsection{The SanDiego Dataset}

\begin{figure}[t]
	\centering
	\subfloat[]{\includegraphics[width=0.3\linewidth]{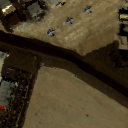}}
	\hfil
	\subfloat[]{\includegraphics[width=0.3\linewidth]{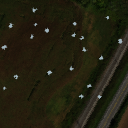}}
	\hfil
	\subfloat[]{\includegraphics[width=0.3\linewidth]{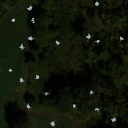}}
	\hfil
	\caption{The SanDiego dataset. (a) The test image. (b)(c) Simulated training images.}
	\label{fig:SanDiego}
\end{figure}

The SanDiego Dataset is captured by the AVIRIS over the SanDiego airport, with a size of 400×400 pixels and a spatial resolution of 3.5 meters. 
As shown in Fig.~\ref{fig:SanDiego}(a), we extract a 128$\times$128 test image focusing on 3 airplanes as objects.
As depicted in  Fig.~\ref{fig:object_label}, we convert the per-pixel annotations provided in literature \cite{shen2023hyperspectral} into the label format for point object detection.
Since the SanDiego Dataset is the atmospherically corrected reflectance data, we use the reflectance data of the flight line 0920-1631 from the SHARE 2012 data campaign to generate 100 training images, each containing 10 airplanes, as illustrated in Fig.~\ref{fig:SanDiego}(b)(c).
After removing bad bands and those not covered by the SHARE 2012 data, we retain 180 bands for the experiment.

\subsection{The Gulfport Dataset}

\begin{figure}[t]
	\centering
	\subfloat{\includegraphics[width=1\linewidth]{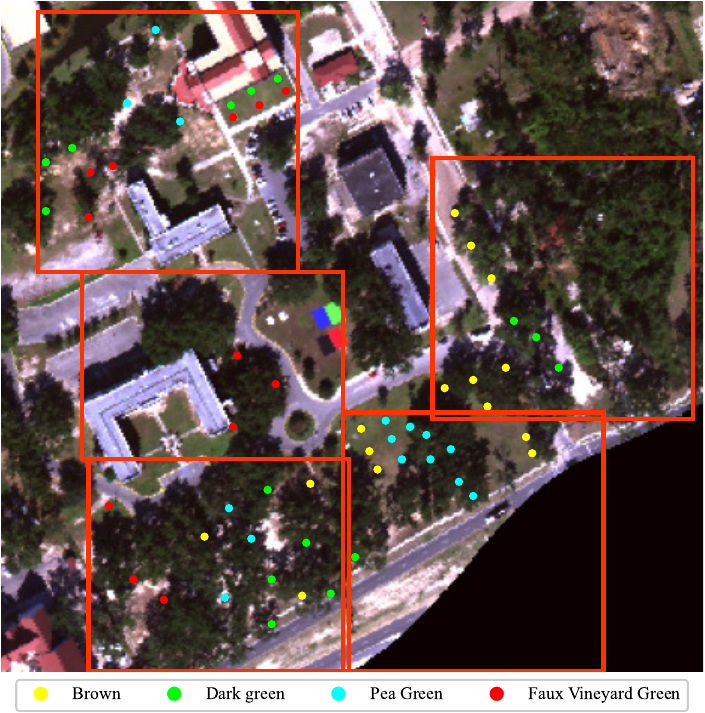}}
	\hfil
	\caption{The Gulfport Dataset. Red bboxs represent the cropped test images. }
	\label{fig:Gulfport}
\end{figure}

\begin{figure}[t]
	\centering
	\subfloat[Visible]{\includegraphics[width=0.24\linewidth]{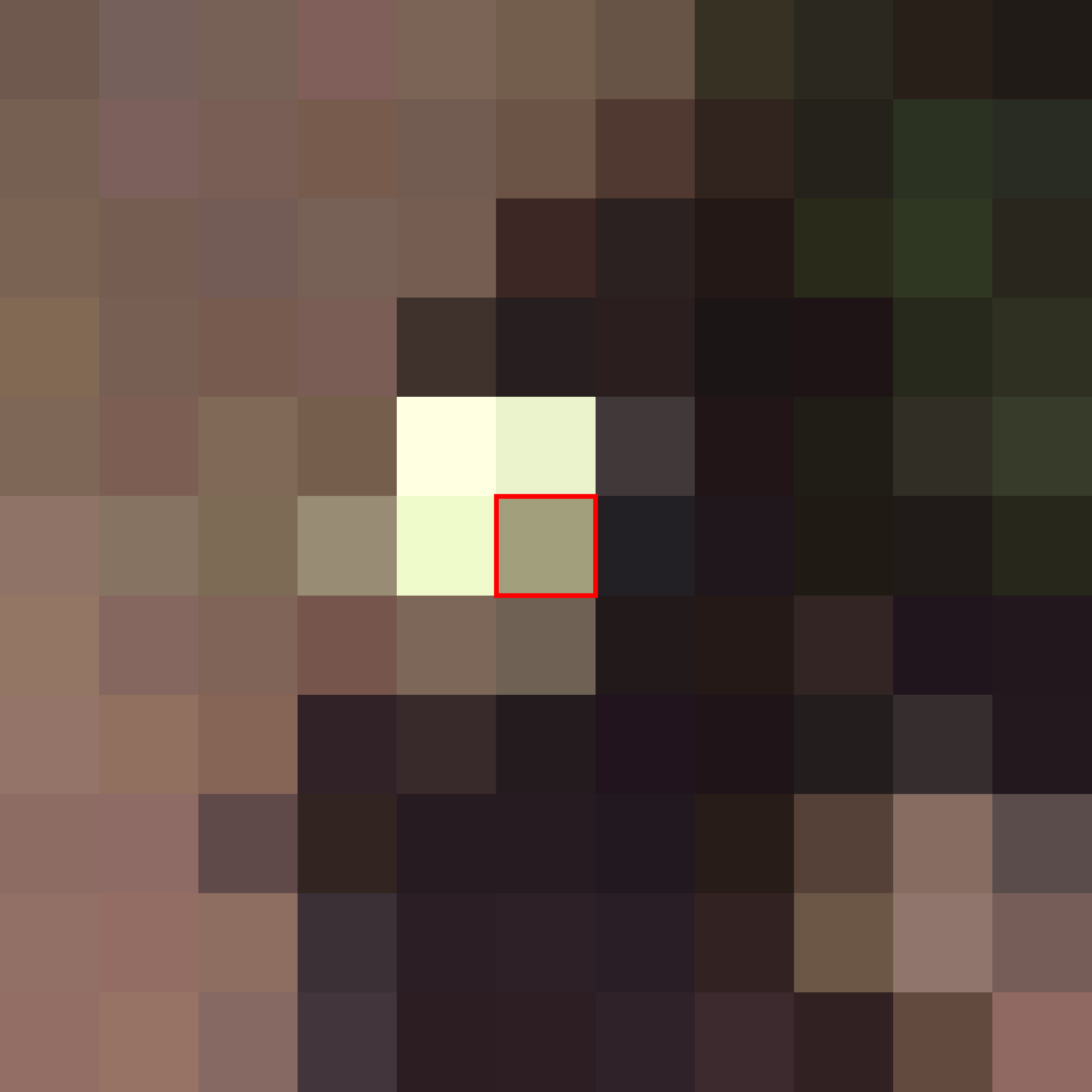}}
	\hfil
	\subfloat[Probably visible]{\includegraphics[width=0.24\linewidth]{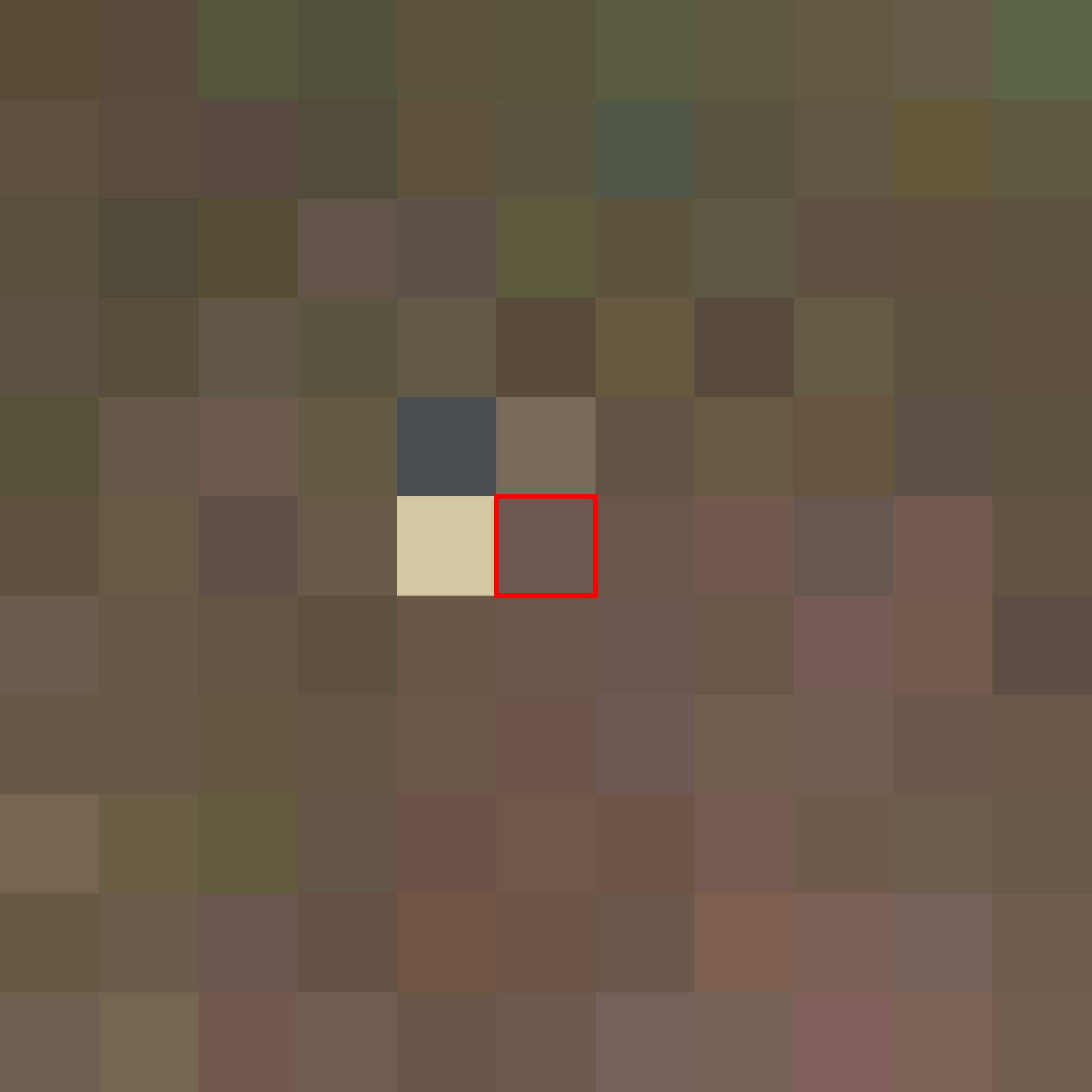}}
	\hfil
	\subfloat[Possibly visible]{\includegraphics[width=0.24\linewidth]{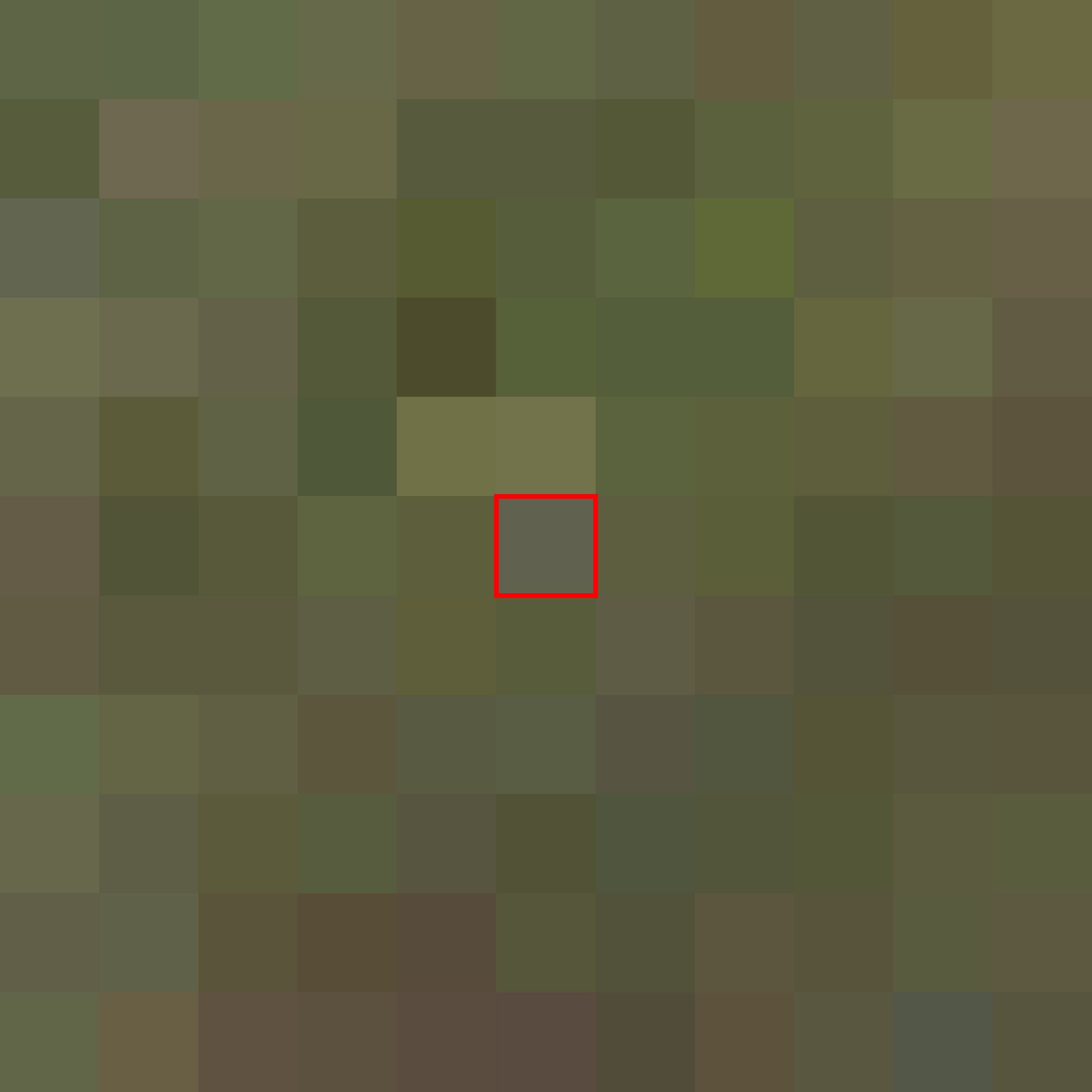}}
	\hfil
	\subfloat[Not visible]{\includegraphics[width=0.24\linewidth]{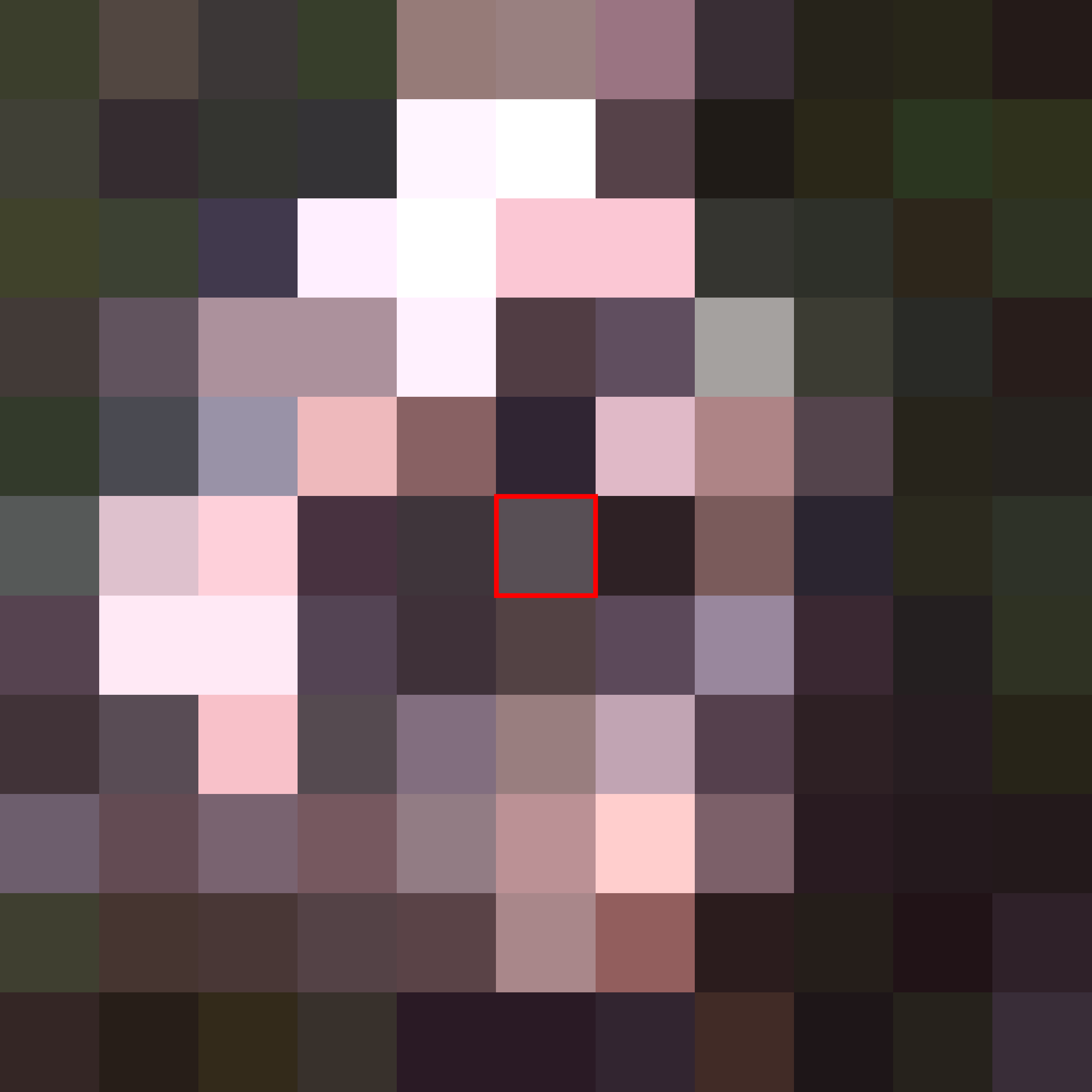}}
	\hfil
	\caption{Pea green panels at four confidence levels. Red bboxs represent the cropped test images. }
	\label{fig:GulfportConf}
\end{figure}

The Gulfport dataset \footnote{https://github.com/GatorSense/MUUFLGulfport/} was collected by a CASI-1500 sensor over the Gulf Park campus of the University of Southern Mississippi, encompassing data from five flights. The images have a spatial resolution of 1 meter and consist of 72 bands between 367.7-1043.4 nm.
This dataset includes 57 artificial objects made of cloth panels in four different colors: brown (15 panels), dark green (15 panels), pea green (15 panels), and faux vineyard green (12 panels).
The object sizes includes 3m$\times$3m, 1m$\times$1m, and 0.5m$\times$0.5m. 
As depicted in Fig.~\ref{fig:Gulfport}, we crop five images from the flight 3 to serve as test images, each padded with zeros to a size of 128$\times$128 pixels.
As show in Fig.~\ref{fig:GulfportConf}, objects are classified into four confidence levels based on the ability of human annotators to identify them in the data: visible (Vis.),  probably visible (Pro.Vis.),  possibly visible (Pos.Vis.), and not visible (NotVis.), with counts of 9, 7, 8, and 33, respectively.
The official GPS GT labels have a certain degree of positioning error, as can also be observed from Fig.~\ref{fig:GulfportConf}.
It is noted that there is a certain degree of position error in the GPS ground truth provided by the official sources, as can also be observed in Figure 3.
Additionally, the flight 4 data where the object areas are masked is used to generate 200 training images. Each image contains five simulated panels of each type panels.
The reference spectra for the simulated objects is the central spectra of the 3m$\times$3m objects from the flight 3.

\subsection{Evaluation Metrics}

\subsubsection{Hyperspectral Point Object Detection}

In the evaluation of the general object detection task, the IoU is calculated as the area of overlap between the predicted bbox and the GT bbox, divided by the area of their union. A prediction is considered correct if the IoU between a predicted box and a GT box exceeds a specified threshold and the class prediction is accurate. Precision (Pr) and Recall (Re) are two fundamental metrics for evaluation. Pr represents the ratio of correctly detected objects to all detected objects, while Re indicates the ratio of correctly detected objects to the total number of GT objects.
﻿

Given the precise GT labels in the SPOD dataset and Avon dataset, we adopt the evaluation methodology of the COCO dataset \cite{lin2014microsoft} for these datasets, primarily reporting the average precision (AP) of each class, and mean AP (mAP) and mean average recall (mAR) across all classes. 
In the COCO evaluation, a fixed number of predictions are selected based on confidence scores for each test image, which we set to 100 in our experiments.
For a specified object class, with a fixed IoU threshold, detection results are divided into positive and negative samples using different confidence thresholds. The precision and recall are calculated for each confidence threshold, and the precision-recall curve is plotted. 
The area under this curve represents the AP value for that class. 
The AP in COCO evaluation is typically the average of AP values over IoU thresholds ranging from 0.5 to 0.95, and the average recall (AR) is the average of recall values over the same IoU range.
For the general object detection task, the object sizes are usually larger, and predictions with smaller IoU values tend to perform poorly in predicting object boundaries. However, for point objects, the ability to locate the object is far more important than predicting the object boundaries, and the bbox annotations of point object are prone to deviation, making smaller IoU thresholds also meaningful. Therefore, we also evaluate the mAP ($\text{mAP}_{\text{25}}$) and mean recall ($\text{mRe}_{\text{25}}$) at an IoU threshold of 0.25.
﻿In the SPOD dataset, subscripts C1 to C8 denote the eight distinct object classes, whereas in the Avon dataset, BL and BR signify the two tarp classes.

However, the GT labels for the test images in the SanDiego dataset and the Gulfport dataset are not precise. The labels for the SanDiego dataset are derived from widely used mask labels within the field. However, unlike the Avon dataset, the object edge pixels in the SanDiego dataset exhibit characteristics of nonlinear mixing. The training set for the SanDiego dataset is based on a linear mixing model simulation, therefore, the prediction boxes of the object detection networks tend to be smaller than the label boxes. For very small objects, even a one or two pixel deviation in the prediction box can lead to a significant drop in IoU.
The labels for the Gulfport dataset are based on GPS point coordinates, which have a certain degree of location error and lack  bbox information. Moreover, many objects in the Gulfport dataset are obscured, and may not have any information in the image, making manual annotation impossible. The COCO evaluation approach, which emphasizes the fitting ability of  bboxes, is not suitable for the SanDiego Dataset and the Gulfport dataset.
To address this, we have fixed the criteria for correct predictions, no longer using the IoU traversal approach. For the SanDiego Dataset, a prediction is considered correct if the IoU between the predicted box and the GT box is greater than 0.25. For the Gulfport Dataset, we use a 5$\times$5 rectangle centered on the pixel corresponding to the object's GPS location as the inner GT bbox, and a 9$\times$9 rectangle as the outer GT bbox. A prediction is considered correct if the predicted box overlaps with the inner GT bbox and does not exceed the outer GT bbox.
With the criteria for correct predictions fixed, we report the AP and Recall metrics for the SanDiego Dataset and the Gulfport Dataset.

\subsubsection{Hyperspectral Target Detection}

Current HTD methods typically produce detection score maps, from which segmentation result maps can be derived through thresholding. In contrast, the proposed SpecDETR provides bbox predictions for objects. To enable a direct comparison between SpecDETR and HTD methods, it is necessary to convert their predictions into a uniform format.
Transforming from segmentation masks to instance bboxes is a well-posed problem, whereas the reverse process is ill-posed. Therefore, we convert the output of current HTD methods into  bboxes for instance-level evaluation.
Specifically, we normalize the score maps for each object class from every HTD compared method using min-max normalization, scaling them to the range [0, 1]. Subsequently, we perform binary segmentation on these score maps. Following the approach illustrated in Fig.~\ref{fig:object_label}, we convert the binary segmentation maps into predicted bboxes, setting the confidence score as the maximum score within the corresponding predicted bbox on the score maps. This transformation of HTD method predictions into object detection predictions allows for the application of object detection evaluation metrics.

Current HTD methods typically use the area under the ROC curve (AUC) \cite{chang2020effective} of the detection score maps as the evaluation metric, without considering the evaluation of segmentation maps. Consequently, many studies do not address the determination of segmentation thresholds. In the field of infrared small target detection, the segmentation IoU of the detection score map \cite{li2022dense} is currently widely adopted, which is the ratio of overlapping pixels between the predicted target pixels and the GT target pixels to the number of pixels in their union. For the normalized score maps of HTD methods for each object class, we traverse the segmentation thresholds at a step size of 0.01, selecting the segmentation map with the highest IoU to maximize the effectiveness of HTD methods in object-level evaluation. Moreover, in our experiments on the SPOD and AVON datasets, we evaluate the HTD methods based on the mean AUC (mAUC) and mean segmentation IoU (mIoU) across all classes.

\section{Experiments}
\label{sec:experiments}

\begin{table*}[t]
	\caption{Performance comparison of SpecDETR and compared visual object detection networks on the SPOD dataset. Size $\times$4 indicates that the input images are enlarged by a factor of 4, while $\times$1 represents the original size.}
	\label{tab:spodnetresult}
	\begin{center}
		\renewcommand\arraystretch{1.0}
		\scriptsize
		\setlength{\tabcolsep}{0.1mm}
		\resizebox{\textwidth}{!}
		{
			\begin{tabular}{c|c|c|c|cccccccccccc}
				\hline
				Detector & Backbone & Epochs & Size  & mAP$\uparrow$   & $\text{mAP}_{\text{25}}$$\uparrow$  & mAR$\uparrow$    & $\text{mRe}_{\text{25}}$$\uparrow$  & $\text{AP}_{\text{C1}}$$\uparrow$  & $\text{AP}_{\text{C2}}$$\uparrow$  & $\text{AP}_{\text{C3}}$$\uparrow$  & $\text{AP}_{\text{C4}}$$\uparrow$  & $\text{AP}_{\text{C5}}$$\uparrow$  & $\text{AP}_{\text{C6}}$$\uparrow$  & $\text{AP}_{\text{C7}}$$\uparrow$  & $\text{AP}_{\text{C8}}$$\uparrow$ \\
				\hline
				Faster R-CNN\cite{girshick2015fast} & ResNet50\cite{he2016deep} & 100   &$\times$4   & 0.197 & 0.377 & 0.245 & 0.419 & 0.000 & 0.000 & 0.000 & 0.035 & 0.026 & 0.537 & 0.430 & 0.550  \\
				Faster R-CNN\cite{girshick2015fast}  & RegNetX\cite{RegnetX} & 100   &  $\times$4  & 0.227 & 0.379 & 0.267 & 0.399 & 0.000 & 0.000 & 0.000 & 0.042 & 0.043 & 0.631 & 0.522 & 0.578  \\
				Faster R-CNN\cite{girshick2015fast}  & ResNeSt50\cite{ResNeSt}& 100   &$\times$4  & 0.246 & 0.316 & 0.269 & 0.319 & 0.000 & 0.000 & 0.000 & 0.008 & 0.003 & 0.644 & 0.564 & 0.747 \\
				Faster R-CNN\cite{girshick2015fast}  & ResNeXt101\cite{ResNeXt} & 100   &  $\times$4    & 0.220 & 0.368 & 0.253 & 0.376 & 0.000 & 0.000 & 0.000 & 0.016 & 0.010 & 0.618 & 0.518 & 0.596  \\
				Faster R-CNN\cite{girshick2015fast}  & HRNet\cite{HRNet}& 100   &   $\times$4  & 0.320 & 0.404 & 0.364 & 0.434 & 0.000 & 0.000 & 0.000 & 0.107 & 0.076 & 0.849 & 0.731 & 0.793  \\
				TOOD\cite{feng2021tood} & ResNeXt101\cite{ResNeXt} & 100   &   $\times$4    & 0.304 & 0.464 & 0.401 & 0.570 & 0.000 & 0.000 & 0.000 & 0.181 & 0.194 & 0.743 & 0.648 & 0.663  \\
				CentripetalNet\cite{CentripetalNet} & HourglassNet104\cite{HourglassNet} & 100   & $\times$4    & 0.695 & 0.829 & 0.840 & 0.956 & 0.831 & 0.888 & 0.915 & 0.373 & 0.367 & 0.810 & 0.655 & 0.725 \\
				CornerNet\cite{Cornernet} & HourglassNet104\cite{HourglassNet} & 100   &   $\times$4    & 0.626 & 0.736 & 0.855 & 0.969 & 0.797 & 0.751 & 0.855 & 0.328 & 0.308 & 0.768 & 0.554 & 0.644 \\
				RepPoints\cite{Reppoints} & ResNet50\cite{he2016deep}& 100   &  $\times$4  & 0.207 & 0.691 & 0.346 & 0.934 & 0.043 & 0.143 & 0.269 & 0.071 & 0.073 & 0.372 & 0.265 & 0.417  \\
				RepPoints\cite{Reppoints}& ResNeXt101\cite{ResNeXt} & 100   &  $\times$4  & 0.485 & 0.806 & 0.635 & 0.961 & 0.373 & 0.561 & 0.632 & 0.242 & 0.253 & 0.658 & 0.508 & 0.649  \\
				RetinaNet\cite{RetinaNet} & EfficientNet\cite{tan2019efficientnet} & 100   &   $\times$4    & 0.462 & 0.836 & 0.611 & 0.971 & 0.566 & 0.602 & 0.530 & 0.182 & 0.210 & 0.566 & 0.471 & 0.569 \\
				RetinaNet\cite{RetinaNet} & PVTv2-B3\cite{wang2022pvt} & 100   &  $\times$4  & 0.426 & 0.757 & 0.650 & 0.987 & 0.356 & 0.478 & 0.458 & 0.209 & 0.232 & 0.563 & 0.470 & 0.644 \\
				DeformableDETR\cite{zhu2020deformable} & ResNet50\cite{he2016deep} & 100   & $\times$4   & 0.231 & 0.692 & 0.385 & 0.883 & 0.230 & 0.316 & 0.234 & 0.077 & 0.070 & 0.289 & 0.238 & 0.395 \\
				DINO\cite{zhang2022dino}  & ResNet50\cite{he2016deep} & 100   &   $\times$4  & 0.168 & 0.491 & 0.368 & 0.763 & 0.020 & 0.047 & 0.277 & 0.080 & 0.064 & 0.286 & 0.213 & 0.360 \\
				DINO\cite{zhang2022dino}  & Swin-L\cite{liu2021swin} & 100   &   $\times$4   & 0.757 & 0.852 & \textbf{0.909} & \textbf{0.983} & 0.915 & 0.912 & 0.951 & 0.483 & 0.497 & 0.847 & 0.728 & 0.721 \\
				\hline
				SpecDETR			&  -     & 100   & $\times$1      & \textbf{0.856} & \textbf{0.938} & 0.897 & 0.955 & \textbf{0.963} & \textbf{0.969} & \textbf{0.970} & \textbf{0.698} & \textbf{0.648} & \textbf{0.905} & \textbf{0.844} & \textbf{0.850} \\
				\hline
			\end{tabular}%
		}
	\end{center}
\end{table*}

\begin{table*}[t]
	\caption{Computational Efficiency comparison of SpecDETR and compared visual object detection networks on the SPOD dataset.}
	\label{tab:spodnetresult_time}
	\begin{center}
		\renewcommand\arraystretch{1.0}
		\scriptsize
		\resizebox{\textwidth}{!}
		{
			\begin{tabular}{cc|c|ccccc}
				\hline
				\multirow{2}[0]{*}{Detector} & \multirow{2}[0]{*}{Backbone} & Image & Training & Inference   & FLOPs & Params & GPUMem. \\
				&       & Size  & time (min) & Speed (FPS) & (G)   & (M)   &   (MB) \\
				\hline
				Faster R-CNN\cite{girshick2015fast} & ResNet50\cite{he2016deep} & $\times$4 & \textbf{8.8} & \textbf{40.3} & 68.8  & 41.5  & 309 \\
				Faster R-CNN\cite{girshick2015fast}  & RegNetX\cite{RegnetX} & $\times$4 & 9.0   & 33.2  & 57.7  & 31.6  & 270 \\
				Faster R-CNN\cite{girshick2015fast}  & ResNeSt50\cite{ResNeSt} & $\times$4 & 12.8  & 23.9  & 185.1 & 44.6  & 467 \\
				Faster R-CNN\cite{girshick2015fast}  & ResNeXt101\cite{ResNeXt} & $\times$4 & 14.6  & 26.3  & 128.4 & 99.4  & 528 \\
				Faster R-CNN\cite{girshick2015fast}  & HRNet\cite{HRNet} & $\times$4 & 13.9  & 17.4  & 104.4 & 63.2  & 407 \\
				TOOD\cite{feng2021tood} & ResNeXt101\cite{ResNeXt} & $\times$4 & 17.3  & 11.2  & 114.3 & 97.7  & 485 \\
				CentripetalNet\cite{CentripetalNet} & HourglassNet104\cite{HourglassNet} & $\times$4 & 30.2  & 14.7  & 501.3 & 205.9 & 1098 \\
				CornerNet\cite{Cornernet} & HourglassNet104\cite{HourglassNet} & $\times$4 & 25.0  & 16.3  & 462.6 & 201.1 & 981 \\
				RepPoints\cite{Reppoints} & ResNet50\cite{he2016deep} & $\times$4 & 10.4  & 33.7  & 54.1  & 36.9  & \textbf{238} \\
				RepPoints\cite{Reppoints}& ResNeXt101\cite{ResNeXt} & $\times$4 & 17.6  & 14.7  & 75.0  & 58.1  & 329 \\
				RetinaNet\cite{RetinaNet} & EfficientNet\cite{tan2019efficientnet} & $\times$4 & 9.0   & 30.5  & \textbf{36.1} & 18.5  & 215 \\
				RetinaNet\cite{RetinaNet} & PVTv2-B3\cite{wang2022pvt} & $\times$4 & 13.4  & 22.3  & 71.3  & 52.4  & 313 \\
				DeformableDETR\cite{zhu2020deformable} & ResNet50\cite{he2016deep} & $\times$4 & 13.8  & 24.6  & 58.7  & 41.2  & 252 \\
				DINO\cite{zhang2022dino}  & ResNet50\cite{he2016deep} & $\times$4 & 17.3  & 21.3  & 86.3  & 47.6  & 301 \\
				DINO\cite{zhang2022dino}  & Swin-L\cite{liu2021swin} & $\times$4 & 53.8  & 10.1  & 203.9 & 218.3 & 1168 \\
				\hline
				SpecDETR &     -   & $\times$1 & 17.8  & 24.7  & 139.7 & \textbf{16.1} & 287 \\
				\hline
			\end{tabular}
		}
	\end{center}
\end{table*}

\begin{table*}[t]
	\caption{Performance comparison of SpecDETR and compared HTD methods on the SPOD dataset.}
	\begin{center}
		\renewcommand\arraystretch{1.0}
		\scriptsize
		\setlength{\tabcolsep}{1mm}
		\resizebox{\textwidth}{!}{
			\begin{tabular}{c|cc|cccccccccccc}
				\hline
				Method  & mAUC$\uparrow$  & mIoU$\uparrow$  & mAP$\uparrow$   & $\text{mAP}_{\text{25}}$$\uparrow$  & mAR$\uparrow$    & $\text{mRe}_{\text{25}}$$\uparrow$  & $\text{AP}_{\text{C1}}$$\uparrow$  & $\text{AP}_{\text{C2}}$$\uparrow$  & $\text{AP}_{\text{C3}}$$\uparrow$  & $\text{AP}_{\text{C4}}$$\uparrow$  & $\text{AP}_{\text{C5}}$$\uparrow$  & $\text{AP}_{\text{C6}}$$\uparrow$  & $\text{AP}_{\text{C7}}$$\uparrow$  & $\text{AP}_{\text{C8}}$$\uparrow$   \\
				\hline
				ASD \cite{ASD99-TD}   & \textbf{0.991} & \textbf{0.359} & 0.182 & 0.286 & 0.712 & 0.939 & 0.061 & 0.068 & 0.587 & 0.078 & 0.065 & 0.298 & 0.101 & 0.194 \\
				CEM \cite{CEM-TD}   & 0.981 & 0.166 & 0.040 & 0.122 & 0.262 & 0.672 & 0.011 & 0.062 & 0.215 & 0.009 & 0.005 & 0.008 & 0.010 & 0.002 \\
				OSP \cite{OSP-TD}   & 0.969 & 0.159 & 0.031 & 0.108 & 0.202 & 0.594 & 0.008 & 0.039 & 0.181 & 0.002 & 0.002 & 0.006 & 0.009 & 0.000 \\
				KOSP \cite{KOSP}  & 0.961 & 0.116 & 0.017 & 0.083 & 0.165 & 0.539 & 0.000 & 0.002 & 0.119 & 0.002 & 0.002 & 0.009 & 0.001 & 0.002 \\
				SMF \cite{SMF-TD}   & 0.904 & 0.039 & 0.003 & 0.016 & 0.078 & 0.301 & 0.000 & 0.000 & 0.020 & 0.000 & 0.000 & 0.000 & 0.000 & 0.000 \\
				KSMF \cite{kwon2006comparative}  & 0.891 & 0.041 & 0.003 & 0.015 & 0.059 & 0.206 & 0.000 & 0.000 & 0.024 & 0.000 & 0.000 & 0.000 & 0.000 & 0.000 \\
				TCIMF \cite{TCIMF-TD} & 0.975 & 0.096 & 0.009 & 0.061 & 0.160 & 0.547 & 0.004 & 0.001 & 0.052 & 0.003 & 0.001 & 0.004 & 0.005 & 0.003 \\
				CR \cite{CR-AD}    & 0.939 & 0.160 & 0.031 & 0.138 & 0.242 & 0.685 & 0.008 & 0.000 & 0.160 & 0.017 & 0.015 & 0.016 & 0.020 & 0.012 \\
				KSR \cite{KSR-C}   & 0.877 & 0.140 & 0.015 & 0.091 & 0.133 & 0.530 & 0.008 & 0.001 & 0.094 & 0.002 & 0.003 & 0.006 & 0.004 & 0.001 \\
				KSRBBH \cite{KSRBBH-TD} & 0.785 & 0.144 & 0.007 & 0.087 & 0.105 & 0.466 & 0.000 & 0.000 & 0.000 & 0.007 & 0.007 & 0.028 & 0.008 & 0.010 \\
				LSSA \cite{zhu2023learning}  & 0.702 & 0.119 & 0.041 & 0.093 & 0.163 & 0.302 & 0.001 & 0.002 & 0.272 & 0.001 & 0.000 & 0.051 & 0.000 & 0.000 \\
				IRN \cite{shen2023hyperspectral}  & 0.625 & 0.004 & 0.000 & 0.002 & 0.005 & 0.033 & 0.000 & 0.000 & 0.000 & 0.000 & 0.000 & 0.000 & 0.000 & 0.000 \\
				TSTTD \cite{jiao2023triplet} & 0.964 & 0.056 & 0.044 & 0.057 & 0.745 & 0.918 & 0.005 & 0.022 & 0.207 & 0.017 & 0.018 & 0.020 & 0.013 & 0.049 \\
				\hline
				SpecDETR &  -   &   -    & \textbf{0.856} & \textbf{0.938} &  \textbf{0.897} &  \textbf{0.955} & \textbf{0.963} & \textbf{0.969} & \textbf{0.970} & \textbf{0.698} & \textbf{0.648} & \textbf{0.905} & \textbf{0.844} & \textbf{0.850} \\
				\hline
			\end{tabular}%
		}
	\end{center}
	\label{tab:spodhtdresult}
\end{table*}

\subsection{Comparison Experiments on the SPOD and Avon Dataset}
\label{sec:ep_SPOD}

\subsubsection{Implementation Details}

We use the MMDetection project \cite{chen2019mmdetection} to implement SpecDETR.
To ensure the diversity of generated samples within the denoising approach, the parameters $\tau _1$ and $\tau _2$ for CCDN are set to 0.5 and 1.5, respectively, while the number of DN queries is set to 200 pairs.
In the matching approach, to guarantee the quality of the assigned positive samples with the hybrid label assigner, the IoU threshold is set to 0.95, and the number threshold $T$ is fixed at 9. 
The learning rate is initialized at 0.001, consistent with DINO, and is reduced by a factor of 0.1 at the 20th, 30th, and 90th epochs with the setting of 24, 36, and 100 training epochs, respectively. 
During the inference stage, we retain the top 300 prediction boxes with the highest confidence scores on each image and subsequently apply NMS to handle overlapping boxes. The IoU threshold for NMS is set to 0.01, indicating that the output predicted boxes are considered non-overlapping. All other parameters are kept identical to those used in DINO.

Given that the hyperspectral point object detection task developed in this paper spans the domains of HTD and visual object detection, the proposed SpecDETR is compared with both visual object detection networks and HTD methods on the SPOD dataset.
Visual object detection networks typically use backbone networks to extract downsampled feature maps from input images. Directly inputting images of the original size without altering the network structure can be inefficient for extracting features of point objects, so we upscale the input HSIs by a factor of 4 by replicating each pixel 16 times before feeding them into the compared visual object detection networks. Additionally, the input channels of these networks are adjusted from 3 to match the band number of the input HSIs.The number of training epochs is set to 100 to ensure full convergence of the networks, while all other parameters are retained at their default settings. SpecDETR and the compared networks are trained with a batch size of 4 on an RTX 3090 GPU.

We compare SpecDETR with current HTD methods on the SPOD dataset and the Avon dataset, both of which provide fine, per-pixel annotations for multi-class objects.
Current HTD methods take a test image along with prior spectral signatures as input and typically treat the predictions as a single class. To obtain multi-class predictions, the compared methods \cite{ASD99-TD, CEM-TD, OSP-TD, KOSP, SMF-TD, kwon2006comparative, TCIMF-TD,CR-AD, KSR-C,SRBBH-TD,shen2023hyperspectral,jiao2023triplet} sequentially use the prior spectra of different object classes for detection, yielding predictions for each object class individually.
LSSA \cite{zhu2023learning} detects sub-pixel targets by directly learning the target abundances, so we adapt LSSA into a multi-class detector by inputting all classes' prior spectra and simultaneously outputting abundance predictions for all classes.
Moreover, IRN \cite{shen2023hyperspectral} and TSTTD \cite{jiao2023triplet} are the SOTA deep learning-based HTD methods, and we also make improvements to them. 
The modified IRN and TSTTD are trained on a set of training images and then directly infer on the test images.
The original version of TSTTD generates training samples by mixing prior spectral signatures with background spectra, whereas our improved version trains directly on the target spectra from the image set, ensuring that TSTTD and SpecDETR are exposed to an equivalent level of object prior information.
The SPOD dataset contains multi-spectral objects thus providing 20 prior spectra for each object class. In contrast, the Avon dataset consists of single-spectral objects, with only 1 prior spectrum provided for each object class.

In the absence of further specifications, SpecDETR defaults to 100 training epochs on the SPOD dataset and 36 epochs on the Avon dataset.
Since the comparison on the Avon dataset is exclusively with HTD methods, the output bounding boxes produced by SpecDETR are rounded to the nearest integer.

\subsubsection{Result Analysis}
Table~\ref{tab:spodnetresult} provides the performance comparison between SpecDETR and the compared visual object detection networks  in point object detection. 
SpecDETR achieves the highest mAP of 0.856 after 100 epochs among all compared networks. 
SpecDETR exhibits remarkable detection capability for subpixel objects and achieves $\text{AP}_{\text{C1}}$, $\text{AP}_{\text{C2}}$, and $\text{AP}_{\text{C3}}$ of 0.963, 0.969, and 0.97, respectively.
Table~\ref{tab:spodnetresult} further demonstrates the viability of current visual object detection networks in directly extracting  spatial-spectral joint features from hyperspectral cube data. Upon substituting the input of these networks from a three-channel RGB image to the complete hyperspectral cube, they continue to operate effectively without any structural alterations. Notably, both C1 and C2, which occupy merely a single pixel and possess object abundances below 0.2, exhibit AP performance that is remarkably close to that of multi-pixel objects like C7 and C8 across several compared networks. This observation suggests that the features extracted by these networks indeed incorporate spectral information.
We also observe that those backbones that can offer high-resolution shallow features are more valuable in point object detection. Followed by Faster R-CNN, HRNet, which achieves multi-scale feature interaction, outperforms ResNeXt. CentripetalNet and CornerNet with HourglassNet achieve 0.626 mAP and 0.691 mAP, respectively.

\begin{figure*}[htpb]
	\centering
	\subfloat{\includegraphics[width=0.22\linewidth]{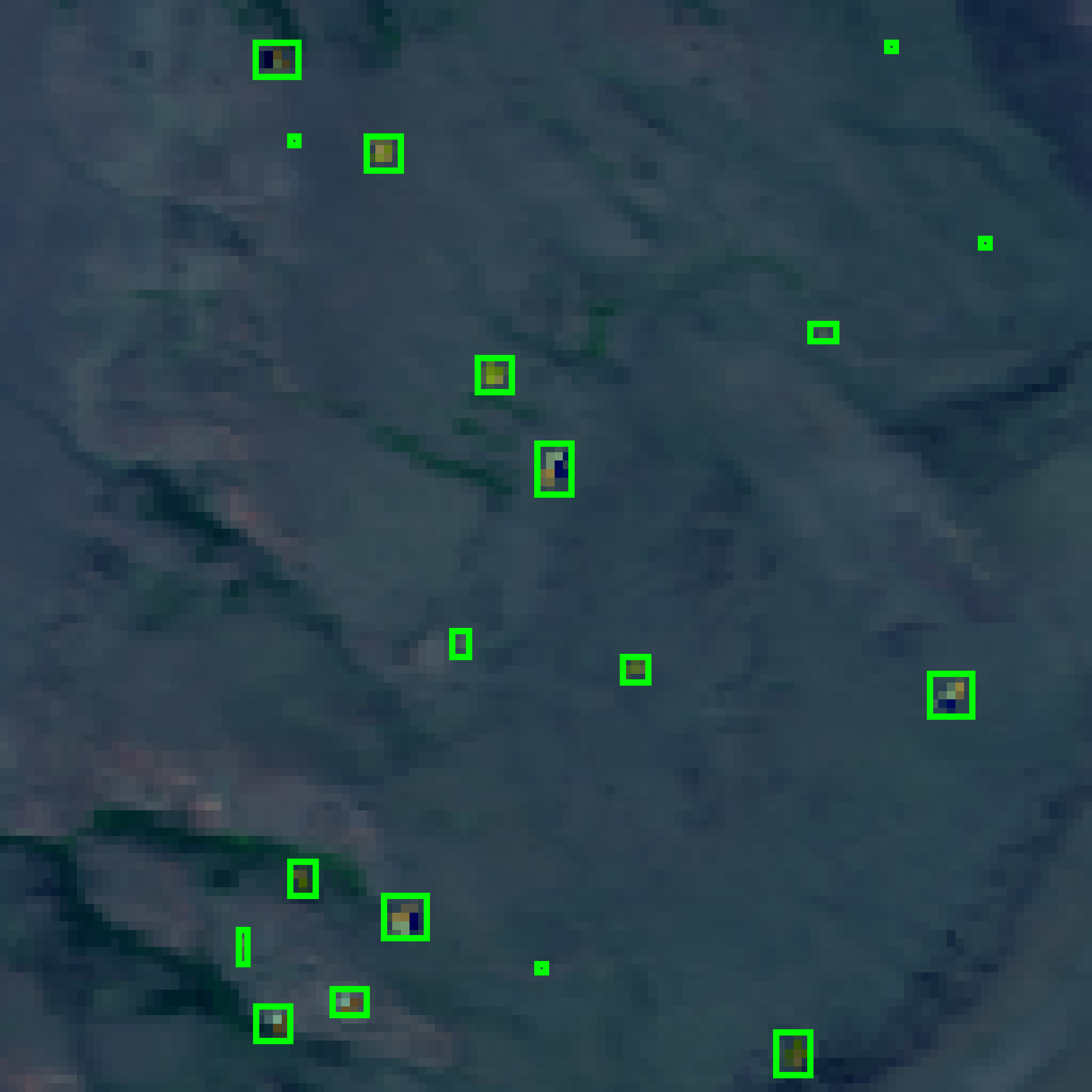}}
	\hfil
	\subfloat{\includegraphics[width=0.22\linewidth]{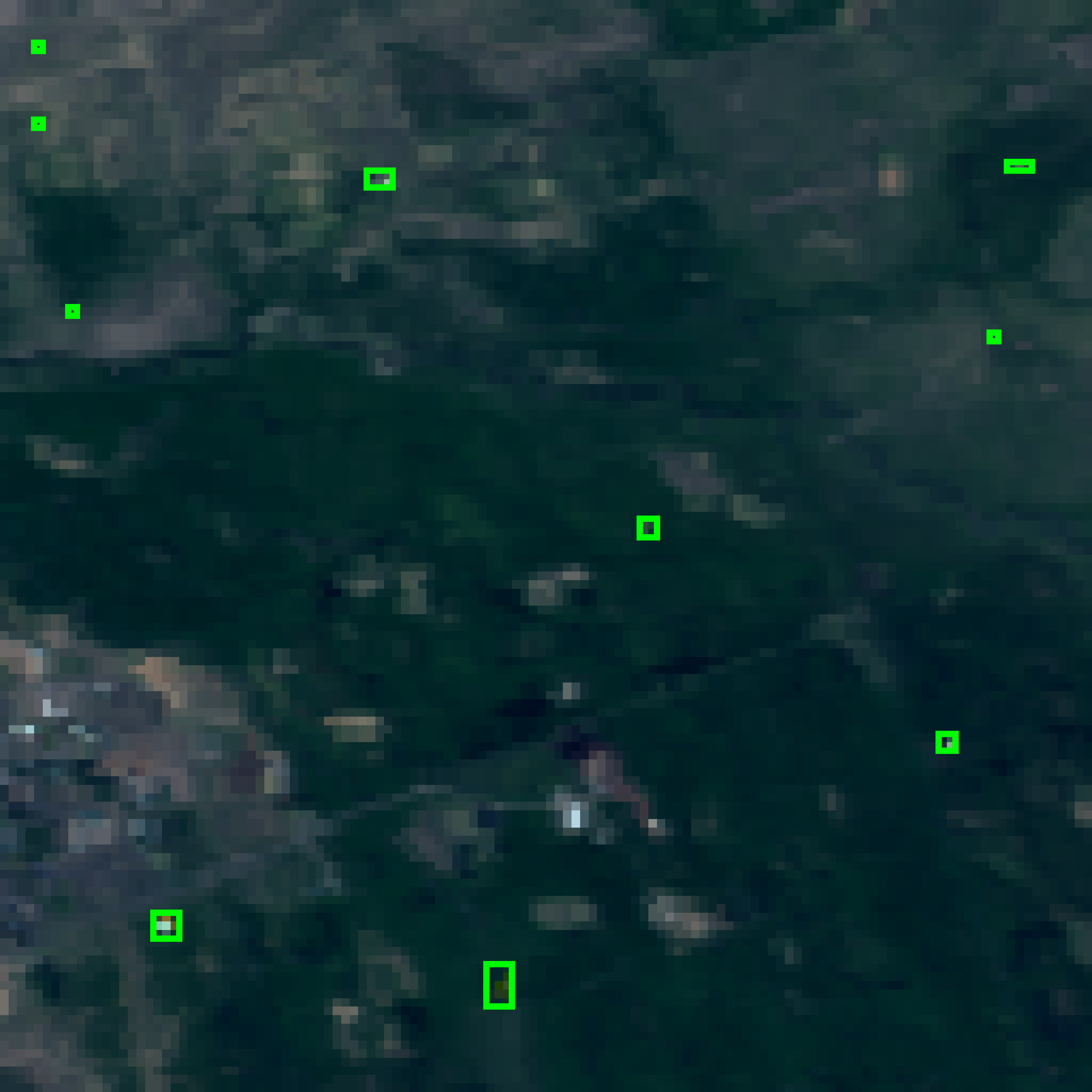}}
	\hfil
	\subfloat{\includegraphics[width=0.22\linewidth]{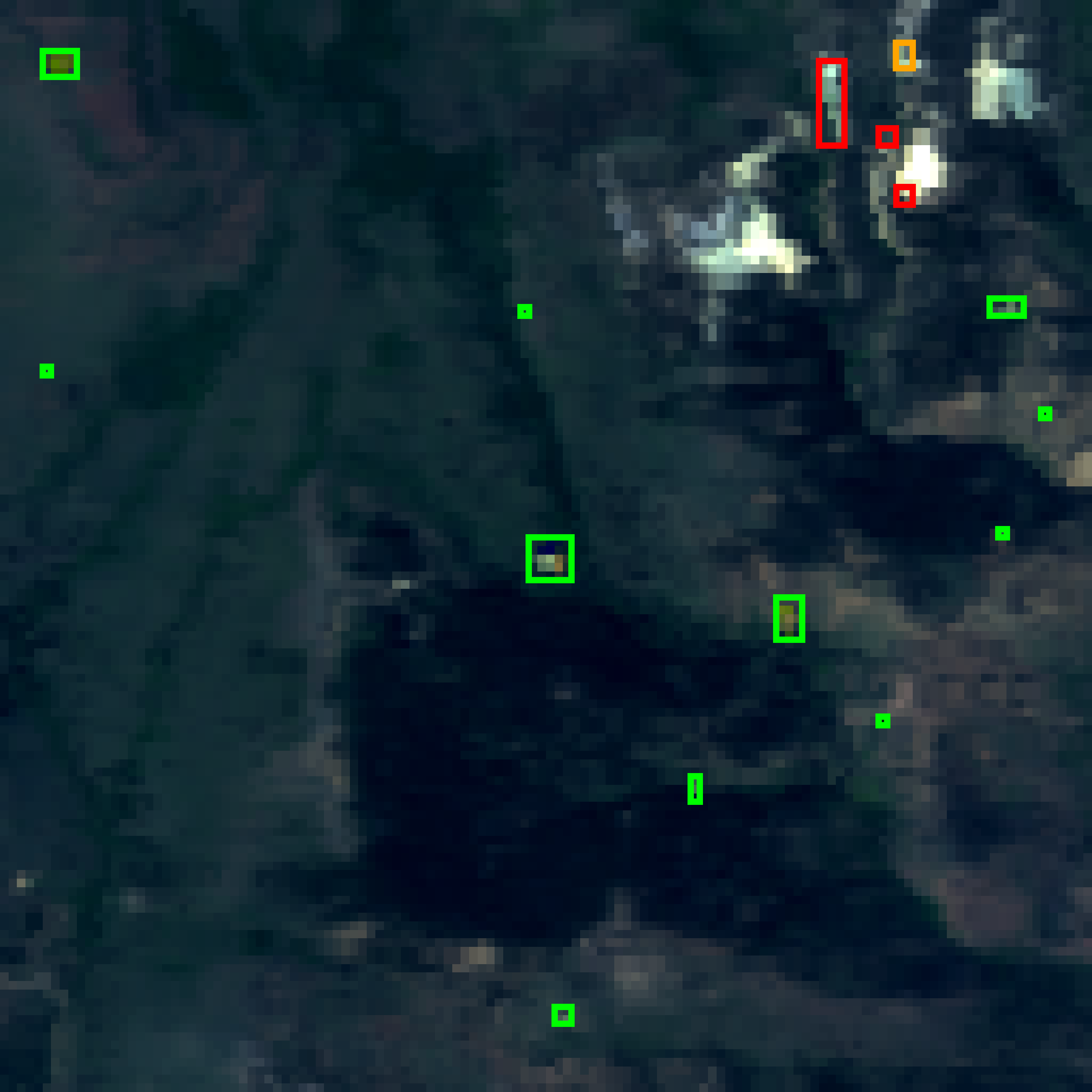}}
	\hfil
	\subfloat{\includegraphics[width=0.22\linewidth]{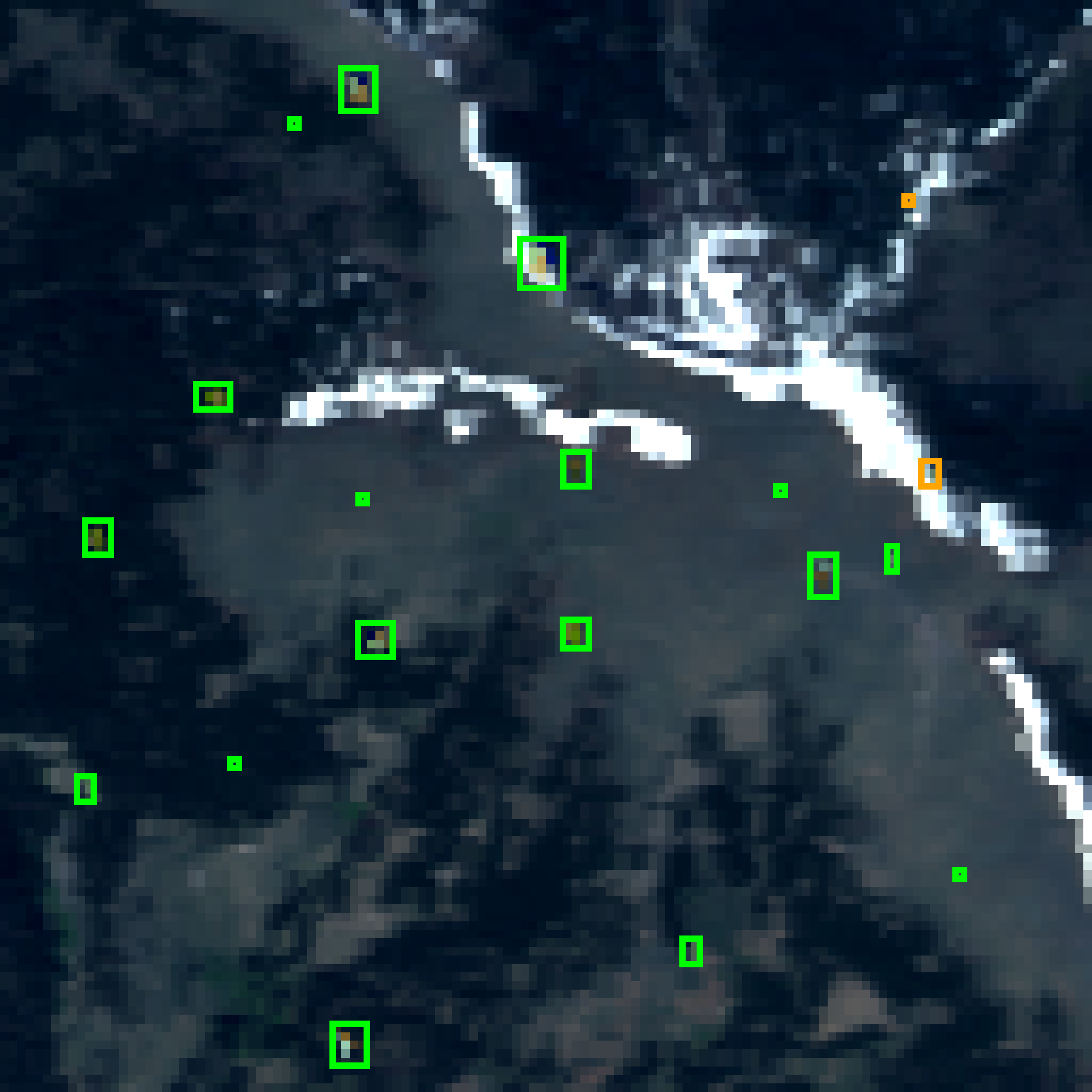}}
	\hfil
	\\
	\subfloat{\includegraphics[width=0.22\linewidth]{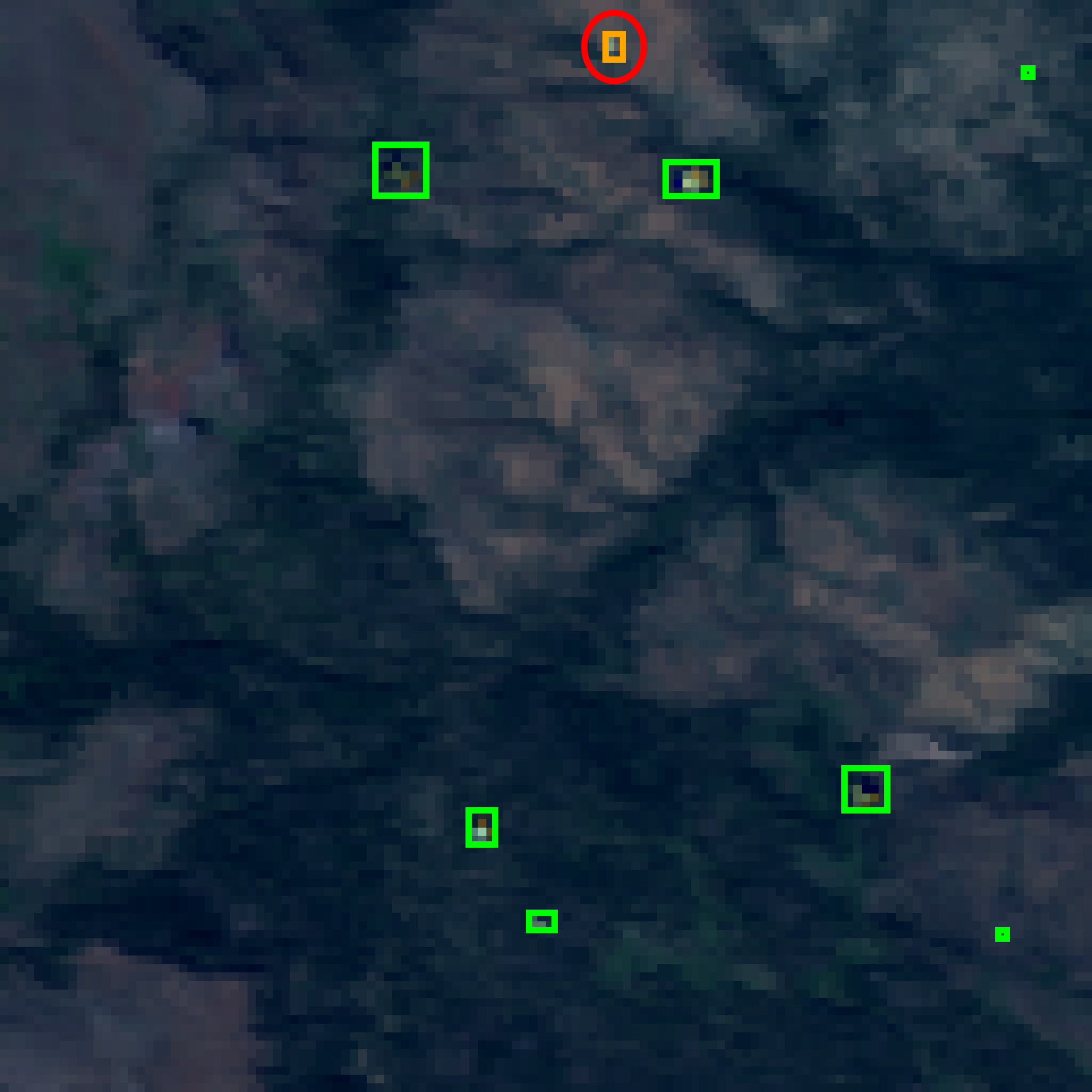}}
	\hfil
	\subfloat{\includegraphics[width=0.22\linewidth]{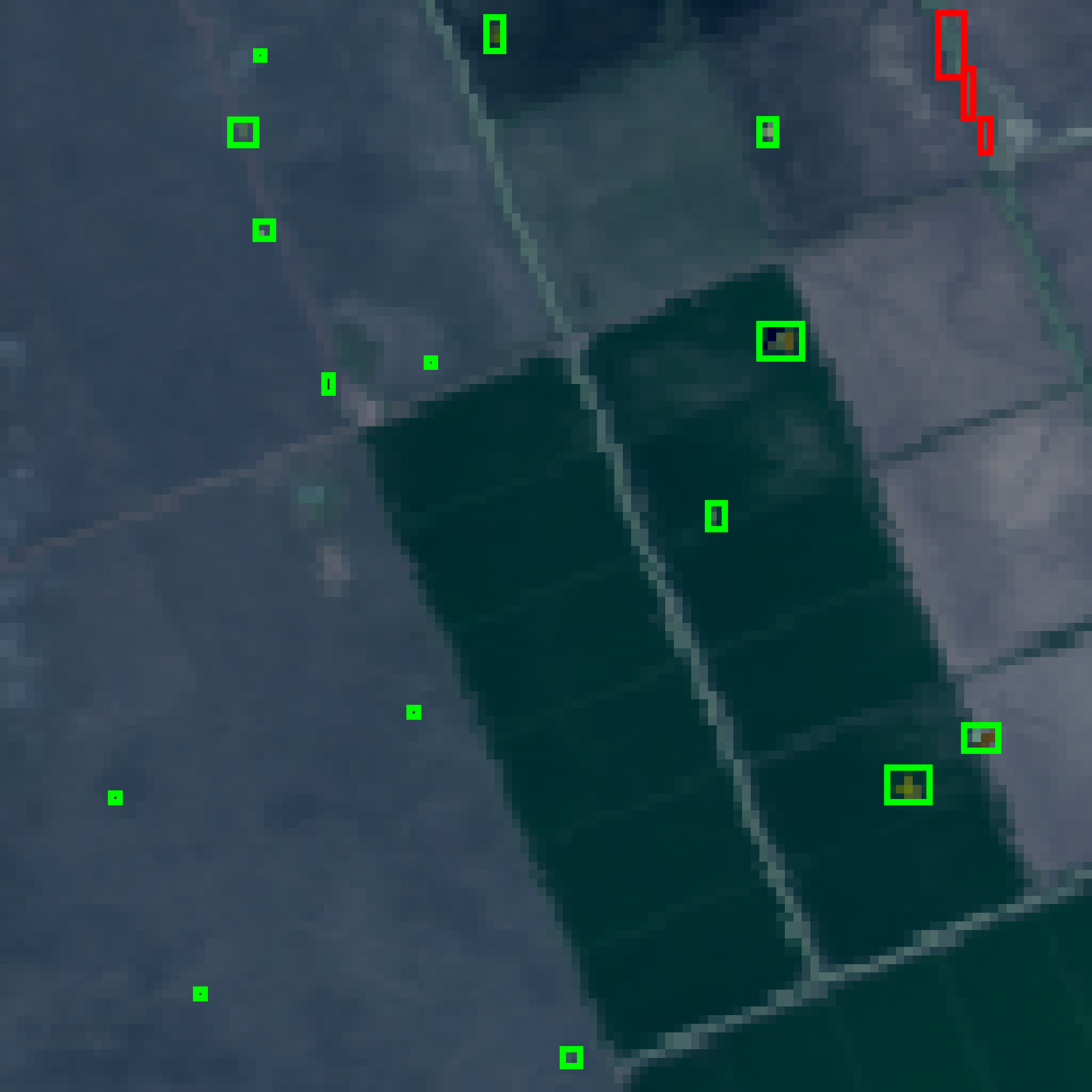}}
	\hfil
	\subfloat{\includegraphics[width=0.22\linewidth]{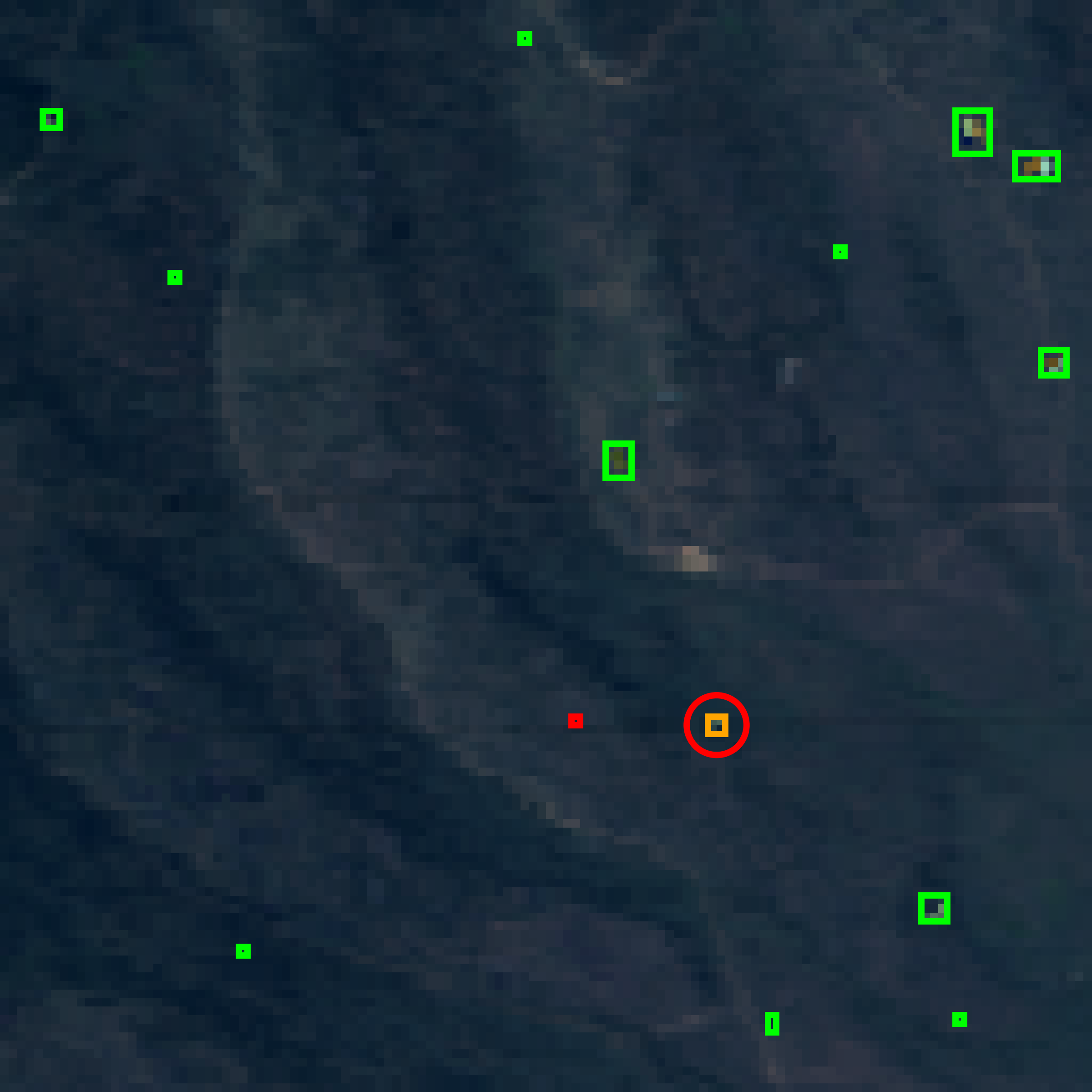}}
	\hfil
	\subfloat{\includegraphics[width=0.22\linewidth]{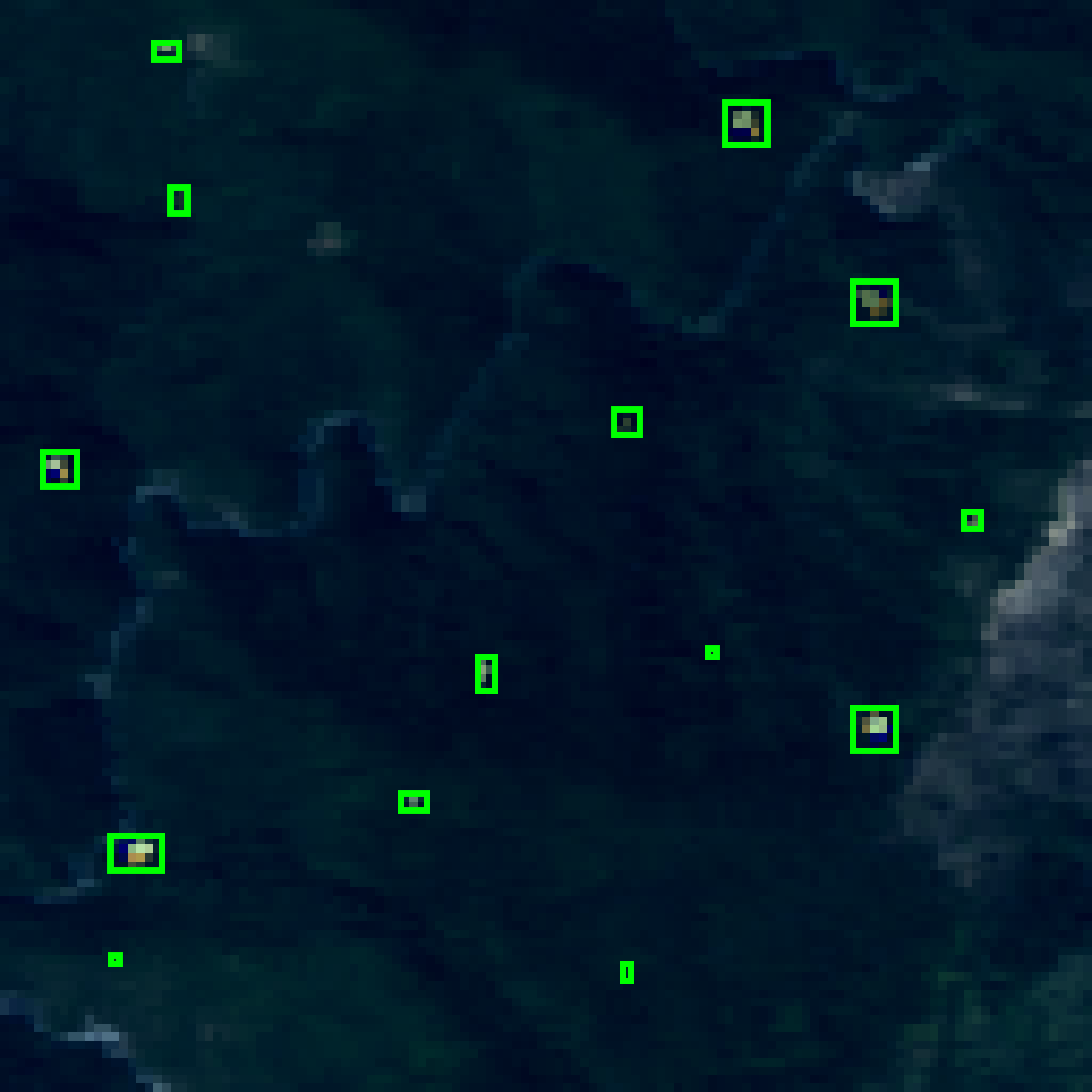}}
	\hfil
	\\
	\subfloat{\includegraphics[width=0.22\linewidth]{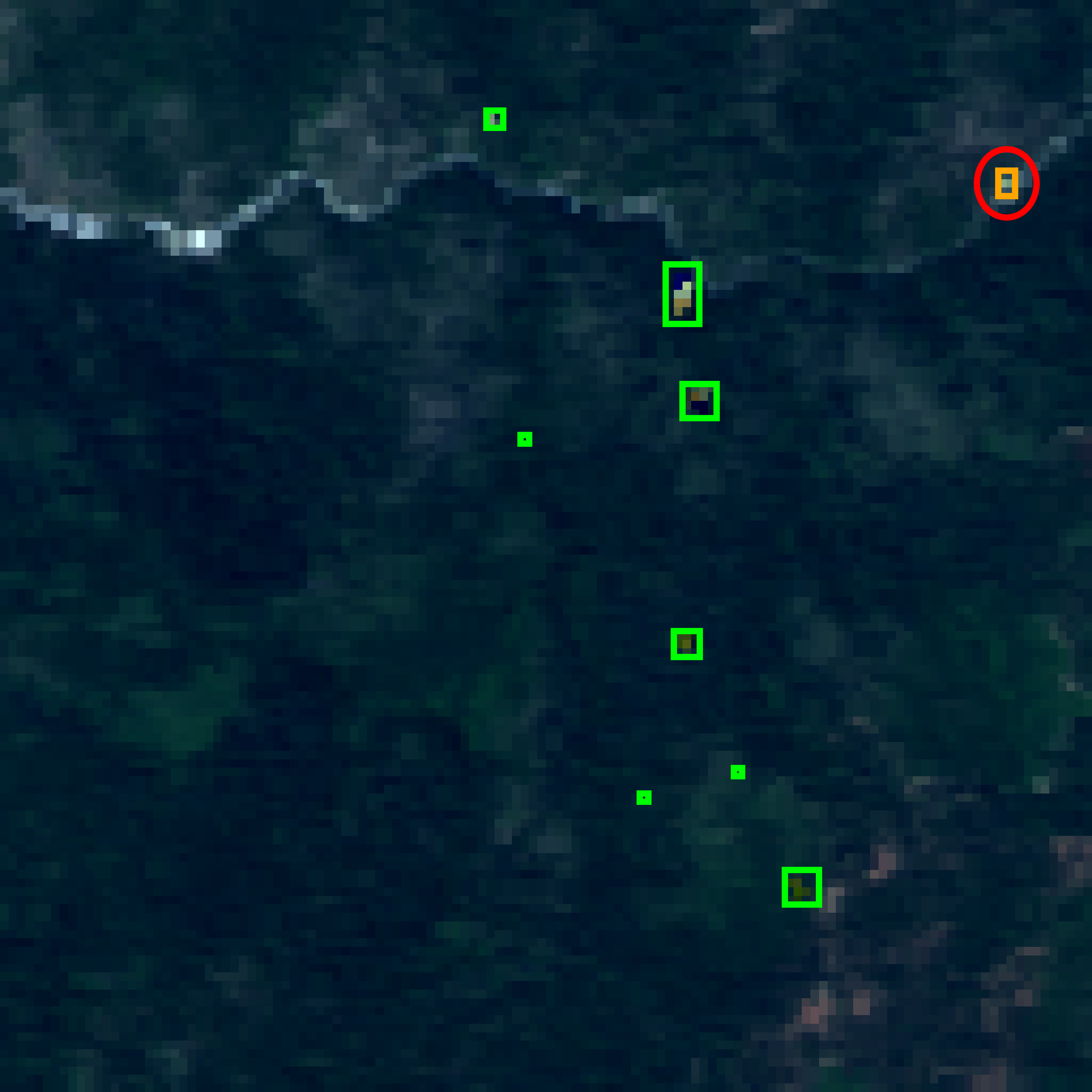}}
	\hfil
	\subfloat{\includegraphics[width=0.22\linewidth]{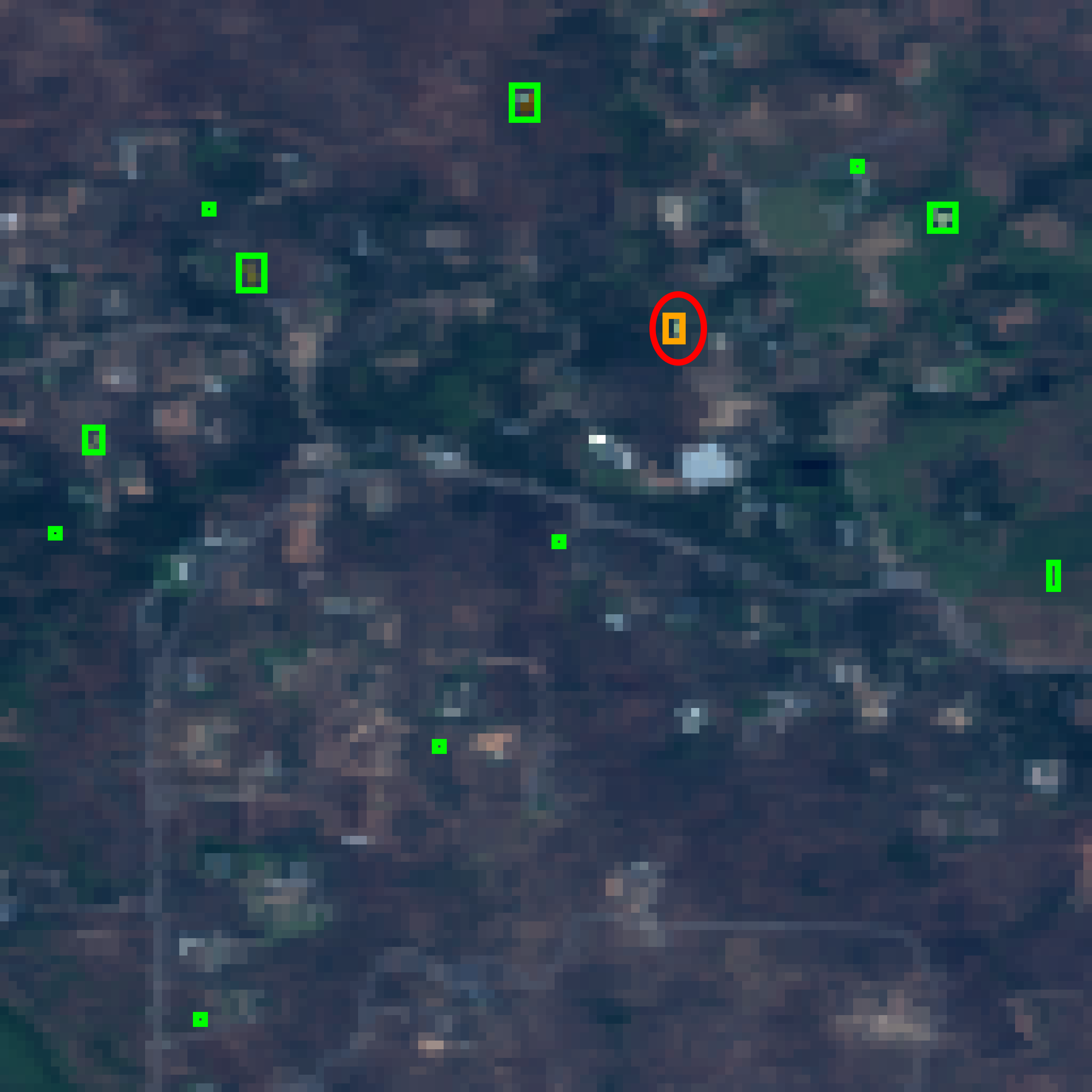}}
	\hfil
	\subfloat{\includegraphics[width=0.22\linewidth]{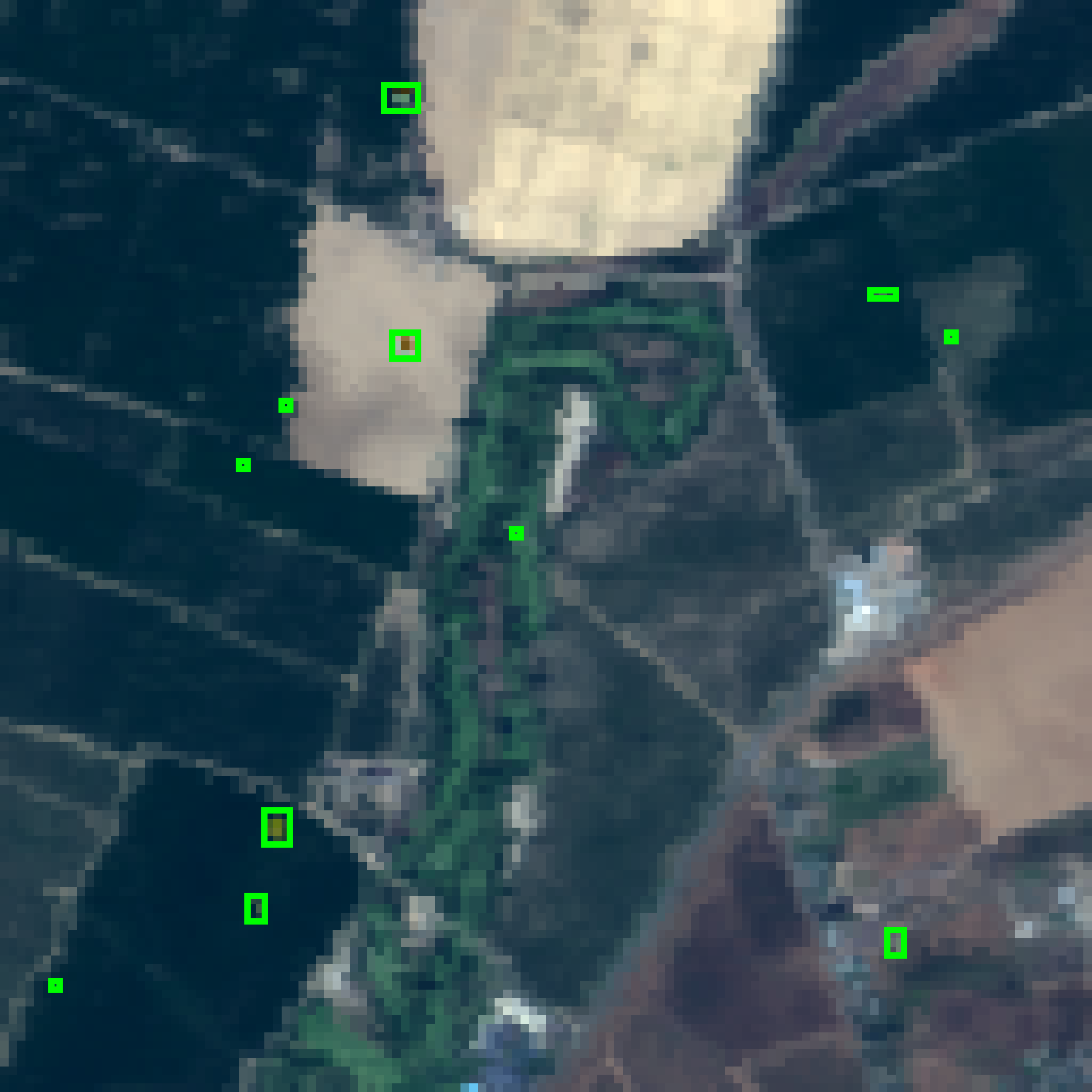}}
	\hfil
	\subfloat{\includegraphics[width=0.22\linewidth]{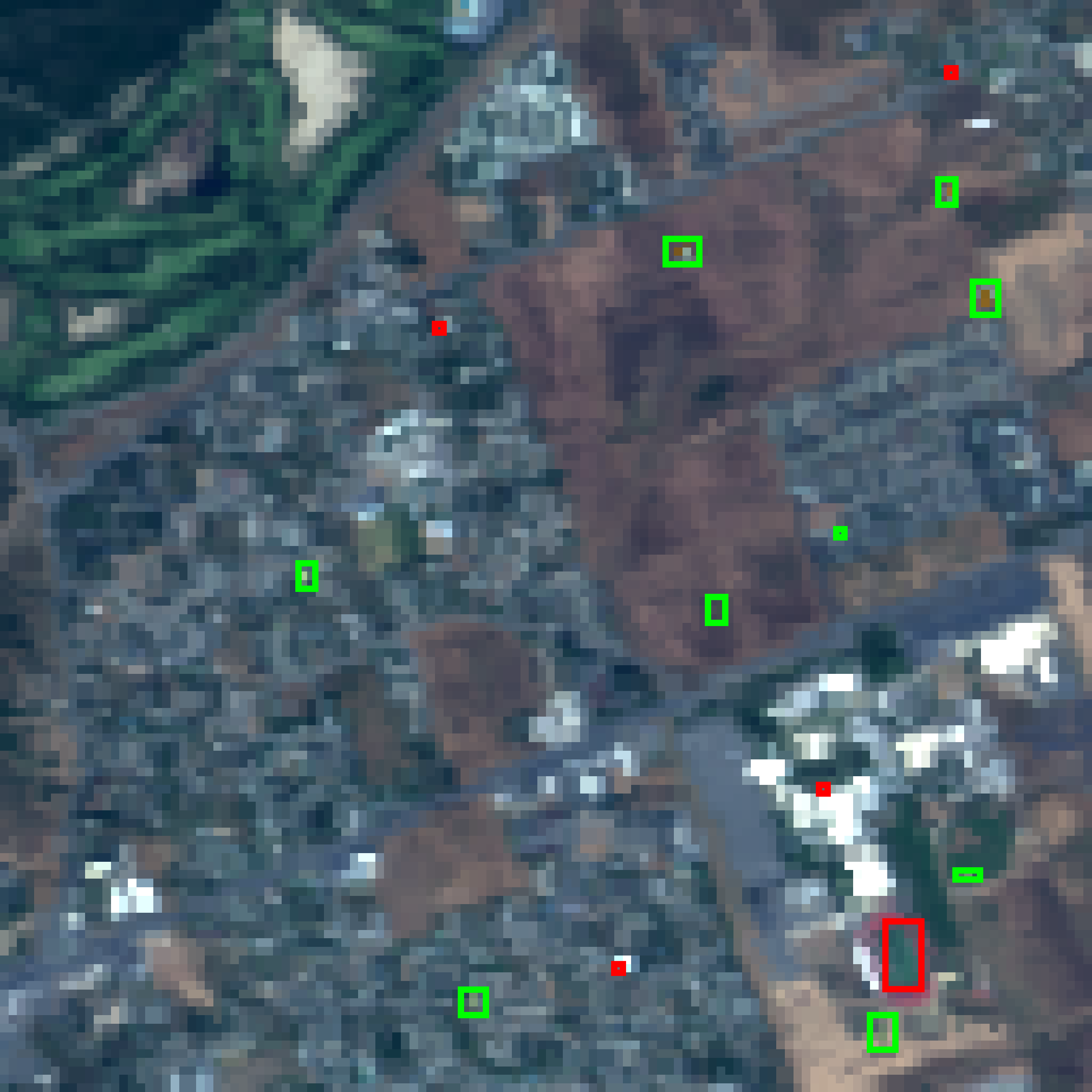}}
	\hfil
	\\
	\subfloat{\includegraphics[width=0.22\linewidth]{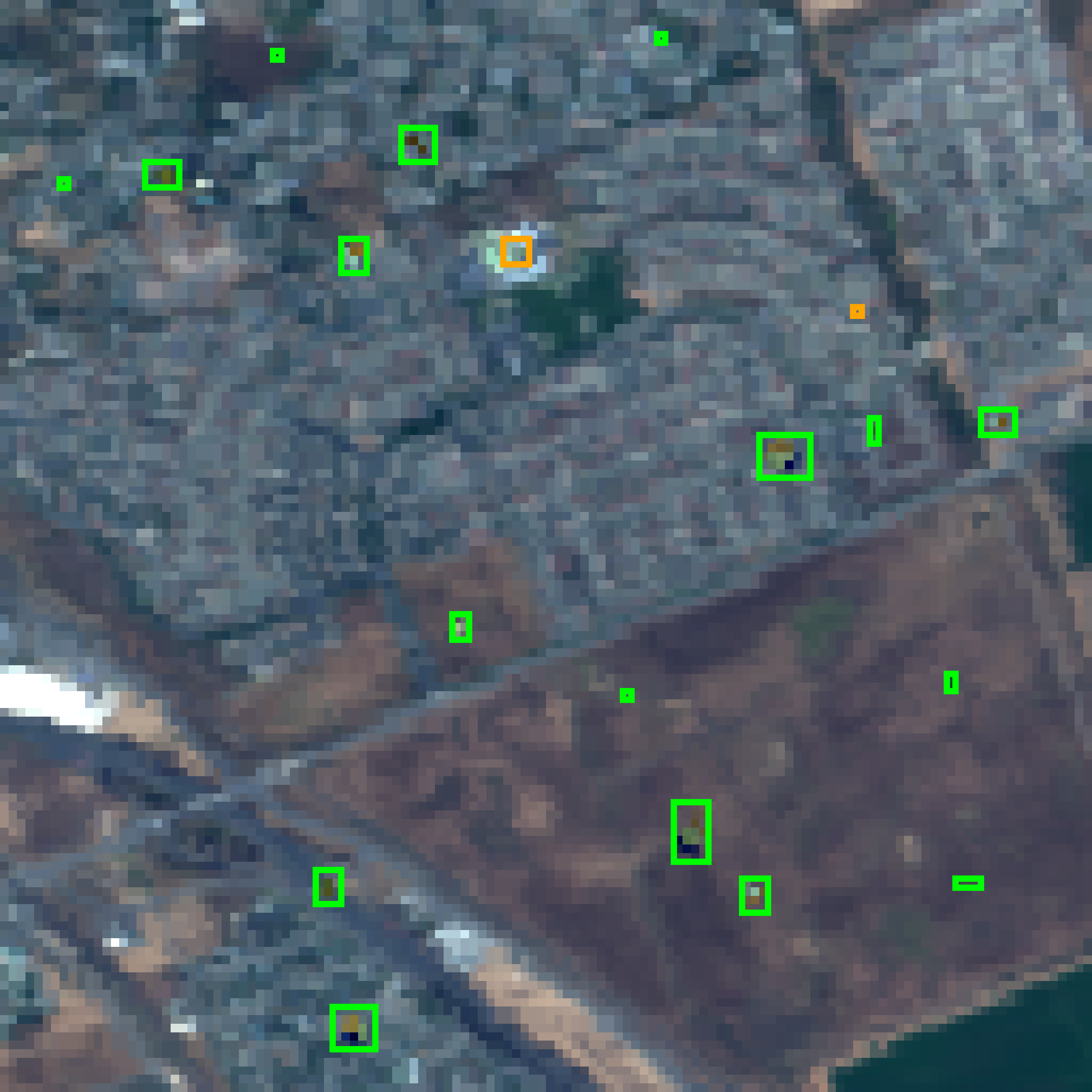}}
	\hfil
	\subfloat{\includegraphics[width=0.22\linewidth]{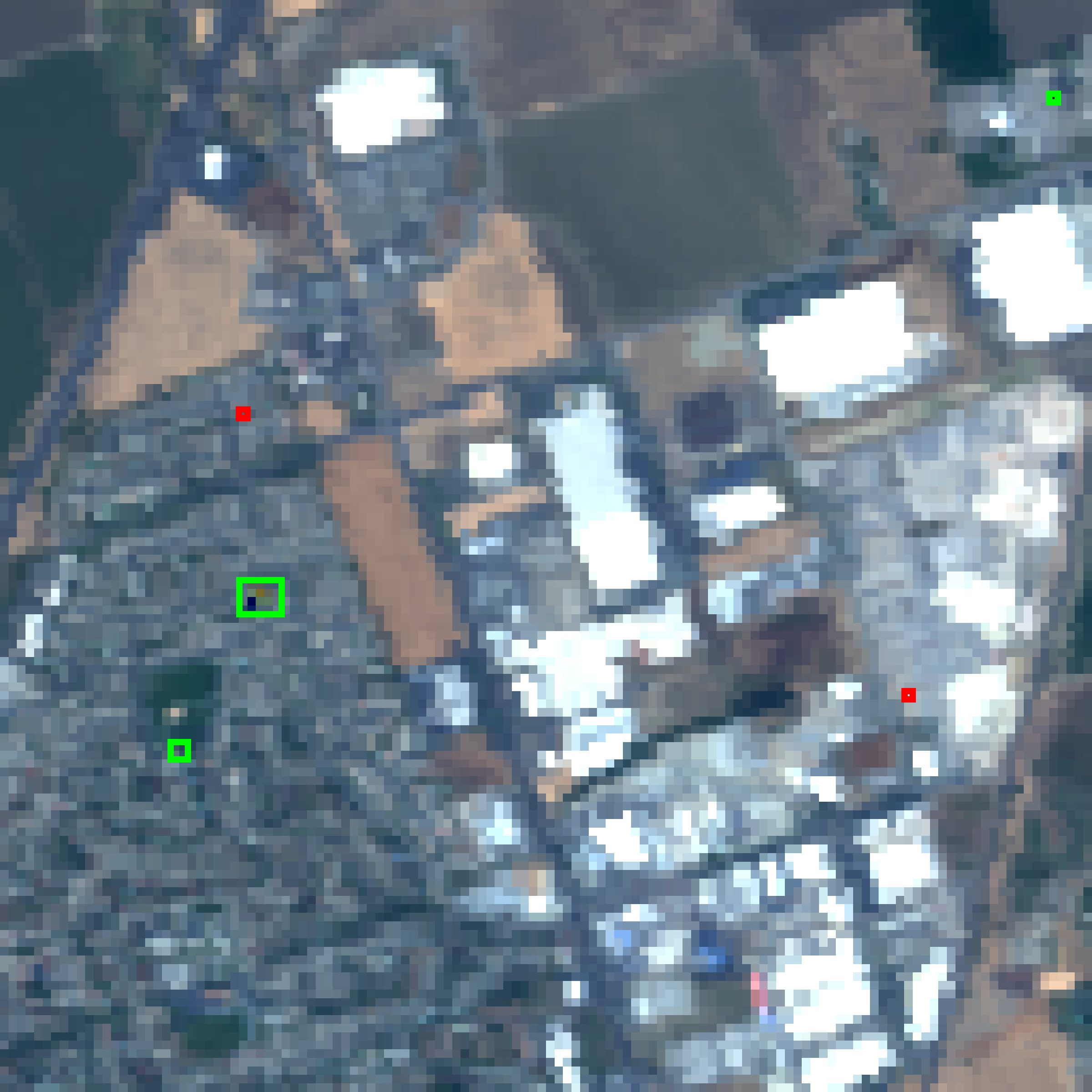}}
	\hfil
	\subfloat{\includegraphics[width=0.22\linewidth]{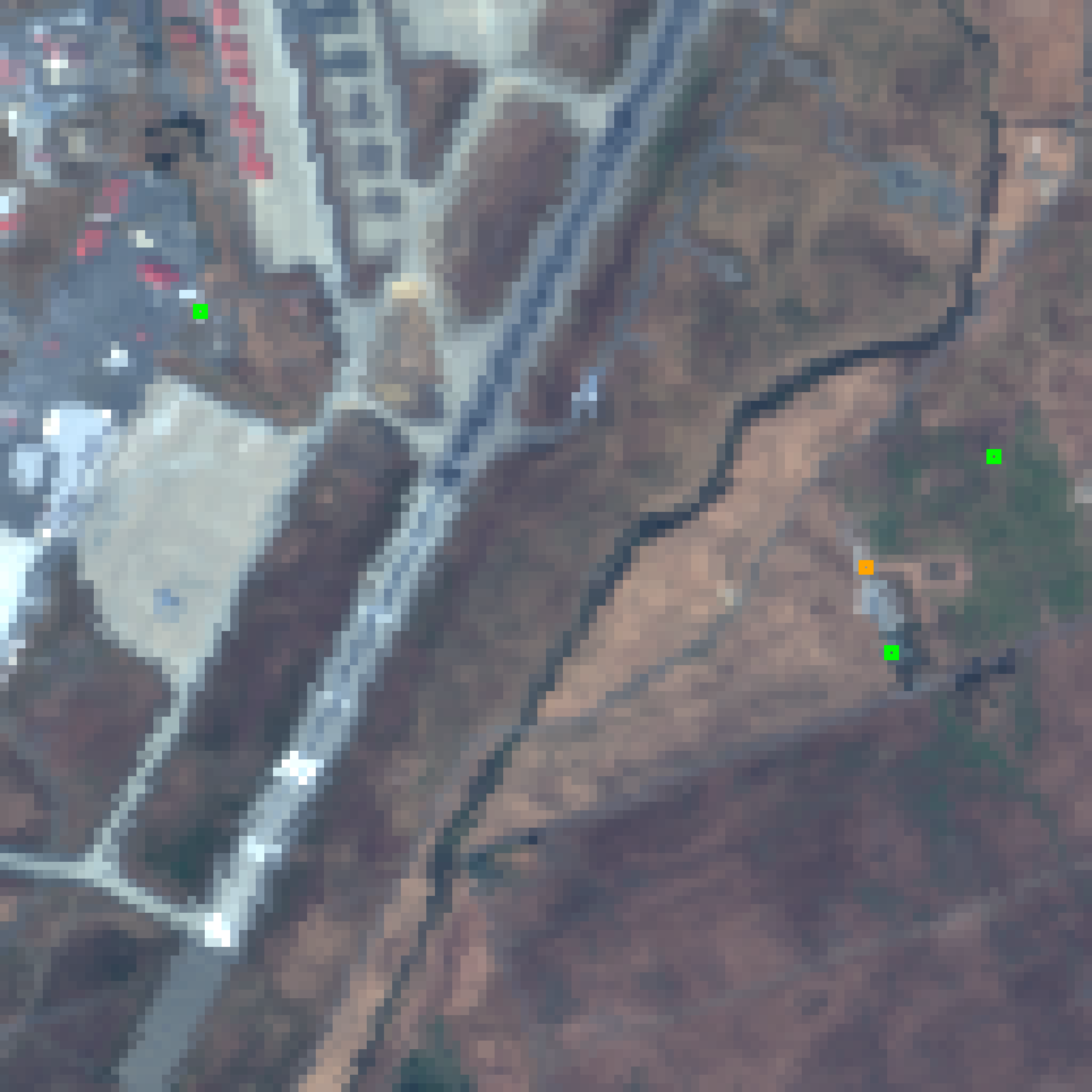}}
	\hfil
	\subfloat{\includegraphics[width=0.22\linewidth]{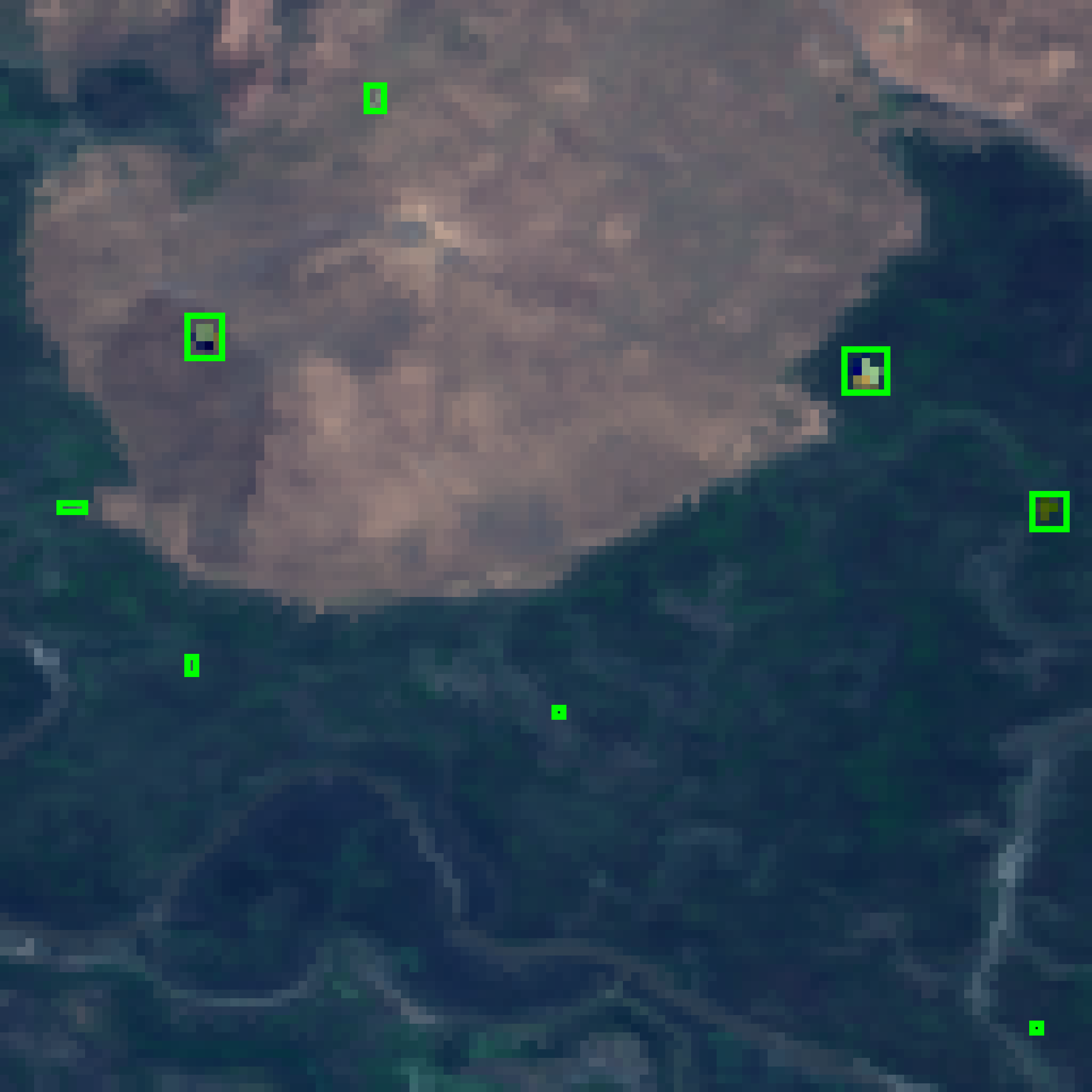}}
	\hfil
	\\
	\subfloat{\includegraphics[width=0.22\linewidth]{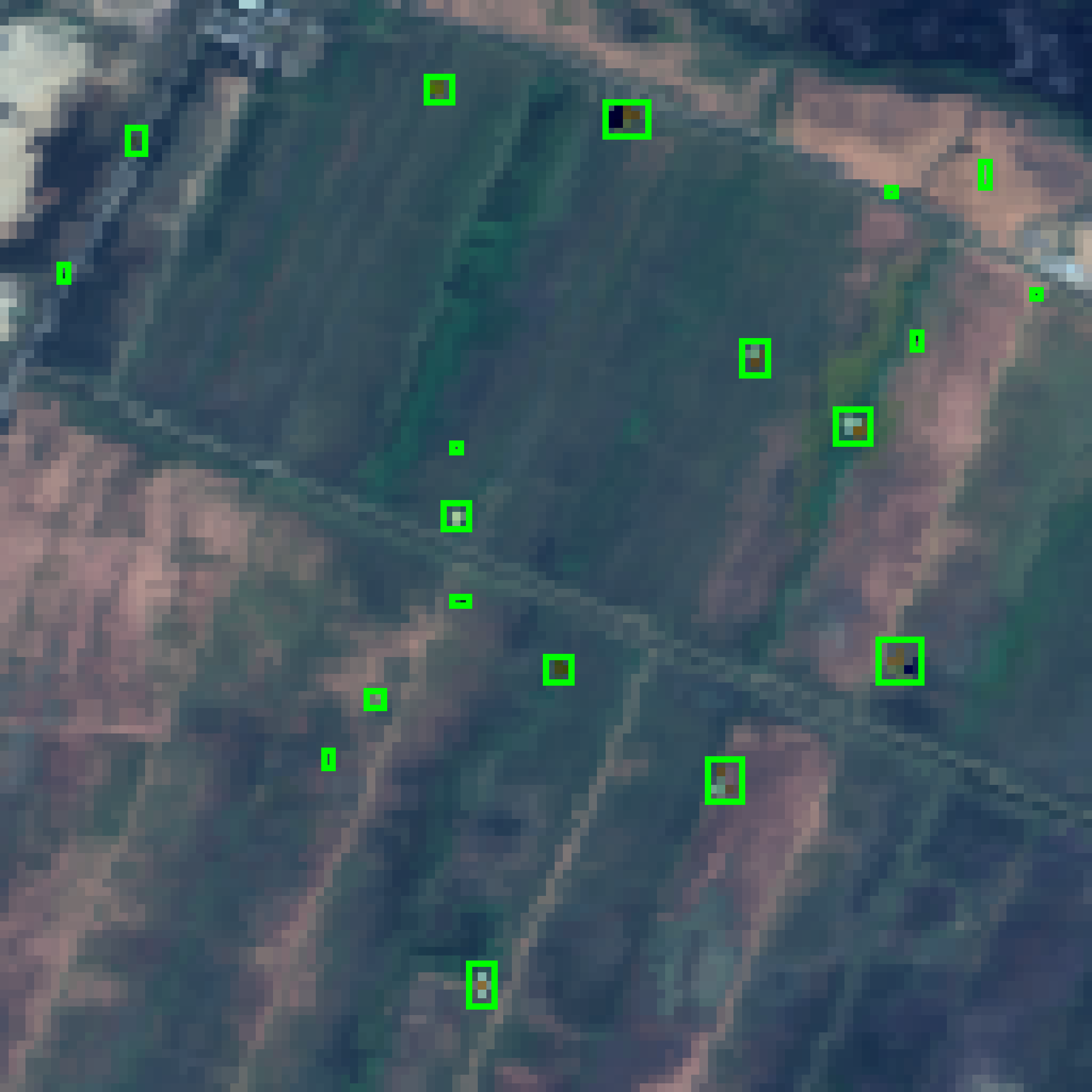}}
	\hfil
	\subfloat{\includegraphics[width=0.22\linewidth]{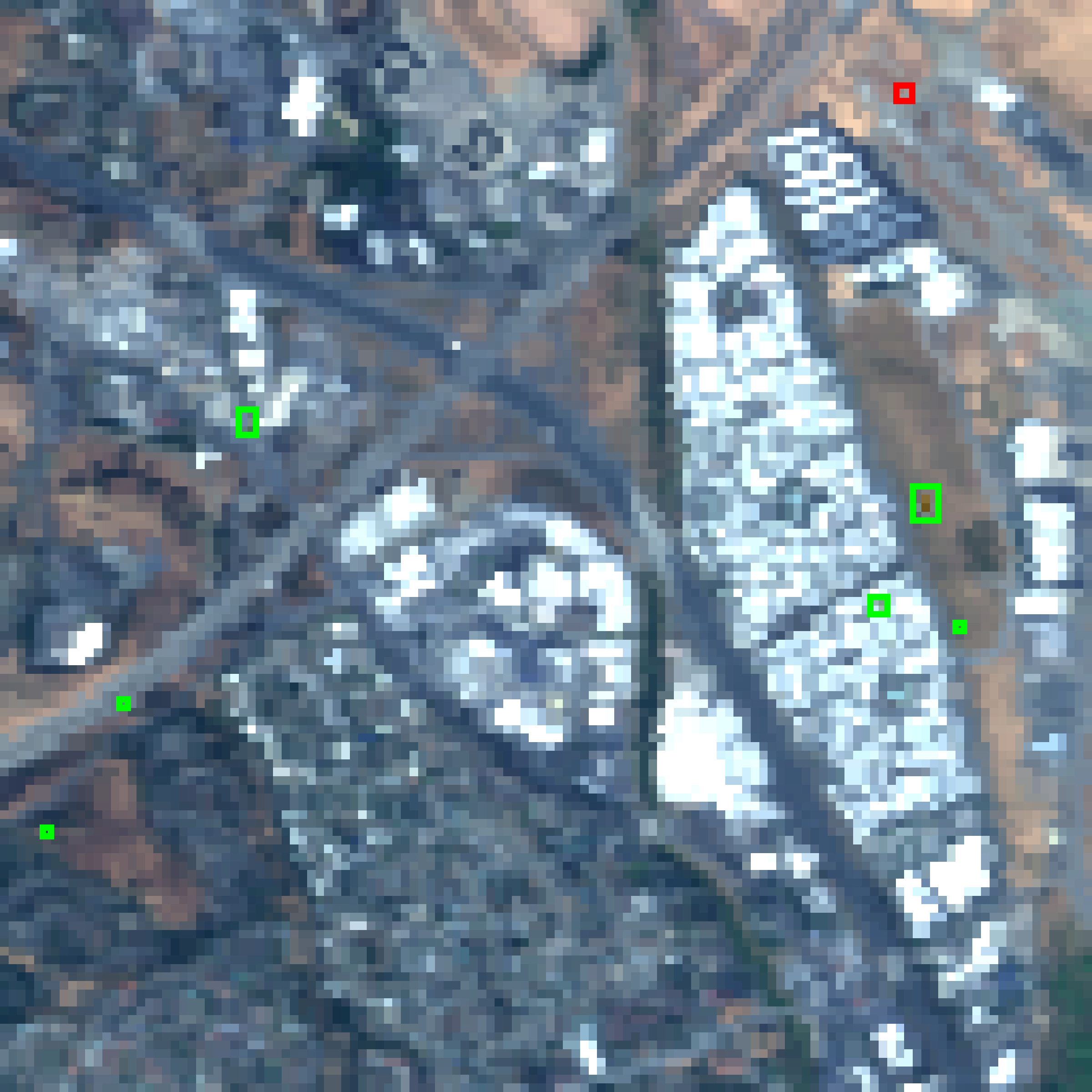}}
	\hfil
	\subfloat{\includegraphics[width=0.22\linewidth]{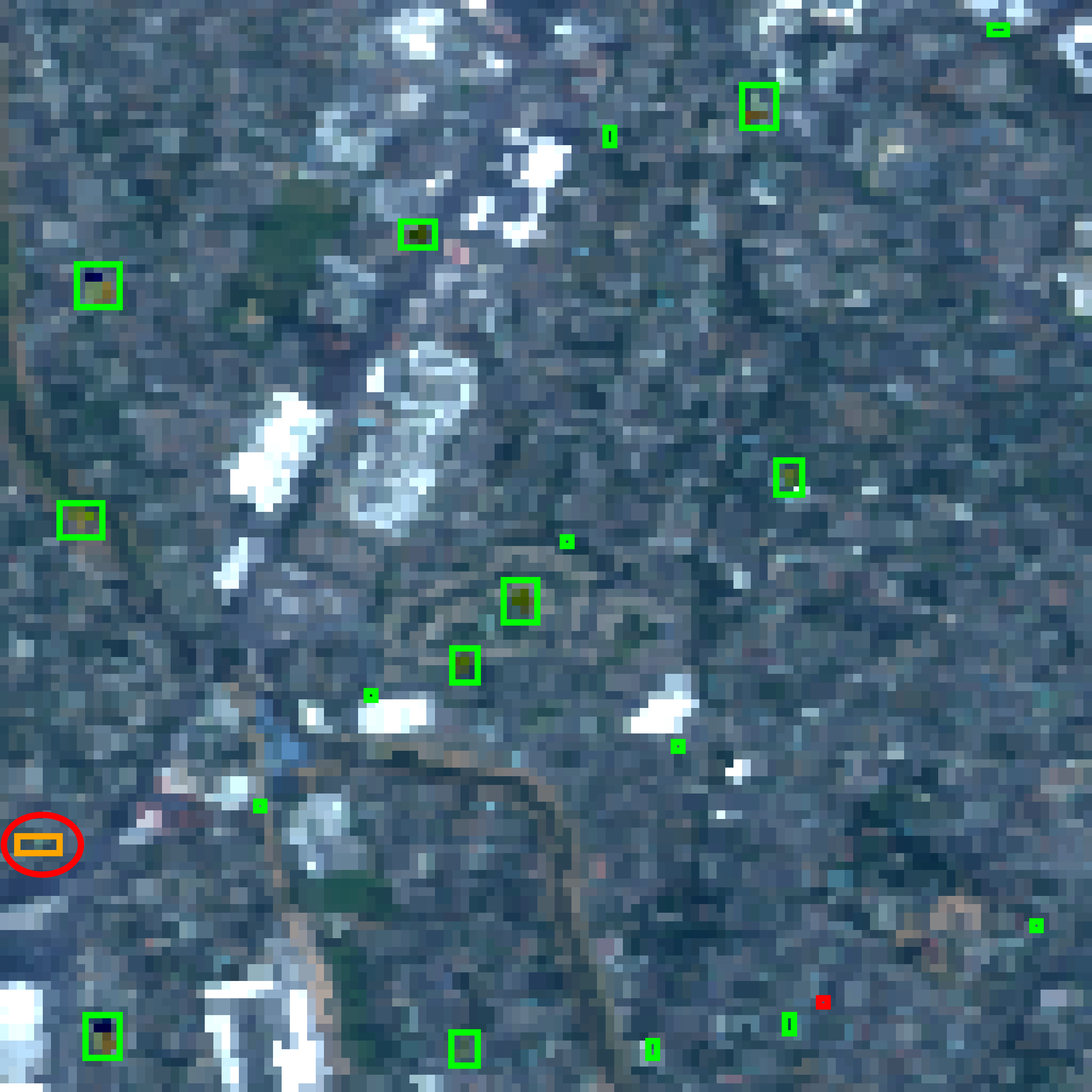}}
	\hfil
	\subfloat{\includegraphics[width=0.22\linewidth]{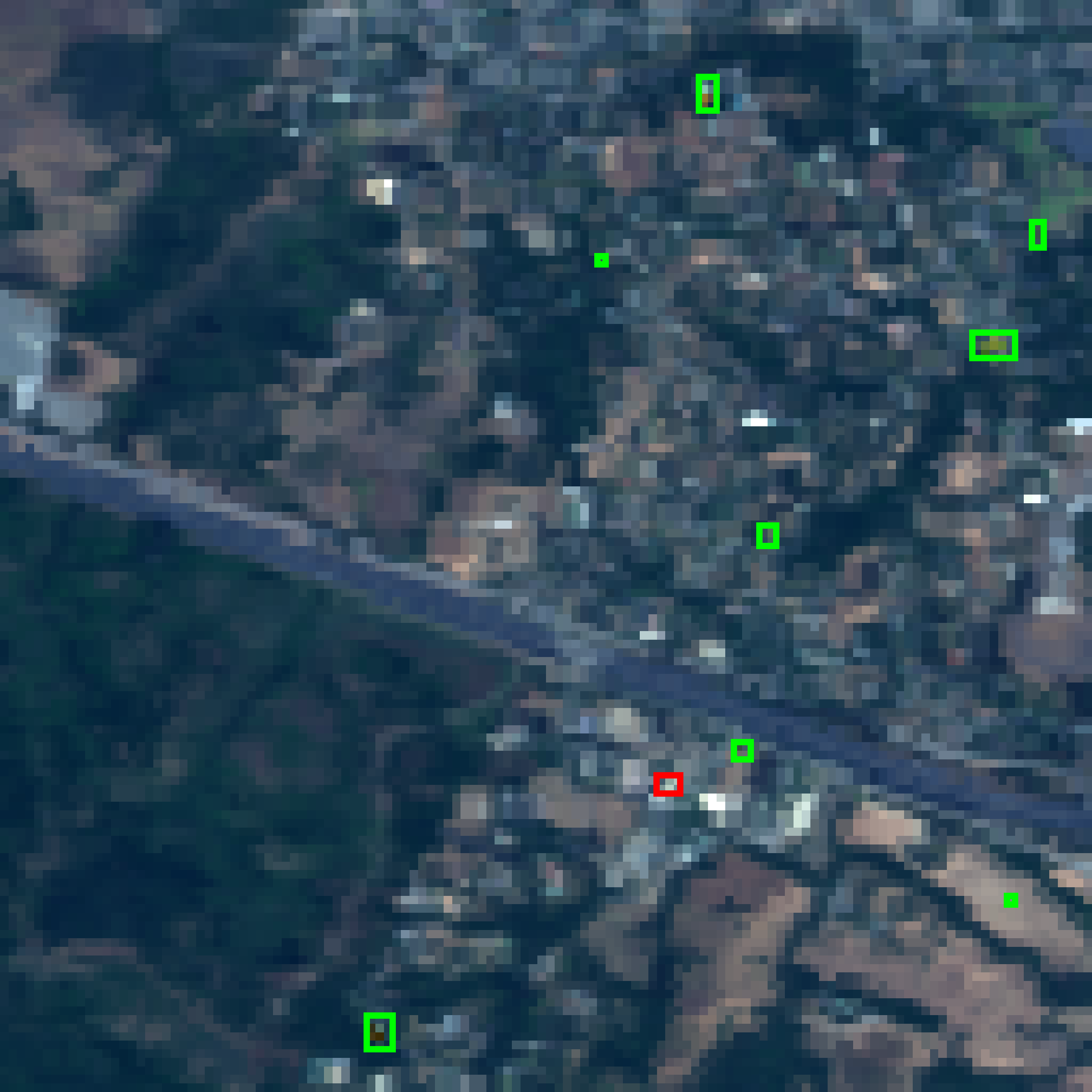}}
	\hfil
	\caption{Visualization Result Examples of SpecDETR on the SPOD dataset. 
		We filter out predictions with confidence scores below 0.2 and set the IoU threshold for GT matching to 0.25.
		Green boxes indicate true positive predictions are denoted, red boxes indicate false positive predictions, while yellow boxes or circles indicate undetected objects. Additionally, red circles represent cases where the localization is accurate, but the category prediction is incorrect, also classified as false negatives.}
	\label{fig:specdetr_spod}
\end{figure*}

Fig.~\ref{fig:specdetr_spod} presents some visualization results of SpecDETR. It is evident that SpecDETR demonstrates effective detection of objects that are challenging for human eyes to identify in the SPOD dataset. However, it is also shown that SpecDETR can accurately localize objects of C4 and C5 but sometimes provides incorrect class predictions, resulting in lower $\text{AP}_{\text{C4}}$ and $\text{AP}_{\text{C5}}$. Enhancing the network's ability to distinguish between hard-to-distinguish object categories remains a critical direction for future research.

Table~\ref{tab:spodnetresult_time} presents a quantitative evaluation of the computational efficiency performance of SpecDETR and compared visual object detection networks. The floating-point operations (FLOPs) are used to quantify the time complexity of the object detection networks, while  the parameters (Params) and memory consumption are employed to assess the spatial complexity. Here, we report the maximum GPU memory occupied during the inference of a single image\footnote{https://pytorch.org/docs/stable/generated/torch.cuda.max\_memory\_allocated\\.html}. Additionally, we provide the training time with 100 training epochs and the inference speed which is the reciprocal of the single-image test time for each network.
Owing to the removal of the backbone and the optimization of the decoder structure, SpecDETR has a relatively small number of Params, totaling only 16.1M, which is superior to the compared visual object detection networks. Although SpecDETR exhibits better performance in terms of spatial complexity, its performance in terms of time complexity is mediocre.
Although the input images of SpecDETR are not upsampled as in other networks, SpecDETR achieves 24.7 FPS and 139.7 GFLOPs, whereas Faster R-CNN with ResNet50 achieves 40.3 FPS and 68.8 GFLOPs. This is primarily due to two reasons: first, SpecDETR is based on a Transformer architecture, which is inherently at a disadvantage in terms of computational speed compared to pure CNN-based networks; second, SpecDETR does not downsample the feature maps during the feature extraction process.
Enhancing the computational efficiency of the proposed point object detection network could be a promising direction for future exploration.

\begin{table}[t]
	\caption{Performance comparison of SpecDETR and HTD methods on the Avon dataset.}
	\label{tab:avonresult}
	\begin{center}
		\renewcommand\arraystretch{1.0}
		\scriptsize
		\setlength{\tabcolsep}{0.5mm}{
			\begin{tabular}{c|cc|cccccc}
				\hline
				Method &mAUC$\uparrow$  & mIoU$\uparrow$  & mAP$\uparrow$   & $\text{mAP}_{\text{25}}$$\uparrow$  & mAR$\uparrow$    & $\text{mRe}_{\text{25}}$$\uparrow$  & $\text{AP}_{\text{BL}}$$\uparrow$ &$\text{AP}_{\text{BR}}$$\uparrow$ \\
				\hline
				ASD \cite{ASD99-TD}  & 0.961 & \textbf{0.654} & 0.324 & 0.853 & 0.538 & \textbf{1.000} & 0.312 & 0.336 \\
				CEM  \cite{CEM-TD} & 0.967 & 0.460 & 0.178 & 0.566 & 0.471 & \textbf{1.000} & 0.218 & 0.139 \\
				OSP \cite{OSP-TD}  & 0.967 & 0.457 & 0.164 & 0.526 & 0.471 & \textbf{1.000} & 0.202 & 0.126 \\
				KOSP \cite{KOSP} & 0.977 & 0.374 & 0.111 & 0.380 & 0.454 & 0.958 & 0.132 & 0.090 \\
				SMF \cite{SMF-TD}  & 0.957 & 0.155 & 0.047 & 0.240 & 0.350 & 0.958 & 0.002 & 0.092 \\
				KSMF \cite{kwon2006comparative} & 0.965 & 0.211 & 0.068 & 0.229 & 0.463 & 0.958 & 0.064 & 0.073 \\
				TCIMF \cite{TCIMF-TD} & \textbf{0.982} & 0.433 & 0.181 & 0.456 & 0.538 & \textbf{1.000} & 0.220 & 0.142 \\
				CR \cite{CR-AD}   & 0.806 & 0.313 & 0.013 & 0.551 & 0.046 & 0.875 & 0.023 & 0.002 \\
				KSR \cite{KSR-C}  & 0.944 & 0.269 & 0.003 & 0.329 & 0.008 & 0.750 & 0.004 & 0.001 \\
				KSRBBH \cite{KSRBBH-TD} & 0.868 & 0.495 & 0.181 & 0.578 & 0.483 & \textbf{1.000} & 0.148 & 0.213 \\
				LSSA \cite{zhu2023learning} & 0.923 & 0.388 & 0.097 & 0.407 & 0.292 & 0.833 & 0.187 & 0.008 \\
				IRN \cite{shen2023hyperspectral}  & 0.750 & 0.391 & 0.302 & 0.500 & 0.367 & 0.500 & 0.605 & 0.000 \\
				TSTTD \cite{jiao2023triplet} & 0.951 & 0.463 & 0.312 & 0.570 & 0.579 & \textbf{1.000} & 0.560 & 0.065 \\
				\hline
				SpecDETR  & -     & -    & \textbf{0.885} & \textbf{0.990} & \textbf{0.925} & \textbf{1.000} & \textbf{0.924} & \textbf{0.847} \\
				\hline
			\end{tabular}%
		}
	\end{center}
\end{table}

\begin{figure*}[ht]
	\centering
	\subfloat[ASD]{\includegraphics[width=0.14\linewidth]{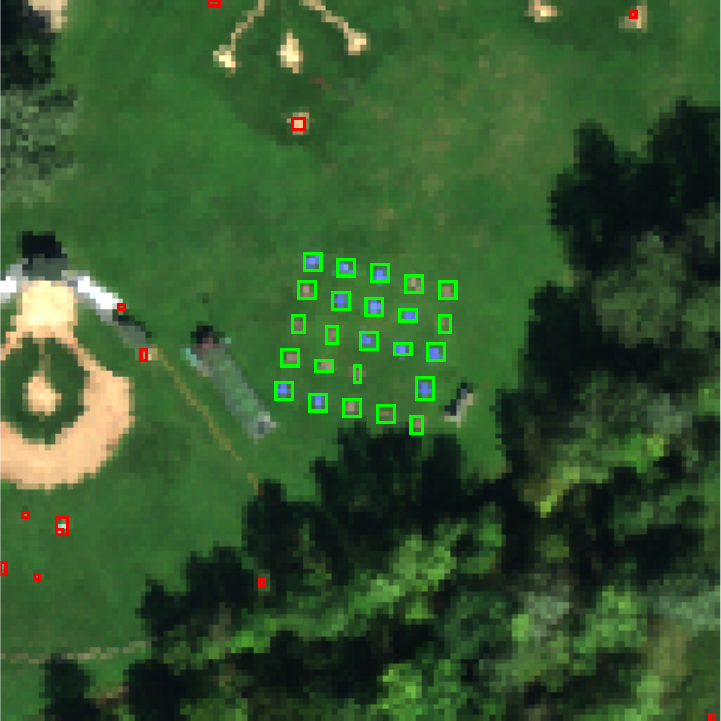}}
	\hfil
	\subfloat[CEM]{\includegraphics[width=0.14\linewidth]{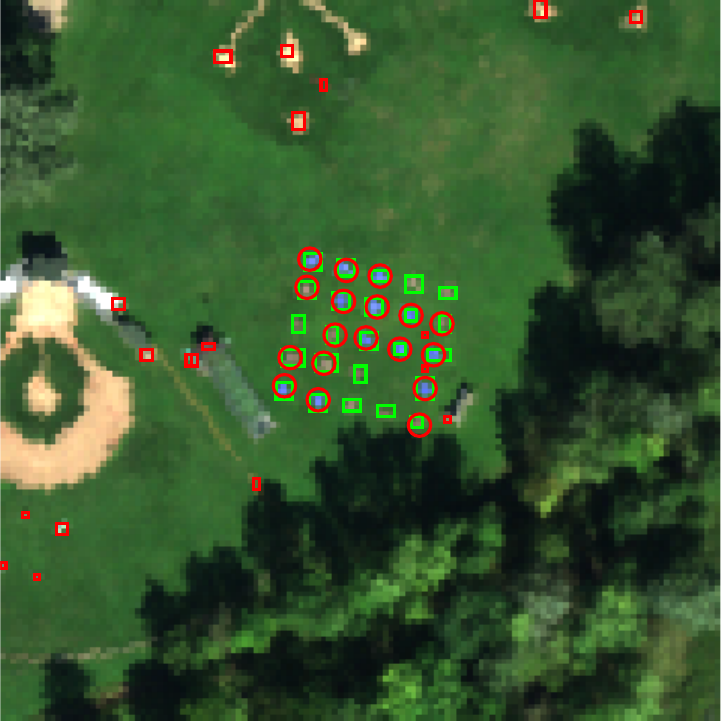}}
	\hfil
	\subfloat[OSP]{\includegraphics[width=0.14\linewidth]{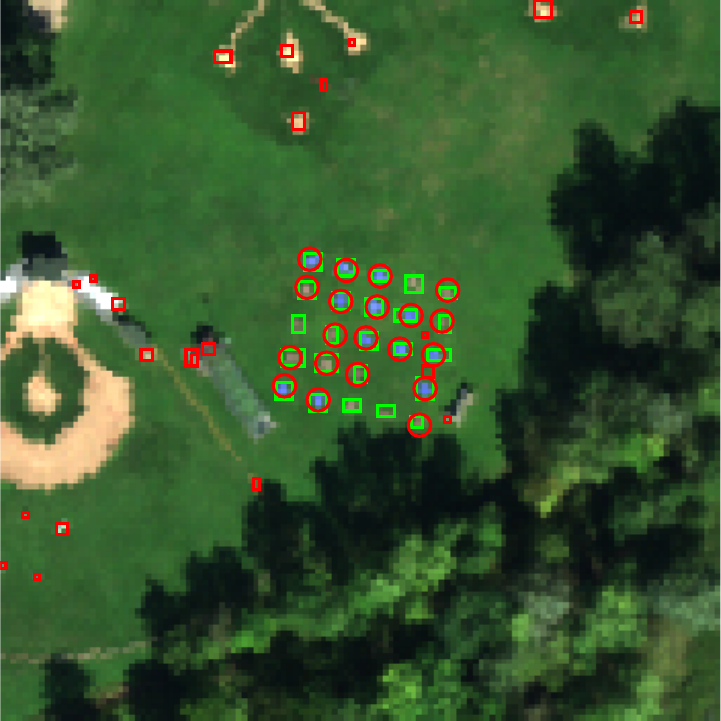}}
	\hfil
	\subfloat[KOSP]{\includegraphics[width=0.14\linewidth]{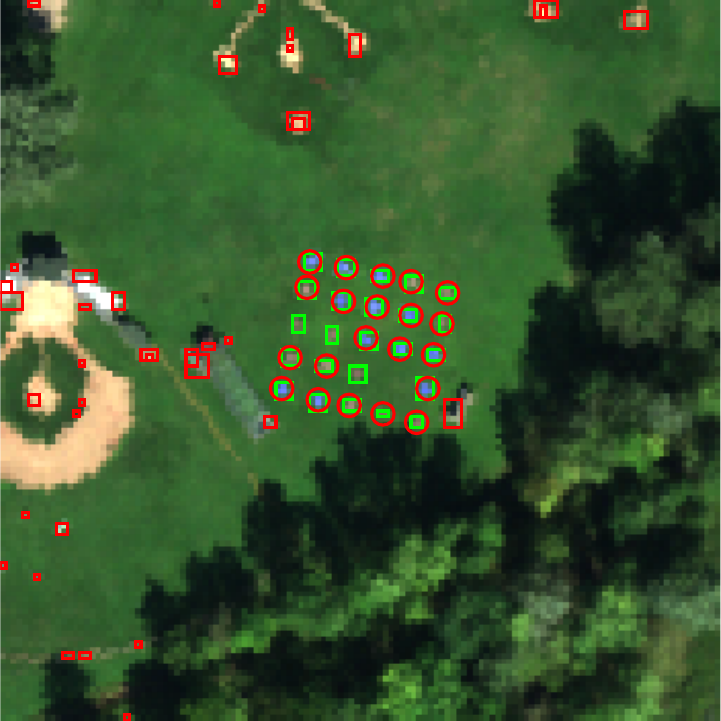}}
	\hfil
	\subfloat[SMF]{\includegraphics[width=0.14\linewidth]{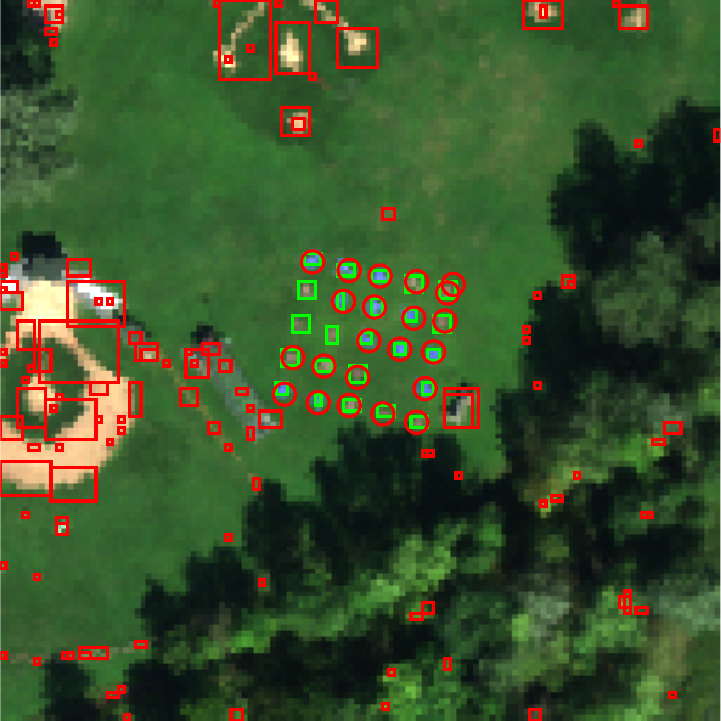}}
	\hfil
	\subfloat[KSMF]{\includegraphics[width=0.14\linewidth]{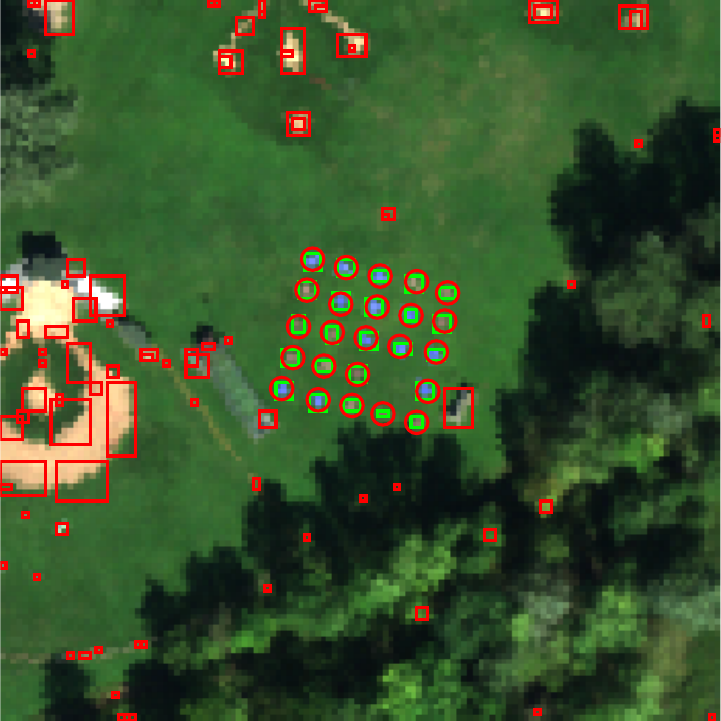}}
	\hfil
	\subfloat[TCIMF]{\includegraphics[width=0.14\linewidth]{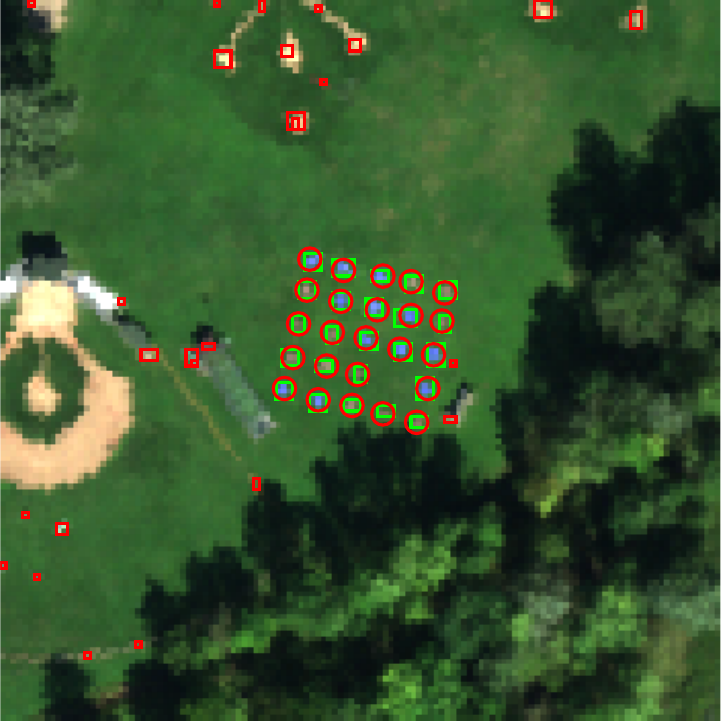}}
	\hfil
	\\
	\subfloat[CR]{\includegraphics[width=0.14\linewidth]{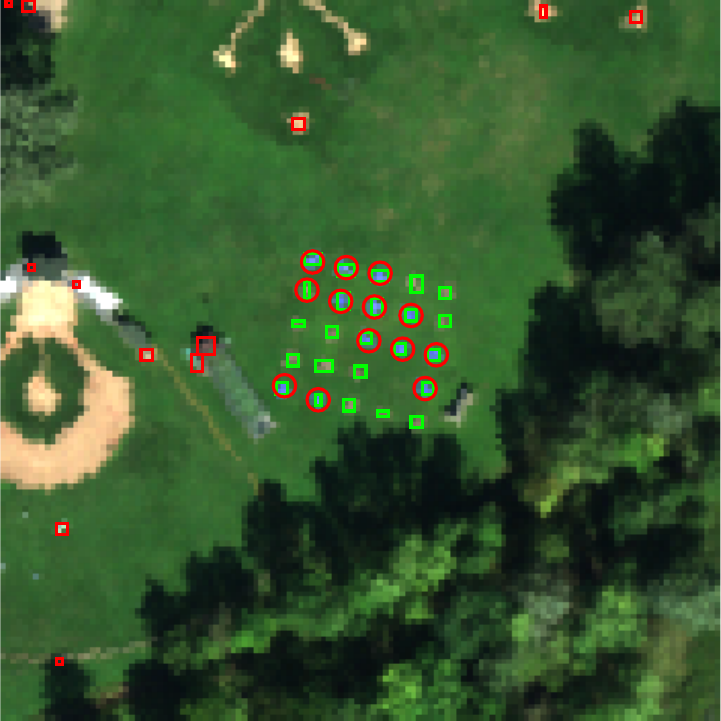}}
	\hfil
	\subfloat[KSR]{\includegraphics[width=0.14\linewidth]{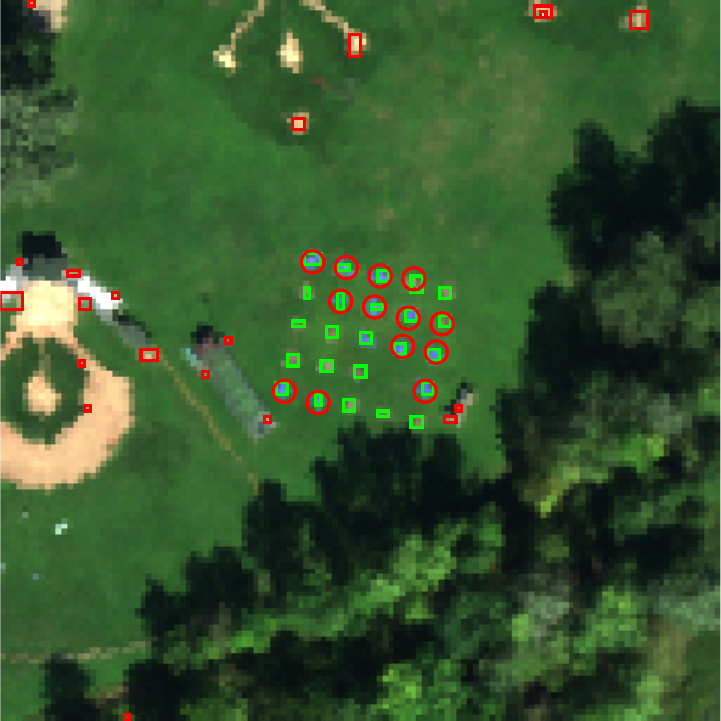}}
	\hfil
	\subfloat[KSRBBH]{\includegraphics[width=0.14\linewidth]{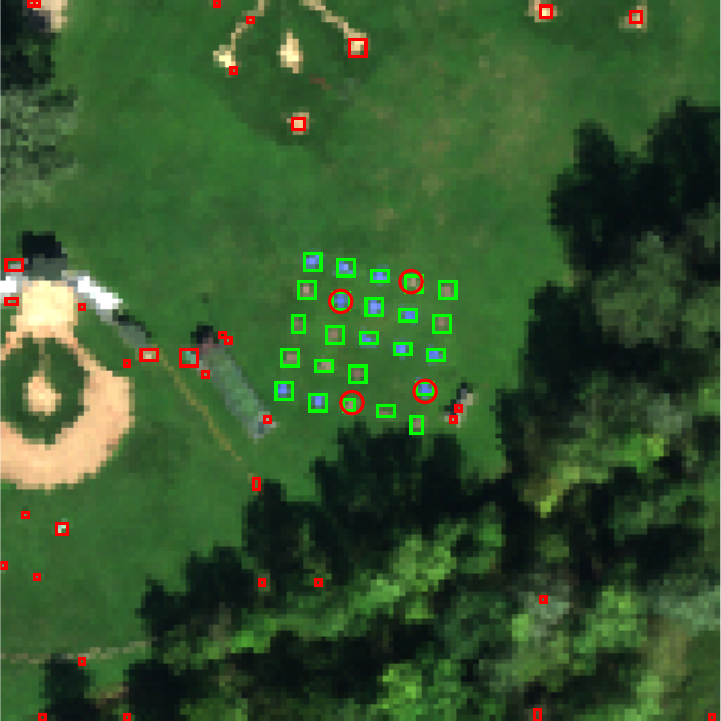}}
	\hfil
	\subfloat[LSSA]{\includegraphics[width=0.14\linewidth]{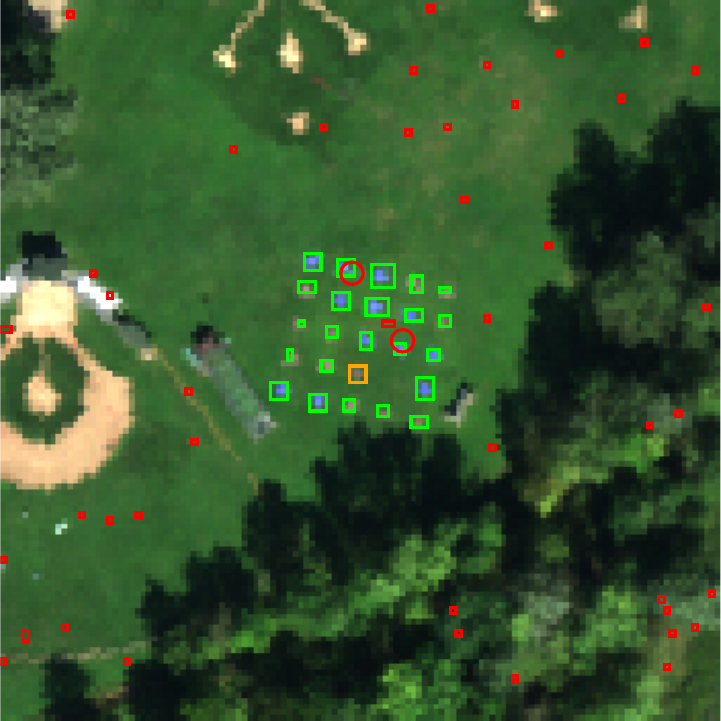}}
	\hfil
	\subfloat[IRN]{\includegraphics[width=0.14\linewidth]{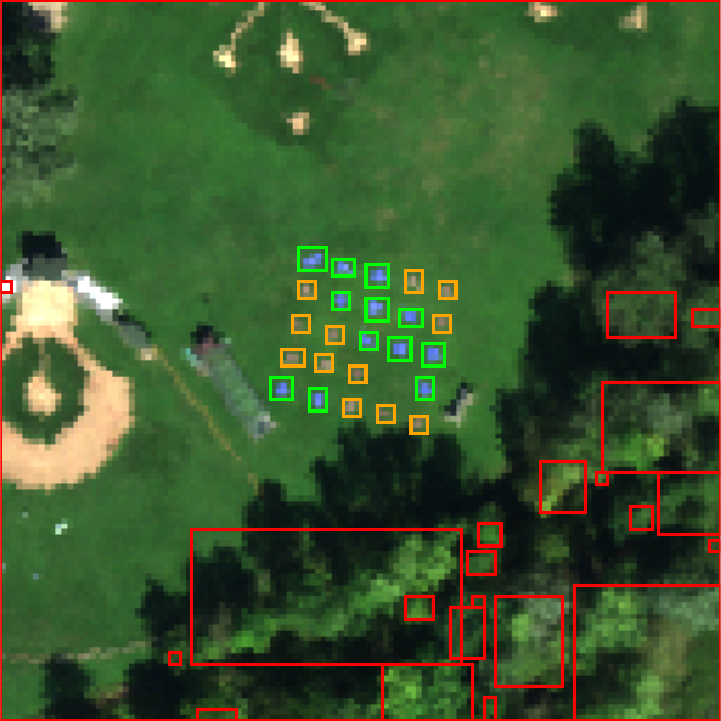}}
	\hfil
	\subfloat[TSTTD]{\includegraphics[width=0.14\linewidth]{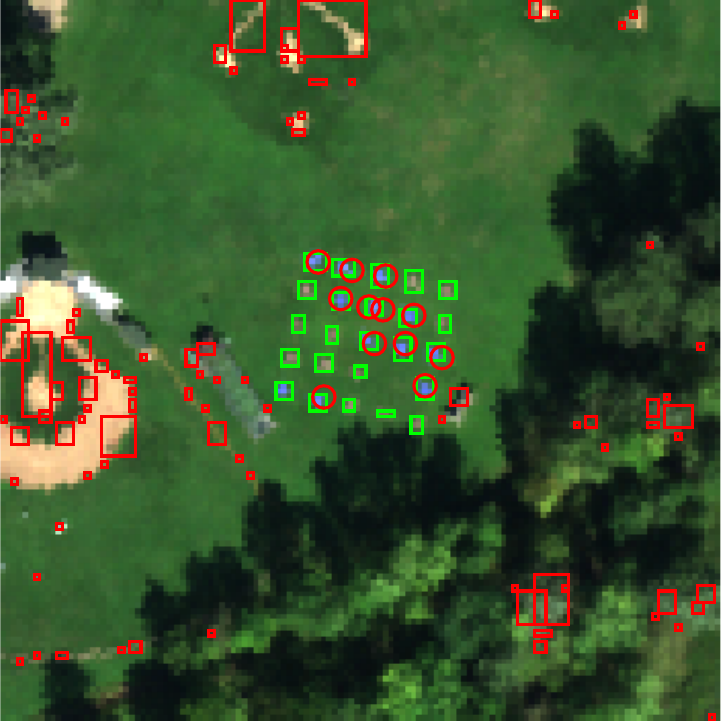}}
	\hfil
	\subfloat[SpecDETR]{\includegraphics[width=0.14\linewidth]{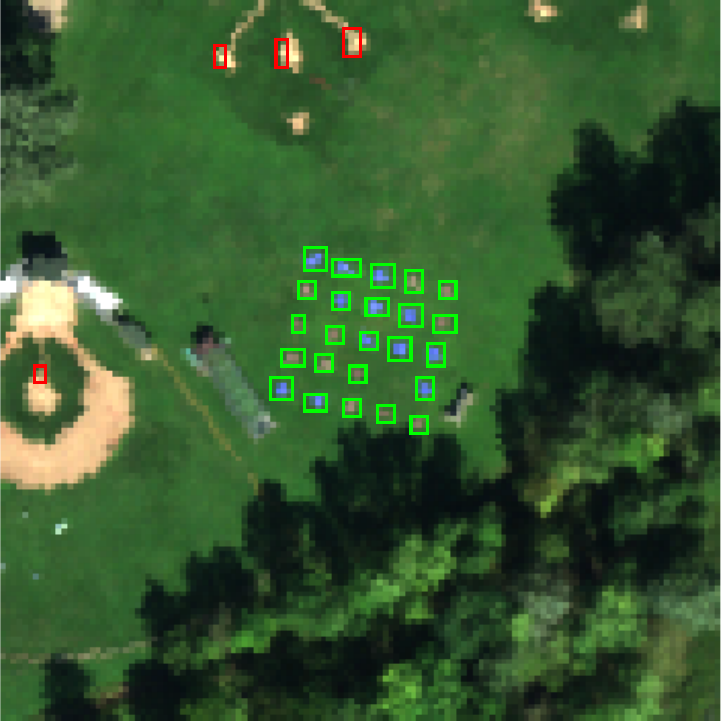}}
	\hfil
	\caption{Visualization results of SpecDETR and HTD methods on the Avon dataset. The confidence threshold is 0, the IoU threshold is 0.01, and the remaining visualization settings are consistent with Fig.~\ref{fig:specdetr_spod}.}
	\label{fig:avon_result}
\end{figure*}

\begin{table*}[t]
	\centering
	\caption{
		Performance comparison of SpecDETR and compared visual object detection networks on the SanDiego and Gulfport datasets.
		The best results are marked in red, and the second-best results are marked in blue.
		B., D.G., P.G., and F.V.G. represent brown, dark green, pea green, and faux vineyard green cloth panels, respectively.
	}	
	\renewcommand\arraystretch{1} 
	\scriptsize  
	\setlength{\tabcolsep}{1mm}{   
		\begin{tabular}{c|cc|cc|cc|cc|cc|cc}
			\hline
			& \multicolumn{2}{c|}{\multirow{2}[0]{*}{SanDiego}} & \multicolumn{8}{c}{Gulfport}                            &       &  \\
			\cline{4-13} 
			Method & \multicolumn{2}{c}{} & \multicolumn{2}{|c}{Mean} & \multicolumn{2}{c}{B.} & \multicolumn{2}{c}{D.G.} & \multicolumn{2}{c}{P.G.} & \multicolumn{2}{c}{F.V.G.} \\
			\cline{2-13} 
			& AP    & Re    & AP    & Re    & AP    & Re    & AP    & Re    & AP    & Re    & AP    & Re \\
			\hline
			CentripetalNet & \textcolor[rgb]{ 0,  0,  1}{0.776} & \textcolor[rgb]{ 1,  0,  0}{1.000} & \textcolor[rgb]{ 0,  0,  1}{0.178} & 0.213 & 0.208 & 0.200 & \textcolor[rgb]{ 0,  0,  1}{0.192} & \textcolor[rgb]{ 0,  0,  1}{0.267} & 0.139 & 0.133 & \textcolor[rgb]{ 1,  0,  0}{0.171} & 0.250 \\
			CornerNet & 0.750 & \textcolor[rgb]{ 1,  0,  0}{1.000} & 0.160 & 0.271 & 0.203 & 0.267 & 0.069 & 0.133 & \textcolor[rgb]{ 1,  0,  0}{0.242} & 0.267 & 0.125 & \textcolor[rgb]{ 0,  0,  1}{0.417} \\
			DINO  & 0.210 & \textcolor[rgb]{ 1,  0,  0}{1.000} & 0.174 & \textcolor[rgb]{ 1,  0,  0}{0.408} & \textcolor[rgb]{ 0,  0,  1}{0.220} & \textcolor[rgb]{ 1,  0,  0}{0.333} & \textcolor[rgb]{ 1,  0,  0}{0.232} & \textcolor[rgb]{ 1,  0,  0}{0.400} & 0.166 & \textcolor[rgb]{ 1,  0,  0}{0.400} & 0.077 & \textcolor[rgb]{ 1,  0,  0}{0.500} \\
			SpecDETR & \textcolor[rgb]{ 1,  0,  0}{1.000} & \textcolor[rgb]{ 1,  0,  0}{1.000} & \textcolor[rgb]{ 1,  0,  0}{0.213} & \textcolor[rgb]{ 0,  0,  1}{0.321} & \textcolor[rgb]{ 1,  0,  0}{0.277} & \textcolor[rgb]{ 1,  0,  0}{0.333} & 0.191 & 0.200 & \textcolor[rgb]{ 0,  0,  1}{0.218} & \textcolor[rgb]{ 0,  0,  1}{0.333} & \textcolor[rgb]{ 0,  0,  1}{0.167} & \textcolor[rgb]{ 0,  0,  1}{0.417} \\
			\hline
		\end{tabular}%
	}
	\label{tab:SanDiegoandGulfport}%
\end{table*}%

\begin{figure*}[t]
	\centering
	\subfloat[CentripetalNet]{\includegraphics[width=0.23\linewidth]{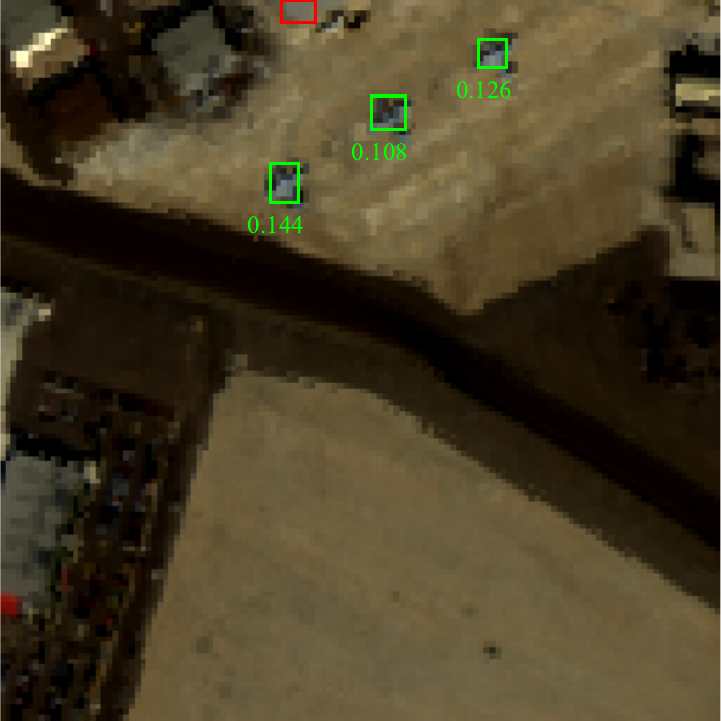}}
	\hfil
	\subfloat[Cornernet]{\includegraphics[width=0.23\linewidth]{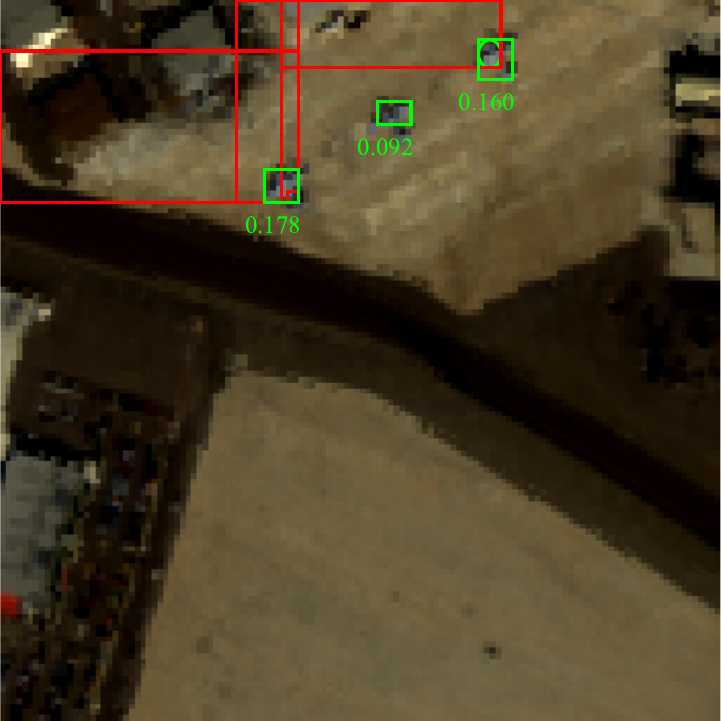}}
	\hfil
	\subfloat[Dino]{\includegraphics[width=0.23\linewidth]{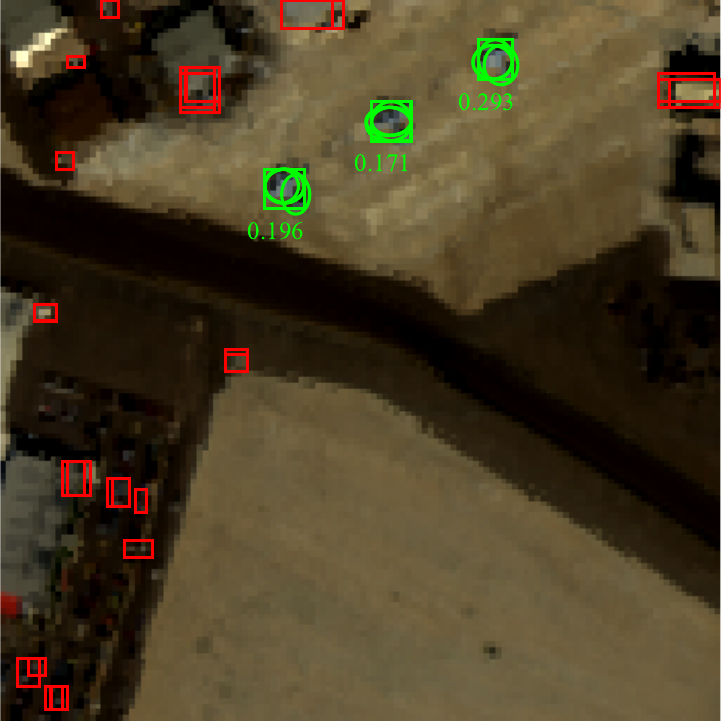}}
	\hfil
	\subfloat[Specdetr]{\includegraphics[width=0.23\linewidth]{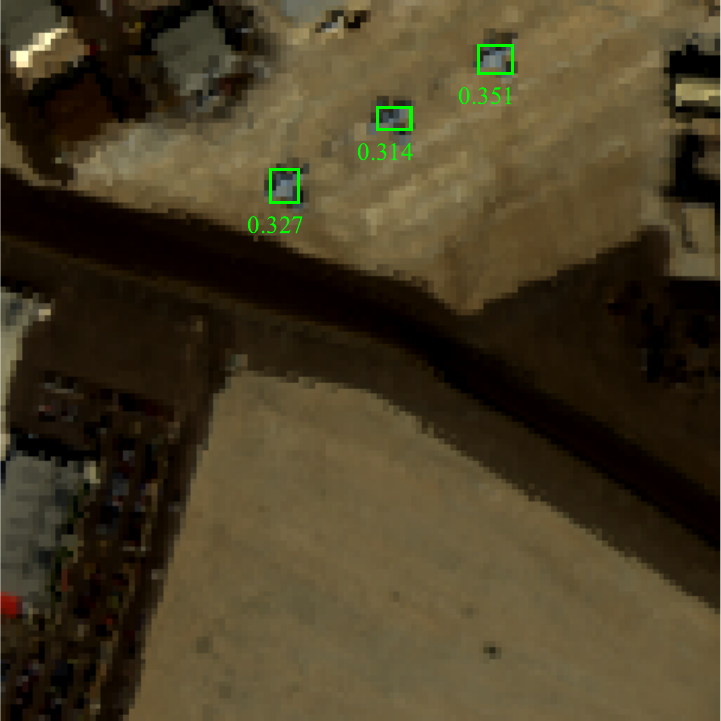}}
	\hfil
	\caption{Visualization results of SpecDETR and compared visual object detection networks  on the SanDiego dataset.
		The confidence threshold is 0.09, object confidence scores are shown, and the remaining visualization settings are consistent with Fig.~\ref{fig:specdetr_spod}.
	}
	\label{fig:SanDiego_detect}
\end{figure*}	

\begin{figure*}[t]
	\centering
	\subfloat[]{\includegraphics[width=0.18\linewidth]{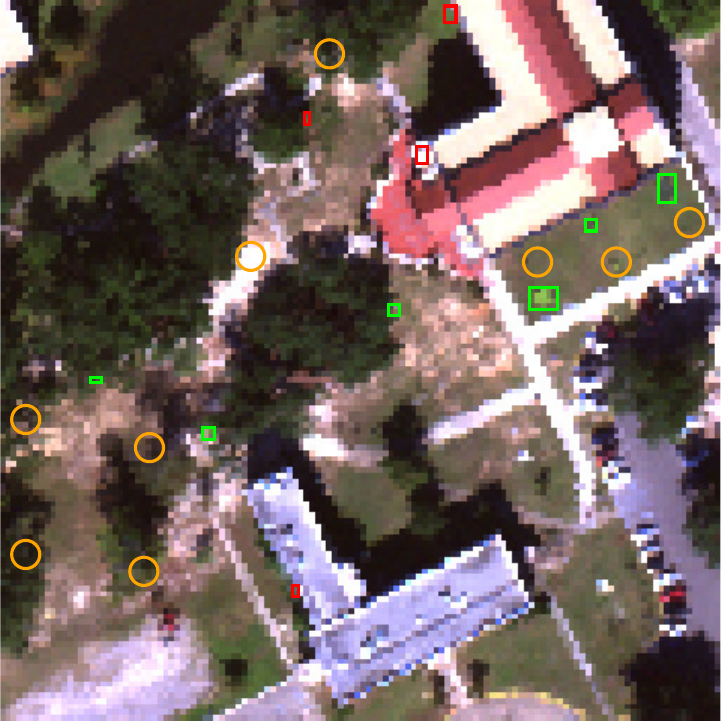}}
	\hfil
	\subfloat[]{\includegraphics[width=0.18\linewidth]{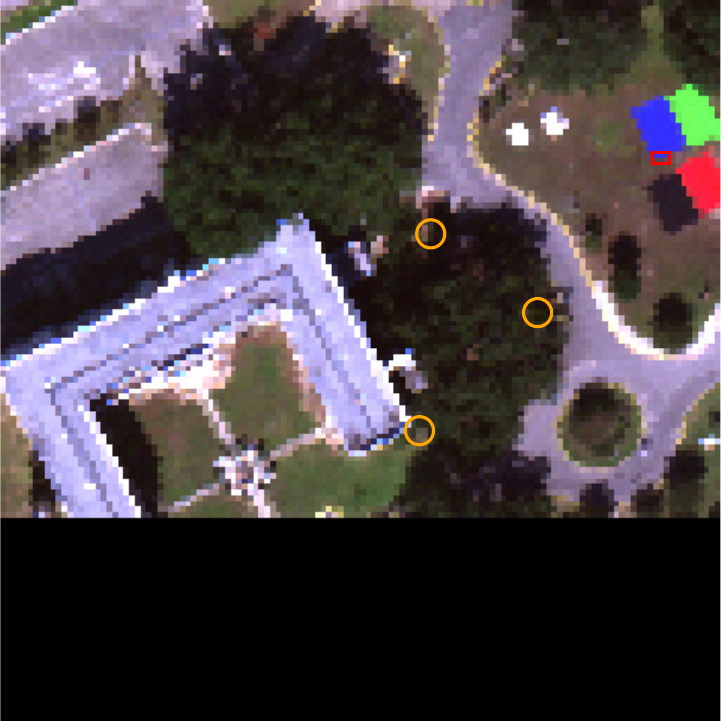}}
	\hfil
	\subfloat[]{\includegraphics[width=0.18\linewidth]{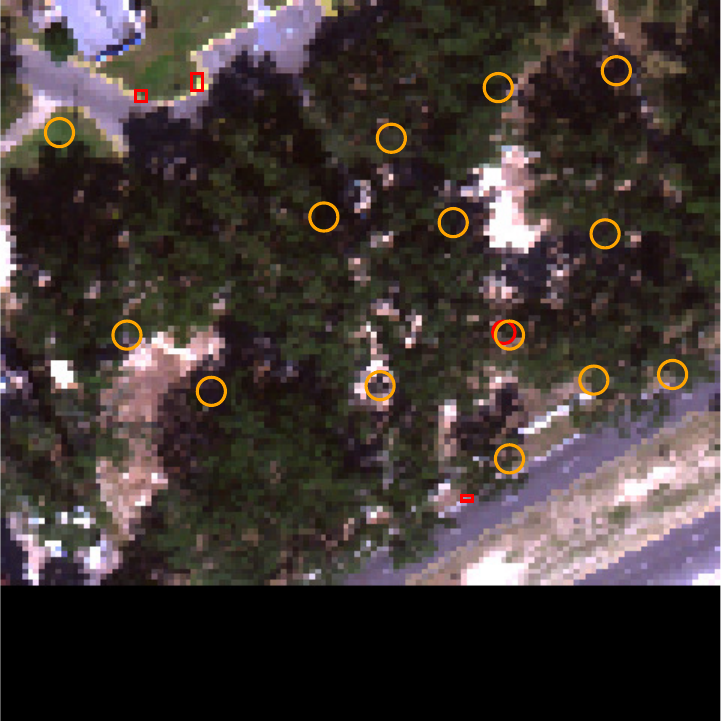}}
	\hfil
	\subfloat[]{\includegraphics[width=0.18\linewidth]{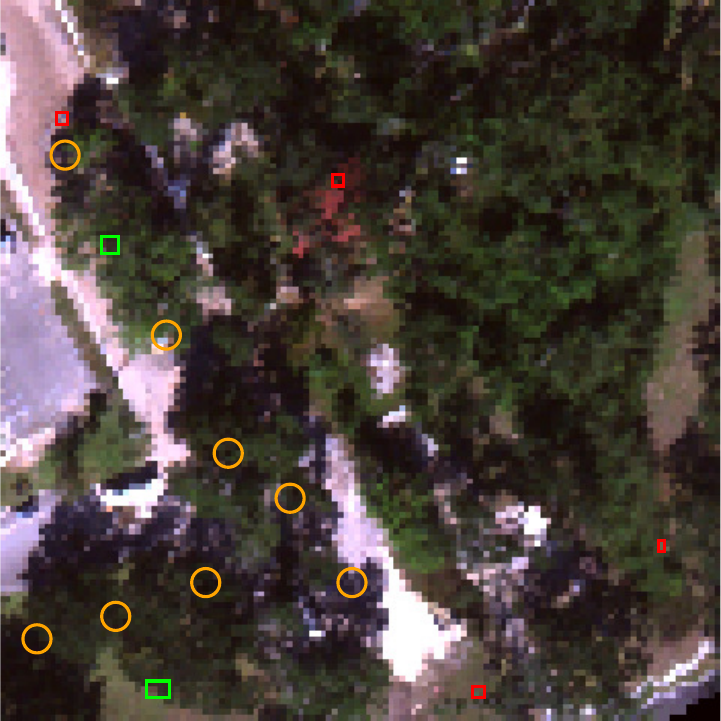}}
	\hfil
	\subfloat[]{\includegraphics[width=0.18\linewidth]{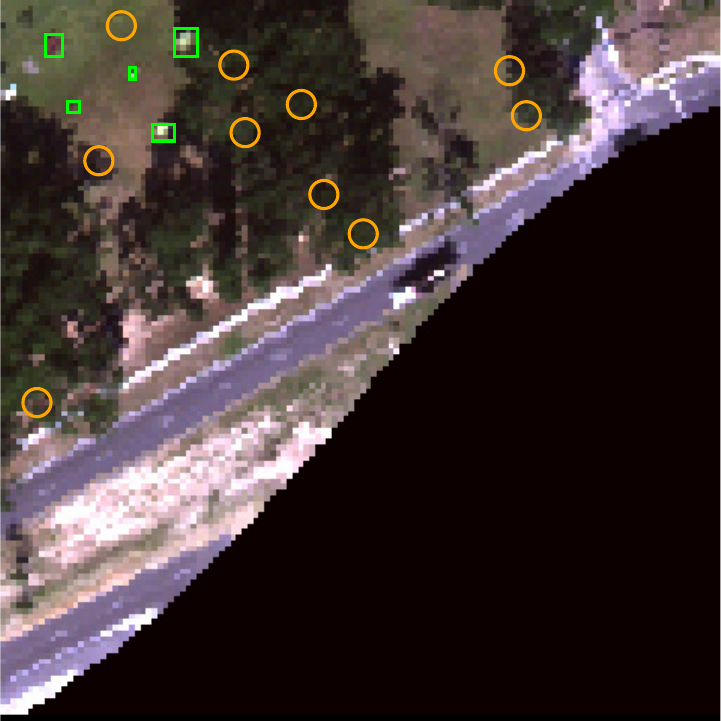}}
	\hfil
	\caption{Visualization results of SpecDETR on the Gulfport dataset. Visualization settings are consistent with Fig.~\ref{fig:specdetr_spod}.}
	\label{fig:Gulfport_detec}
\end{figure*}

\begin{table}[t]
	\centering
	\caption{
		The performance of SpecDETR on the Gulfport dataset under different confidence thresholds. Num. represents the number of predictions after confidence filtering, and TP stands for True Positives.
	}	
	\renewcommand\arraystretch{1} 
	\scriptsize  
	\setlength{\tabcolsep}{1mm}{   
		\begin{tabular}{c|cccc|cc|cc|cc|cc}
			\hline
			Confidence  & \multicolumn{4}{c|}{Total}      & \multicolumn{2}{c|}{Vis.} & \multicolumn{2}{c|}{Pro.Vis.} & \multicolumn{2}{c|}{Pos.Vis.} & \multicolumn{2}{c}{NotVis.} 
			\\
			\cline{2-13}
			Threshold & Num.   & TP    & Pr    & Re    & TP    & Re    & TP    & Re    & TP    & Re    & TP    & Re \\
			\hline
			
			0     & 380   & 18    & 0.047 & 0.316 & 9     & 1.000 & 4     & 0.571 & 3     & 0.375 & 2     & 0.061 \\
			0.05  & 97    & 15    & 0.155 & 0.263 & 9     & 1.000 & 4     & 0.571 & 1     & 0.125 & 1     & 0.030 \\
			0.1   & 52    & 15    & 0.289 & 0.263 & 9     & 1.000 & 4     & 0.571 & 1     & 0.125 & 1     & 0.030 \\
			0.2   & 26    & 13    & 0.500 & 0.228 & 8     & 0.889 & 4     & 0.571 & 1     & 0.125 & 0     & 0.000 \\
			0.3   & 17    & 12    & 0.706 & 0.211 & 7     & 0.778 & 4     & 0.571 & 1     & 0.125 & 0     & 0.000 \\
			0.4   & 13    & 9     & 0.692 & 0.158 & 6     & 0.667 & 3     & 0.429 & 0     & 0.000 & 0     & 0.000 \\
			0.5   & 9     & 6     & 0.667 & 0.105 & 4     & 0.444 & 2     & 0.286 & 0     & 0.000 & 0     & 0.000 \\
			
			\hline
		\end{tabular}%
	}
	\label{tab:Gulfport}%
\end{table}%

\begin{figure*}[t]
	\centering
	\subfloat[GT]{\includegraphics[width=0.24\linewidth]{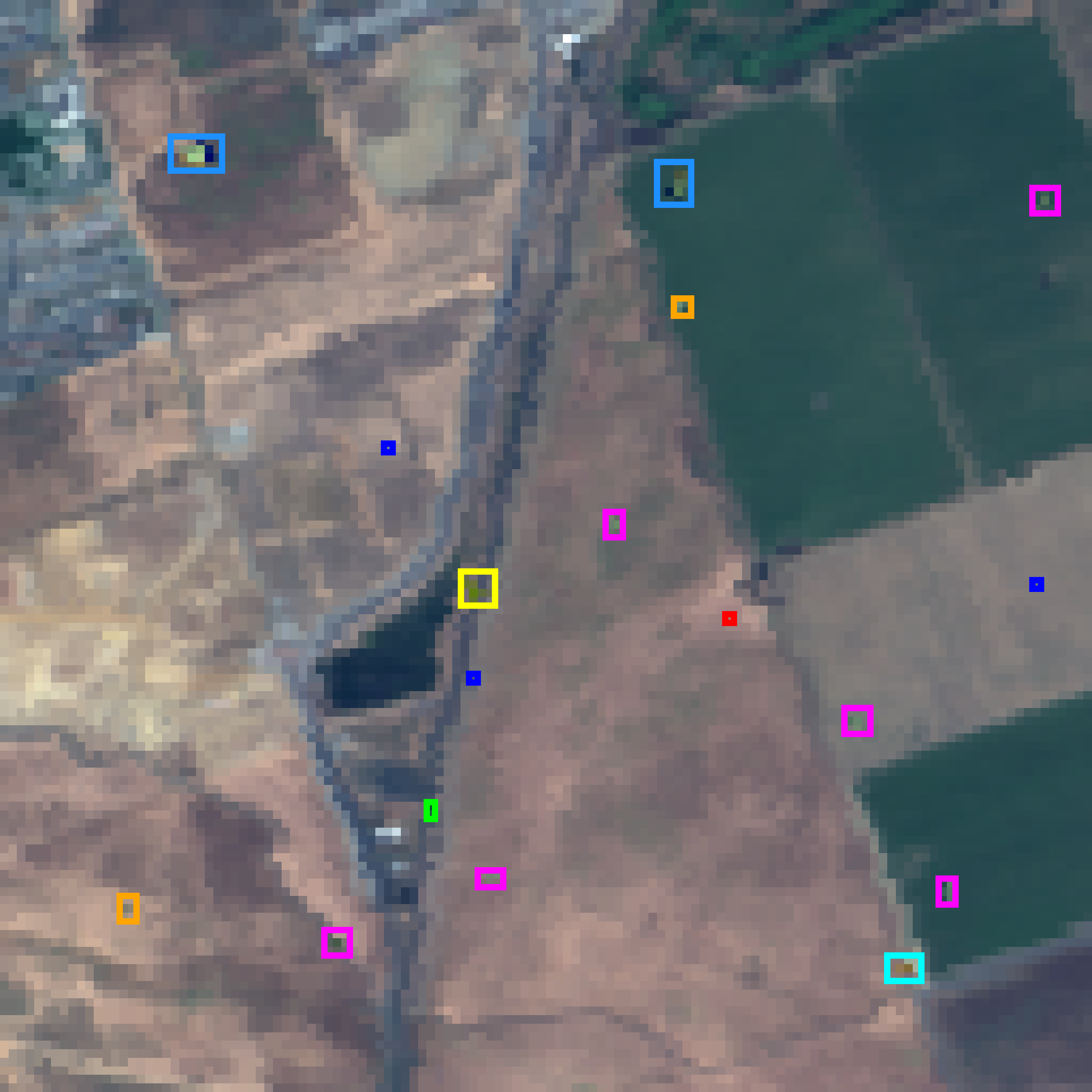}}
	\hfil
	\subfloat[SpecDETR]{\includegraphics[width=0.24\linewidth]{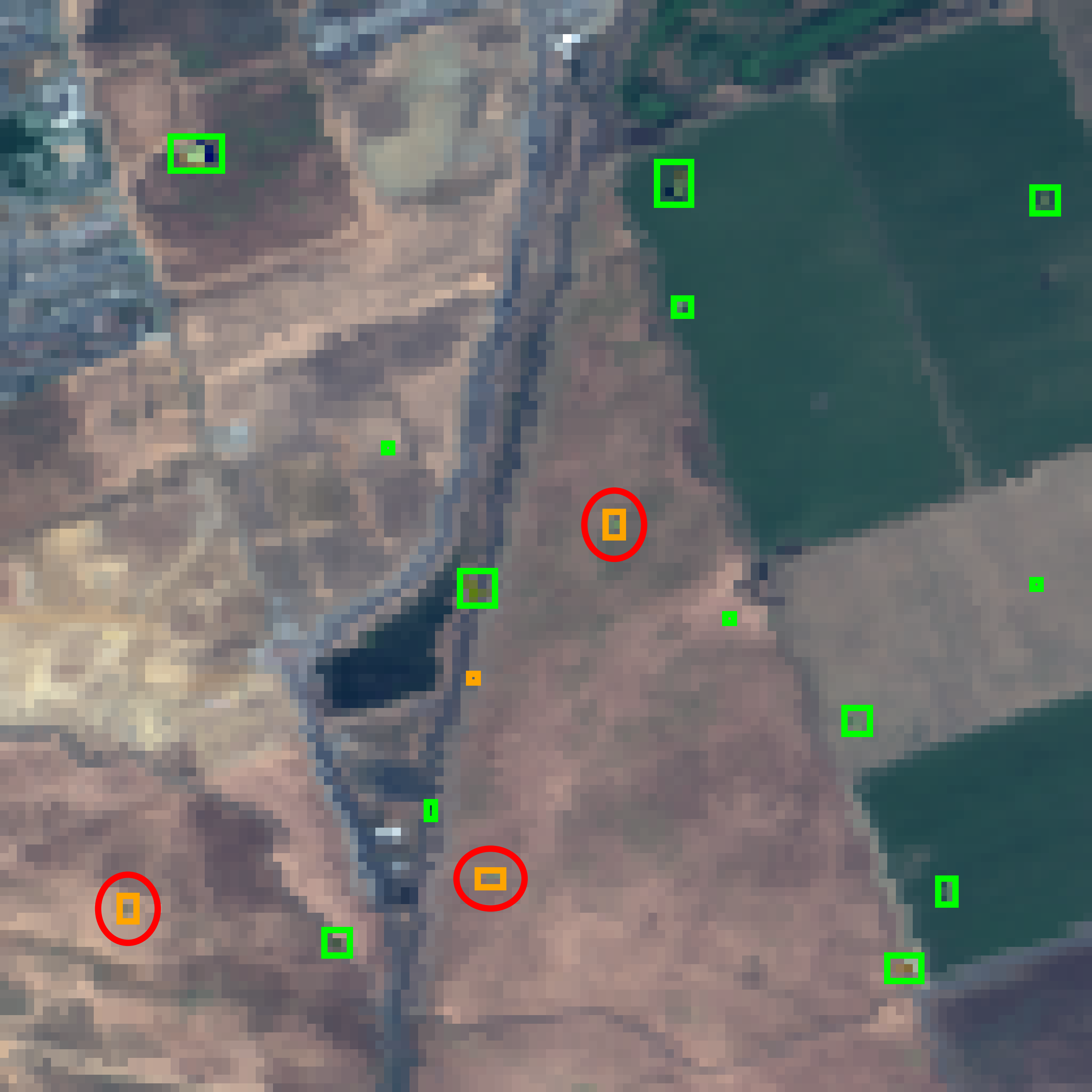}}
	\hfil
	\subfloat[DeformableDETR]{\includegraphics[width=0.24\linewidth]{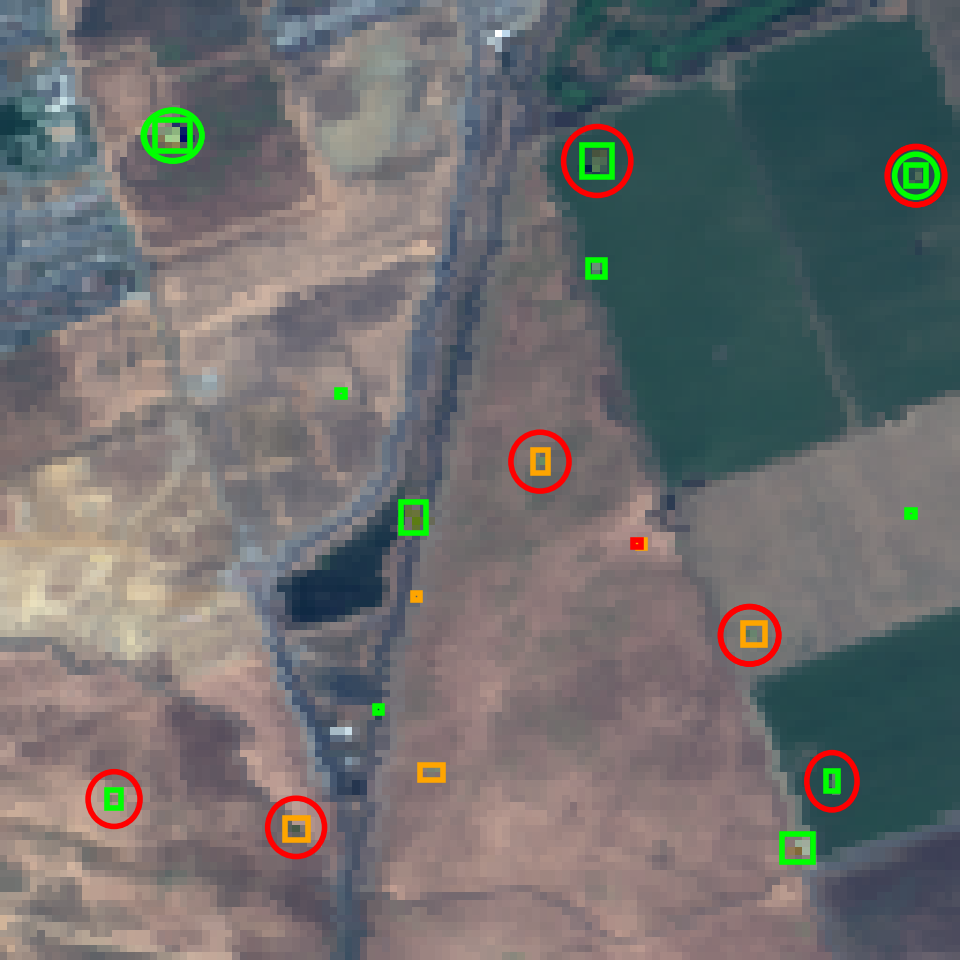}}
	\hfil
	\subfloat[DINO]{\includegraphics[width=0.24\linewidth]{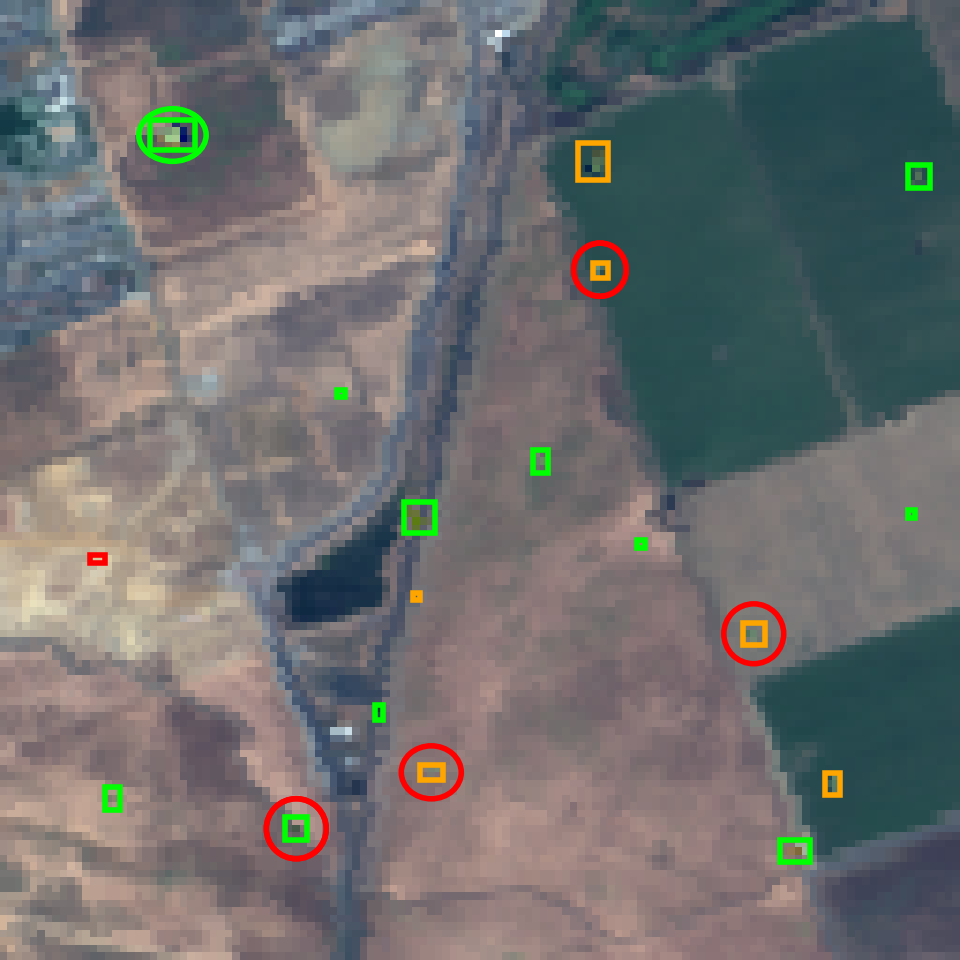}}
	\hfil
	\caption{Visual comparison of SpecDETR and other DETR-like detectors. (a) Ground truth labels, with color meanings consistent with Fig.~\ref{fig:spotobject}. (b)-(d) Visualization of results, with visualization settings consistent with Fig.~\ref{fig:specdetr_spod}. Green circles indicate redundant true positives which are actually considered as false positives in the evaluation.}
	\label{fig:DETR_spod}
\end{figure*}

On the Avon dataset, SpecDETR demonstrates superior performance in real-world single-spectral point object detection, achieving 0.885 mAP and 0.925 mAR. 
As illustrated in Tables~\ref{tab:spodhtdresult} and~\ref{tab:avonresult}, SpecDETR exhibits a significant advantage over the compared HTD methods in object-level evaluation on both the SPOD and Avon datasets.
In the conversion process from HTD results to object detection outcomes, we apply the optimal segmentation threshold for each object class to ensure the recall of objects.
Some HTD methods achieve high $\text{mAR}_{\text{25}}$, but perform poorly in terms of $\text{mAP}_{\text{25}}$.
This is because current HTD methods primarily rely on the one-dimensional spectral information of objects, which tends to produce high detection responses in background regions with similar spectral characteristics or in other classes of similar objects.
As shown in Fig.~\ref{fig:avon_result}, most HTD methods generate numerous false alarms on soil, which has spectra similar to brown tarps.
The poor performance in $\text{mAP}_{\text{25}}$ is actually due to some false alarms having higher detection responses than true positives.
In contrast, SpecDETR can learn the  spatial-spectral joint features of objects from a large number of training samples, which is superior in the dimension of object feature perception compared to existing HTD methods.
As shown in Fig.~\ref{fig:avon_result}(n), after NMS processing of the predictions, SpecDETR reports only 4 false alarms on the soil.
These results demonstrate that the point object detection framework we propose is superior to the current  per-pixel classification framework based on prior spectral information.

\subsection{Experiments on the SanDiego and Gulfport Datasets}

We select the three methods that performed best on the SPOD dataset, CentripetalNet, CornerNet, and DINO,  for further comparative evaluation with SpecDETR on the SanDiego and Gulfport datasets. The training epochs for the SanDiego and Gulfport datasets are set to 12 and 24, respectively, with the learning rate reduced by a factor of 0.1 at the 10th and 20th epochs, respectively. All other settings are kept consistent with those used for the SPOD dataset.

Table~\ref{tab:SanDiegoandGulfport} presents the performance comparison of SpecDETR and the compared visual object detection networks on the SanDiego and Gulfport datasets. SpecDETR achieves an AP of 1.0 and a Re of 1.0 on the SanDiego dataset, while CentripetalNet, CornerNet, and DINO achieve AP values of 0.750, 0.776, and 0.210, respectively. Additionally, SpecDETR achieves the best mean AP of 0.213 and the second-best mean Recall of 0.321 on the Gulfport dataset.
Fig.~\ref{fig:SanDiego_detect} shows the visualization results of SpecDETR and the compared visual object detection networks on the SanDiego dataset. Since the lowest confidence score for object prediction boxes among the four methods is 0.092, Fig.~\ref{fig:SanDiego_detect} filters out prediction boxes with confidence scores below 0.09. Although all four methods can detect three airplanes, the other three comparison methods have false predictions with higher confidence scores than the object prediction boxes. Furthermore, despite the prediction boxes being filtered by confidence, DINO has many repeated predictions, indicating that the current DETR-like detectors' one-to-one prediction is not very suitable for hyperspectral point object detection. In contrast, SpecDETR successfully detects all three airplanes with confidence scores of 0.327, 0.314, and 0.351, respectively. Other predictions output by SpecDETR, with confidence scores below 0.05, are filtered out in Fig.~\ref{fig:SanDiego_detect}.

Drawing inspiration from the task of moving object detection in satellite videos\cite{10549838}, Table~\ref{tab:Gulfport} presents the precision and recall performance of SpecDETR on the Gulfport dataset under fixed confidence thresholds.
At a confidence threshold of 0.1, SpecDETR detects 15 objects, achieving a Re of 0.26.3 and a Pr of 0.289. When the confidence threshold is raised to 0.3, SpecDETR detects 12 objects, with a Re of 0.706 and a Pr of 0.105.
Fig.~\ref{fig:Gulfport_detec} visualizes the detection results of SpecDETR on the Gulfport dataset, where a large number of undetected objects were obscured by trees or shadows. 
Consequently, in Table~\ref{tab:Gulfport}, we analyze the detection performance of SpecDETR under different levels of object observational confidence. It is observed that the lower the  object observational confidence level, the lower the Re of SpecDETR for detection. At a 0.1 confidence threshold, SpecDETR  detects all 9 Vis. objects, 4 out of 7 Pro.Vis. objects, but only 1 each for the Pos.Vis. and NotVis. objects.

The predicted confidence scores for objects by SpecDETR are predominantly below 0.5, which can be attributed to the domain gap between the simulated training images and the real test images. Moreover, the extremely low Re for Pos.Vis. and NotVis. objects may be due, in part, to some objects being completely obscured. Additionally, it is possible that the linear mixing model on which our simulation is based does not accurately represent the actual imaging phenomena when objects are obscured by trees or shadows.

\subsection{Model Analysis}

\subsubsection{Robustness Analysis under Varying Lighting Conditions}

\begin{table}[t]
	\centering
	\caption{Detection Performance of SpecDETR on the Avon dataset under Varying Lighting Conditions.}
	\renewcommand\arraystretch{1.2}
	\scriptsize
	\setlength{\tabcolsep}{1mm}{
		\begin{tabular}{c|cc|cc|cc}
			\hline
			Imaging  & \multicolumn{2}{c|}{Mean} & \multicolumn{2}{c|}{Blue Tarp} & \multicolumn{2}{c}{Brown Tarp} \\
			\cline{2-7}
			time  &  $\text{mAP}_{\text{25}}$ &  $\text{mRe}_{\text{25}}$ &   $\text{AP}_{\text{25}}$ &   $\text{Re}_{\text{25}}$ &  $\text{AP}_{\text{25}}$ &  $\text{Re}_{\text{25}}$ \\
			\hline
			12:01 & 0.990 & 1.000 & 1.000 & 1.000 & 0.981 & 1.000 \\
			12:06 & 0.974 & 1.000 & 1.000 & 1.000 & 0.949 & 1.000 \\
			13:44 & 0.981 & 1.000 & 1.000 & 1.000 & 0.962 & 1.000 \\
			13:51 & 0.987 & 1.000 & 1.000 & 1.000 & 0.974 & 1.000 \\
			13:57 & 1.000 & 1.000 & 1.000 & 1.000 & 1.000 & 1.000 \\
			\hline
		\end{tabular}
	}
	\label{tab:avon_light}%
\end{table}%

\begin{figure*}[t]
	\centering
	\subfloat[12:06]{\includegraphics[width=0.24\linewidth]{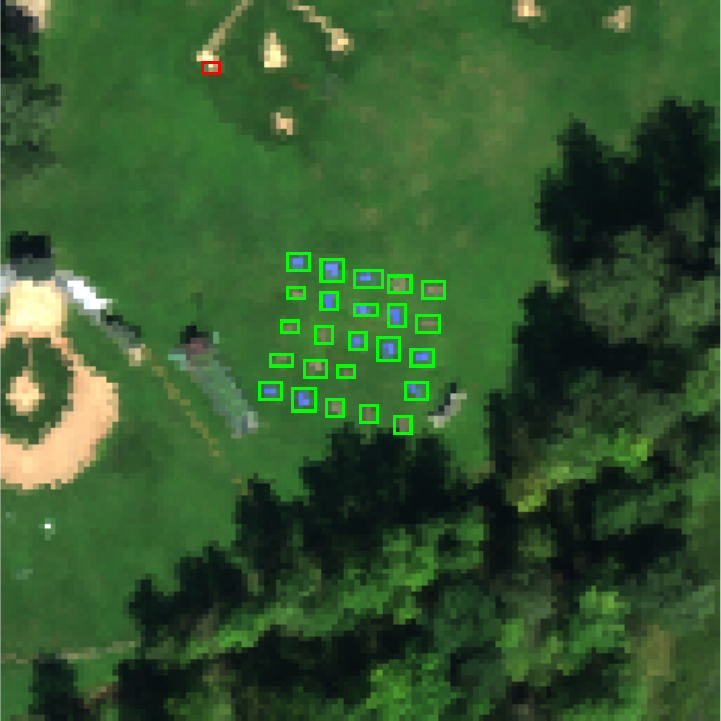}}
	\hfil
	\subfloat[13:44]{\includegraphics[width=0.24\linewidth]{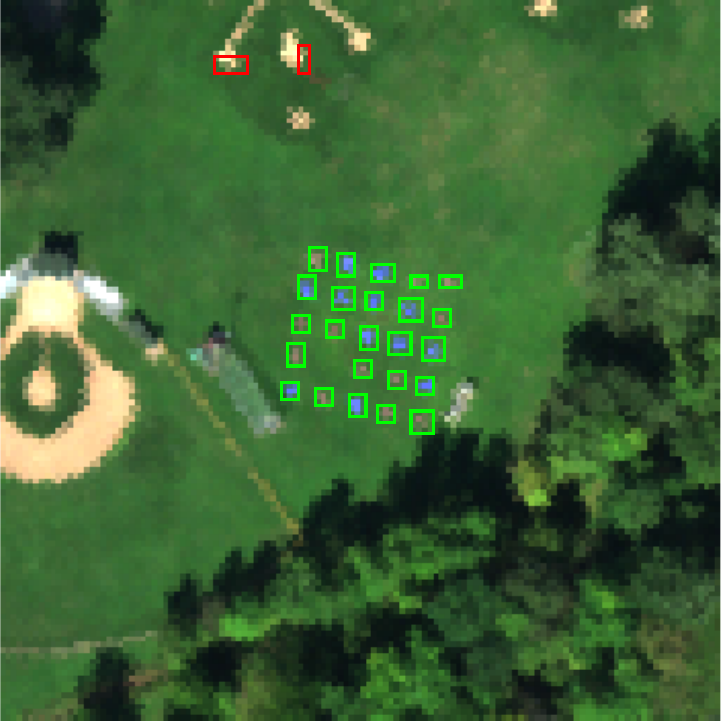}}
	\hfil
	\subfloat[13:51]{\includegraphics[width=0.24\linewidth]{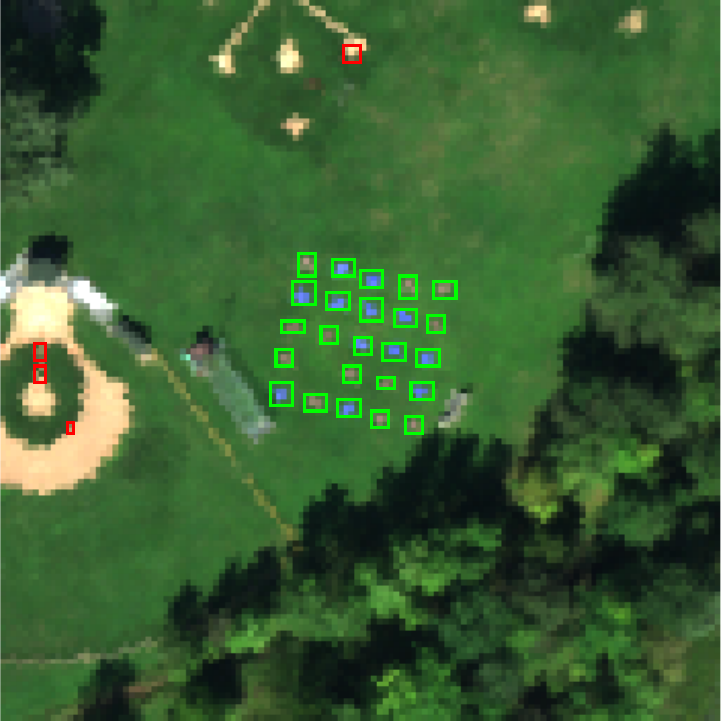}}
	\hfil
	\subfloat[13:57]{\includegraphics[width=0.24\linewidth]{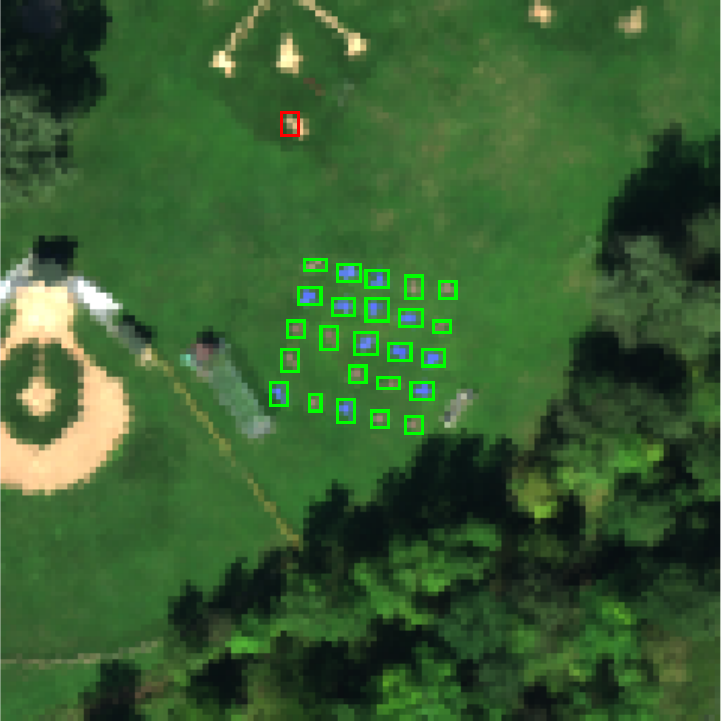}}
	\hfil
	\caption{Visualization results of SpecDETR on the Avon dataset under varying lighting conditions. The confidence threshold is 0, and the remaining visualization settings are consistent with Fig.~\ref{fig:specdetr_spod}.}
	\label{fig:avon_light}
\end{figure*}

In the experiments on the Avon dataset in Section~\ref{subsec:datasets_gt}, the imaging time for the training data was at 11:31, while the test image was captured at 12:01. Additionally, we extract the same area as the test image from four other flights with imaging times at 12:06, 13:44, 13:51, and 13:57, respectively, to analyze and assess the robustness of SpecDETR under varying lighting conditions. Both the training and test images consist of uncorrected radiometric data, which are susceptible to solar illumination effects.
The training and test images both consist of radiance data without atmospheric correction, which are influenced by solar illumination.
As shown in Table~\ref{tab:avon_light}, SpecDETR achieves a $\text{mRe}_{\text{25}}$ of 1.0 and above 0.97 $\text{mRe}_{\text{25}}$ across the five test images with different lighting conditions, and it also attains a 1.0 $\text{AP}_{\text{25}}$ and 1.0 $\text{Re}_{\text{25}}$ for blue tarps. Figs.~\ref{fig:avon_result}(n) and ~\ref{fig:avon_light} present the visualization results, showing only a minimal number of false positives for brown tarps on soil. The robustness of SpecDETR to lighting conditions suggests that the differences in data distribution under various lighting conditions on the Avon dataset have a negligible impact on the point object detection task.

\begin{table}[t]
	\centering
	\caption{Detection Performance of SpecDETR on the Avon dataset under Different Noise Levels.}
	\renewcommand\arraystretch{1.2}
	\scriptsize
	\setlength{\tabcolsep}{1mm}{
		\begin{tabular}{c|cc|cc|cc}
			\hline
			Noise & \multicolumn{2}{c|}{Mean} & \multicolumn{2}{c|}{Blue Tarp} & \multicolumn{2}{c}{Brown Tarp} \\
			\cline{2-7}
			level &  $\text{mAP}_{\text{25}}$ &  $\text{mRe}_{\text{25}}$ &   $\text{AP}_{\text{25}}$ &   $\text{Re}_{\text{25}}$ &  $\text{AP}_{\text{25}}$ &  $\text{Re}_{\text{25}}$ \\
			\hline
			$\text{SNR}_{\text{dB}}$=10 & 0.255 & 0.458 & 0.477 & 0.583 & 0.032 & 0.333 \\
			$\text{SNR}_{\text{dB}}$=15 & 0.416 & 0.708 & 0.752 & 0.750 & 0.079 & 0.667 \\
			$\text{SNR}_{\text{dB}}$=20 & 0.520 & 0.833 & 0.911 & 0.917 & 0.128 & 0.750 \\
			$\text{SNR}_{\text{dB}}$=25 & 0.481 & 0.917 & 0.842 & 0.917 & 0.120 & 0.917 \\
			$\text{SNR}_{\text{dB}}$=30 & 0.614 & 0.958 & 1.000 & 1.000 & 0.227 & 0.917 \\
			$\text{SNR}_{\text{dB}}$=35 & 0.761 & 1.000 & 1.000 & 1.000 & 0.523 & 1.000 \\
			w.o. noise & 0.990 & 1.000 & 1.000 & 1.000 & 0.981 & 1.000 \\
			\hline
		\end{tabular}%
	}
	\label{tab:avon_noise}%
\end{table}%

\begin{figure}[t]
	\centering
	\subfloat[$\text{SNR}_{\text{dB}}$=10]{\includegraphics[width=0.33\linewidth]{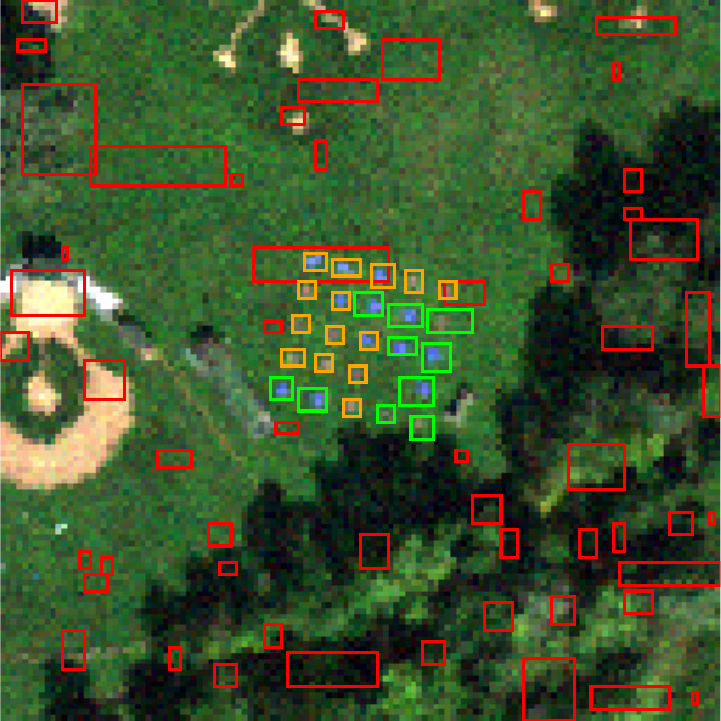}}
	\hfil
	\subfloat[$\text{SNR}_{\text{dB}}$=15]{\includegraphics[width=0.33\linewidth]{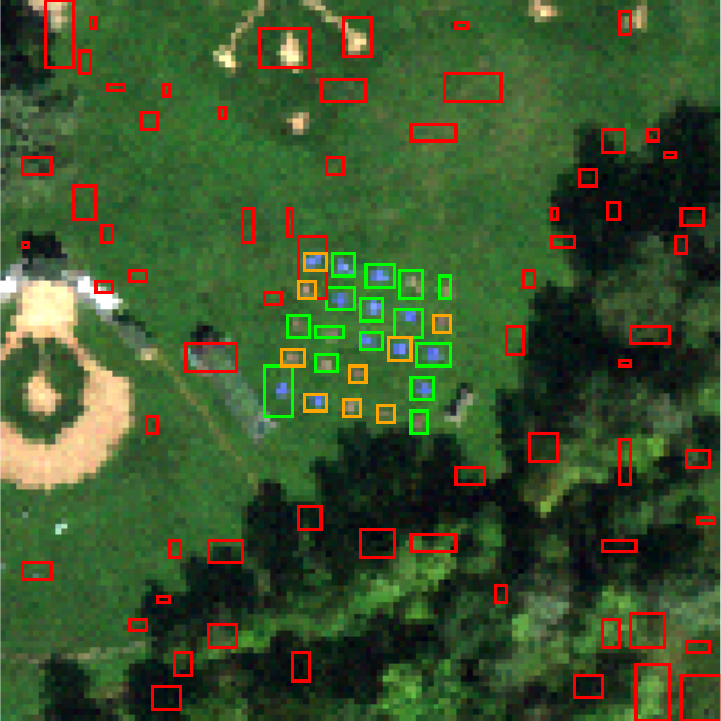}}
	\hfil
	\subfloat[$\text{SNR}_{\text{dB}}$=20]{\includegraphics[width=0.33\linewidth]{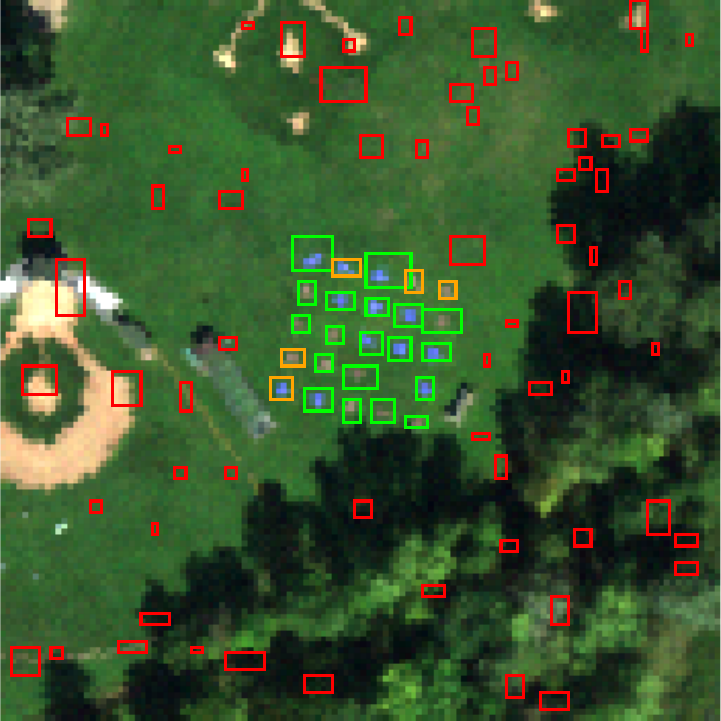}}
	\hfil
	\\
	\subfloat[$\text{SNR}_{\text{dB}}$=25]{\includegraphics[width=0.33\linewidth]{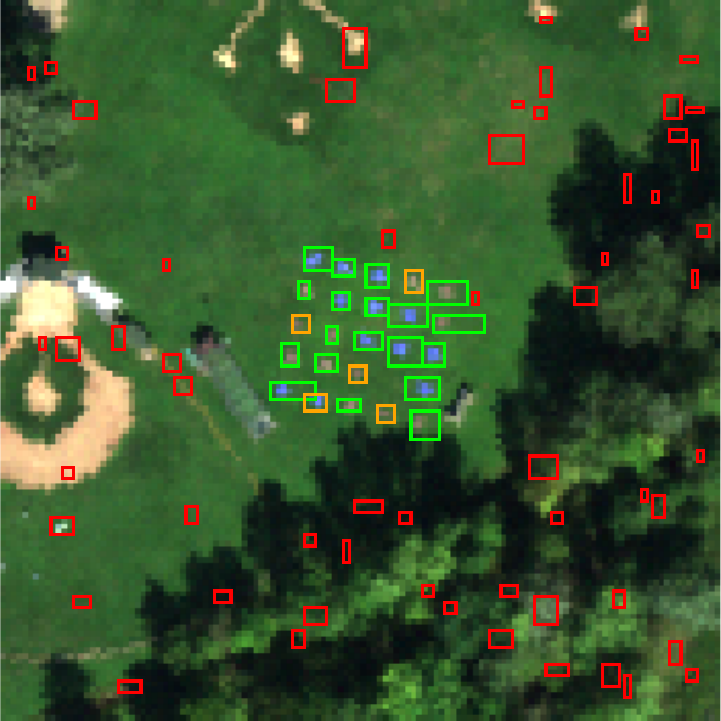}}
	\hfil
	\subfloat[$\text{SNR}_{\text{dB}}$=30]{\includegraphics[width=0.33\linewidth]{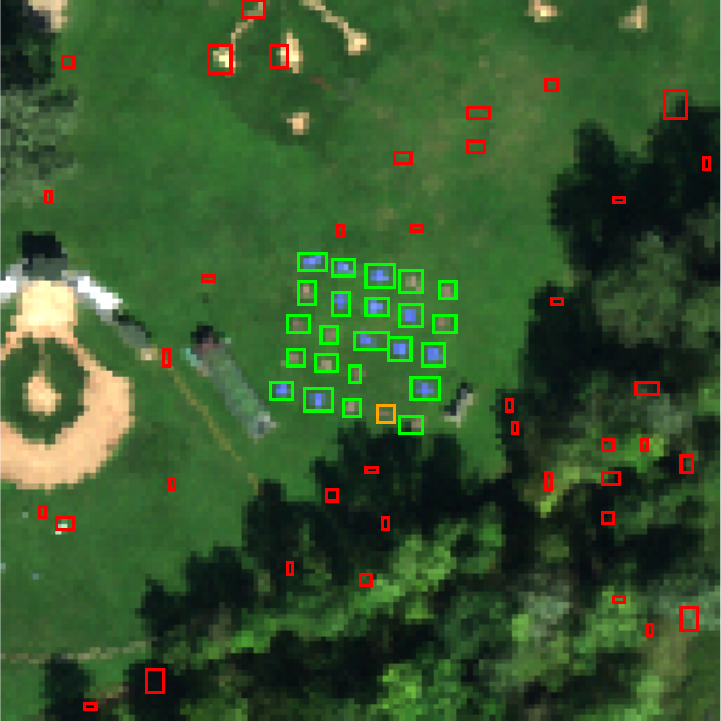}}
	\hfil
	\subfloat[$\text{SNR}_{\text{dB}}$=35]{\includegraphics[width=0.33\linewidth]{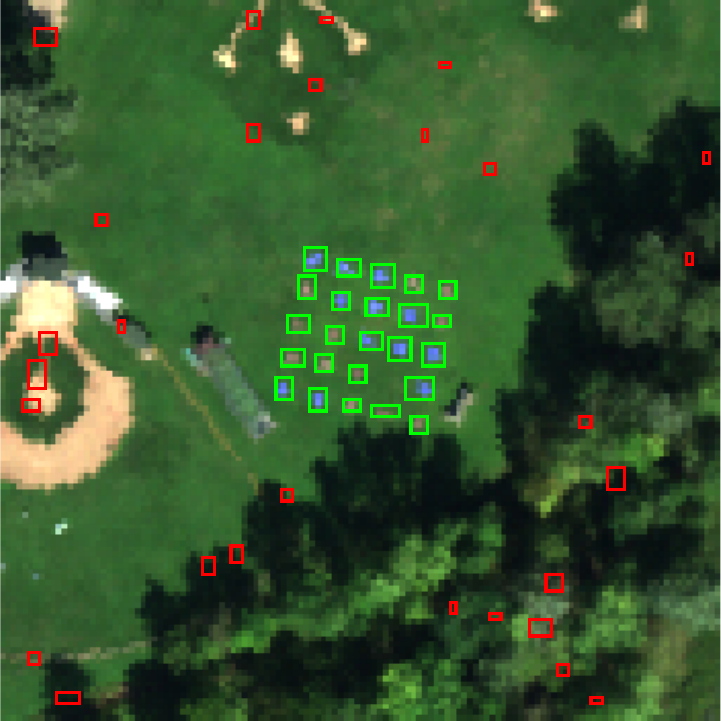}}
	\hfil
	\caption{Visualization results of SpecDETR on the Avon dataset under different noise levels.  The visualization settings are consistent with Fig.~\ref{fig:specdetr_spod}.}
	\label{fig:avon_noise}
\end{figure}

\subsubsection{Robustness Analysis  under Different Noise Levels}

Following the classic work \cite{zhang2013sparse} in the field, we  add additive Gaussian white noise to the Avon test image to assess the robustness of SpecDETR under different noise levels. As described in literature \cite{zhang2013sparse}, zero-mean Gaussian noise with different variances are added to each spectral band, with the noise level quantified by $\text{SNR}_{\text{dB}}$, defined as:
\begin{equation}
	\label{snr}
	\begin{aligned}
		\text{SNR}_{\text{dB}}=10\log _{10}\left( \frac{\sigma _{signal,i}^{2}}{\sigma _{noise,i}^{2}} \right) , \text{for } i=1,2,3,\cdots ,C,
	\end{aligned}
\end{equation}
where $i$ denotes the band index, $\sigma_{signal,i}^{2}$ is the variance of the $i$-th band, and $\sigma_{noise,i}^{2}$ is the variance of the added Gaussian noise. We set $\text{SNR}_{\text{dB}}$ to 10 dB, 15 dB, 20 dB, 25 dB, 30 dB, and 35 dB, respectively, and add the corresponding $\sigma_{noise,i}^{2}$ Gaussian noise into the Avon test image. Subsequently, we employ SpecDETR, trained on the noise-free training set, for direct inference without fine-tuning.
Table~\ref{tab:avon_noise} provides a quantitative evaluation of the results, while Fig.~\ref{fig:avon_noise} visualizes the predicted  bboxes with confidence scores above 0.2. It can be observed that the detection performance is positively correlated with the magnitude of the added noise variance. Moreover, SpecDETR demonstrates superior noise robustness for blue tarps compared to brown tarps, attributed to the greater spectral distinctiveness of blue tarps from the background. At $\text{SNR}_{\text{dB}}$ of 10 dB, SpecDETR achieves a 0.477 $\text{AP}_{\text{25}}$ for blue tarps, which improves to 0.911 $\text{AP}_{\text{25}}$ when $\text{SNR}_{\text{dB}}$ is 20 dB. However, at $\text{SNR}_{\text{dB}}$ of 30 dB, SpecDETR attains only a 0.227 $\text{AP}_{\text{25}}$ for brown tarps.
As depicted in Fig.~\ref{fig:avon_noise}, the introduction of Gaussian noise to the test image not only leads to false positives on soil but also on grass. This occurs because the data distribution of the noisy test images significantly diverges from that of the noise-free training images. The domain gap between training and test sets is a well-recognized challenge in the field of object detection and has given rise to the  cross-domain few-shot object detection task \cite{fu2025cross}.

\subsubsection{Analysis of the Necessity of Non-Maximum Suppression}
Fig.~\ref{fig:DETR_spod} presents the detection results of SpecDETR, Deformable DETR, and DINO on a test HSI from the SPOD dataset. Despite being end-to-end detectors, both Deformable DETR and DINO still exhibit redundant predictions, indicating the necessity of reintroducing NMS in SpecDETR. 

\subsubsection{Feature Extraction in SpecDETR Encoder}
\begin{figure*}[t]
	\centering
	\subfloat[The 1st layer]{\includegraphics[width=0.15\linewidth]{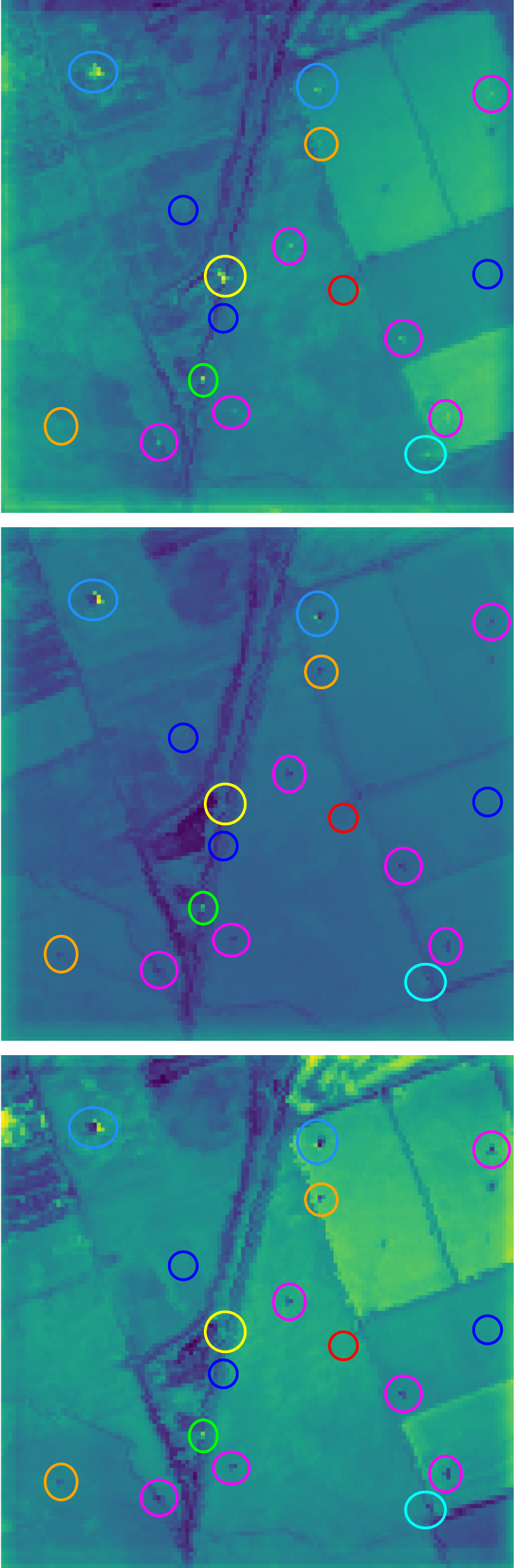}}
	\hfil
	\subfloat[The 2nd layer]{\includegraphics[width=0.15\linewidth]{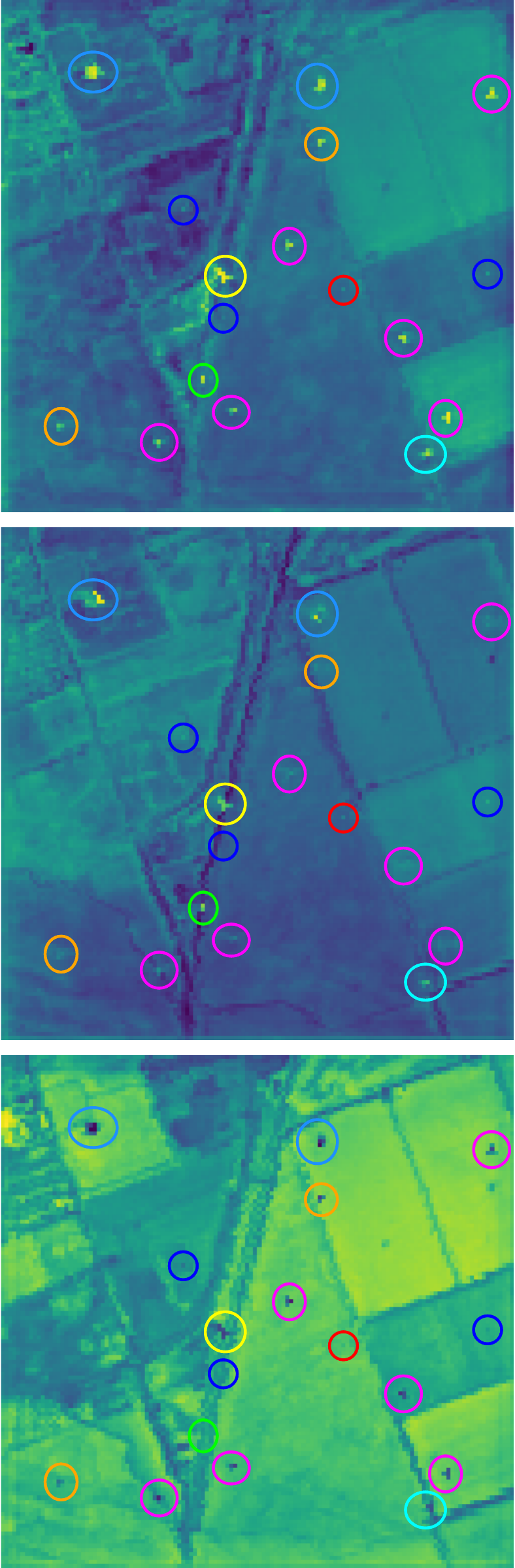}}
	\hfil
	\subfloat[The 3rd layer]{\includegraphics[width=0.15\linewidth]{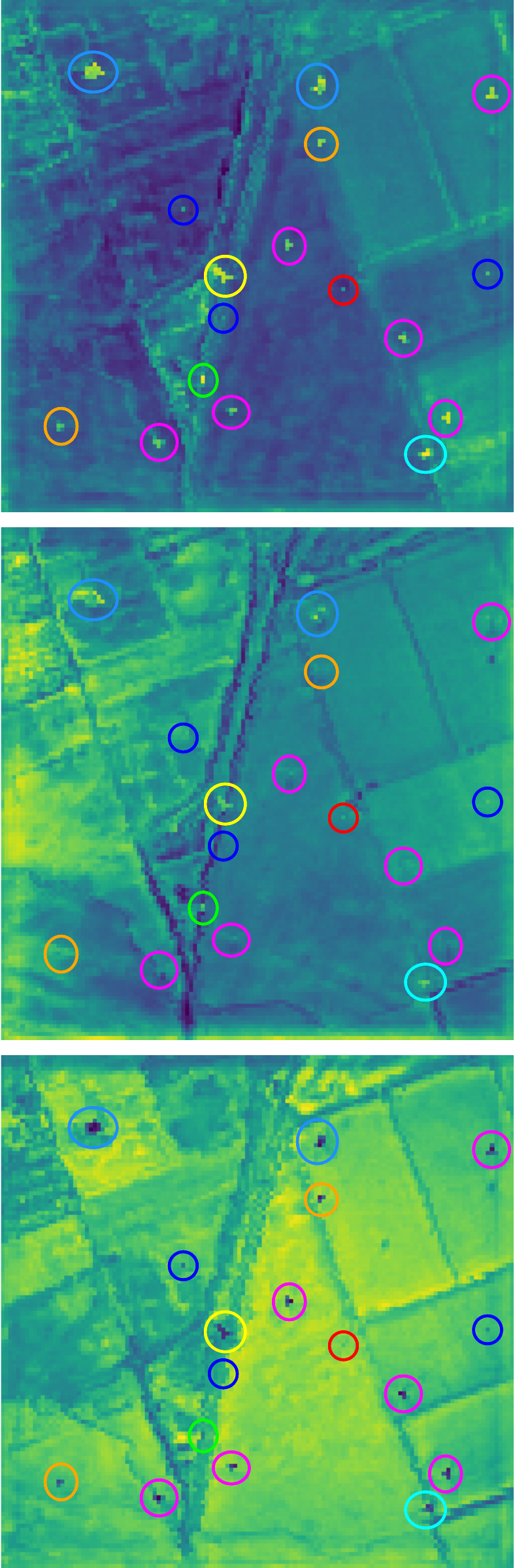}}
	\hfil
	\subfloat[The 4th layer]{\includegraphics[width=0.15\linewidth]{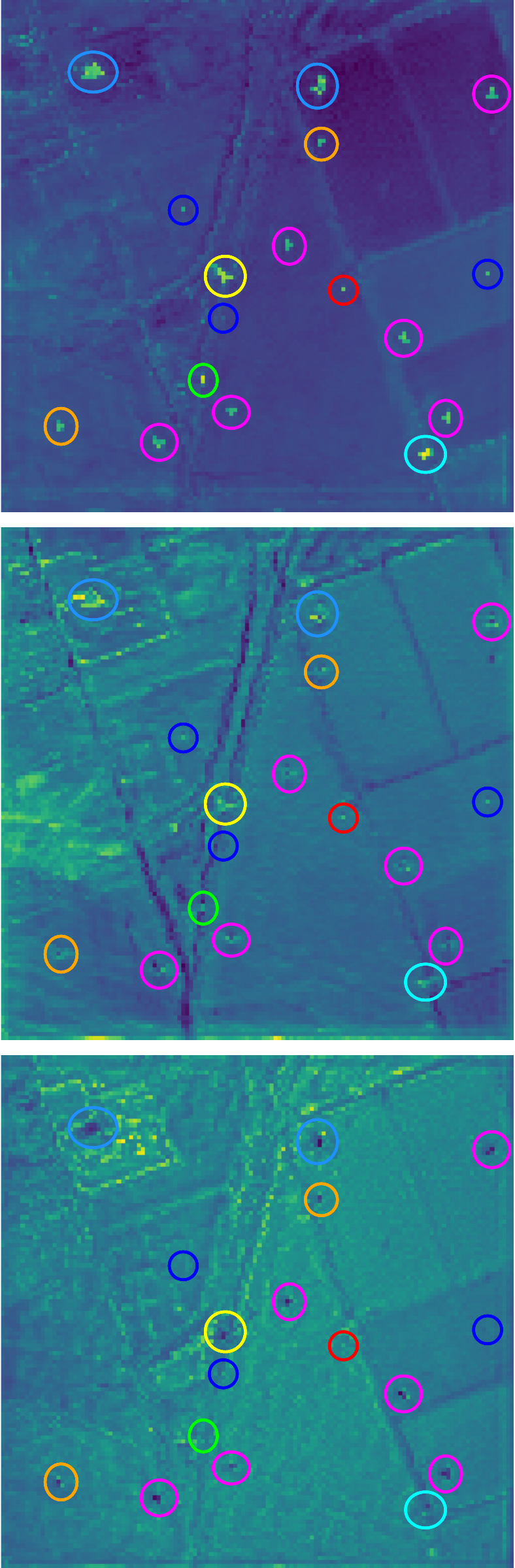}}
	\hfil
	\subfloat[The 5th layer]{\includegraphics[width=0.15\linewidth]{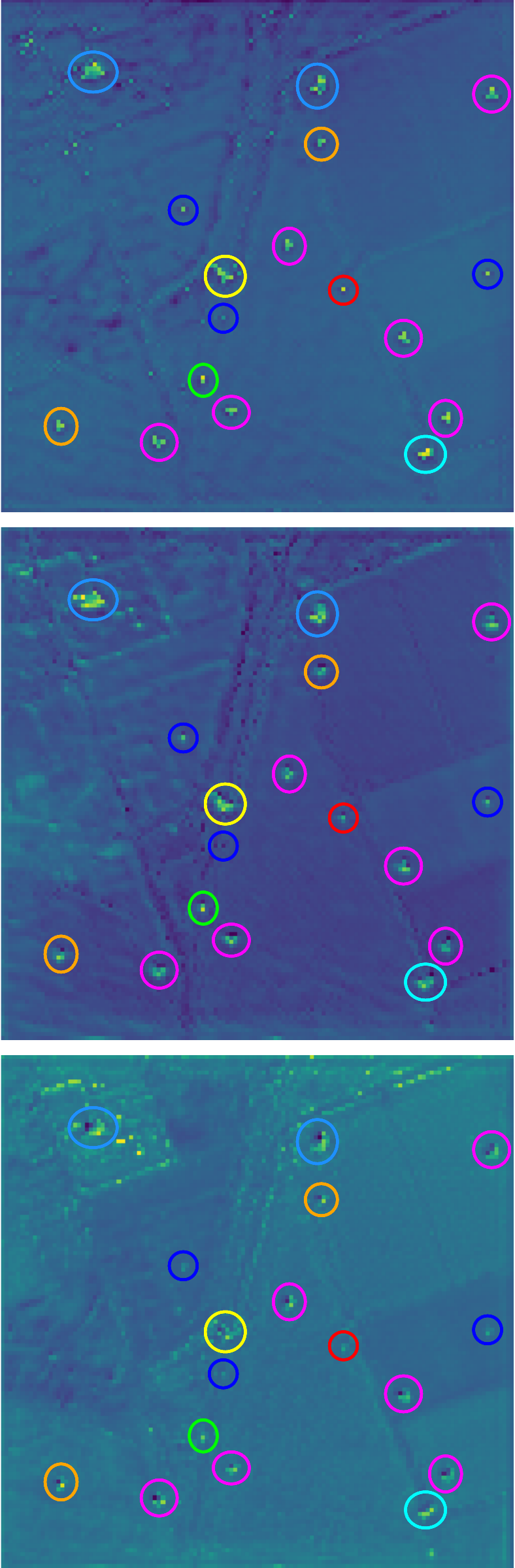}}
	\hfil
	\subfloat[The 6th layer]{\includegraphics[width=0.15\linewidth]{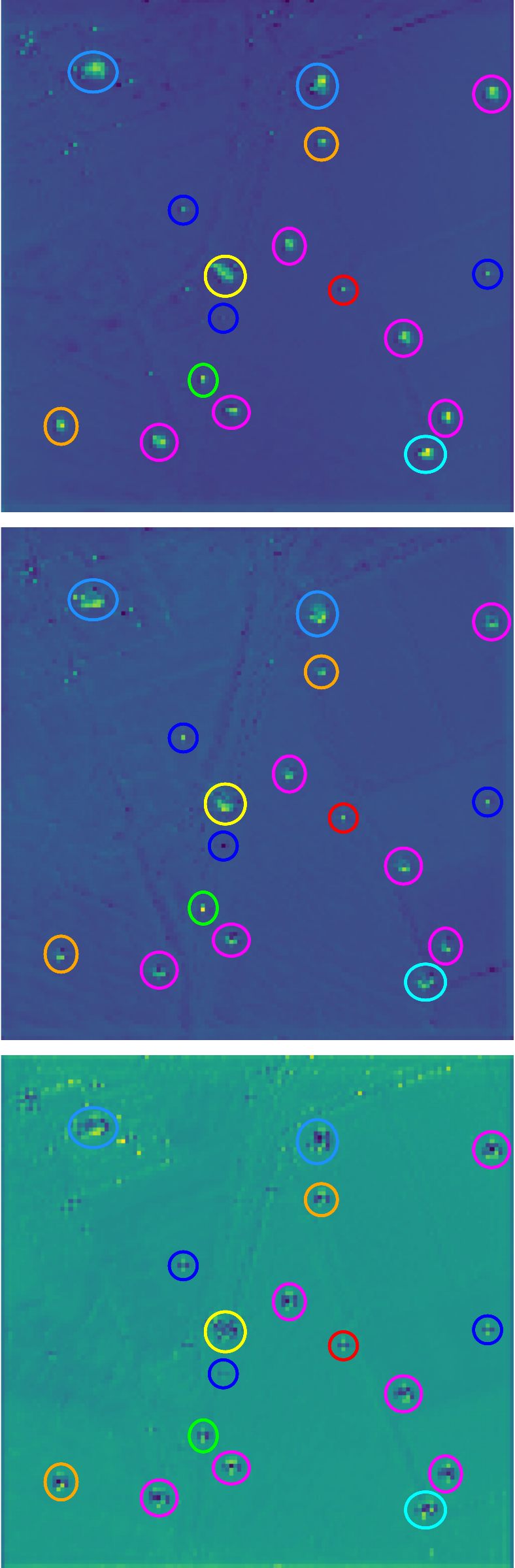}}
	\hfil
	\caption{
		Feature maps output by each encoder layer of SpecDETR on Fig.~\ref{fig:DETR_spod}(a). From top to bottom, the channels 50, 100, and 150 of the output feature maps are displayed, with objects marked by circles. The meaning of the circle colors is consistent with Fig.~\ref{fig:spotobject}.
	}
	\label{fig:featuremap}
\end{figure*}	

\begin{figure*}[t]
	\centering
	\subfloat[]{\includegraphics[width=0.15\linewidth]{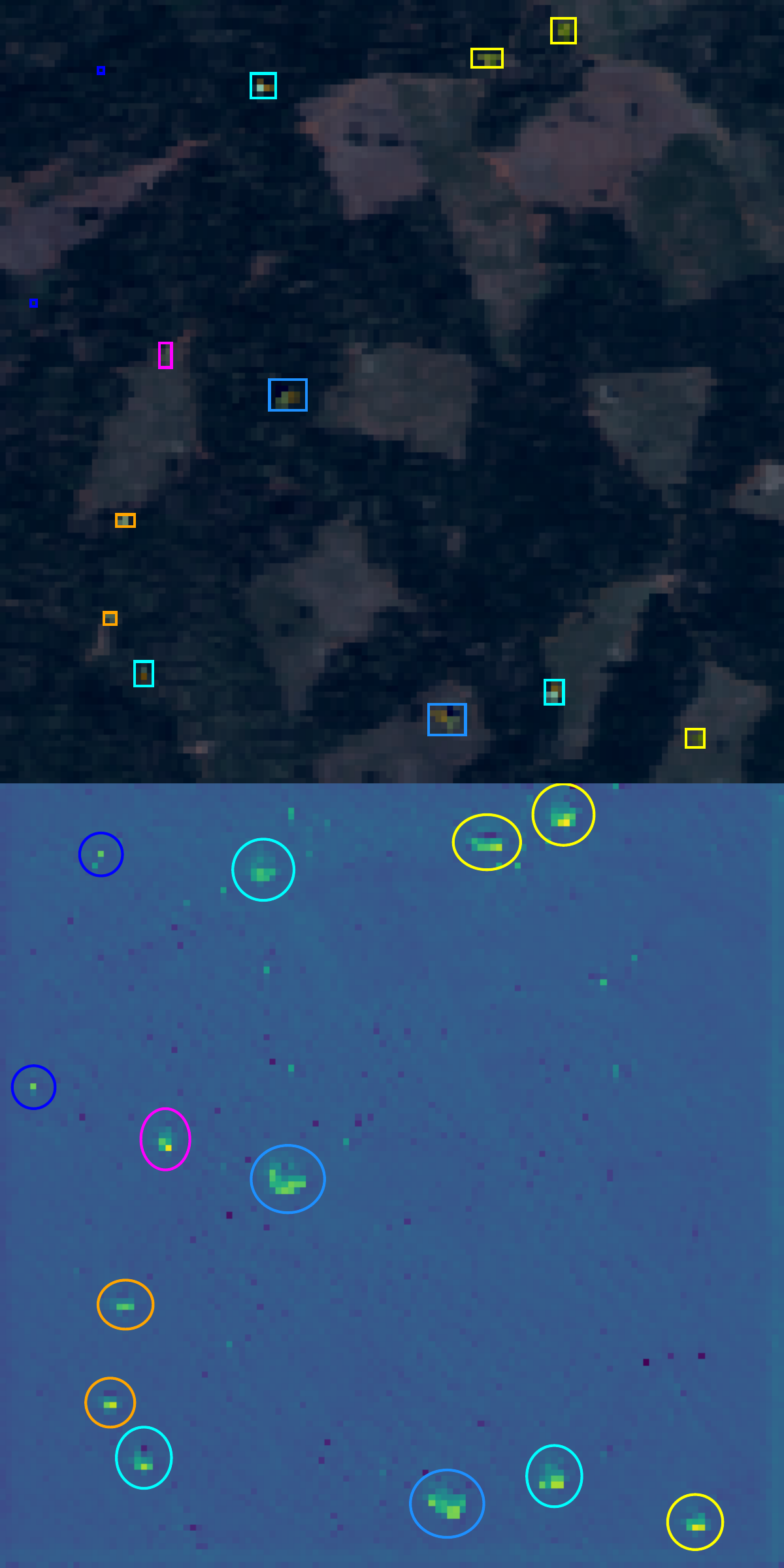}}
	\hfil
	\subfloat[]{\includegraphics[width=0.15\linewidth]{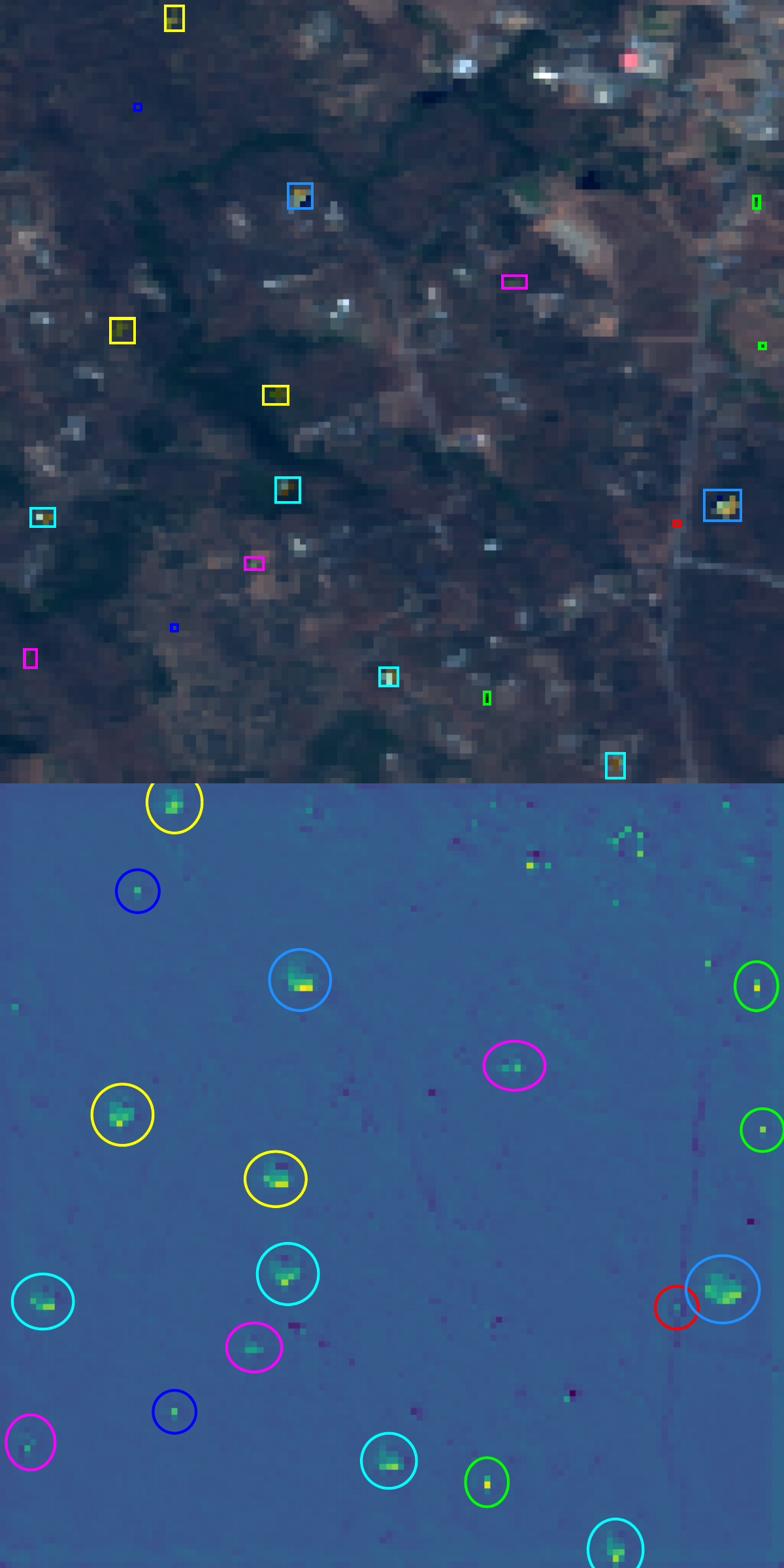}}
	\hfil
	\subfloat[]{\includegraphics[width=0.15\linewidth]{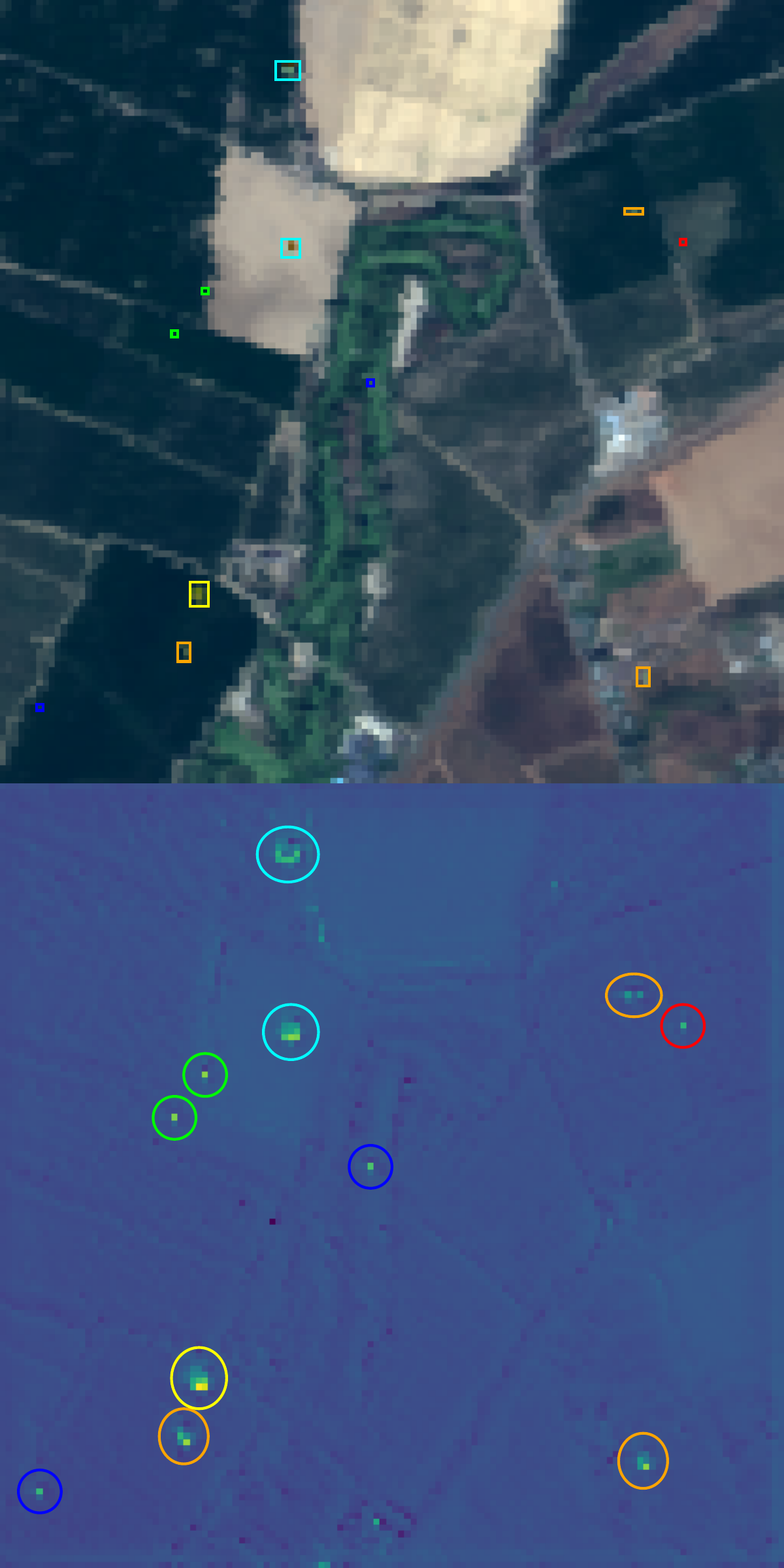}}
	\hfil
	\subfloat[]{\includegraphics[width=0.15\linewidth]{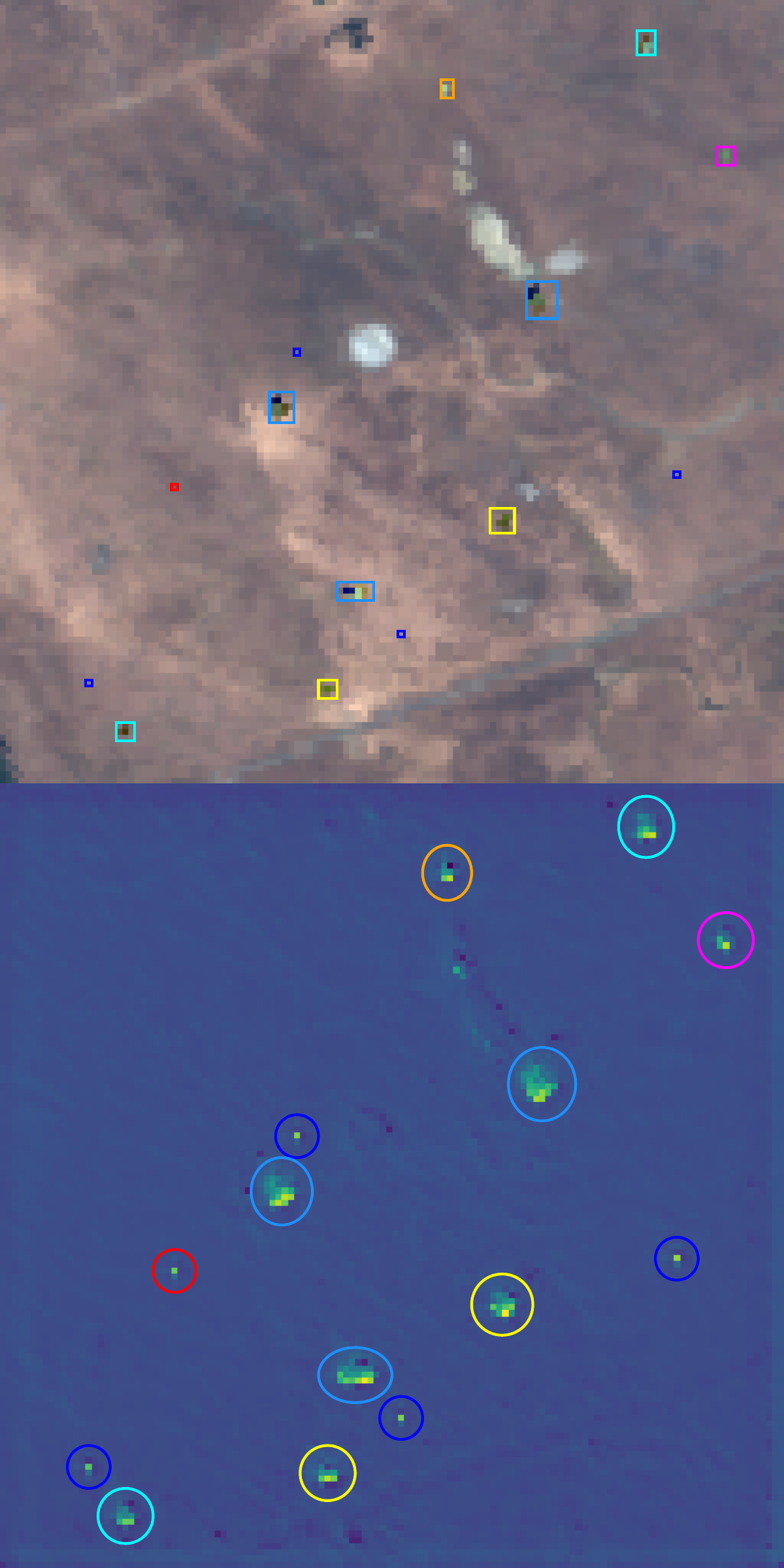}}
	\hfil
	\subfloat[]{\includegraphics[width=0.15\linewidth]{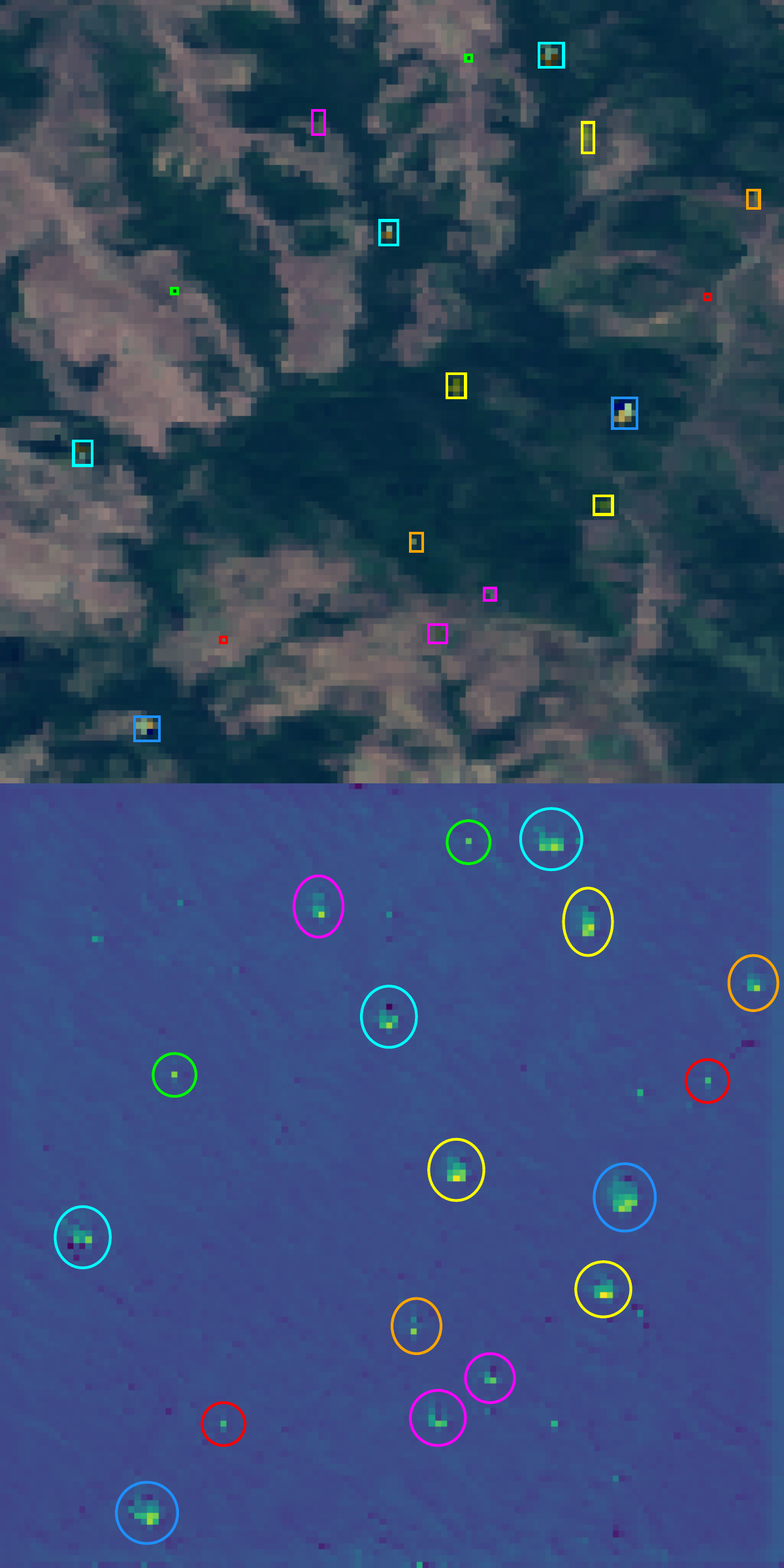}}
	\hfil
	\subfloat[]{\includegraphics[width=0.15\linewidth]{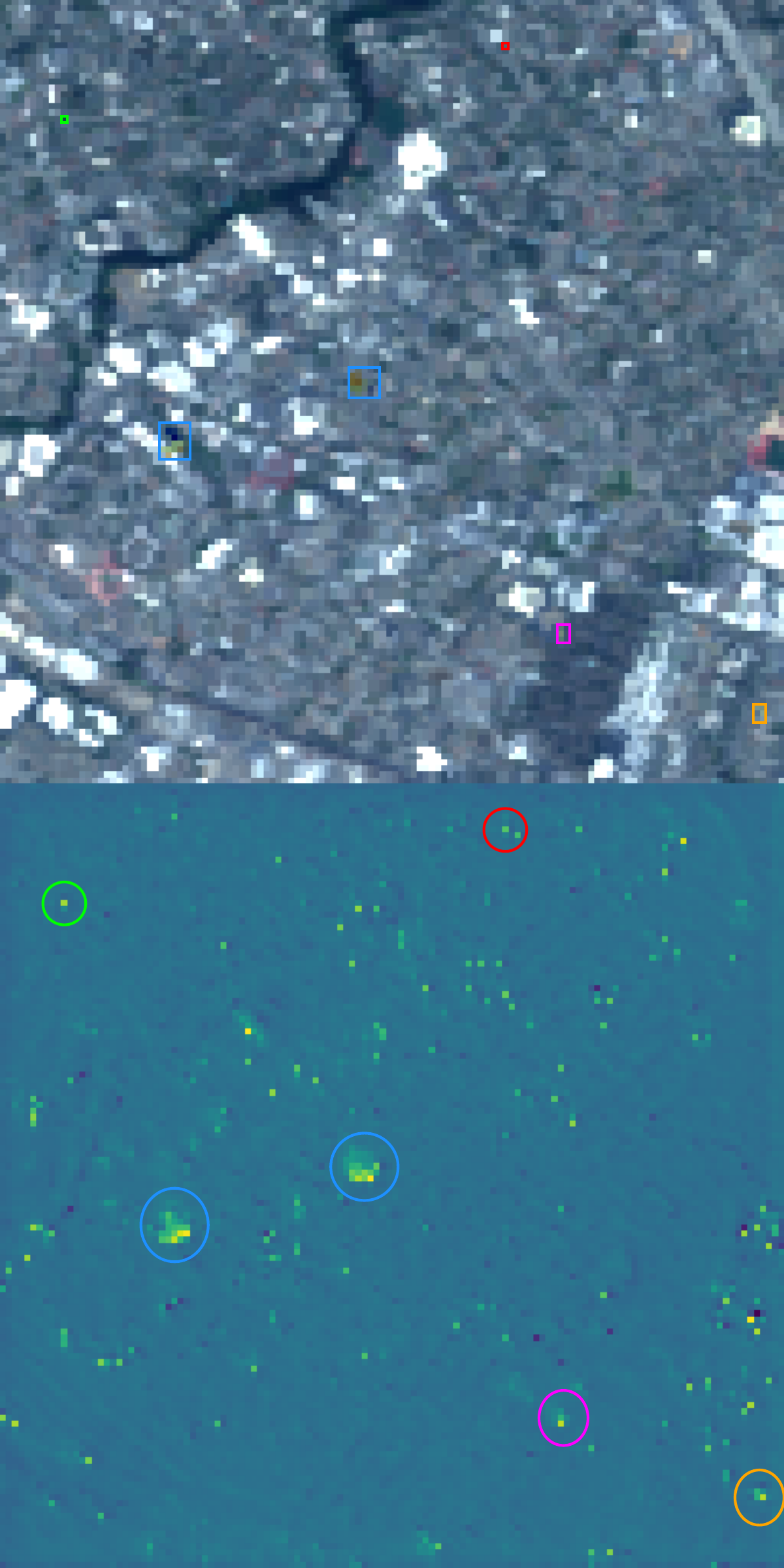}}
	\hfil
	\caption{Feature maps output by the encoder of SpecDETR on some test images from the SPOD dataset. The upper part shows the color images of the input, while the lower part displays the feature maps. The meaning of the circle and box colors is consistent with Fig.~\ref{fig:spotobject}.}
	\label{fig:featuremap2}
\end{figure*}

Following the conventional setup of DETR-like detectors, the number of SpecDETR encoder layers is set to 6 by default.
Fig.~\ref{fig:featuremap} visualizes the feature maps output by each encoder layer of SpecDETR for Fig.~\ref{fig:DETR_spod}(a).
As the number of encoder layers increases, the background area with complex texture structure becomes progressively smoother, while objects emerge more distinctly from the background.
Even for objects with extremely low abundance, C1 and C2, SpecDETR generates pronounced responses on the feature maps output by the final encoder layer.
Moreover, Fig.~\ref{fig:featuremap2} illustrates the feature maps output by the encoder of SpecDETR on various background test images, where the objects also exhibit clear features.
This demonstrates that SpecDETR can effectively extract the correct object features rather than noise or proxy features.
Additionally, in scenes with simpler background types such as farmland and woodland, the background regions in the feature maps are smooth and clean. However, in complex scenes like urban areas, the feature maps generate many isolated response points, which leads to some false alarms in the urban background regions as shown in  Fig.~\ref{fig:specdetr_spod}.

\subsubsection{Analysis of the Self-S2A Module}

\begin{figure}[t]
	\centering
	\subfloat[Background]{\includegraphics[width=0.33\linewidth]{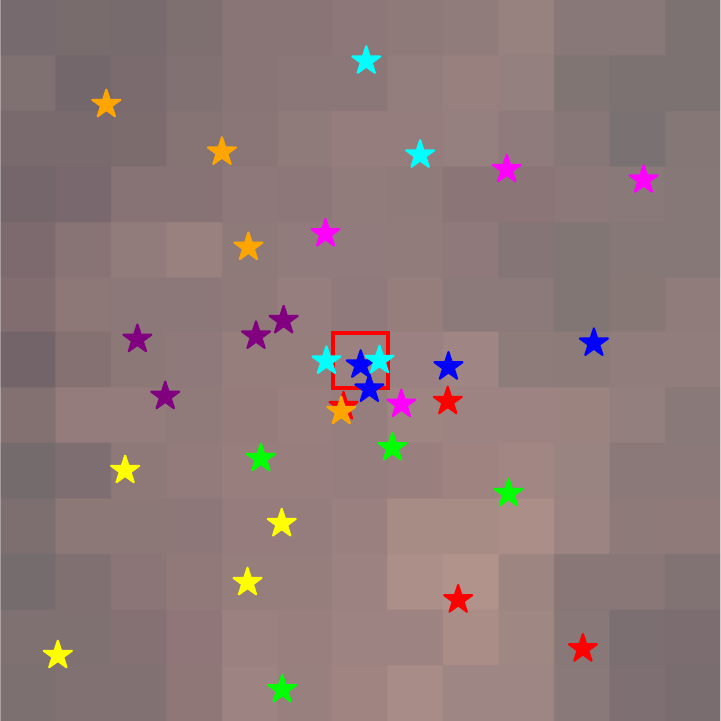}}
	\hfil
	\subfloat[C1]{\includegraphics[width=0.33\linewidth]{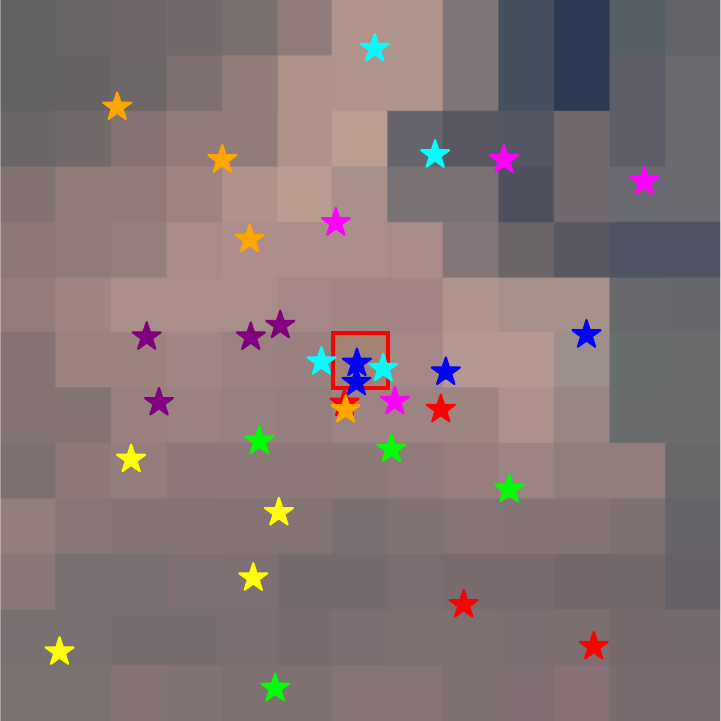}}
	\hfil
	\subfloat[C8]{\includegraphics[width=0.33\linewidth]{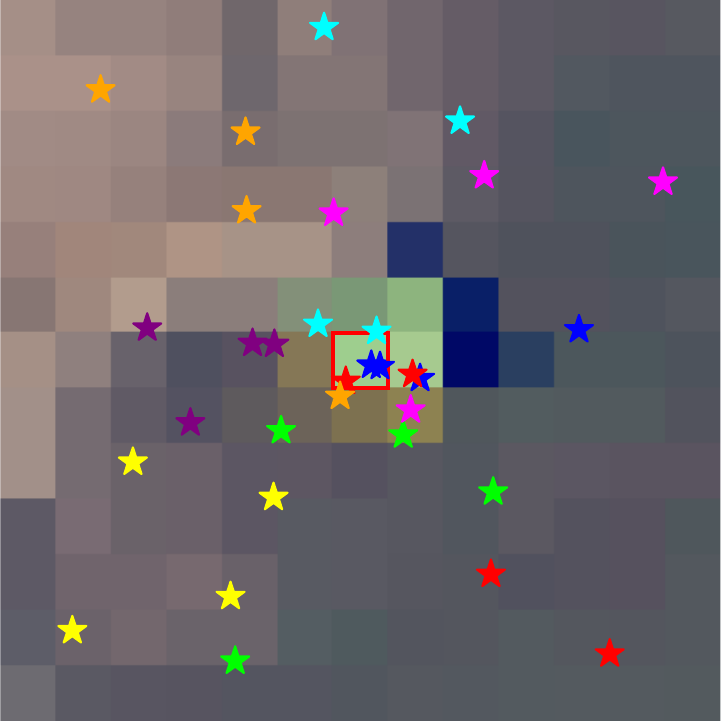}}
	\hfil
	\caption{Examples of sampling point distributions of the subpixel-scale deformable sampling operator in the first encoder layer of SpecDETR on Fig.~\ref{fig:DETR_spod}(a).
		The red box indicates the position of the query element. Sampling points associated with different attention heads are distinguished by various colors.                                                                                                      
	}
	\label{fig:feat1stl1}
\end{figure}

\begin{figure}[h]
	\centering
	\subfloat[Background]{\includegraphics[width=0.33\linewidth]{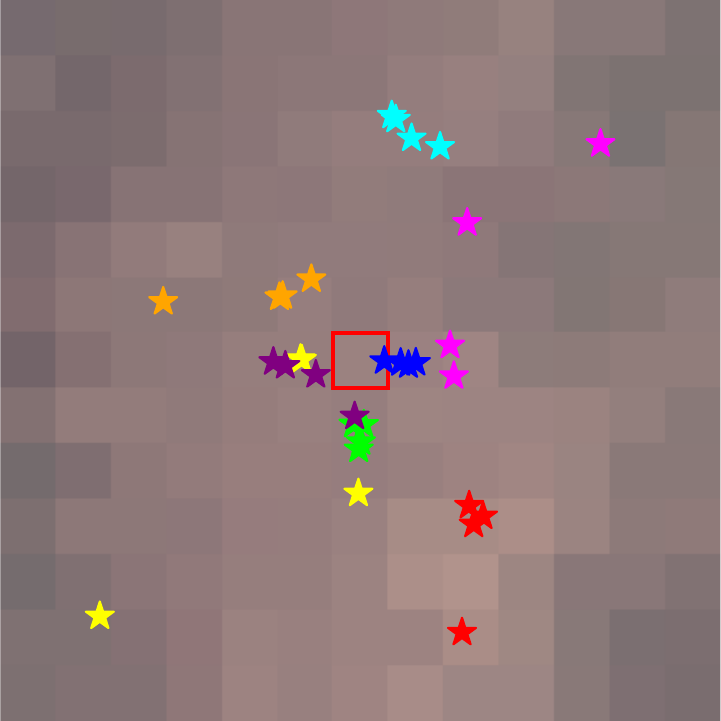}}
	\hfil
	\subfloat[C1]{\includegraphics[width=0.33\linewidth]{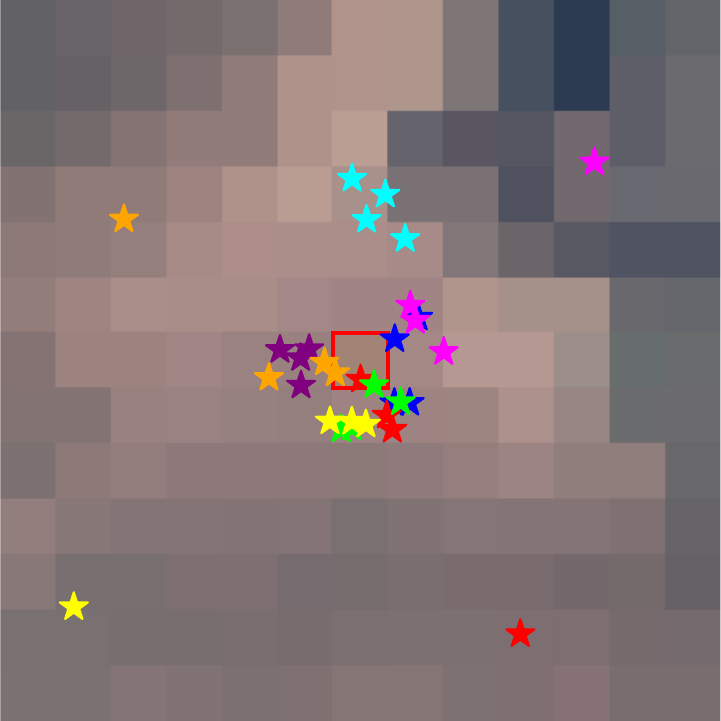}}
	\hfil
	\subfloat[C8]{\includegraphics[width=0.33\linewidth]{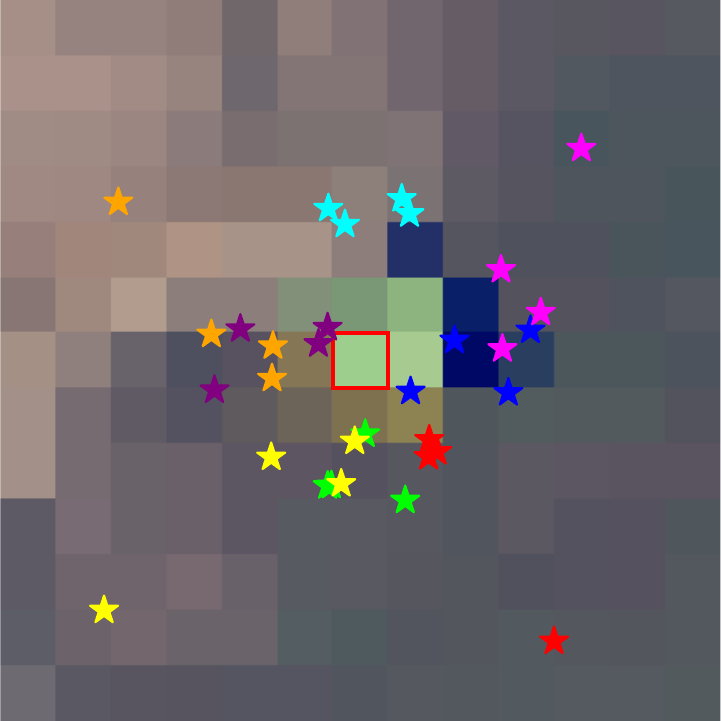}}
	\hfil
	\caption{Examples of sampling point distributions of the subpixel-scale deformable sampling operator in the last encoder layer of SpecDETR on Fig.~\ref{fig:DETR_spod}(a).}
	\label{fig:feat1stl6}
\end{figure}

\begin{figure}[t]
	\centering
	\subfloat[Background]{\includegraphics[width=0.33\linewidth]{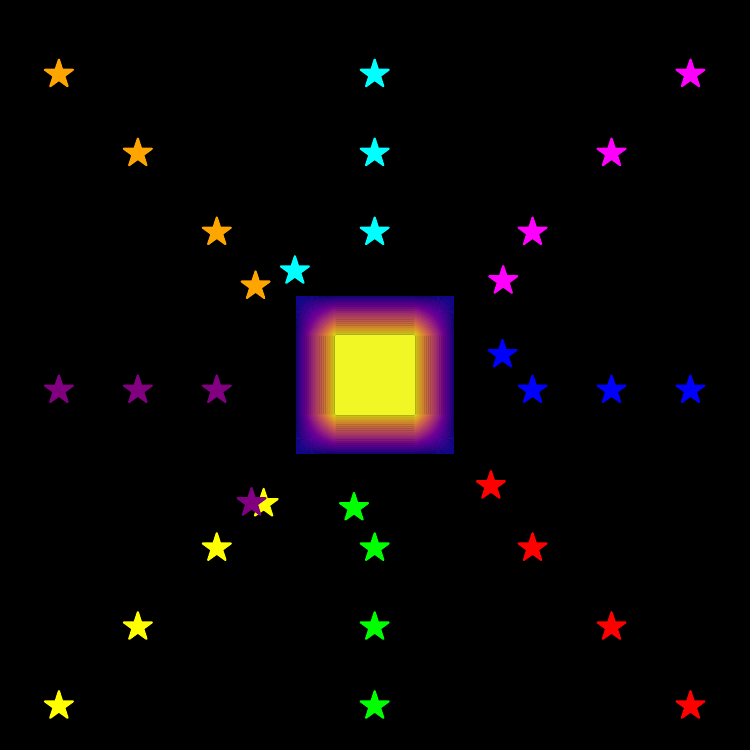}}
	\hfil
	\subfloat[C1]{\includegraphics[width=0.33\linewidth]{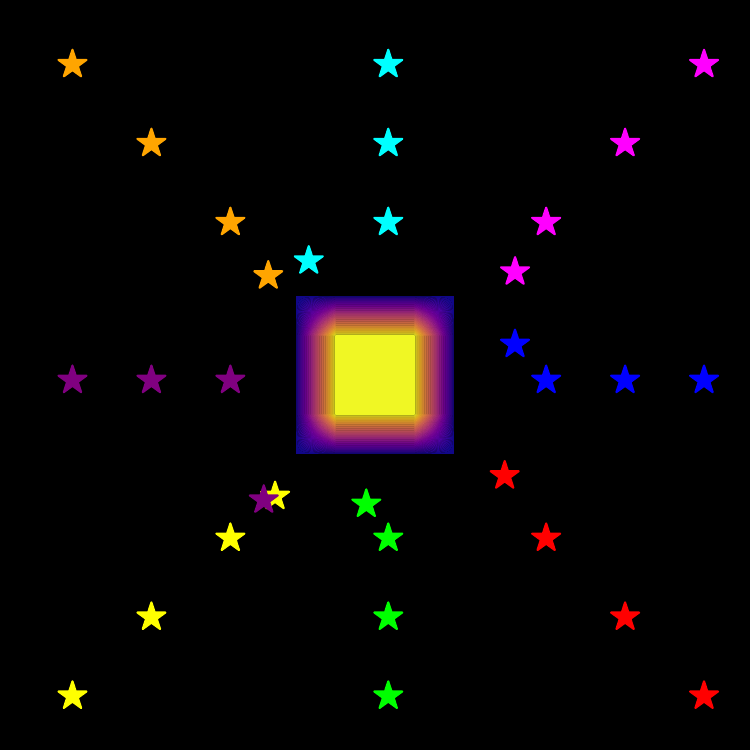}}
	\hfil
	\subfloat[C8]{\includegraphics[width=0.33\linewidth]{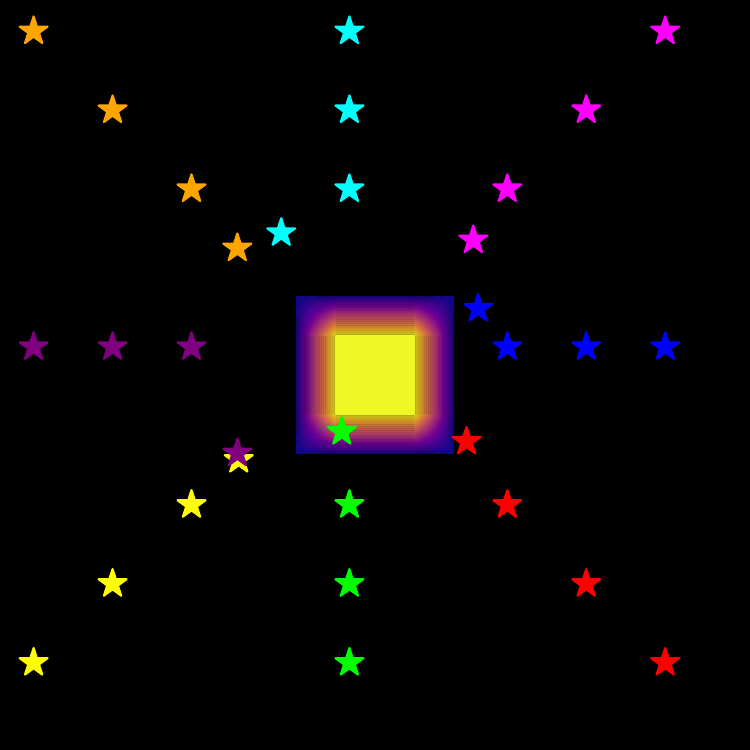}}
	\hfil
	\caption{Examples of sampling point distributions of the self-excited operator in the first encoder layer of SpecDETR on Fig.~\ref{fig:DETR_spod}(a). }
	\label{fig:feat2ndl1}
\end{figure}
\begin{figure}[h]
	\centering
	\subfloat[Background]{\includegraphics[width=0.33\linewidth]{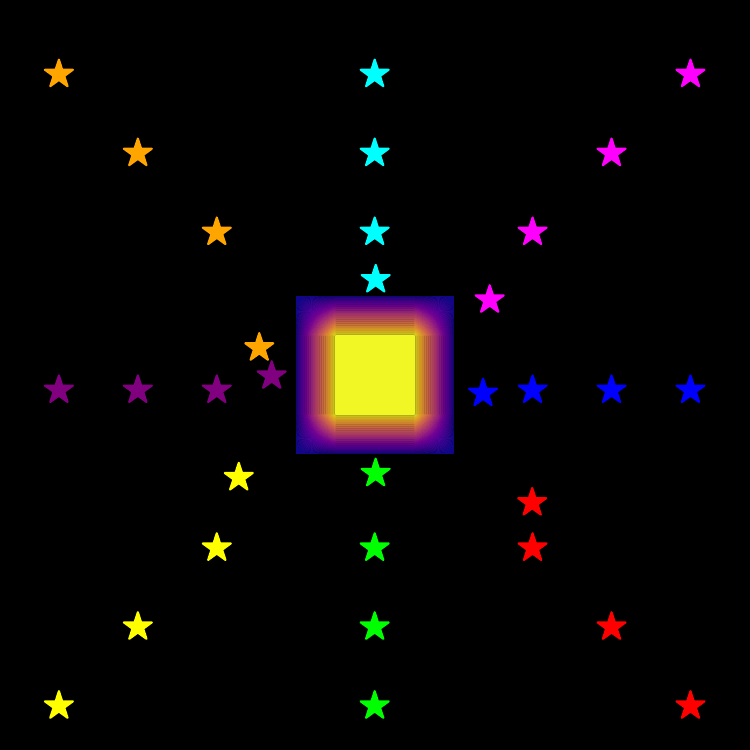}}
	\hfil
	\subfloat[C1]{\includegraphics[width=0.33\linewidth]{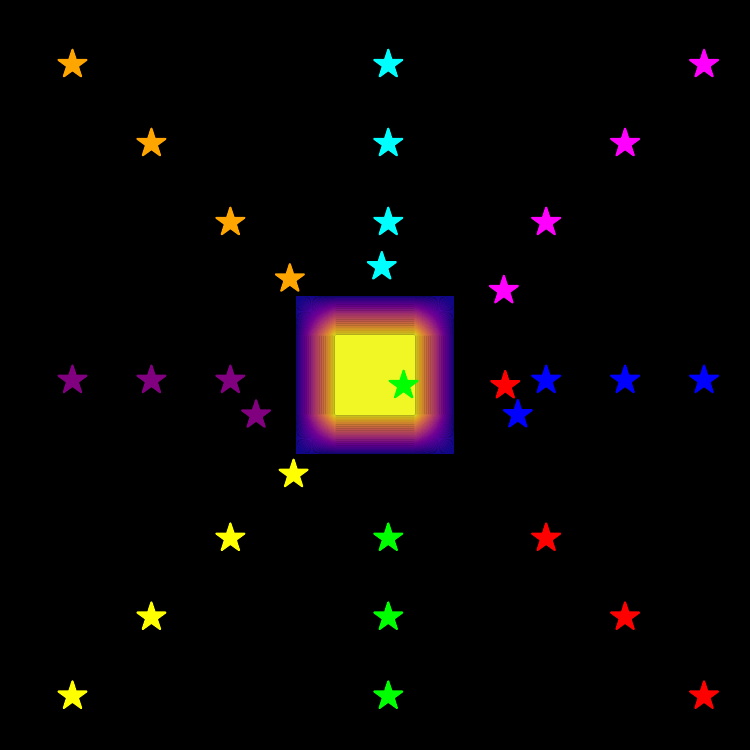}}
	\hfil
	\subfloat[C8]{\includegraphics[width=0.33\linewidth]{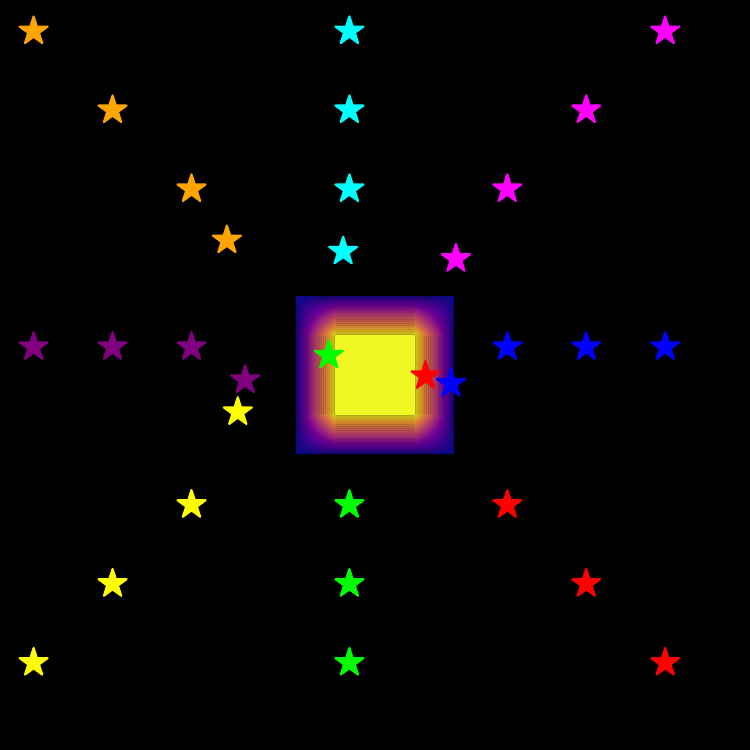}}
	\hfil
	\caption{Examples of sampling point distributions of the self-excited operator in the last encoder layer of SpecDETR on Fig.~\ref{fig:DETR_spod}(a).}
	\label{fig:feat2ndl6}
\end{figure}

\begin{figure}[h]
	\centering
	\subfloat[Background]{\includegraphics[width=0.33\linewidth]{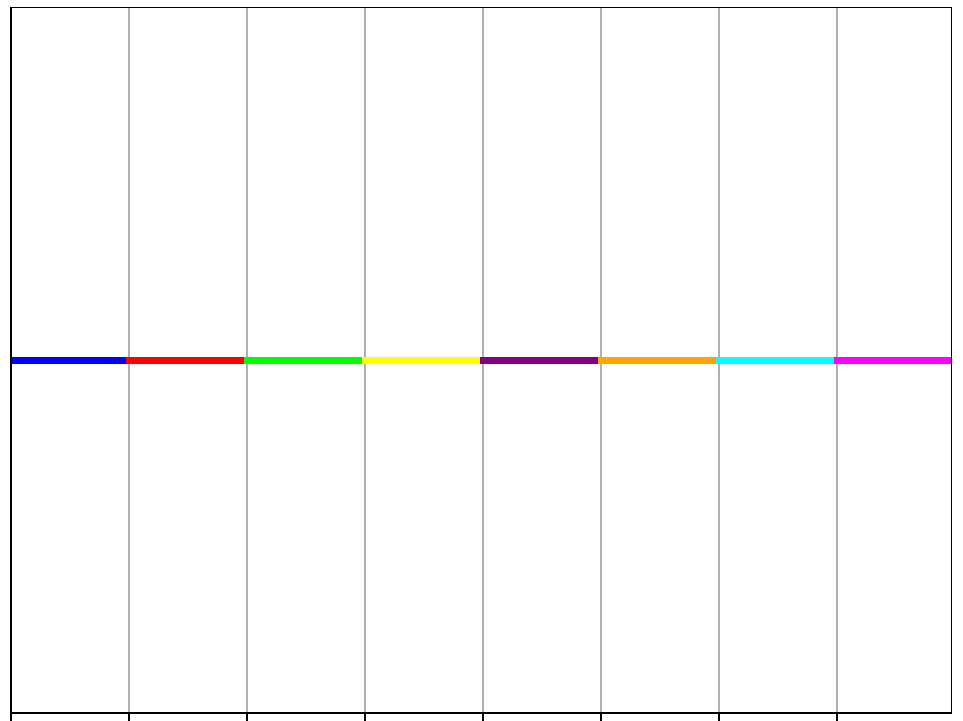}}
	\hfil
	\subfloat[C1]{\includegraphics[width=0.33\linewidth]{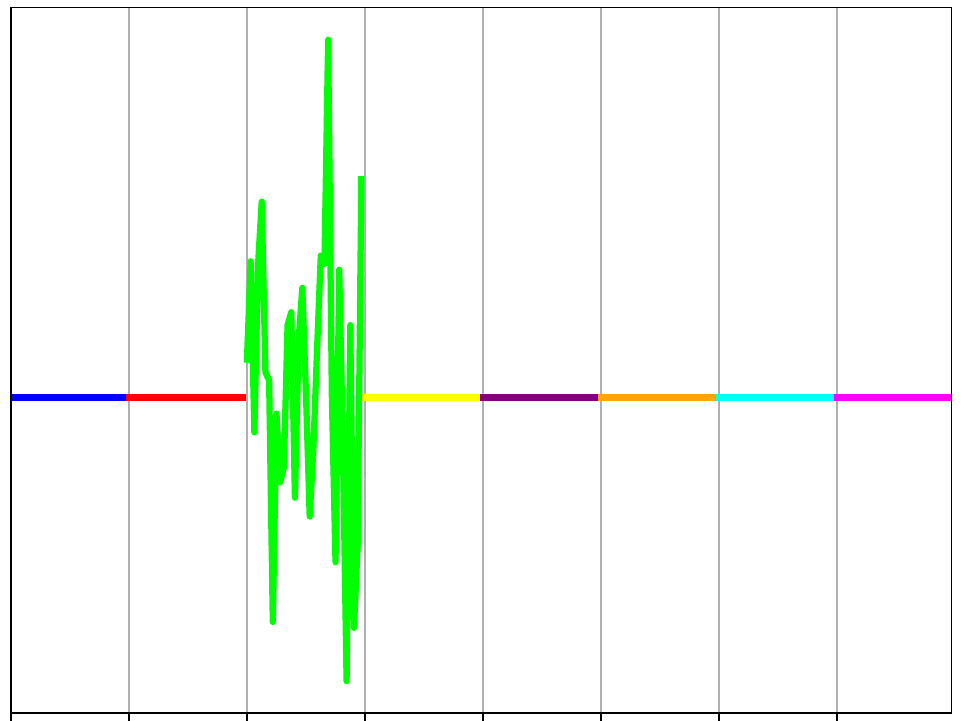}}
	\hfil
	\subfloat[C8]{\includegraphics[width=0.33\linewidth]{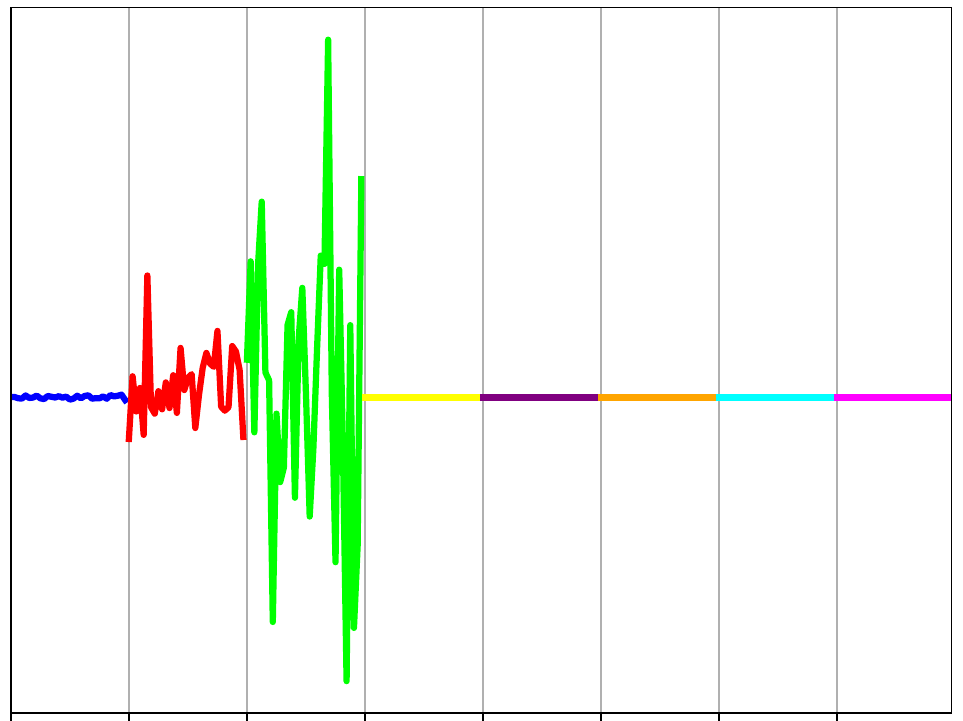}}
	\hfil
	\caption{Examples of sampling output of the self-excited operator in the last encoder layer of SpecDETR on Fig.~\ref{fig:DETR_spod}(a).}
	\label{fig:sp2ndl6}
\end{figure}

\begin{figure}[t]
	\centering
	\subfloat[]{\includegraphics[width=0.45\linewidth]{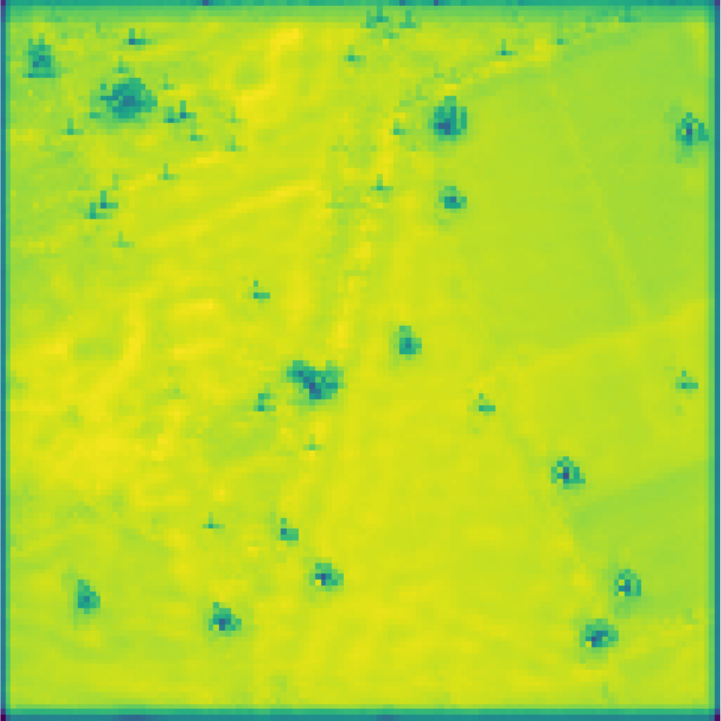}}
	\hfil
	\subfloat[]{\includegraphics[width=0.45\linewidth]{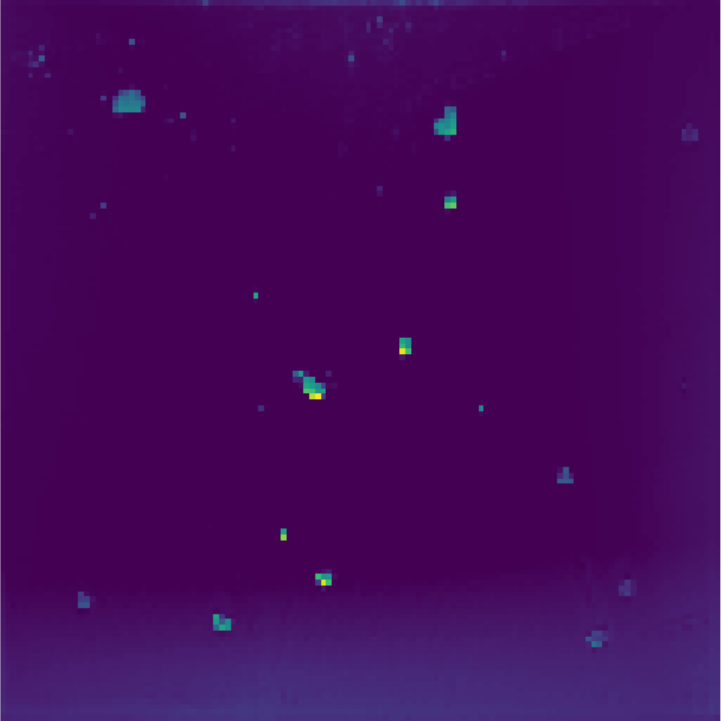}}
	\hfil
	\caption{Output feature maps of two deformable sampling operators from the S2A module in the last encoder layer of SpecDETR, corresponding to Fig.~\ref{fig:DETR_spod}(a). (a) The subpixel-scale deformable sampling operator. (b) The self-excited operator. Both (a) and (b) are the sum of the absolute values across all channels of the feature maps. }
	\label{fig:feat_2df}
\end{figure}

We visualize the sampling point distributions of the self-S2A module when using two object pixels and a background pixel as query elements in Fig.~\ref{fig:DETR_spod}(a). The two object pixels belong to C1 and C8, respectively, with the former being a low-abundance single-pixel object and the latter a high-abundance pixel within a multi-pixel object. The selected background pixel shares the same background component class as the C1 pixel,  resulting in similar spectral characteristics between the background pixel and the C1 pixel.
Fig.~\ref{fig:feat1stl1} and Fig.~\ref{fig:feat1stl6} visualize the sampling point distributions of the subpixel-scale deformable sampling operator in the first and last encoder layers, respectively. As shown in Fig.~\ref{fig:feat1stl1}, in the first encoder layer, the sampling point distributions for different types of query elements is largely consistent. Several sampling points are concentrated around the pixel location of the query element, while the remaining points are evenly distributed in the local spatial neighborhood.
In contrast, Fig.~\ref{fig:feat1stl6} illustrates that in the last encoder layer, the sampling point distribution varies significantly among different types of query elements. The sampling points for the C1 pixel are mainly focused around its own pixel location, whereas those for the C8 pixel are primarily distributed along the object boundary. This indicates that the pixel tokens updated in the penultimate encoder layer already possess distinct class information.
Fig.~\ref{fig:feat2ndl1} and Fig.~\ref{fig:feat2ndl6} depict the sampling point distributions of the self-excited operator in the first and last encoder layers, respectively. As depicted in Fig.~\ref{fig:feat2ndl1}, in the first encoder layer, the attention heads for both the C1 pixel and the background pixel remain inactive, while the green attention head for the C8 pixel is activated. This is due to the lower object abundance of the C1 pixel compared to the higher abundance of the C8 pixel.
However, as shown in Fig.~\ref{fig:feat2ndl6}, in the last encoder layer, the attention heads for the background pixel remain inactive, while both the C1 and C8 pixels have activated attention heads. This suggests that the preceding multiple encoder layers effectively extract the object features of C1.
The output vectors are generated through concatenation of the sampling vectors from different attention heads,  and the C1 and C8 pixels exhibit different combinations of activated attention heads. Consequently, the output vectors of the self-excited operator for the C1 and C8 pixels are significantly distinct, as illustrated in Fig.~\ref{fig:sp2ndl6}.
Fig.~\ref{fig:feat_2df} presents the sum of the absolute values across all channels of the output feature maps from these two sampling operators. In the output map of the self-excited operator, the background regions are either zero or close to zero, while high-response regions typically correspond to objects. This indicates that the self-excited operator helps object pixels to amplify the their own features.

\subsection{Ablation Experiments}

\subsubsection{Network Architecture}
\begin{table*}[t]
	\centering
	\caption{Performance Comparison Under Different Network Module Combinations on the SPOD Dataset.}
	\renewcommand\arraystretch{1.2}
	\scriptsize
	\setlength{\tabcolsep}{1mm}{
		\begin{tabular}{cccc|ccc|cc}
			\hline
			Image & \multirow{2}[0]{*}{Backbone} & \multirow{2}[0]{*}{Encoder} & \multirow{2}[0]{*}{Decoder} & \multicolumn{3}{c|}{mAP} & FLOPs & Params \\
			\cline{5-7}
			Size &       &       &       & 24 epochs & 36 epochs & 100 epochs & (G)   & (M) \\
			\hline
			$\times$4 & Swin-L & DINO  & DINO  & 0.391 & 0.605 & 0.757 & 203.9 & 218.3 \\
			$\times$1 &    -  & SpecDETR & DINO  & 0.663 & 0.759 & 0.849 & 143.6 & 17.9 \\
			$\times$1 &    -  & SpecDETR & SpecDETR & 0.706 & 0.799 & 0.856 & 139.7 & 16.1 \\
			\hline
		\end{tabular}%
	}
	\label{tab:decoder}%
\end{table*}%

DINO stands as one of the most successful object detectors to date. Co-DETR \cite{zong2023detrs}, which inherits the DINO decoder structure, achieves the highest mAP of 0.66 on the COCO test-dev dataset\footnote{https://paperswithcode.com/sota/object-detection-on-coco}. Our experiments on the SPOD dataset also demonstrate that DINO, equipped with Swin-L as the backbone network, significantly outperforms other compared object detectors in terms of detection accuracy.
SpecDETR builds upon DINO, specifically tailored for the point object detection task, by refining the feature extraction pattern and decoder structure. As shown in Table~\ref{tab:decoder}, we compare the performance of different network module combinations on the SPOD dataset to validate the effectiveness of our improvements.
DINO employs a backbone network to initially extract multi-scale feature maps from the input images, and then uses the Transformer encoder module for further feature extraction. In contrast, SpecDETR dispenses with the backbone network, directly utilizing the Transformer encoder to extract deep feature maps that match the input image size.
While maintaining the DINO decoder unchanged, replacing Swin-L and the DINO encoder with the SpecDETR encoder results in an mAP increase from 0.391 to 0.663 under 24 training epochs, and from 0.757 to 0.849 under 100 training epochs. Additionally, the number of parameters decreases from 218.3M to 17.9M. However, due to the absence of downsampling in the feature maps extracted by the SpecDETR encoder, the advantage in FLOPs is not as pronounced as that in parameters.
Furthermore, replacing the DINO decoder with the SpecDETR decoder also validates the superior performance of the SpecDETR decoder for the point object detection task through various quantitative metrics. Compared to the DINO decoder, the SpecDETR decoder achieves a 6.5\% and 5.3\% increase in mAP at 24 and 36 epoch settings, respectively. Meanwhile, there is a 2.7\% reduction in FLOPs and a 10.6\% decrease in parameters.

\subsubsection{Network Scale}

\begin{table}[t]
	\centering
	\caption{Performance Comparison of SpecDETR with Different Encoder and Decoder Layer Settings on the SPOD Dataset under 36 Training Epochs.}
	\renewcommand\arraystretch{1.2}
	\scriptsize
	\setlength{\tabcolsep}{1mm}{
		\begin{tabular}{cc|ccc}
			\hline
			Encoder & Decoder & \multirow{2}[0]{*}{mAP} & FLOPs & Params \\
			Layers & Layers &       & (G)   & (M) \\
			\hline
			\multirow{3}[0]{*}{2} & 2     & 0.470 & 48.9  & 5.7 \\
			& 4     & 0.577 & 54.0  & 8.4 \\
			& 6     & 0.579 & 59.0  & 11.2 \\
			\hline
			\multirow{3}[0]{*}{4} & 2     & 0.628 & 89.2  & 8.2 \\
			& 4     & 0.703 & 94.3  & 10.9 \\
			& 6     & 0.737 & 99.4  & 13.6 \\
			\hline
			\multirow{3}[0]{*}{6} & 2     & 0.704 & 129.6 & 10.6 \\
			& 4     & 0.750 & 134.7 & 13.4 \\
			& 6     & 0.799 & 139.7 & 16.1 \\
			\hline
		\end{tabular}
	}
	\label{tab:layernumber}%
\end{table}%

SpecDETR primarily consists of two modules: the Transformer encoder and decoder. In the default setting, we adopt the classical Transformer configuration, setting both the encoder and decoder layers to 6.
To explore the potential redundancy of each module, we compare the detection accuracy and complexity under different combinations of encoder and decoder layers in Table~\ref{tab:layernumber}.
When both the encoder and decoder layers are set to 2, SpecDETR achieves an mAP of 0.47 on the SPOD dataset under 36 training epochs.
Increasing either the encoder or decoder layers results in improved detection accuracy for SpecDETR. The gain in detection accuracy from increasing the encoder layers is greater than that from increasing the decoder layers.
SpecDETR with 2 encoder layers and 6 decoder layers achieves an mAP of 0.579, whereas SpecDETR with 6 encoder layers and 2 decoder layers achieves an mAP of 0.704.
However, the gain in accuracy from increasing encoder layers comes at a cost. The increase in encoder layers results in an almost linear increase in FLOPs, while the increase in encoder layers has a smaller impact on FLOPs. This is because the number of queries in the encoder is equal to the number of image pixels, whereas in the decoder, it is equal to the number of anchor boxes, with the former being significantly larger than the latter.
Additionally, increasing the number of decoder layers leads to a slightly higher increase in Params compared to encoder layers. This is due to the fact that while the Transformer compute operators in the SpecDETR encoder and decoder are largely consistent, each decoder layer includes an additional class prediction head and bbox regression head.

\subsubsection{Positive and Negative Sample Setting}
SpecDETR is a one-to-many set prediction problem, primarily reflected in the positive and negative sample setting, which includes the label assignment in the matching branch and the GT box noise addition in the denoising branch.
As shown in Table~\ref{tab:sample}, we investigate the impact of different positive and negative sample settings on the performance of SpecDETR. The label assignment considers one-to-one assignment (Forced Matching) and one-to-many assignment (Forced Matching + Dynamic Matching). For the denoising branch, we consider DN \cite{li2022dn}, CDN \cite{zhang2022dino}, our CCDN, and the case without denoising training.
Under the three denoising ways, SpecDETR performs significantly better than the case without denoising training.
This indicates that even when the matching part adopts one-to-many label assignment, SpecDETR still needs to rely on denoising training to accelerate convergence.
Under one-to-many matching, CCDN outperforms CDN and DN. Additionally, under one-to-one label assignment, CDN is lower than DN by 2.2\% and 1.7\% in AP at 36 epochs and 100 epochs, respectively, because CDN generates negative samples that are highly similar to the GT, which interferes with the classification head of SpecDETR.
Our one-to-many label assignment introduces more high-quality positive samples into the classification head from the matching branch. It improves detection performance regardless of whether DN, CDN, or CCDN is adopted.

\subsubsection{Attention Module}
As demonstrated in Table~\ref{tab:attention}, the removal of the self-excited operator leads to a significant decline in the detection performance of SpecDETR. Specifically, SpecDETR without the self-excited mechanism achieves only 0.228 mAP, 0.540 mAP, and 0.804 mAP at 24 epochs, 36 epochs, and 100 epochs, respectively, on the SPOD dataset. Furthermore, when SpecDETR operates without deformable sampling, the sampling points remain in their initial distribution and the self-excited mechanism is deactivated. Under this configuration, SpecDETR achieves 0.652 mAP at the 24-epoch setting on the SPOD dataset.
These results highlight two insights. First, the use of the subpixel-scale deformable sampling operator alone can disrupt the convergence of SpecDETR, indicating that deformable sampling, without proper mechanisms, may introduce instability. Second, the proposed self-excited mechanism plays a pivotal role in ensuring the convergence of SpecDETR, as it helps stabilize the training process when deformable sampling is employed. This underscores the importance of the self-excited mechanism in enhancing the robustness and performance of SpecDETR.

\subsubsection{Content Query Initialization}
As shown in Table~\ref{tab:query}, we explore the impact of different content query initialization for SpecDETR. ``Constant" represents initializing all content queries as all one vectors, ``Random" represents initializing queries with random values each time, and ``Embedding" represents the index embedding in DINO \cite{zhang2022dino}.
SpecDETR converges fastest with the constant initialization. 
At 24 epochs, the constant initialization achieved an mAP that is 9.1\% higher than the embedding initialization and 15.4\% higher than the random initialization.
DINO is regarded as a one-to-one set prediction problem, requiring embedding initialization for content queries to differentiate them.
We consider SpecDETR as a one-to-many set prediction problem, so the initialization of content queries should be consistent.

\begin{table}[t]
	\centering
	\caption{Performance Comparison of SpecDETR with Different Positive and Negative Sample Settings on the SPOD Dataset.}
	\renewcommand\arraystretch{1.2}
	\scriptsize
	\setlength{\tabcolsep}{1mm}{
		\begin{tabular}{ccccc}
			\hline
			\multirow{2}[2]{*}{DeNosing} & \multicolumn{2}{c}{Label Assigner} & \multicolumn{2}{c}{mAP} \\
			\cmidrule(r){2-3}	\cmidrule(r){4-5}        & Forced Matching & Dynamic Matching & 36epochs  & 100epochs  \\
			\hline
			\multirow{2}[0]{*}{\ding{55}} & \ding{51}   &       & 0.611 & 0.796 \\
			& \ding{51}   & \ding{51}   & 0.595 & 0.800 \\
			\hline
			\multirow{2}[0]{*}{DN\cite{li2022dn}} & \ding{51}   &       & 0.773 & 0.836 \\
			& \ding{51}   & \ding{51}   & 0.785 & 0.844 \\
			\hline
			\multirow{2}[0]{*}{CDN\cite{zhang2022dino}} & \ding{51}   &       & 0.756 & 0.822 \\
			& \ding{51}   & \ding{51}   & 0.791 & 0.846 \\
			\hline
			\multirow{2}[0]{*}{CCDN} & \ding{51}   &       & 0.780 & 0.829 \\
			& \ding{51}   & \ding{51}   & 0.799 & 0.856 \\
			\hline
		\end{tabular}%
	}
	\label{tab:sample}%
\end{table}%

\begin{table}[t]
	\centering
	\caption{Performance Comparison of SpecDETR with Variants of the S2A Module on the SPOD Dataset. \ding{52}\rotatebox[origin=c]{-9.2}{\kern-0.7em\ding{55}} denotes cases where the structure exists but fails to perform its intended function.}
	\renewcommand\arraystretch{1.2}
	\scriptsize
	\setlength{\tabcolsep}{2mm}{
		\begin{tabular}{ccccc}
			\hline
			Self-excited  & Deformable  & \multicolumn{3}{c}{mAP} \\
			\cline{3-5}  operator & sampling  & 24epochs  & 36epochs  & 100epochs  \\
			\hline
			\ding{55}   & \ding{51}   & 0.228 & 0.540  & 0.804 \\
			\ding{52}\rotatebox[origin=c]{-9.2}{\kern-0.7em\ding{55}}    &  \ding{55}   & 0.652 & 0.748 & 0.842 \\
			\ding{51}   &  \ding{51}   & 0.706 & 0.799 & 0.856 \\
			\hline
		\end{tabular}%
	}
	\label{tab:attention}%
\end{table}%

\begin{table}[t]
	\centering
	\caption{AP Performance Comparison of SpecDETR with Different Content Query Initialization on the SPOD Dataset.}
	\renewcommand\arraystretch{1.2}
	\scriptsize
	\setlength{\tabcolsep}{2mm}{
		\begin{tabular}{c|ccc}
			\hline
			Content Query	& 24epochs & 36epochs & 100epochs \\
			\hline
			Embedding & 0.647 & 0.771 & 0.845 \\
			Random & 0.612 & 0.790 & 0.855 \\
			Constant   & 0.706 & 0.799 & 0.856 \\
			\hline
		\end{tabular}%
	}
	\label{tab:query}%
\end{table}%

\section{Conclusion}
\label{sec:Cnl}

In this paper, we extend the hyperspectral target detection task to the hyperspectral point object detection task,
and propose the first specialized hyperspectral point object detection network,  termed SpecDETR. Unlike the traditional HTD methods that rely on a per-pixel binary classification framework, our proposed SpecDETR is capable of learning instance-level  spatial-spectral joint object feature representations from large-scale training images, and can directly provide end-to-end instance-level predictions of object location and classification. Extensive experiments verifies that our SpecDETR significantly outperforms both the SOTA visual object detection networks and HTD methods in  hyperspectral object detection. However, our framework necessitates a substantial amount of labeled training data, which poses a challenge due to the difficulty in annotating point objects. To address this, we employ simulation techniques to construct synthetic training datasets for single-spectral point objects within three publicly HTD datasets. SpecDETR, trained on these simulation-based datasets, demonstrates robust detection performance for real-world multi-pixel single-spectral objects. Nevertheless, its performance diminishes when confronted with occlusions such as trees or shadows, or when additional noise is artificially introduced into the test images. We attribute this decline in performance to the domain gap between the training and test data. In the future, we plan to explore cross-domain few-shot object detection within the hyperspectral point object detection task, a direction we believe holds potential for mitigating the detection performance degradation  caused by the domain gap and  the dilemma of insufficient real point object training samples.

\ifCLASSOPTIONcaptionsoff
\newpage
\fi


\bibliographystyle{IEEEtran}
\bibliography{IEEEabrv, AETNET, CSRBBH-TD, main}

\end{document}


\title{Supplementary Materials for \\
	SpecDETR: A Transformer-based Hyperspectral Point Object Detection Network}
\author{Zhaoxu Li, Wei An, Gaowei Guo, Longguang Wang, Yingqian Wang, and Zaiping Lin  
}

\markboth{}{}


\maketitle

\IEEEpeerreviewmaketitle

\section{Supplements to the SPOD Dataset}
\label{sec:Cnl}
\subsection{Spectral Fluctuation Model}

To simulate spectral fluctuations more realistically, we modeled the spectral fluctuations of real hyperspectral data captured by the AVIRIS sensor. 
We chose these water areas as the modeling objects because land background pixels are often mixed, while water areas such as oceans and lakes have a single spectral composition.

\begin{figure}[h]
	\centering
	\includegraphics[width=0.5\linewidth]{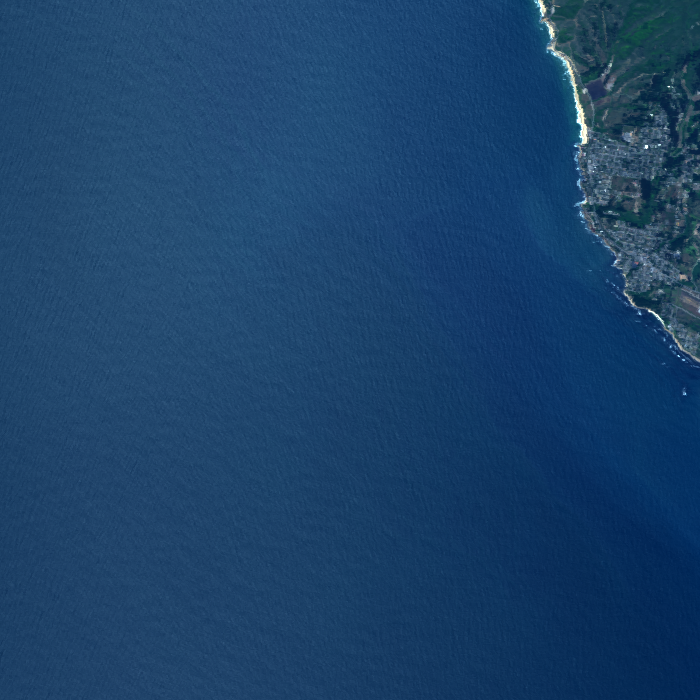}
	\hfil
	\caption{Golden Gate.} 
	\hfil
	\label{fig:GoldenGate}
\end{figure}

As shown in Fig.~\ref{fig:GoldenGate}, a $880\times460$ region of seawater in the Golden Gate, California, is selected as the reference area for analysis. 
First, the coefficient of variation of the radiances in each band is calculated for the reference area:

\begin{equation}
	\gamma _i=\frac{\mu _i}{\sigma _i}
\end{equation}
where $\gamma _i$ represents the coefficient of variation of the radiances in the $i$-th band in the reference area, $\mu _i$ represents the mean of the radiances in the $i$-th band, and $\sigma _i$ represents the standard deviation of  the radiances  in the $i$-th band.

\begin{figure}[h]
	\centering
	\includegraphics[width=0.8\linewidth]{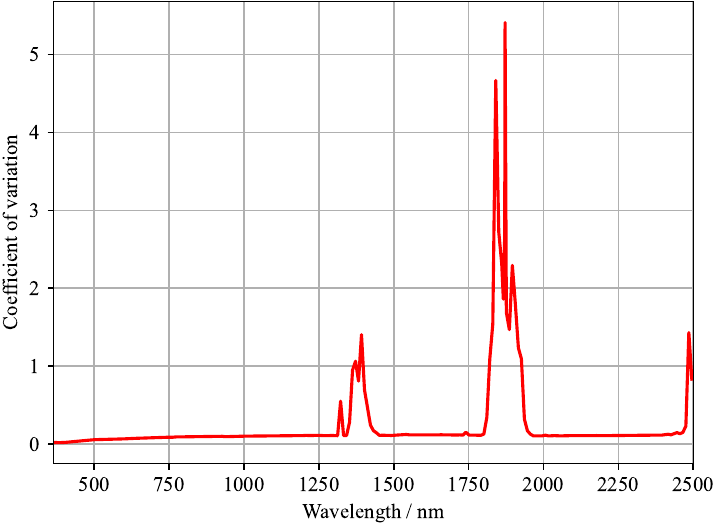}
	\vspace{0cm}
	\caption{The coefficient of variation curve of the reference area.} 
	\label{fig:CV}
\end{figure}

Fig.~\ref{fig:CV} shows the coefficient curve of variation in the reference area with wavelength. 
It can be observed that except for the atmospheric absorption band at 1400nm, 1800nm, and 2500nm, the coefficient of variation in the reference area is continuous and smooth.
For adjacent pixels of the same category on the same image, their imaging conditions, such as observation angle and lighting, can be considered consistent
To describe the fluctuations of  these pixels of the same category in a  small-scale area, we introduce a local fluctuation factor:
\begin{equation}
	\alpha _{i}^{j}=\frac{s_{i}^{j}-\mu _i}{\mu _i}
\end{equation}
where $s_{i}^{j}$ represents the radiance value, and $\alpha _{i}^{j}$ represents the local fluctuation factor of the $j$-th pixel in the $i$-th band within the reference area.

\begin{figure}[h]
	\centering
	\includegraphics[width=0.8\linewidth]{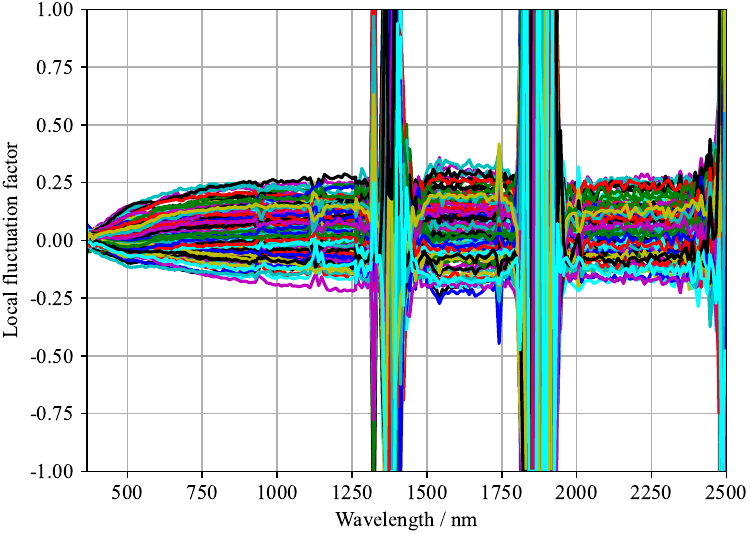}
	\vspace{0cm}
	\caption{The local fluctuation factor curves in the reference area.} 
	\label{fig:local_fluctuation_factor}
\end{figure}

Fig.~\ref{fig:local_fluctuation_factor} shows the variation of the local fluctuation factor of different pixels in the reference area with wavelength. 
It can be observed that except for the atmospheric absorption bands at 1400nm, 1800nm,  and 2500nm,  the fluctuation level of the local fluctuation factor increases with increasing wavelength, similar to the coefficient curve of variation. Therefore, the standardized local fluctuation factor is defined as:
\begin{equation}
	\bar{\alpha}_{i}^{j}=\frac{\alpha _{i}^{j}}{\gamma _i}
\end{equation}
where $\bar{\alpha}_{i}^{j}$ represents the standardized local fluctuation factor of the $j$-th pixel in the $i$-th band in the reference area.

\begin{figure}[h]
	\centering
	\includegraphics[width=0.8\linewidth]{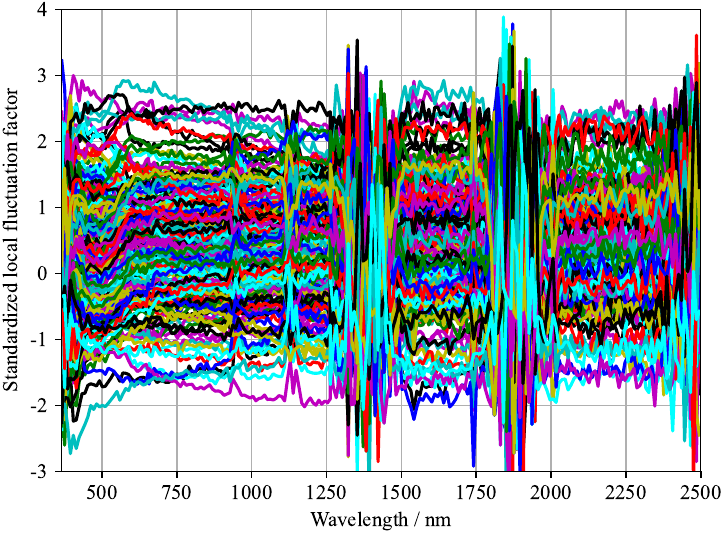}
	\caption{The standardized local fluctuation factor curves in the reference area.} 
	\label{fig:standardized_local_fluctuation_factor}
\end{figure}

Fig.~\ref{fig:standardized_local_fluctuation_factor}  shows the standardized local fluctuation factor variation of different pixels in the reference area with wavelength. 
It can be observed that except for the atmospheric absorption bands, the curves of the standardized local fluctuation factor in the reference area are relatively stable in the visible to short-wave infrared spectral range. Hence, the standardized local fluctuation factor can be modeled as:
\begin{equation}
	\bar{\alpha}_{i}^{j}=a^j+\upsilon _{i}^{j}
\end{equation}
where $a^j$ represents the baseline value of the standardized local fluctuation factor for the $j$-th pixel in the reference area, and $\upsilon _{i}^{j}$ represents the noise of the standardized local fluctuation factor for the $j$-th pixel in the $i$-th band. The baseline value $a^j$ can be approximated by the mean of the standardized local fluctuation factors in each band.

\begin{figure}[h]
	\centering
	\includegraphics[width=0.8\linewidth]{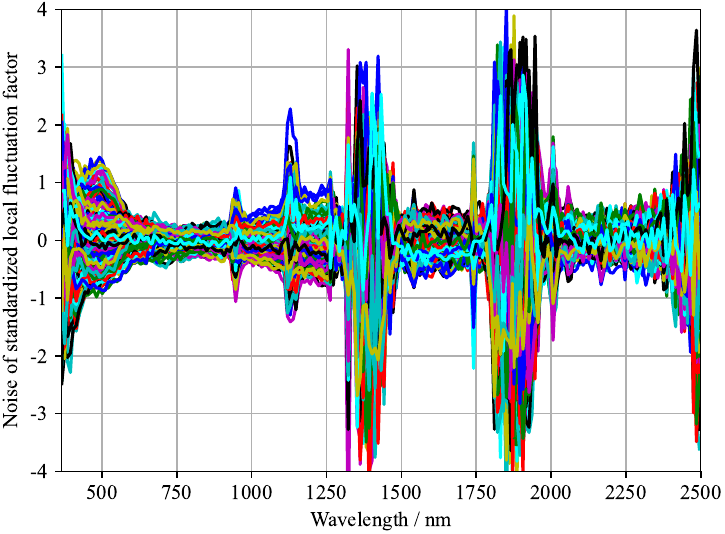}
	\caption{The noise curves of the standardized local fluctuation factor in the reference area.} 
	\label{fig:noise_standardized_local_fluctuation_factor}
\end{figure}
Fig.~\ref{fig:noise_standardized_local_fluctuation_factor}  shows the noise curves of the standardized local fluctuation factor for different pixels in the reference area. 
It can be observed that the noise intensity of the standardized local fluctuation factor is greatly influenced by wavelength.

\begin{figure}[h]
	\centering
	\includegraphics[width=0.8\linewidth]{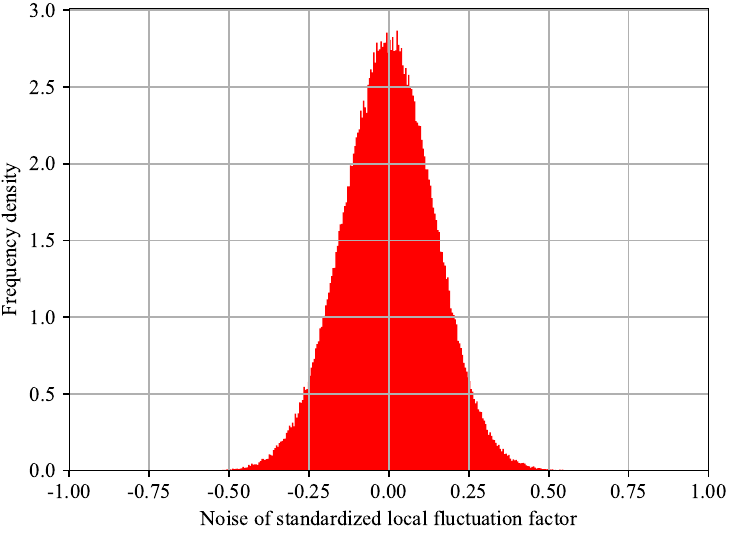}
	\caption{The noise distribution of the standardized local fluctuation factor at  654nm in the reference area.} 
	\label{fig:654nm_SH_NSLFF}
\end{figure}
\begin{figure}[h]
	\centering
	\includegraphics[width=0.8\linewidth]{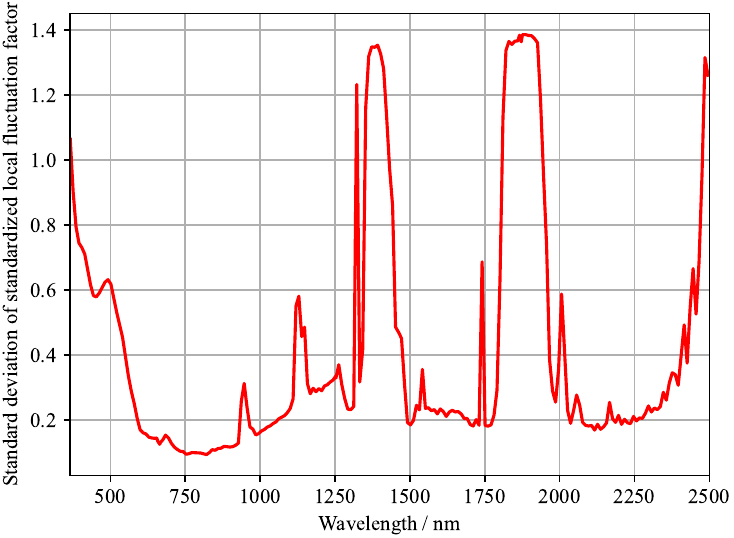}
	\vspace{0cm}
	\caption{The standard deviation curves of the standardized local fluctuation factor in the reference area.} 
	\label{fig:STD_standardized_local_fluctuation_factor}
\end{figure}

Fig.~\ref{fig:654nm_SH_NSLFF} displays the noise distribution of the standardized local fluctuation factor at  654nm in the reference area. 
It can be observed that at a specific wavelength, the noise of the standardized local fluctuation factor can be considered as Gaussian noise.
Fig.~\ref{fig:STD_standardized_local_fluctuation_factor}  shows the standard deviation variation of the standardized local fluctuation factor in the reference area with wavelength.

\begin{figure}[h]
	\centering
	\includegraphics[width=0.8\linewidth]{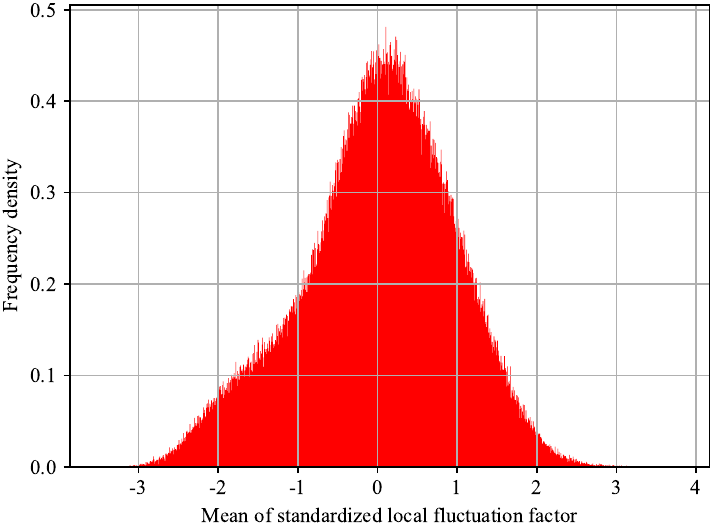}
	\vspace{0cm}
	\caption{The  baseline value  distribution of the standardized local fluctuation factor for different pixels in the reference area.} 
	\label{fig:SH_MSLFF}
\end{figure}

From Fig.~\ref{fig:SH_MSLFF}, it can be concluded that the baseline value of the standardized local fluctuation factor in the reference area can be approximated as a normal distribution with a mean of 0.
Based on the above analysis, in a small-scale area on the same image, the spectra of adjacent pixels of the same category  in non-atmospheric-absorption bands can be modeled numerically as:
\begin{equation}
	\label{eql}
	\boldsymbol{s}\approx \boldsymbol{\gamma }\circ \left( a+\boldsymbol{\upsilon } \right)\circ \bar{\boldsymbol{s}}+\bar{\boldsymbol{s}}
\end{equation}
where $\circ$ denotes element-wise multiplication, $\boldsymbol{s}$ represents the observed spectral curve of a certain pixel, $\bar{\boldsymbol{s}}$ represents the baseline spectral curve under the current observation conditions and can be approximated by the average radiance vector $\boldsymbol{\mu }$,  $\boldsymbol{s}$ and $\bar{\boldsymbol{s}}$ are both $N\times 1$ column vectors, where $N$ is the number of bands. $a$ represents the baseline value of the standardized local fluctuation factor, which follows a normal distribution with mean 0 and standard deviation $\sigma _a$. $\boldsymbol{\upsilon }$ represents the noise of the standardized local fluctuation factor, which follows an $N$-dimensional normal distribution with mean 0 and standard deviation vector $\boldsymbol{\sigma }_{\upsilon}$.

\begin{figure}[h]
	\label{fig:6water}
	\centering
	\subfloat[]{\includegraphics[width=0.3\linewidth]{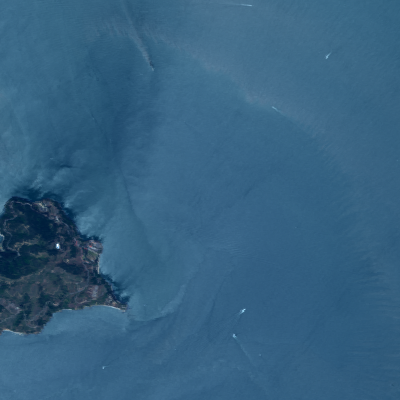}}
	\hfil
	\subfloat[]{\includegraphics[width=0.3\linewidth]{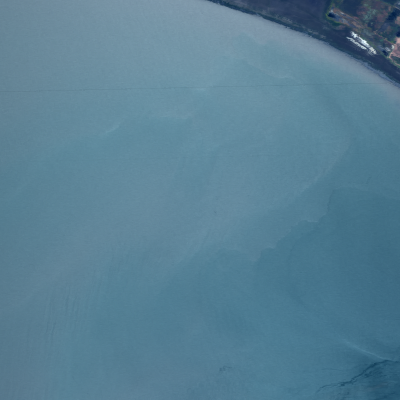}}
\hfil
	\subfloat[]{\includegraphics[width=0.3\linewidth]{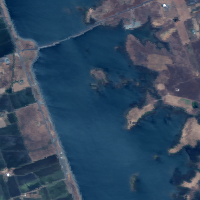}}
\hfil
\\
	\subfloat[]{\includegraphics[width=0.3\linewidth]{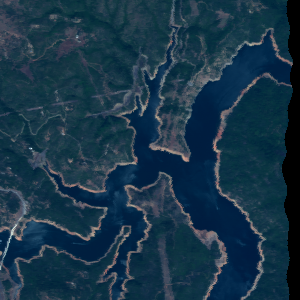}}
\hfil
	\subfloat[]{\includegraphics[width=0.3\linewidth]{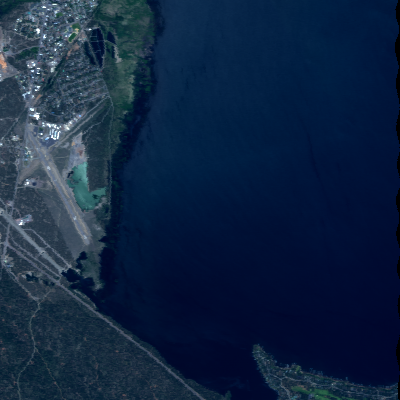}}
\hfil
	\subfloat[]{\includegraphics[width=0.3\linewidth]{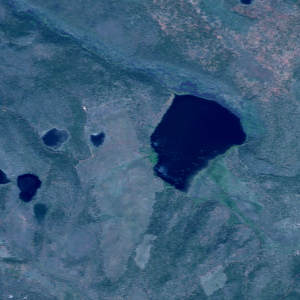}}
\hfil
	\caption{Six water areas near the Golden Gate. (a) San Francisco Bay. (b) San Pablo Bay. (c) Thermalito Afterbay. (d) West Branch Feather River. (e) Lake Almanor. (f) Big Jack Lake.  }
\end{figure}

Based on the local fluctuation characteristics,  we further analyzed the wide-area spectral fluctuation characteristics of six different water areas shown in Fig.~\ref{fig:6water}. 
These water areas are far apart with longer imaging time intervals, resulting in differences in imaging conditions such as observation distance, observation angle, and lighting.
To describe the fluctuations of  spectra  of the same category under different imaging conditions, we introduce a wide-area fluctuation factor:
\begin{equation}
	\beta _{i}^{k}=\frac{\mu _{i}^{k}-\mu _i}{\mu _i}
\end{equation}
where $\beta _{i}^{k}$ represents the wide-area fluctuation factor of the $k$-th water area in the $i$-th band, $\mu _{i}^{k}$ represents the average radiance of the $k$-th water area in the $i$-th band,
and $\mu _{i}$ represents the average radiance of the reference area in the $i$-th band.

\begin{figure}[h]
	\label{fig:Wide-area-fluctuation-factor}
	\centering
	\includegraphics[width=0.8\linewidth]{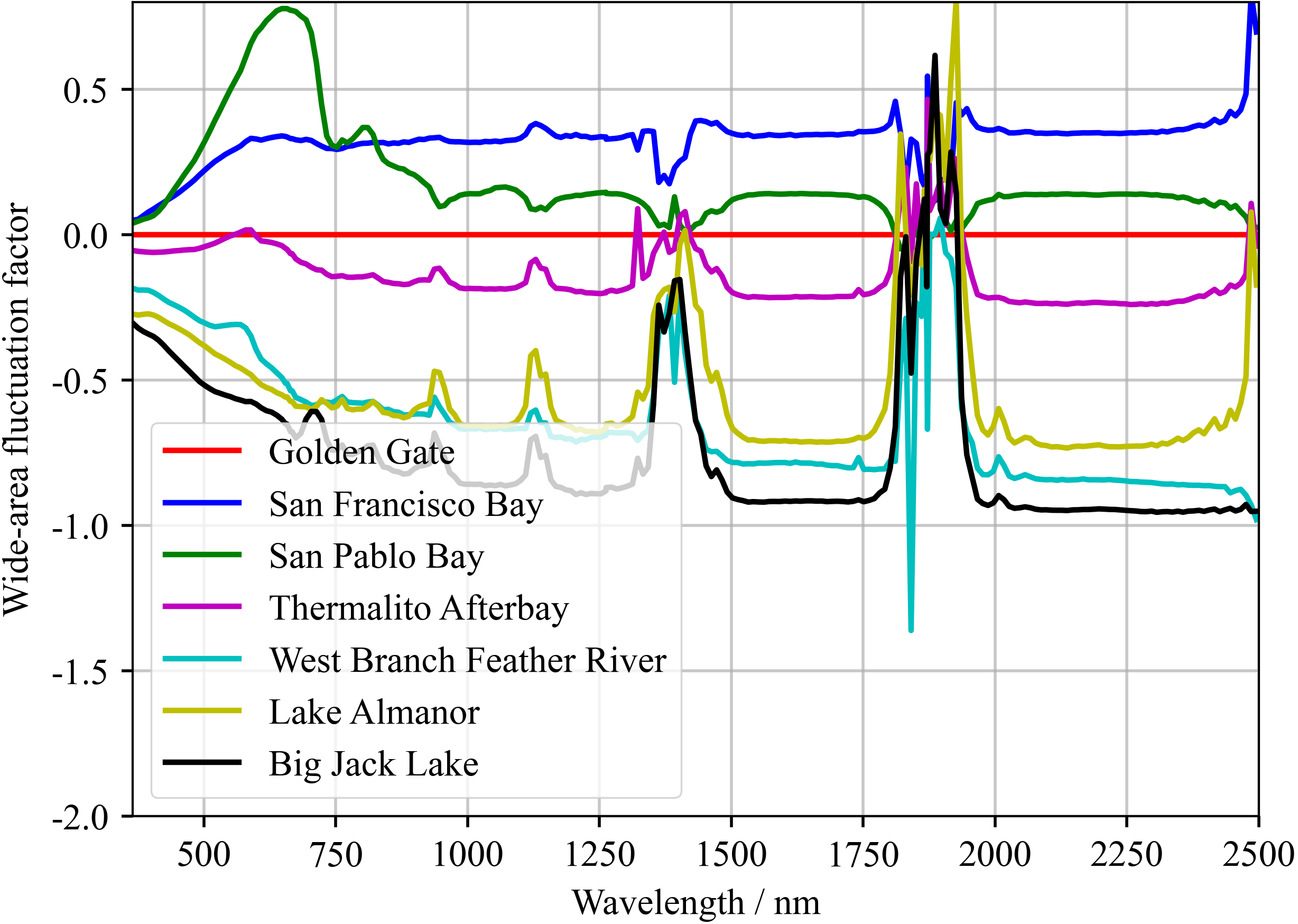}
	\caption{The wide-area fluctuation factor curves of different water areas.} 
\end{figure}

Fig.~\ref{fig:Wide-area-fluctuation-factor} shows the variation of the wide-area fluctuation factor of different water areas with wavelength.
It can be observed that, except for the atmospheric absorption bands at 1400nm, 1800nm, and 2500nm, the wide-area fluctuation factor exhibits a similar trend to the local fluctuation factor.
Based on this, we define a standardized wide-area fluctuation factor as:
\begin{equation}
	\bar{\beta}_{i}^{k}=\frac{\beta _{i}^{k}}{\gamma _i}
\end{equation}
where $\bar{\beta}_{i}^{k}$ represents the standardized wide-area fluctuation factor of the $k$-th water area in the $i$-th  band.

\begin{figure}[h]
	\centering
	\includegraphics[width=0.8\linewidth]{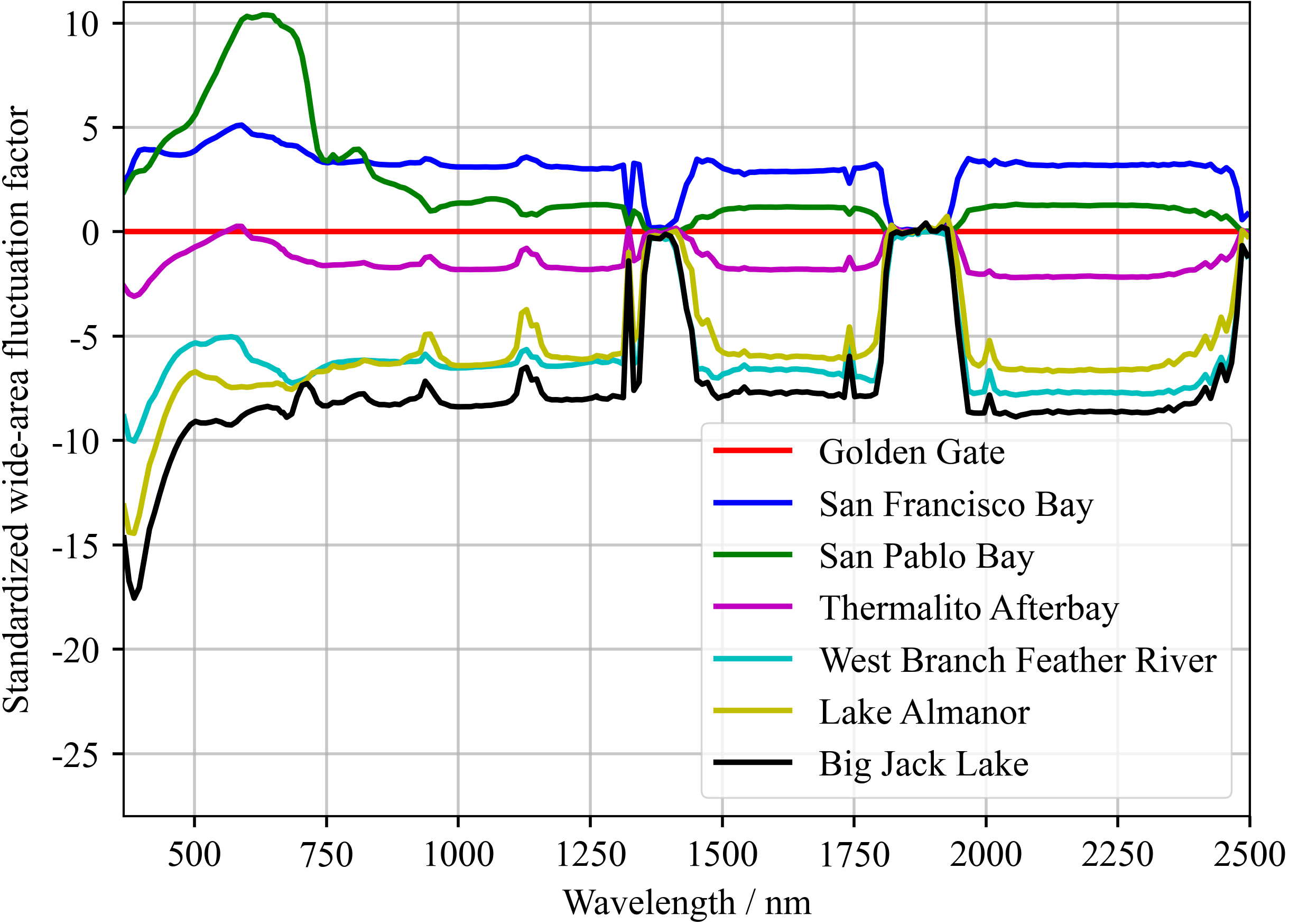}
	\vspace{0cm}
	\caption{The standardized wide-area fluctuation factor curves of different  water areas.} 
	\label{fig:Standardized_wide-area_fluctuation_factor}
\end{figure}

Fig.~\ref{fig:Standardized_wide-area_fluctuation_factor}  shows the variation of the standardized wide-area fluctuation factor of different water areas with wavelength. The standardized wide-area fluctuation factor of different water areas remains stable on the near-infrared to short-wave infrared spectral range outside the atmospheric absorption bands. 
However, in the visible spectral range, some water areas exhibit more significant fluctuations in the wide-area fluctuation factor.
This is because the reflectivity of water areas in the visible spectral range is greatly influenced by the chlorophyll content, which varies among different water areas.
In the near-infrared and short-wave infrared spectral range, the influence of chlorophyll on spectral reflectivity is minimal, and these water areas have similar spectral reflection characteristics.
By ignoring factors such as chlorophyll in the near-infrared to short-wave infrared spectral range outside the atmospheric absorption bands, the average spectral curve of the $k$-th water area can be approximated as:

\begin{equation}
	\label{eqw}
	\boldsymbol{\mu }^k\approx b^k\boldsymbol{\gamma }\circ \boldsymbol{\mu }+\boldsymbol{\mu }
\end{equation}
where $b^k$ represents the baseline value of the standardized wide-area fluctuation factor for the $k$-th water area.

Combining \eqrefnew{eql} and \eqrefnew{eqw}, the fluctuations of  spectra of the same category  in non-atmospheric-absorption bands can be modeled as:

\begin{equation}
	\boldsymbol{s}=\left( b\boldsymbol{\gamma }+1 \right)\circ \left( \left( a+\boldsymbol{\upsilon } \right)\circ \boldsymbol{\gamma }+1 \right)\circ \bar{\boldsymbol{s}}
\end{equation}
where $\boldsymbol{s}$ represents the measured spectral curve, $\bar{\boldsymbol{s}}$ represents the average spectral curve of the reference area, $\gamma $ represents the coefficient of variation curve of the reference area. $b$ represents the standardized wide-area fluctuation factor, and $a$ represents the baseline value of the standardized local fluctuation factor, which follows a normal distribution with mean 0 and standard deviation $\sigma _a$. 
$\boldsymbol{\upsilon }$ represents the noise of the standardized local fluctuation factor, which follows an $N$-dimensional normal distribution with mean 0 and standard deviation $\boldsymbol{\sigma }_{\upsilon}$.

\subsection{Supplements to Implementation Details}
We use the  AVIRIS  raw data that has not been corrected with radiance gains to generate the SPOD dataset.
To ensure the discriminability of simulated object spectra in the SPOD dataset, we select the spectral reflectance curves of 12 artificial materials from the USGS spectral library to generate simulated spectral radiance curves.
The spectral reflectance curve refers to the ratio of the light flux reflected by an object to the light flux incident on the object. 
Accurately simulating radiance curves from spectral reflectance curves requires careful consideration of the entire process of light propagation in the atmosphere, including factors such as atmospheric absorption, atmospheric scattering, and surface reflection.
We only need to validate the methods for hyperspectral object detection, so it is unnecessary to precisely approximate the actual radiance of specific land cover types for the simulated object spectra.
Therefore, we simplify the simulation process by treating the transformation from spectral reflectance curves to radiance curves as a linear transformation. 
The simulated radiance curves can be obtained by :
\begin{equation}
	\boldsymbol{s}_t=\frac{\boldsymbol{s}_w}{\boldsymbol{r}_w}\circ \boldsymbol{r}_t
\end{equation}
where $\boldsymbol{s}_t$ represents the simulated object radiance curve, $\boldsymbol{s}_w$ represents the average radiance curve of the reference water area, $\boldsymbol{r}_w$ is the spectral reflectance curve of the seawater provided by the USGS spectral library, and $\boldsymbol{r}_t$ stands for the spectral reflectance curve of selected artificial materials provided by the USGS spectral library. As 
$\boldsymbol{r}_w$ is the water first surface reflection curve and has relatively low values, the values of $\boldsymbol{s}_t$ are excessively high. 
Therefore, linear scaling is required to map the numerical range of $\boldsymbol{s}_t$ to the normal range of the AVIRIS  data.
The simulated object spectra can be added to the hyperspectral images by :
\begin{equation}
	\boldsymbol{s}=\frac{M_t}{\max \left( \boldsymbol{s}_t \right)}\left( b\boldsymbol{\gamma }+1 \right)\circ \left( \left( a+\boldsymbol{\upsilon } \right)\circ \boldsymbol{\gamma }+1 \right)\circ \boldsymbol{s}_t
\end{equation}
where $M_t$ represents the baseline of the maximum value of the simulated object spectra. 
Considering the numerical range of the  AVIRIS data used in the SPOD dataset, The value range of $M_t$ for each type of simulated spectrum is $\left[ 2000, 3000 \right]$, which corresponds to the distribution range of maximum digital numbers for most spectra in the AVIRIS data. 
$\boldsymbol{\gamma}$, $\sigma _a$, and $\sigma _\upsilon$ are the statistical values of the reference water area.
It can be assumed that their observation conditions are consistent for different pixels of the same object, and  $b$  is the same for all of them.
We set $b$ to follow a uniform distribution in the range $[-0.3, 0.3]$, ensuring significant spectral fluctuations between the same type of objects.

\section{Supplements to Experiments on the SPOD Dataset}

Tables.~\ref{tabspodnet_apresult} and \ref{tabspodnet_arresult}  and serve as the supplements to Table~\uppercase\expandafter{\romannumeral2} by providing AP and AR performance of more object detection networks \cite{kim2020probabilistic,zhou2019objects,tian2019fcos,zhou2019objects,zhang2019freeanchor,zhang2020bridging,sun2021sparse,carion2020end,meng2021conditional,liu2022dab} on the SPOD dataset, respectively. 
All object detection networks use parameter configurations of the COCO dataset \cite{lin2014microsoft} from the MMDetection framework \cite{chen2019mmdetection}, with variations limited to training epochs, image sizes, and input channels. 
Due to the domain gap between the SPOD and COCO datasets, many networks face challenges in detecting point objects in the SPOD dataset. 
However, our SpecDETR, as well as other networks such as DINO and CentripetalNet, demonstrate the feasibility of point object detection.
In addition, SpecDETR performs the best across all APs. 
Despite not providing redundant predictions, which can significantly impact AR performance, the mAR of SpecDETR is second only to that of DINO.

\section{Experimental Setup for Compared HTD Methods}

The compared HTD methods, LSSA \cite{zhu2023learning}, IRN \cite{shen2023hyperspectral}, and TSTTD \cite{jiao2023triplet}, are adapted based on their official codes to suit the characteristics of our dataset. Among these, IRN and TSTTD are implemented in Python, while LSSA is implemented in MATLAB. The original LSSA takes a single target spectrum as input and outputs one abundance map, whereas the improved LSSA processes multiple target spectra simultaneously and output multiple abundance maps for different target classes. The original IRN and TSTTD utilize background pixels from a single test image for training, while the improved versions use background pixels from all training images. Additionally, the original TSTTD employs a single target prior spectrum for data augmentation, whereas the improved TSTTD directly uses all target pixels from the training images for training. Other parameters are retained at their default settings as provided in the official implementations.

\begin{table*}[!th]
	\caption{Simulation Parameter Settings for Training Sets of the Avon, SanDiego, and Gulfport Datasets.}
	\label{tab:HTDParam}
	\begin{center}
		\renewcommand\arraystretch{1.0}
    \begin{tabular}{c|cccccccc|cc}
    		\hline
	Dataset & \multicolumn{8}{c|}{SPOD Dataset}                             & \multicolumn{2}{c}{Avon Dataset} \\
	\hline
	Object type & C1    & C2    & C3    & C4    & C5    & C6    & C7    & C8    & Blue tarp & Brown tarp \\
		\hline
	$\omega _{in}$   & 1     & 1     & 3     & 11    & 11    & 13    & 13    & 13    & 5     & 5 \\
	$\omega _{out}$   & 3     & 3     & 5     & 13    & 13    & 15    & 15    & 15    & 7     & 7 \\
	\hline
\end{tabular}%
	\end{center}
\end{table*}
\begin{table*}[!thp]
	\caption{Simulation Parameter Settings for Training Sets of the Avon, SanDiego, and Gulfport Datasets.}
	\label{tab:ParameterSettings}
	\begin{center}
		\renewcommand\arraystretch{1.0}
		\begin{tabular}{ccccccc}
			\hline
			Dataset & Test data & Training data & Object pixels  & Maximum abundance & Training images & Training samples \\
			\hline
			Avon  & Avon Flight 0920-1701 & Avon Flight 0920-1631 & 11-16 & 0.95-1 & 150   & 1500/1500 \\
			SanDiego & SanDiego & Avon Flight 0920-1631 & 11-20 & 0.95-1 & 100   & 2000 \\
			Gulfport & Gulfport  Flight3 & Gulfport  Flight4 & 3-10  & 0.7-1 & 200   & 1000/1000/1000/1000 \\
			\hline
		\end{tabular}%
	\end{center}
\end{table*}

We implement several compared  HTD methods in Python, including ASD \cite{ASD99-TD}, CEM \cite{CEM-TD}, OSP \cite{OSP-TD}, KOSP \cite{KOSP}, SMF \cite{SMF-TD}, KSMF \cite{kwon2006comparative}, TCIMF \cite{TCIMF-TD}, CR \cite{CR-AD}, KSR \cite{KSR-C}, and KSRBBH \cite{KSRBBH-TD}. These methods follow the classic HTD pipeline, where each type of object spectra is sequentially used as the prior spectra, and each pixel in the all test images is sequentially treated as the pixel to be tested. The background pixels are selected from a dual-window centered on the pixel to be tested.
The dual-window sizes $\left( \omega_{in}, \omega_{out} \right)$ are configured based on the size of each type of object, as detailed in Table~\ref{tab:HTDParam}. 
We tune the other hyperparameters of these algorithms using the test image of the Avon dataset.

%

\section{Generation of Simulated Training Data for Public HTD Datasets}
Traditional HTD  typically provides only a single test image and a few, or even just one, prior spectra. 
In contrast, the proposed hyperspectral point object detection framework requires a large number of annotated training images and training samples. 
To address this, we adopt the data simulation method from the SPOD dataset to generate simulated training sets for three public HTD datasets: Avon, SanDiego, and Gulfport. These datasets contain single-spectrum objects of 2, 1, and 4 classes, respectively. For each objects class, we extract a pure target spectrum from the test image for simulation.
Table~\ref{tab:ParameterSettings} presents the simulation parameter settings for the training sets of the three datasets. Taking the Avon dataset as an example, we crop 150 training images of size 128$\times$128 from the flight line data numbered 0920-1631. Each training image contains 10 simulated blue tarps and 10 simulated brown tarps randomly placed. The number of pixels for each simulated object is randomly selected from the range (11, 16), and the pixels are combined in a random clustered pattern. The object pixels not adjacent to background pixels are considered pure pixels, with object spectral abundance set to 1. The object pixels adjacent to background pixels are treated as mixed pixels, with object spectral abundance randomly generated from the range (0.1, 1). The abundance of mixed pixels is assigned based on their distance to the object center, with closer pixels receiving higher abundance values. Additionally, if  all the object pixels are mixed pixels, the abundance of the center pixel is randomly selected from the maximum abundance range (0.95, 1).
For the Gulfport dataset, where objects are generally smaller and many consist entirely of mixed pixels, the pixel number range of simulated objects is set to  (3, 10), and the maximum abundance range is adjusted to (0.3, 1).

\begin{algorithm}[!t]
	\caption{Hyperspectral Point Object Detection Network}
	\label{alg:object_detection}
	\begin{algorithmic}[1]
		\REQUIRE Training hyperspectral image set $D_{train}$, test hyperspectral image set $D_{test}$, point object detection model $M$, learning rate $\eta$, number of epochs $E$, batch size $B$
		\ENSURE Trained model $M^*$, detection results $R$ on the test set
		\STATE \textbf{Training Phase:}
		\FOR{each epoch $e = 1$ to $E$}
		\STATE Shuffle the training dataset $D_{train}$
		\FOR{each batch $b = 1$ to $\lceil |D_{train}| / B \rceil$}
		\STATE Sample a batch of training hyperspectral images $X_b$ and corresponding labels $Y_b$ from $D_{train}$
		\STATE Forward pass: Compute model output $\hat{Y}_b = M(X_b)$
		\STATE Compute loss $L_b = \text{Loss}(\hat{Y}_b, Y_b)$
		\STATE Backward pass: Compute gradients $\nabla L_b$
		\STATE Update model parameters: $M \leftarrow M - \eta \nabla L_b$
		\ENDFOR
		\ENDFOR
		\STATE Save the trained model $M^* = M$
		\STATE \textbf{Inference Phase:}
		\FOR{each test hyperspectral image $x \in D_{test}$}
		\STATE Forward pass: Compute model output $\hat{y} = M^*(x)$
		\STATE Apply post-processing (e.g., non-maximum suppression) to $\hat{y}$ to obtain detection result $r$
		\STATE Add detection result $r$ to the result set $R$
		\ENDFOR
		\RETURN $M^*$, $R$
	\end{algorithmic}
\end{algorithm}

\begin{algorithm}[!t]
	\caption{Forward Pass Process of SpecDETR}
	\label{alg:specdetr}
	\begin{algorithmic}[1]
		\REQUIRE Hyperspectral image cube $\boldsymbol{X} \in \mathbb{R}^{H \times W \times N}$, number of encoder layers $E$, number of decoder layers $D$, number of object queries for matching $Q_{\text{match}}$, number of object queries for denoising training $Q_{\text{DN}}$, normalization constant $V$
		\ENSURE Predicted bboxes $\boldsymbol{B}^{\text{pred}}$, class confidence scores $\boldsymbol{C}^{\text{pred}}$
		\STATE \textbf{Data Tokenization:}
		\STATE Initialize pixel tokens: $\boldsymbol{P}_0=\mathrm{LN}\left( \mathrm{linear}\left( \boldsymbol{X}/V \right) \right) $
		\STATE Initialize global token: $\boldsymbol{g}_0=\left( \sum_{i=1}^{H\times W}{\boldsymbol{p}_{0,i}} \right) /\left( H\times W \right) $
		\STATE $\boldsymbol{F}_0 = \{\boldsymbol{P}_0, \boldsymbol{g}_0\}$
		\STATE \textbf{Transformer Encoder:}
		\FOR{$j = 1$ to $E$}
		\STATE $\tilde{\boldsymbol{F}}_j = \text{LN}\left(\text{Self-S2A}\left(\boldsymbol{F}_{j-1}\right) + \boldsymbol{F}_{j-1}\right)$ 
		\STATE $\boldsymbol{F}_j = \text{LN}\left(\text{FFN}\left(\tilde{\boldsymbol{F}}_j\right) + \tilde{\boldsymbol{F}}_j\right)$ 
		\ENDFOR
		\STATE \textbf{Transformer Decoder:}
		\STATE Generate initial anchor boxes $\boldsymbol{b}_{0,i}$ and class scores $\boldsymbol{c}_{0,i}$:
		\FOR{$i = 1$ to $H \times W$}
		\STATE $\boldsymbol{b}_{0,i} = \text{Sig}\left(\text{breg}_0\left(\boldsymbol{p}_{E,i}\right) + \text{InSig}\left(\boldsymbol{b}_{\text{int},i}\right)\right)$
		\STATE $\boldsymbol{c}_{0,i} = \text{Sig}\left(\text{cls}_0\left(\boldsymbol{p}_{E,i}\right)\right)$
		\ENDFOR
		\STATE Select top $Q_{\text{match}}$ anchor boxes.
		\IF{\textbf{training}}
		\STATE $Q=Q_{\text{match}}+Q_{\text{DN}}$
		\STATE Generate noised GT boxes $\boldsymbol{b}_{0,i}$ for $i = Q_{\text{match}}+1$ to $Q$
		\ELSE
		\STATE $Q=Q_{\text{match}}$
		\ENDIF
		\STATE Initialize object queries $\boldsymbol{q}_{0,i} = \mathbf{1}$ for $i = 1$ to $Q$
		\FOR{$d = 1$ to $D$}
		\FOR{$i = 1$ to $Q$}
		\STATE $\tilde{\boldsymbol{q}}_{d,i} = \text{LN}\left(\text{Cross-S2A}\left(\boldsymbol{q}_{d-1,i}, \boldsymbol{F}_E\right) + \boldsymbol{q}_{d-1,i}\right)$ 
		\STATE $\boldsymbol{q}_{d,i} = \text{LN}\left(\text{MLP}\left(\tilde{\boldsymbol{q}}_{d,i}\right) + \tilde{\boldsymbol{q}}_{d,i}\right)$ 
		\STATE Update  bbox $\boldsymbol{b}_{d,i}$ and class scores $\boldsymbol{c}_{d,i}$:
		\STATE $\hat{\boldsymbol{b}}_{d,i} = \text{Sig}\left(\text{breg}_d\left(\boldsymbol{q}_{d,i}\right) + \text{InSig}\left(\boldsymbol{b}_{d-1,i}\right)\right)$
		\STATE $\boldsymbol{b}_{d,i} = \text{detach}\left(\hat{\boldsymbol{b}}_{d,i}\right)$
		\STATE $\boldsymbol{b}_{d,i}^{\text{pred}} = \text{Sig}\left(\text{breg}_d\left(\boldsymbol{q}_{d,i}\right) + \text{InSig}\left(\hat{\boldsymbol{b}}_{d-1,i}\right)\right)$
		\STATE $\boldsymbol{c}_{d,i} = \text{Sig}\left(\text{cls}_d\left(\boldsymbol{q}_{d,i}\right)\right)$
		\ENDFOR
		\ENDFOR
		\RETURN  $\boldsymbol{B}^{\text{pred}} = \{\boldsymbol{b}_{D,i}^{\text{pred}}\}_{i=1}^Q$ ,$\boldsymbol{C}^{\text{pred}} = \{\boldsymbol{c}_{D,i}\}_{i=1}^Q$ 
	\end{algorithmic}
\end{algorithm}

\begin{algorithm}[!t]
	\setlength{\parskip}{1.0pt}
	\caption{Hybrid Label Assigner of SpecDETR}
	\label{alg:Assigner}
	\begin{algorithmic}[1]
		\REQUIRE  $ $ \\
		$\mathcal{G} $: Set of ground GT boxes, $\mathcal{G} = \{g_1, g_2, \dots, g_N\} $. \\
		$\mathcal{P} $: Set of predicted boxes, $\mathcal{P} = \{p_1, p_2, \dots, p_M\} $. \\
		$\tau_{\text{IoU}} $: IoU threshold for dynamic matching.\\
		$T$: Maximum number of positive samples per GT box for dynamic matching.  \\
		\ENSURE	$ $ \\
		$\mathcal{A} $: Set of assigned positive samples, $\mathcal{A} = \{(g_i, p_j)\} $  \\
		\STATE 	\textbf{Initialize:} $\mathcal{A} \leftarrow \emptyset $.\\
		\STATE \textbf{Step 1: Forced Matching}
		\STATE  Compute the combined loss $\mathcal{L}_{\text{comb}} $ for all $(g_i, p_j) $ pairs:\\
		$\mathcal{L}_{\text{comb}}(g_i, p_j) = \mathcal{L}_{\text{GIoU}}(g_i, p_j) + \mathcal{L}_{\text{L1}}(g_i, p_j) + \mathcal{L}_{\text{cls}}(g_i, p_j)$
		\STATE	Perform bipartite matching using the Hungarian algorithm:  \\
		$\mathcal{M}_{\text{forced}} \leftarrow \text{Hungarian}(\mathcal{L}_{\text{comb}})$
		\STATE	Assign one predicted box to each GT box:  \\
		$\mathcal{A} \leftarrow \mathcal{A} \cup \mathcal{M}_{\text{forced}}$
		\STATE  \textbf{Step 2: Dynamic Matching}
		\STATE Compute the IoU matrix $\mathbf{IoU} $ between $\mathcal{G} $ and $\mathcal{P} $:  \\
		$\mathbf{IoU}_{i,j} = \text{IoU}(g_i, p_j)$
		\FOR{each $g_i \in \mathcal{G} $}
		\STATE Find candidate predicted boxes with $\mathbf{IoU}_{i,j} > \tau_{\text{IoU}} $: \\ 
		$
		\mathcal{C}_i \leftarrow \{p_j \mid \mathbf{IoU}_{i,j} > \tau_{\text{IoU}}\}
		$
		\STATE Sort $\mathcal{C}_i $ by IoU in descending order:  \\
		$\mathcal{C}_i^{\text{sorted}} \leftarrow \text{Sort}(\mathcal{C}_i, \text{by } \mathbf{IoU}_{i,j})$
		\STATE Retain the top $T$ predicted boxes:   \\
		$
		\mathcal{C}_i^{\text{top}} \leftarrow \mathcal{C}_i^{\text{sorted}}[:T]
		$
		\STATE  Assign additional positive samples: \\ 
		$
		\mathcal{A} \leftarrow \mathcal{A} \cup \{(g_i, p_j) \mid p_j \in \mathcal{C}_i^{\text{top}}\}
		$
		\ENDFOR
		\RETURN $\mathcal{A} $
	\end{algorithmic}
\end{algorithm}

\section{Algorithmic Pseudocode}
To more clearly illustrate the differences between hyperspectral point object detection and the traditional HTD, we present the processing flow of the hyperspectral point object detection framework, as shown in Algorithm \ref{alg:object_detection}.
Additionally, Algorithm \ref{alg:specdetr} outlines the forward propagation process of SpecDETR on a single hyperspectral image, while Algorithm \ref{alg:Assigner} introduces the steps of the hybrid label assigner in SpecDETR.

\section{Comparison of SpecDETR and Sparse Representation-Based HTD Methods for Multi Classes}

Classic sparse representation-based methods in HTD are also adept at simultaneously detecting sub-pixel targets across multiple classes, where different classes correspond to different atoms. Sparse representation-based HTD methods can be primarily categorized into two types. The first type involves performing two sparse reconstructions on the test pixels: one using only background spectra as atoms, and the other using target spectra or a combination of target and background spectra as atoms. The difference between the reconstruction errors from these two processes is used as the detection score. The second type, represented by LSSA \cite{zhu2023learning}, performs a single sparse reconstruction using a combination of target and background spectra as atoms, and regards the coefficients of the target atoms as the detection scores. The first type of sparse representation-based method cannot directly identify the class of the detected target; therefore, in the comparative experiments on the SPOD and Avon datasets, we conduct separate detection for each object class. To compare the performance of this type of methods with SpecDETR in multi-class target detection, we treat all objects in the SPOD dataset as a single class and re-run this type of HTD method on the SPOD dataset, and also re-evaluate SpecDETR under the single-class object setting. As shown in Table~\ref{tab:spod_sr}, even when all objects are treated as the same class, this type of sparse representation method still underperforms SpecDETR in object-level evaluation. Furthermore, we improve LSSA to obtain prediction results for all object classes in a single run. However, in the comparative experiments on the SPOD and Avon datasets, LSSA also perform less effectively than SpecDETR in object-level evaluation.

Although sparse representation-based methods can utilize multi-class object spectral information, they still struggle to leverage multi-class object spatial information. For instance, in the SanDiego dataset, despite the low spatial resolution, aircraft still possess certain morphological information. On our developed SPOD dataset, C7 and C8 are combination objects consisting of two and three single-spectrum objects, respectively, meaning they actually have spatial characteristics. Current sparse representation-based HTD methods are designed from the perspective of prior target spectra, making it difficult for them to learn object spatial characteristics like SpecDETR.

\section{Loss Function}

SpecDETR uses the same loss function as DINO, which includes L1 loss and GIoU loss \cite{GIOULoss} for box regression, as well as focal loss \cite{FocalLoss} for classification. For each layer of the decoder, bbox L1 loss, bbox GIoU loss, and classification loss are computed individually for both the denoising and matching branches. Furthermore, these losses are also calculated for the candidate anchor boxes derived from the encoder's output features. The total loss employed for backpropagation is a weighted sum of all these individual losses. We use the same loss coefficients as those used in DINO, setting 1.0 for classification loss,  5.0 for bbox L1 loss, 2.0 for GIoU loss.

\section{Generalization Analysis}

The training set and test set of the SPOD dataset we developed are both generated using the same simulation mechanism. The training set comprises 100 images, while the test set includes 500 images. Table.~\ref{tab:spodtraintest} presents the detection performance of SpecDETR on the training and test sets through cross-validation.
As illustrated in Fig.~\ref{fig:loss_epoch}, SpecDETR has been sufficiently fitted after 100 training epochs. Regardless of whether the training set or test set is used as the training images, when the training images are used as validation images, the average metrics across all categories are essentially consistent and close to 1, with only slight differences in the AP metrics for each category. The mAP values are 0.979 and 0.980, $\text{mAP}{\text{25}}$ are 0.997 and 1.000, mAR are 0.984 and 0.986, and $\text{mRe}{\text{25}}$ are both 1.000. This phenomenon indicates that SpecDETR has a small bias on the SPOD Dataset.
When validated on unseen images during training, the detection performance of SpecDETR decreases, showing a clear variance. Additionally, increasing the scale of the training data reduces the variance of SpecDETR and enhances the network's generalization capability. When trained on the training set and validated on the test set, SpecDETR's mAP drops to 0.856, and mAR drops to 0.897. Conversely, when trained on the test set and validated on the training set, SpecDETR's mAP is 0.922, and mAR is 0.942.

The test set of the Avon dataset only consists of a single image containing 24 real single-spectrum point objects, each with at least one pixel of high object spectral abundance, while the training set comprises 150 simulated images. Given that the test set of the Avon dataset comprises only one image, we solely analyze the performance of SpecDETR using the training set of the Avon Dataset as the training data. Table.~\ref{tab:avontraintest} presents the detection performance of SpecDETR on the Avon dataset when validated on the training and test sets.
The detection metrics of SpecDETR on the training set are nearly 1, indicating a small bias of SpecDETR on the AVON dataset. Considering the potential errors in manually annotated labels in the test set, we focus on the metrics $\text{mAP}{\text{25}}$ and $\text{mRe}{\text{25}}$ calculated at an IoU threshold of 0.25. SpecDETR achieves $\text{mAP}{\text{25}}$ of 0.990 and $\text{mRe}{\text{25}}$ of 1.000 on the test set, demonstrating good generalization capability of SpecDETR trained on simulated samples to real objects in the Avon dataset.


\begin{table*}[!t]
	\caption{Performance Comparison of SpecDETR on the SPOD Dataset under Different Training and Validation Data Combinations.}
	\label{tab:spodtraintest}
	\begin{center}
		\renewcommand\arraystretch{1.0}
		\scriptsize
		{
	\begin{tabular}{cc|cccccccccccc}
				\hline
				
	Training  & 		Validation  & mAP$\uparrow$   & $\text{mAP}_{\text{25}}$$\uparrow$  & mAR$\uparrow$    & $\text{mRe}_{\text{25}}$$\uparrow$  & $\text{AP}_{\text{C1}}$$\uparrow$  & $\text{AP}_{\text{C2}}$$\uparrow$  & $\text{AP}_{\text{C3}}$$\uparrow$  & $\text{AP}_{\text{C4}}$$\uparrow$  & $\text{AP}_{\text{C5}}$$\uparrow$  & $\text{AP}_{\text{C6}}$$\uparrow$  & $\text{AP}_{\text{C7}}$$\uparrow$  & $\text{AP}_{\text{C8}}$$\uparrow$ \\
	\hline
	\multirow{2}[0]{*}{Training set}	&	Training set  & 0.980 & 1.000 & 0.984 & 1.000 & 1.000 & 1.000 &0.989 & 0.967 & 0.958 & 1.000 & 0.990 & 0.937 \\
		&	Test set  & 0.856 & 0.938 & 0.897 & 0.955 & 0.963 & 0.969 & 0.970 & 0.698 & 0.648 & 0.905 & 0.844 & 0.850 \\
	\hline
	\multirow{2}[0]{*}{Test set}	&		Test set & 0.979 & 0.997 & 0.986 & 1.000 & 0.990 & 1.000 & 1.000 & 0.975 & 0.959 & 0.977 & 0.967 & 0.967 \\
		&	Training set & 0.922 & 0.974 & 0.942 & 0.980 &  0.947 & 0.990 & 0.980 & 0.895 & 0.823 & 0.964 & 0.883 & 0.891 \\
	\hline
	\end{tabular}%
		}
	\end{center}
\end{table*}

\begin{table}[!t]
	\caption{Performance Comparison of SpecDETR on the A Dataset under Different Validation Data.}
	\label{tab:avontraintest}
	\begin{center}
		\renewcommand\arraystretch{1.0}
		\scriptsize
		{
			\begin{tabular}{c|cccccc}
				\hline
				Validation  & mAP$\uparrow$   & $\text{mAP}_{\text{25}}$$\uparrow$  & mAR$\uparrow$    & $\text{mRe}_{\text{25}}$$\uparrow$  & $\text{AP}_{\text{BL}}$$\uparrow$ &$\text{AP}_{\text{BR}}$$\uparrow$ \\
				\hline
				Training set & 0.979 & 0.998 & 0.991 & 1.000 & 0.987 & 0.971 \\
				Test set  & 0.885 & 0.990 & 0.925 & 1.000 & 0.924 & 0.847 \\
				\hline
			\end{tabular}%
		}
	\end{center}
\end{table}
\begin{table}[!t]
	\centering
	\caption{Performance comparison of SpecDETR and Sparse Representation-Based HTD Methods  on the SPOD Dataset under the Single-Class Object Setting.}
	\renewcommand\arraystretch{1.2}
	\scriptsize
	\setlength{\tabcolsep}{1mm}{
		\begin{tabular}{c|cc|cccc}
			\hline
			Method &  AUC  &  IoU  &  AP   &  $\text{AP}_{\text{25}}$ &  AR   &  $\text{Re}_{\text{25}}$ \\
			\hline
			CR \cite{CR-AD}    & 0.943 & 0.446 & 0.031 & 0.346 & 0.144 & 0.566 \\
			KSR \cite{KSR-C}   & 0.961 & 0.342 & 0.018 & 0.202 & 0.143 & 0.556 \\
			KSRBBH \cite{KSRBBH-TD} & 0.915 & 0.506 & 0.073 & 0.341 & 0.279 & 0.640 \\
			\hline
			SpecDETR & - & - & 0.907 & 0.989 & 0.930 & 0.994 \\
			\hline
		\end{tabular}%
	}
	\label{tab:spod_sr}%
\end{table}%

\section{Convergence Analysis}

\begin{figure}[t]
	
	\centering
	\subfloat{\includegraphics[width=0.9\linewidth]{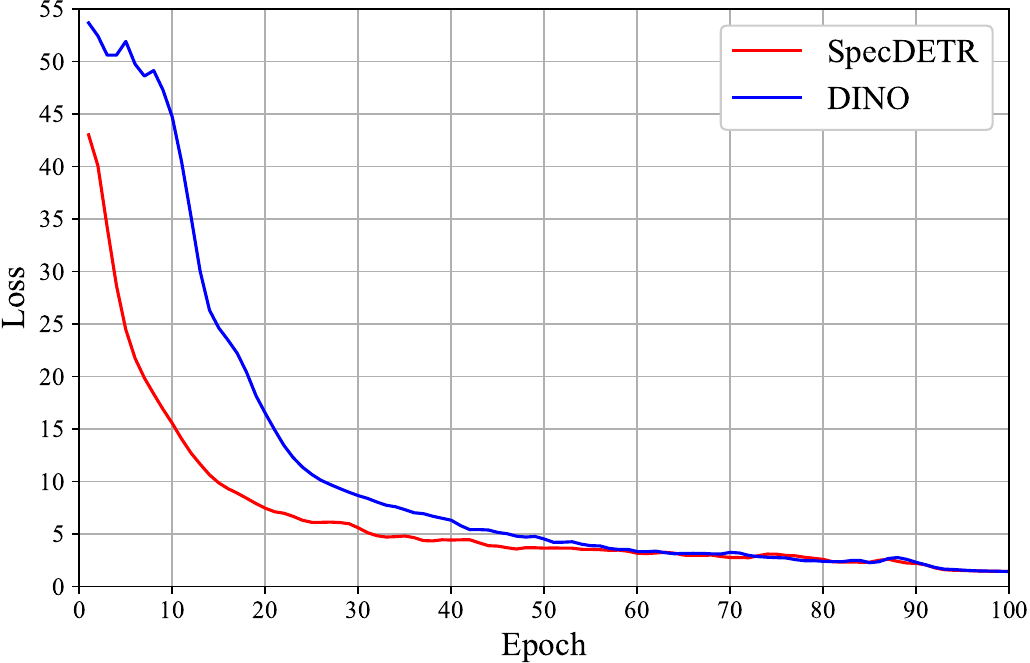}}
	\hfil
	\caption{Training Loss convergence curves of SpecDETR and DINO on the SPOD dataset. \label{fig:loss_epoch}}
\end{figure}

\begin{figure}[t]

	\centering
	\subfloat{\includegraphics[width=0.9\linewidth]{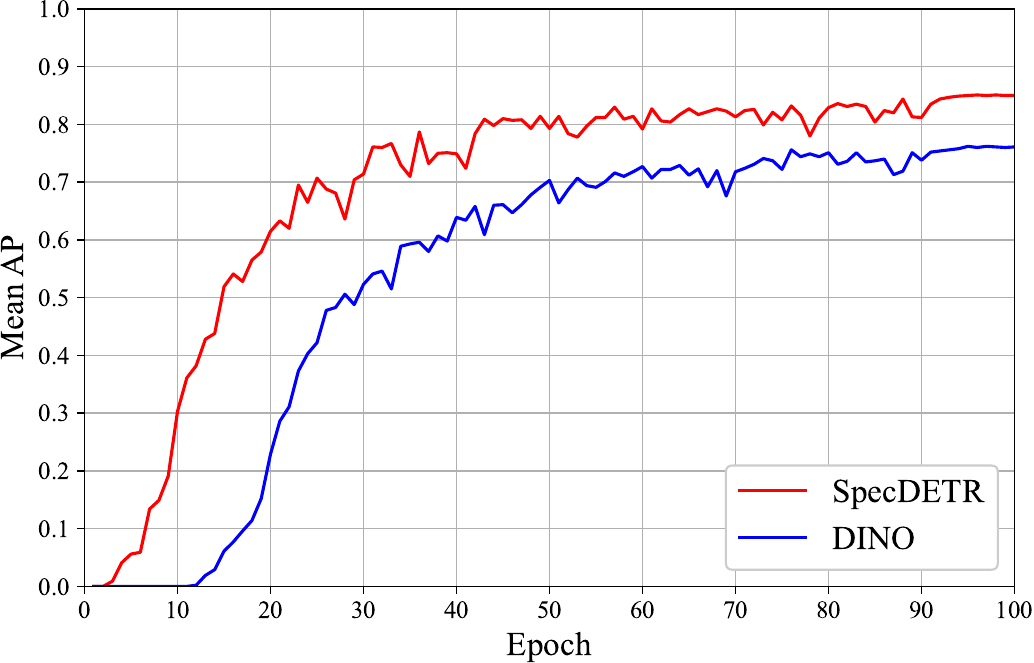}}
	\hfil
	\caption{AP Convergence curves of SpecDETR and DINO on the SPOD dataset.  	\label{fig:ap_epoch}}
\end{figure}

Figs.~\ref{fig:loss_epoch} and ~\ref{fig:ap_epoch} present the training loss and mAP curves over epochs for our proposed SpecDETR and the current SOTA object detection network DINO on the SPOD dataset. Both SpecDETR and DINO are trained for 100 epochs, with the learning rate reduced to 0.1 of its original value at the 90th epoch.
It is evident that during the first 50 epochs, SpecDETR exhibits a faster increase in mAP and a faster decrease in training loss compared to DINO. This indicates that SpecDETR has a superior convergence speed over DINO. In the latter 40 epochs, the training loss curves of SpecDETR and DINO are largely consistent, and the training loss remains essentially unchanged in the final few epochs, suggesting that both SpecDETR and DINO are well-fitted to the training set after training.
However, the final mAP of SpecDETR is significantly higher than that of DINO, demonstrating that SpecDETR has better generalization and convergence properties than DINO.

\begin{table}[t]
	\centering
	\caption{Object-Level Detection Performance of ASD under Different Segmentation Pixel Number Settings on the Avon Dataset.}
	\renewcommand\arraystretch{1.2}
	\scriptsize
	\setlength{\tabcolsep}{1.1mm}{
		\begin{tabular}{c|cccccccccc}
			\hline
			Pixels & 10    & 20    & 30    & 40    & 50    & 60    & 70    & 80    & 90    & 100 \\
			mAP    & 0.000 & 0.000 & 0.009 & 0.029 & 0.101 & 0.198 & 0.262 & 0.294 & 0.318 & 0.299 \\
			\hline
			Pixels & 110   & 120   & 130   & 140   & 150   & 160   & 170   & 180   & 190   & 200 \\
			mAP    & 0.252 & 0.245 & 0.219 & 0.201 & 0.187 & 0.172 & 0.160 & 0.153 & 0.146 & 0.143 \\
			\hline
		\end{tabular}%
	}
	\label{tab:HTDtopkeval}%
\end{table}%

\section{Discussion on the Conversion of Detection Score Maps to Object-Level Results for the HTD Methods}
Current literature related to the HTD methods often evaluates the detection score map of a single detection image using ROC curves and AUC. However, our extended point object detection task employs an object-level evaluation approach, necessitating the binarization of the HTD methods' detection score maps and their conversion into object-level predictions.
In our experiments, to maximize the HTD methods' performance on object-level evaluation metrics, we adopt the segmentation threshold corresponding to the maximum segmentation IoU across all images in the test set for each object class. Segmentation IoU is a widely used metric in segmentation tasks and has also become a benchmark evaluation metric for the infrared small target detection task.
Another conversion approach is to directly select the top K pixels from the HTD methods' detection score maps. However, determining the appropriate number of pixels is challenging. A smaller number of pixels may result in missed objects or undersized object shapes, while a larger number may cause oversized object shapes or the merging of different objects. On the SPOD dataset, the number of pixels for different object classes varies within the same test image, and it also varies across different test images. A fixed number of pixels is difficult to apply simultaneously to different test images and object classes.
Table.~\ref{tab:HTDtopkeval} presents the mAP of ASD under different segmentation pixel number settings on the Avon dataset. The Avon dataset consists of only one test image, and the pixel numbers for the two object classes are similar. Nevertheless, it is evident that the mAP is sensitive to the number of segmentation pixels. When the pixel number is 90, ASD's mAP reaches its best at 0.318, slightly lower than the 0.324 mAP under the optimal segmentation IoU criterion. When the pixel number is 70, the mAP drops to 0.262, and when the pixel number is 110, the mAP drops to 0.252. Therefore, selecting the top K pixels from the HTD methods' detection score maps is not well-suited for object-level evaluation.

\begin{table*}[t]
	\caption{AP performance comparison of SpecDETR and more object detection networks on the SPOD dataset.  Deformable DETR represents the original version, Deformable DETR+ represents Deformable DETR with iterative bounding box refinement, and Deformable DETR++  represents two-stage Deformable DETR  with iterative bounding box refinement. Table~\uppercase\expandafter{\romannumeral2} reports the results of DeformableDETR++.}
	\label{tabspodnet_apresult}
	\begin{center}
		\renewcommand\arraystretch{1.0}
		\scriptsize
		\setlength{\tabcolsep}{0.7mm}{
			
			\begin{tabular}{c|c|c|c|cccccccccccc}
				\hline
				Detector & Bakebone & Epochs & Size  & mAP$\uparrow$   & $\text{mAP}_{\text{25}}$$\uparrow$     &
				$\text{mAP}_{\text{50}}$$\uparrow$ &$\text{mAP}_{\text{75}}$$\uparrow$ & $\text{AP}_{\text{C1}}$$\uparrow$  & $\text{AP}_{\text{C2}}$$\uparrow$  & $\text{AP}_{\text{C3}}$$\uparrow$  & $\text{AP}_{\text{C4}}$$\uparrow$  & $\text{AP}_{\text{C5}}$$\uparrow$  & $\text{AP}_{\text{C6}}$$\uparrow$  & $\text{AP}_{\text{C7}}$$\uparrow$  & $\text{AP}_{\text{C8}}$$\uparrow$  \\
				\hline
				Faster R-CNN\cite{girshick2015fast} & ResNet50\cite{he2016deep} & 100   & $\times$4  & 0.197 & 0.377 & 0.374 & 0.179 & 0.000 & 0.000 & 0.000 & 0.035 & 0.026 & 0.537 & 0.430 & 0.550 \\
				Faster R-CNN\cite{girshick2015fast} & ResNet101\cite{he2016deep} & 100   & $\times$4  & 0.207 & 0.368 & 0.365 & 0.209 & 0.000 & 0.000 & 0.000 & 0.027 & 0.015 & 0.587 & 0.450 & 0.577 \\
				Faster R-CNN\cite{girshick2015fast} & Res2Net101\cite{gao2019res2net} & 100   & $\times$4  & 0.000 & 0.000 & 0.000 & 0.000 & 0.000 & 0.000 & 0.000 & 0.000 & 0.000 & 0.000 & 0.000 & 0.000 \\
				Faster R-CNN\cite{girshick2015fast} & RegNetX\cite{RegnetX} & 100   & $\times$4  & 0.227 & 0.379 & 0.378 & 0.242 & 0.000 & 0.000 & 0.000 & 0.042 & 0.043 & 0.631 & 0.522 & 0.578 \\
				Faster R-CNN\cite{girshick2015fast} & ResNeSt50\cite{ResNeSt} & 100   & $\times$4  & 0.246 & 0.316 & 0.316 & 0.277 & 0.000 & 0.000 & 0.000 & 0.008 & 0.003 & 0.644 & 0.564 & 0.747 \\
				Faster R-CNN\cite{girshick2015fast} & ResNeSt101\cite{ResNeSt} & 100   & $\times$4  & 0.183 & 0.253 & 0.252 & 0.216 & 0.000 & 0.000 & 0.000 & 0.006 & 0.001 & 0.386 & 0.392 & 0.681 \\
				Faster R-CNN\cite{girshick2015fast} & ResNeXt101\cite{ResNeXt} & 100   & $\times$4  & 0.220 & 0.368 & 0.366 & 0.231 & 0.000 & 0.000 & 0.000 & 0.016 & 0.010 & 0.618 & 0.518 & 0.596 \\
				Faster R-CNN\cite{girshick2015fast} & HRNet\cite{HRNet}  & 100   & $\times$4  & 0.320 & 0.404 & 0.402 & 0.345 & 0.000 & 0.000 & 0.000 & 0.107 & 0.076 & 0.849 & 0.731 & 0.793 \\
				PAA\cite{kim2020probabilistic}   & ResNet50\cite{he2016deep}  & 100   & $\times$4  & 0.000 & 0.000 & 0.000 & 0.000 & 0.000 & 0.000 & 0.000 & 0.000 & 0.000 & 0.000 & 0.000 & 0.000 \\
				PAA\cite{kim2020probabilistic}    & ResNet101\cite{he2016deep} & 100   & $\times$4  & 0.000 & 0.000 & 0.000 & 0.000 & 0.000 & 0.000 & 0.000 & 0.000 & 0.000 & 0.000 & 0.000 & 0.000 \\
				TOOD\cite{feng2021tood}   & ResNet50\cite{he2016deep} & 100   & $\times$4  & 0.000 & 0.000 & 0.000 & 0.000 & 0.000 & 0.000 & 0.000 & 0.000 & 0.000 & 0.000 & 0.000 & 0.000 \\
				TOOD\cite{feng2021tood}   & ResNeXt101\cite{ResNeXt} & 100   & $\times$4  & 0.304 & 0.464 & 0.440 & 0.303 & 0.000 & 0.000 & 0.000 & 0.181 & 0.194 & 0.743 & 0.648 & 0.663 \\
				FCOS\cite{tian2019fcos}  & ResNet50\cite{he2016deep} & 100   & $\times$4  & 0.000 & 0.000 & 0.000 & 0.000 & 0.000 & 0.000 & 0.000 & 0.000 & 0.000 & 0.000 & 0.000 & 0.000 \\
				CenterNet\cite{zhou2019objects} & ResNet50\cite{he2016deep} & 100   & $\times$4  & 0.000 & 0.000 & 0.000 & 0.000 & 0.000 & 0.000 & 0.000 & 0.000 & 0.000 & 0.000 & 0.000 & 0.000 \\
				FreeAnchor\cite{zhang2019freeanchor} & ResNet50\cite{he2016deep} & 100   & $\times$4  & 0.000 & 0.000 & 0.000 & 0.000 & 0.000 & 0.000 & 0.000 & 0.000 & 0.000 & 0.000 & 0.000 & 0.000 \\
				CentripetalNet\cite{CentripetalNet} & HourglassNet104\cite{HourglassNet} & 100   & $\times$4  & 0.695 & 0.829 & 0.805 & 0.673 & 0.831 & 0.888 & 0.915 & 0.373 & 0.367 & 0.810 & 0.655 & 0.725 \\
				CornerNet\cite{Cornernet} & HourglassNet104\cite{HourglassNet}  & 100   & $\times$4  & 0.626 & 0.736 & 0.712 & 0.609 & 0.797 & 0.751 & 0.855 & 0.328 & 0.308 & 0.768 & 0.554 & 0.644 \\
				RepPoints\cite{Reppoints} & ResNet50\cite{he2016deep}  & 100   & $\times$4  & 0.207 & 0.691 & 0.572 & 0.074 & 0.043 & 0.143 & 0.269 & 0.071 & 0.073 & 0.372 & 0.265 & 0.417 \\
				RepPoints\cite{Reppoints} & ResNeXt101\cite{ResNeXt} & 100   & $\times$4  & 0.485 & 0.806 & 0.790 & 0.540 & 0.373 & 0.561 & 0.632 & 0.242 & 0.253 & 0.658 & 0.508 & 0.649 \\
				RetinaNet\cite{Reppoints} & ResNet50\cite{he2016deep} & 100   & $\times$4  & 0.000 & 0.000 & 0.000 & 0.000 & 0.000 & 0.000 & 0.000 & 0.000 & 0.000 & 0.000 & 0.000 & 0.000 \\
				RetinaNet\cite{Reppoints} & EfficientNet\cite{tan2019efficientnet} & 100   & $\times$4  & 0.462 & 0.836 & 0.811 & 0.466 & 0.566 & 0.602 & 0.530 & 0.182 & 0.210 & 0.566 & 0.471 & 0.569 \\
				RetinaNet\cite{Reppoints} & PVTv2-B3\cite{wang2022pvt} & 100   & $\times$4  & 0.426 & 0.757 & 0.734 & 0.442 & 0.356 & 0.478 & 0.458 & 0.209 & 0.232 & 0.563 & 0.470 & 0.644 \\
				ATSS\cite{zhang2020bridging}  & ResNet50\cite{he2016deep} & 100   & $\times$4  & 0.000 & 0.000 & 0.000 & 0.000 & 0.000 & 0.000 & 0.000 & 0.000 & 0.000 & 0.000 & 0.000 & 0.000 \\
				ATSS\cite{zhang2020bridging}  & Swin-L\cite{liu2021swin} & 100   & $\times$4  & 0.000 & 0.000 & 0.000 & 0.000 & 0.000 & 0.000 & 0.000 & 0.000 & 0.000 & 0.000 & 0.000 & 0.000 \\
				Sparse R-CNN\cite{sun2021sparse} & ResNet50\cite{he2016deep} & 100   & $\times$4  & 0.011 & 0.109 & 0.048 & 0.001 & 0.000 & 0.000 & 0.000 & 0.003 & 0.003 & 0.015 & 0.023 & 0.042 \\
				Sparse R-CNN\cite{sun2021sparse} & ResNet101\cite{he2016deep} & 100   & $\times$4  & 0.000 & 0.000 & 0.000 & 0.000 & 0.000 & 0.000 & 0.000 & 0.000 & 0.000 & 0.000 & 0.000 & 0.000 \\
				DETR\cite{carion2020end}  & ResNet50\cite{he2016deep}  & 100   & $\times$4  & 0.000 & 0.000 & 0.000 & 0.000 & 0.000 & 0.000 & 0.000 & 0.000 & 0.000 & 0.000 & 0.000 & 0.000 \\
				ConditionalDETR\cite{meng2021conditional} & ResNet50\cite{he2016deep}& 100   & $\times$4  & 0.000 & 0.000 & 0.000 & 0.000 & 0.000 & 0.000 & 0.000 & 0.000 & 0.000 & 0.000 & 0.000 & 0.000 \\
				DAB-DETR\cite{liu2022dab} & ResNet50\cite{he2016deep}  & 100   & $\times$4  & 0.000 & 0.000 & 0.000 & 0.000 & 0.000 & 0.000 & 0.000 & 0.000 & 0.000 & 0.000 & 0.000 & 0.000 \\
				DeformableDETR\cite{zhu2020deformable}  & ResNet50\cite{he2016deep} & 100   & $\times$4  & 0.022 & 0.174 & 0.089 & 0.002 & 0.001 & 0.002 & 0.001 & 0.005 & 0.006 & 0.031 & 0.024 & 0.109 \\
				DeformableDETR+\cite{zhu2020deformable}  & ResNet50\cite{he2016deep} & 100   & $\times$4  & 0.106 & 0.410 & 0.321 & 0.032 & 0.008 & 0.028 & 0.154 & 0.031 & 0.031 & 0.206 & 0.105 & 0.283 \\
				DeformableDETR++\cite{zhu2020deformable} & ResNet50\cite{he2016deep} & 100   & $\times$4  & 0.231 & 0.692 & 0.560 & 0.147 & 0.230 & 0.316 & 0.234 & 0.077 & 0.070 & 0.289 & 0.238 & 0.395 \\
				DINO\cite{zhang2022dino}  & ResNet50\cite{he2016deep} & 100   & $\times$4  & 0.168 & 0.491 & 0.418 & 0.097 & 0.020 & 0.047 & 0.277 & 0.080 & 0.064 & 0.286 & 0.213 & 0.360 \\
				DINO\cite{zhang2022dino}  & Swin-L\cite{liu2021swin} & 100   & $\times$4  & \textcolor[rgb]{ 0, 0, 1}{\textbf{0.757}} & \textcolor[rgb]{ 0, 0, 1}{\textbf{0.852}} & \textcolor[rgb]{ 0, 0, 1}{\textbf{0.842}} & \textcolor[rgb]{ 0, 0, 1}{\textbf{0.764}} & \textcolor[rgb]{ 0, 0, 1}{\textbf{0.915}} & \textcolor[rgb]{ 0, 0, 1}{\textbf{0.912}} & \textcolor[rgb]{ 0, 0, 1}{\textbf{0.951}} & \textcolor[rgb]{ 0, 0, 1}{\textbf{0.483}} & \textcolor[rgb]{ 0, 0, 1}{\textbf{0.497}} & \textcolor[rgb]{ 0, 0, 1}{\textbf{0.847}} & \textcolor[rgb]{ 0, 0, 1}{\textbf{0.728}} & \textcolor[rgb]{ 0, 0, 1}{\textbf{0.721}} 
				\\
				\hline
				\multirow{3}[0]{*}{SpecDETR} & \multirow{3}[0]{*}{-} & 24    & $\times$1  & 0.706 & 0.877 & 0.873 & 0.734 & 0.946 & 0.966 & 0.937 & 0.360 & 0.357 & 0.757 & 0.645 & 0.682 \\
				&       & 36    & $\times$1  & 0.799 & 0.907 & 0.903 & 0.821 & 0.958 & 0.965 & 0.961 & 0.562 & 0.501 & 0.869 & 0.777 & 0.799 \\
				&       & 100   & $\times$1  &
				\textcolor[rgb]{1, 0, 0}{\textbf{0.856}} & \textcolor[rgb]{1, 0, 0}{\textbf{0.938}} & \textcolor[rgb]{1, 0, 0}{\textbf{0.930}} & \textcolor[rgb]{1, 0, 0}{\textbf{0.863}} & \textcolor[rgb]{1, 0, 0}{\textbf{0.963}} & \textcolor[rgb]{1, 0, 0}{\textbf{0.969}} & \textcolor[rgb]{1, 0, 0}{\textbf{0.970}} & \textcolor[rgb]{1, 0, 0}{\textbf{0.698}} & \textcolor[rgb]{1, 0, 0}{\textbf{0.648}} & \textcolor[rgb]{1, 0, 0}{\textbf{0.905}} & \textcolor[rgb]{1, 0, 0}{\textbf{0.844}} & \textcolor[rgb]{1, 0, 0}{\textbf{0.850}}
				\\
				
				\hline
			\end{tabular}%
		}
	\end{center}
\end{table*}

\begin{table*}[t]
	\caption{AR performance comparison of SpecDETR and more  object detection networks on the SPOD dataset.}
	\label{tabspodnet_arresult}
	\begin{center}
		\renewcommand\arraystretch{1.0}
		\scriptsize
		\setlength{\tabcolsep}{0.7mm}{
			
			\begin{tabular}{c|c|c|c|cccccccccccc}
				\hline
				Detector & Bakebone & Epochs & Size  & mAR$\uparrow$   & $\text{mAR}_{\text{25}}$$\uparrow$     &
				$\text{mAR}_{\text{50}}$$\uparrow$ &$\text{mAR}_{\text{75}}$$\uparrow$ & $\text{AR}_{\text{C1}}$$\uparrow$  & $\text{AR}_{\text{C2}}$$\uparrow$  & $\text{AR}_{\text{C3}}$$\uparrow$  & $\text{AR}_{\text{C4}}$$\uparrow$  & $\text{AR}_{\text{C5}}$$\uparrow$  & $\text{AR}_{\text{C6}}$$\uparrow$  & $\text{AR}_{\text{C7}}$$\uparrow$  & $\text{AR}_{\text{C8}}$$\uparrow$  \\
				\hline
				
				Faster R-CNN\cite{girshick2015fast} & ResNet50\cite{he2016deep}  & 100   & $\times$4 & 0.245 & 0.419 & 0.414 & 0.254 & 0.000 & 0.000 & 0.000 & 0.102 & 0.083 & 0.615 & 0.532 & 0.627 \\
				Faster R-CNN\cite{girshick2015fast} & ResNet101\cite{he2016deep}  & 100   & $\times$4  & 0.248 & 0.393 & 0.390 & 0.272 & 0.000 & 0.000 & 0.000 & 0.081 & 0.046 & 0.660 & 0.553 & 0.645 \\
				Faster R-CNN\cite{girshick2015fast} & Res2Net101\cite{gao2019res2net} & 100   & $\times$4   & 0.000 & 0.000 & 0.000 & 0.000 & 0.000 & 0.000 & 0.000 & 0.000 & 0.000 & 0.000 & 0.000 & 0.000 \\
				Faster R-CNN\cite{girshick2015fast} & RegNetX\cite{RegnetX} & 100   & $\times$4   & 0.267 & 0.399 & 0.398 & 0.304 & 0.000 & 0.000 & 0.000 & 0.089 & 0.090 & 0.699 & 0.607 & 0.654 \\
				Faster R-CNN\cite{girshick2015fast} & ResNeSt50\cite{ResNeSt} & 100   & $\times$4  & 0.269 & 0.319 & 0.318 & 0.294 & 0.000 & 0.000 & 0.000 & 0.007 & 0.004 & 0.682 & 0.633 & 0.828 \\
				Faster R-CNN\cite{girshick2015fast} & ResNeSt101\cite{ResNeSt} & 100   & $\times$4 & 0.201 & 0.252 & 0.252 & 0.229 & 0.000 & 0.000 & 0.000 & 0.003 & 0.001 & 0.410 & 0.437 & 0.760 \\
				Faster R-CNN\cite{girshick2015fast} & ResNeXt101\cite{ResNeXt} & 100   & $\times$4   & 0.253 & 0.376 & 0.375 & 0.286 & 0.000 & 0.000 & 0.000 & 0.027 & 0.022 & 0.701 & 0.606 & 0.665 \\
				Faster R-CNN\cite{girshick2015fast} & HRNet\cite{HRNet} & 100   & $\times$4  & 0.364 & 0.434 & 0.430 & 0.386 & 0.000 & 0.000 & 0.000 & 0.185 & 0.155 & 0.899 & 0.818 & 0.859 \\
				PAA\cite{kim2020probabilistic}   & ResNet50\cite{he2016deep}  & 100   & $\times$4  & 0.000 & 0.000 & 0.000 & 0.000 & 0.000 & 0.000 & 0.000 & 0.000 & 0.000 & 0.000 & 0.000 & 0.000 \\
				PAA\cite{kim2020probabilistic}    & ResNet101\cite{he2016deep}  & 100   & $\times$4  & 0.000 & 0.000 & 0.000 & 0.000 & 0.000 & 0.000 & 0.000 & 0.000 & 0.000 & 0.000 & 0.000 & 0.000 \\
				TOOD\cite{feng2021tood}   & ResNet50\cite{he2016deep}  & 100   & $\times$4   & 0.000 & 0.004 & 0.001 & 0.000 & 0.000 & 0.000 & 0.000 & 0.000 & 0.000 & 0.000 & 0.001 & 0.001 \\
				TOOD\cite{feng2021tood}   & ResNeXt101\cite{ResNeXt} & 100   & $\times$4   & 0.401 & 0.570 & 0.532 & 0.391 & 0.000 & 0.000 & 0.000 & 0.429 & 0.442 & 0.816 & 0.756 & 0.764 \\
				FCOS\cite{tian2019fcos}  & ResNet50\cite{he2016deep} & 100   & $\times$4   & 0.000 & 0.003 & 0.002 & 0.000 & 0.000 & 0.000 & 0.000 & 0.000 & 0.000 & 0.000 & 0.000 & 0.004 \\
				CenterNet\cite{zhou2019objects} & ResNet50\cite{he2016deep} & 100   & $\times$4  & 0.000 & 0.000 & 0.000 & 0.000 & 0.000 & 0.000 & 0.000 & 0.000 & 0.000 & 0.000 & 0.000 & 0.000 \\
				FreeAnchor\cite{zhang2019freeanchor} & ResNet50\cite{he2016deep} & 100   & $\times$4  & 0.000 & 0.000 & 0.000 & 0.000 & 0.000 & 0.000 & 0.000 & 0.000 & 0.000 & 0.000 & 0.000 & 0.000 \\
				CentripetalNet\cite{CentripetalNet} & HourglassNet104\cite{HourglassNet} & 100   & $\times$4 & 0.840 & 0.956 & 0.932 & 0.817 & 0.878 & 0.936 & 0.939 & 0.769 & 0.761 & 0.866 & 0.766 & 0.801 \\
				CornerNet\cite{Cornernet} & HourglassNet104\cite{HourglassNet} & 100   & $\times$4  & 0.855 & 0.969 & 0.939 & 0.834 & 0.917 & \textcolor[rgb]{0, 0, 1}{\textbf{0.949}} & 0.940 & \textcolor[rgb]{0, 0, 1}{\textbf{0.804}} & \textcolor[rgb]{0, 0, 1}{\textbf{0.791}} & 0.881 & 0.749 & 0.806 \\
				RepPoints\cite{Reppoints} & ResNet50\cite{he2016deep} & 100   & $\times$4  & 0.346 & 0.934 & 0.804 & 0.229 & 0.185 & 0.279 & 0.379 & 0.275 & 0.272 & 0.467 & 0.402 & 0.505 \\
				RepPoints\cite{Reppoints} & ResNeXt101\cite{ResNeXt}  & 100   & $\times$4  & 0.635 & 0.961 & 0.939 & 0.725 & 0.523 & 0.662 & 0.713 & 0.563 & 0.565 & 0.735 & 0.606 & 0.714 \\
				RetinaNet\cite{Reppoints} & ResNet50\cite{he2016deep} & 100   & $\times$4  & 0.000 & 0.000 & 0.000 & 0.000 & 0.000 & 0.000 & 0.000 & 0.000 & 0.000 & 0.000 & 0.000 & 0.000 \\
				RetinaNet\cite{Reppoints} & EfficientNet\cite{tan2019efficientnet} & 100   & $\times$4  & 0.611 & 0.971 & 0.949 & 0.667 & 0.664 & 0.694 & 0.672 & 0.493 & 0.488 & 0.665 & 0.570 & 0.641 \\
				RetinaNet\cite{Reppoints} & PVTv2-B3\cite{wang2022pvt} & 100   & $\times$4  & 0.650 & \textcolor[rgb]{1, 0, 0}{\textbf{0.987}} & \textcolor[rgb]{0, 0, 1}{\textbf{0.973}} & 0.718 & 0.549 & 0.638 & 0.644 & 0.607 & 0.596 & 0.755 & 0.671 & 0.737 \\
				ATSS\cite{zhang2020bridging}  & ResNet50\cite{he2016deep} & 100   & $\times$4 & 0.001 & 0.004 & 0.003 & 0.000 & 0.000 & 0.000 & 0.000 & 0.000 & 0.000 & 0.000 & 0.000 & 0.005 \\
				ATSS\cite{zhang2020bridging}  & Swin-L\cite{liu2021swin}  & 100   & $\times$4  & 0.001 & 0.003 & 0.002 & 0.001 & 0.000 & 0.000 & 0.000 & 0.000 & 0.000 & 0.000 & 0.000 & 0.004 \\
				Sparse R-CNN\cite{sun2021sparse} & ResNet50\cite{he2016deep} & 100   & $\times$4  & 0.091 & 0.477 & 0.292 & 0.032 & 0.000 & 0.000 & 0.000 & 0.071 & 0.068 & 0.178 & 0.195 & 0.215 \\
				Sparse R-CNN\cite{sun2021sparse} & ResNet101\cite{he2016deep} & 100   & $\times$4   & 0.000 & 0.000 & 0.000 & 0.000 & 0.000 & 0.000 & 0.000 & 0.000 & 0.000 & 0.000 & 0.000 & 0.000 \\
				DETR\cite{carion2020end}  & ResNet50\cite{he2016deep} & 100   & $\times$4 & 0.000 & 0.000 & 0.000 & 0.000 & 0.000 & 0.000 & 0.000 & 0.000 & 0.000 & 0.000 & 0.000 & 0.000 \\
				ConditionalDETR\cite{meng2021conditional} & ResNet50\cite{he2016deep} & 100   & $\times$4  & 0.000 & 0.000 & 0.000 & 0.000 & 0.000 & 0.000 & 0.000 & 0.000 & 0.000 & 0.000 & 0.000 & 0.000 \\
				DAB-DETR\cite{liu2022dab} & ResNet50\cite{he2016deep} & 100   & $\times$4 & 0.000 & 0.015 & 0.001 & 0.000 & 0.000 & 0.000 & 0.000 & 0.000 & 0.000 & 0.000 & 0.000 & 0.001 \\
				DeformableDETR\cite{zhu2020deformable}  & ResNet50\cite{he2016deep} & 100   & $\times$4  & 0.111 & 0.639 & 0.354 & 0.039 & 0.018 & 0.023 & 0.019 & 0.082 & 0.079 & 0.212 & 0.189 & 0.263 \\
				Deformable DETR+\cite{zhu2020deformable}  & ResNet50\cite{he2016deep} & 100   & $\times$4  & 0.265 & 0.753 & 0.636 & 0.173 & 0.094 & 0.139 & 0.331 & 0.186 & 0.199 & 0.430 & 0.308 & 0.429 \\
				Deformable DETR++\cite{zhu2020deformable} & ResNet50\cite{he2016deep} & 100   & $\times$4   & 0.385 & 0.883 & 0.770 & 0.324 & 0.348 & 0.439 & 0.405 & 0.283 & 0.267 & 0.440 & 0.386 & 0.512 \\
				DINO\cite{zhang2022dino}  & ResNet50\cite{he2016deep} & 100   & $\times$4  & 0.368 & 0.763 & 0.684 & 0.339 & 0.120 & 0.177 & 0.436 & 0.328 & 0.336 & 0.499 & 0.469 & 0.577 \\
				DINO\cite{zhang2022dino}  & Swin-L\cite{liu2021swin} & 100   & $\times$4 & \textcolor[rgb]{1, 0, 0}{\textbf{0.909}} & \textcolor[rgb]{0, 0, 1}{\textbf{0.983}} & \textcolor[rgb]{1, 0, 0}{\textbf{0.976}} & \textcolor[rgb]{1, 0, 0}{\textbf{0.914}} & \textcolor[rgb]{0, 0, 1}{\textbf{0.948}} & 0.942 & \textcolor[rgb]{0, 0, 1}{\textbf{0.973}} & \textcolor[rgb]{1, 0, 0}{\textbf{0.869}} & \textcolor[rgb]{1, 0, 0}{\textbf{0.853}} & \textcolor[rgb]{0, 0, 1}{\textbf{0.933}} & \textcolor[rgb]{0, 0, 1}{\textbf{0.888}} & \textcolor[rgb]{0, 0, 1}{\textbf{0.865}}
				\\
				\hline
				\multirow{3}[0]{*}{SpecDETR} & \multirow{3}[0]{*}{-} & 24    & $\times$1  & 0.784 & 0.924 & 0.919 & 0.821 & 0.963 & 0.975 & 0.955 & 0.560 & 0.491 & 0.821 & 0.742 & 0.768 \\
				&       & 36    & $\times$1  & 0.854 & 0.937 & 0.933 & 0.873 & 0.974 & 0.970 & 0.975 & 0.696 & 0.602 & 0.909 & 0.841 & 0.867 \\
				&       & 100   & $\times$1  & \textcolor[rgb]{0, 0, 1}{\textbf{0.897}} & 0.955 & 0.948 & \textcolor[rgb]{0, 0, 1}{\textbf{0.902}} & \textcolor[rgb]{1, 0, 0}{\textbf{0.975}} & \textcolor[rgb]{1, 0, 0}{\textbf{0.974}} & \textcolor[rgb]{1, 0, 0}{\textbf{0.981}} & 0.766 & 0.742 & \textcolor[rgb]{1, 0, 0}{\textbf{0.945}} & \textcolor[rgb]{1, 0, 0}{\textbf{0.892}} & \textcolor[rgb]{1, 0, 0}{\textbf{0.903}} \\
				\hline
			\end{tabular}%
		}
	\end{center}
\end{table*}

\clearpage
\clearpage
\ifCLASSOPTIONcaptionsoff
\newpage
\fi


\bibliographystyle{IEEEtran}
\bibliography{IEEEabrv, AETNET, CSRBBH-TD, main}